\documentclass[a4paper,fleqn]{cas-dc}

\usepackage[numbers, super]{natbib}
\usepackage{amsmath}
\usepackage{amssymb}
\usepackage{amsthm}
\usepackage{multirow,booktabs}
\usepackage{algorithm}
\usepackage{algpseudocode}
\usepackage{amsmath}
\usepackage{geometry}
\usepackage{algorithmicx}
\usepackage{algpseudocode}
\usepackage{longtable}
\usepackage{subfig,graphicx}
\usepackage{hyperref}
\allowdisplaybreaks[4]
\usepackage{colortbl, color}
\definecolor{red}{RGB}{171, 75, 75}
\definecolor{grey}{RGB}{236, 236, 238}
\usepackage{float}
\usepackage{placeins}
\usepackage{stfloats}
\usepackage{bbding}
\usepackage{xurl}

\makeatletter
\@ifundefined{c@rownum}{%
  \let\c@rownum\rownum
}{}
\@ifundefined{therownum}{%
  \def\therownum{\@arabic\rownum}%
}{}
\makeatother

\newcommand{\subsubsubsection}[1]{\paragraph{#1}\mbox{}\\}
\def\tsc#1{\csdef{#1}{\textsc{\lowercase{#1}}\xspace}}
\tsc{WGM}
\tsc{QE}
\tsc{EP}
\tsc{PMS}
\tsc{BEC}
\tsc{DE}

\begin{document}
\let\printorcid\relax
\begin{sloppypar}

\let\WriteBookmarks\relax
\def\floatpagepagefraction{1}
\def\textpagefraction{.001}

\title [mode = title]{UFPS: A unified framework for partially-annotated federated segmentation in heterogeneous data distribution}                      

\author[1]{Le Jiang}
\author[1]{Li Yan Ma}	\cormark[1]	
\address[1]{School of Computer Engineering and Science, Shanghai university, Shanghai, China}

\author[2]{Tie Yong Zeng}
\address[2]{Department of Mathematics, Chinese University of Hong Kong, Hongkong, China}

\author[3]{Shi Hui Ying}
\address[3]{Department of Mathematics, Shanghai university, Shanghai, China}

\cortext[cor1]{Correspondence: \href{liyanma@shu.edu.cn}{liyanma@shu.edu.cn}}

\noindent\textbf{UFPS: A unified framework for partially-annotated federated segmentation in heterogeneous data distribution}

\noindent\textbf{Authors}\\
Le Jiang, Li Yan Ma, Tie Yong Zeng, Shi Hui Ying

\noindent\textbf{Correspondence}\\
\href{liyanma@shu.edu.cn}{liyanma@shu.edu.cn}

\noindent\textbf{eTOC blurb}\\
Partially supervised segmentation is a deep learning paradigm both frugal in label use and limited by data privacy issues and domain gaps in medical practice. This work presents a unified federated partially-labeled segmentation framework. By systematically analyzing challenges in federated partially supervised segmentation, we give more insights into how to generalize on different domains better without class collision based on partially-annotated datasets.

\noindent\textbf{Highlights}
\begin{itemize}
	\item Challenges and solutions for federated partially supervised segmentation are provided
	\item Training a global model via heterogeneous datasets without class intersections
	\item Comprehensive experiments demonstrate the global model is effective on most classes for all domains
\end{itemize}

\noindent\textbf{Graphical abstract}
\begin{figure}[htbp]
	\raggedright
		\includegraphics[width=0.5\columnwidth]{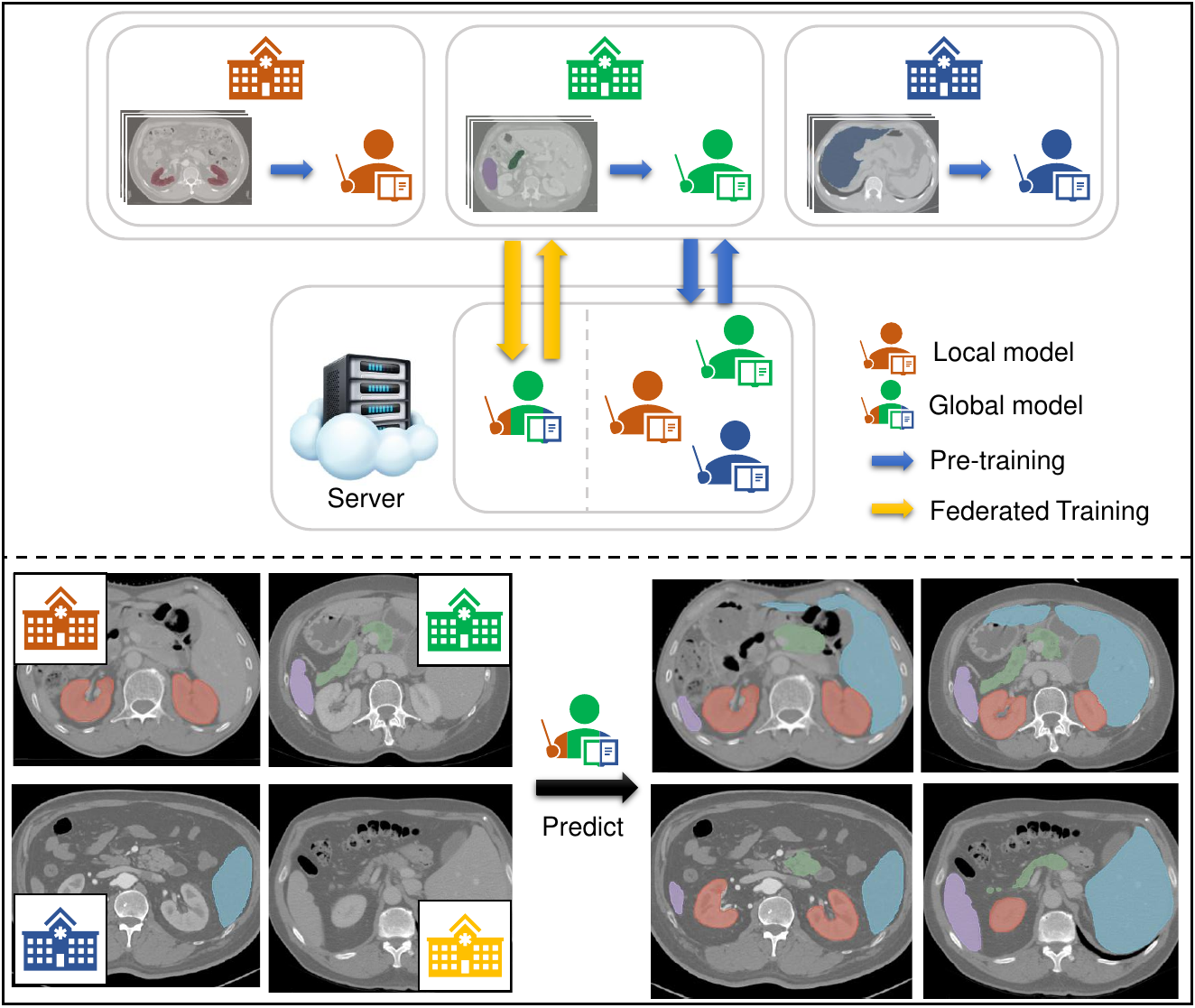}
	\label{Figure graph_abstract}
\end{figure}

\noindent\textbf{THE BIGGER PICTURE} Labeling numerous data for segmentation tasks is labor prohibitive and requires expert knowledge for some classes. Partially supervised segmentation task seeks to solve this issue based on several partially-annotated datasets but is restricted in medical practice due to data privacy and domain gaps. Federated learning can be one solution to resolve privacy concerns but there is hardly any work on the combination of both technologies. The authors analyze underlying challenges and propose a unified federated partially-labeled segmentation framework to conduct federated partially supervised segmentation task. Extensive experiments validate the promising performance of the proposed solution with an application under a heterogeneous real-world setting in a privacy preserving manner where knowledge specific participators cooperate to train a universal global model.

\noindent\textbf{Level 2.} Proof-of-Concept: Data science output has been formulated, implemented, and tested for one domain/problem

\renewcommand{\abstractname}{SUMMARY}
\begin{abstract}
Partially supervised segmentation is a label-saving method based on datasets with fractional classes labeled and intersectant. However, it is still far from landing on real-world medical applications due to privacy concerns and data heterogeneity. As a remedy without privacy leakage, federated partially supervised segmentation (FPSS) is formulated in this work. The main challenges for FPSS are class heterogeneity and client drift. We propose a Unified Federated Partially-labeled Segmentation (UFPS) framework to segment pixels within all classes for partially-annotated datasets by training a totipotential global model without class collision. Our framework includes Unified Label Learning and sparsed Unified Sharpness Aware Minimization for unification of class and feature space, respectively. We find that vanilla combinations for traditional methods in partially supervised segmentation and federated learning are mainly hampered by class collision through empirical study. Our comprehensive experiments on real medical datasets demonstrate better deconflicting and generalization ability of UFPS compared with modified methods. 
\end{abstract}

\begin{keywords}
	Federated Learning \sep Medical Image Segmentation \sep Partial Label
\end{keywords}

\maketitle

\renewcommand{\thefootnote}{\alph{footnote}}
\footnotetext[1]{These authors contributed equally.}
\renewcommand{\thefootnote}{\arabic{footnote}}

\newcommand{\upcite}[1]{\textsuperscript{\cite{#1}}}
\makeatletter
\def\@biblabel#1{#1.~}
\makeatother

\section{INTRODUCTION} \label{introduction}

Promoted by the progress of deep learning \cite{lecun2015deep}, techniques in the field of computer aided diagnosis\cite{DOI199997,DOI2007198} have assisted clinicians with effective routine. Most of these techniques need large-scale data with abundant diversity to maintain reliability\cite{greenwald2022whole,david2019applications}. Nevertheless, it requires specialized knowledge and intensive labor to collect annotations for medical data, especially for dense pixel-level tasks. Recently, partially supervised segmentation (PSS)\cite{zhou2019prior, SHI2021101979, 10.1007/978-3-030-58598-3_23, fang2020multi} has emerged as a label-saving means in allusion to the problem. Unlike traditional learning paradigm, PSS aims to achieve segmentation tasks on datasets with only a subset of all classes annotated for each dataset. Besides, there can be hardly any intersectant labeled classes between datasets in PSS, while the union includes all classes concurrently.

Existing works for PSS \cite{zhou2019prior, SHI2021101979, 10.1007/978-3-030-58598-3_23, fang2020multi} mainly count on centralized datasets, which are against privacy regulations in real-world medical applications\cite{annas2002medical}. Federated learning (FL)\cite{mcmahan2017communication} , a distributed learning framework, can serve as a promising solution to this challenge. It allows all clients (e.g., hospitals or apartments) to cooperate in training a global model with similar function as the one in centralized learning by aggregating model weights or gradients without data disclosure. However, utilizing partially-annotated labels to train a global segmentation model in the FL scenario is under-explored. In this work, we extend the formulation of PSS to an FL manner, i.e., federated partially supervised segmentation (FPSS). Apart from low demands for label integrity and protection for data privacy, this setting has the potential to boost model generalization through knowledge communication as well.

\begin{figure}
	\centering
	\includegraphics[width=\columnwidth]{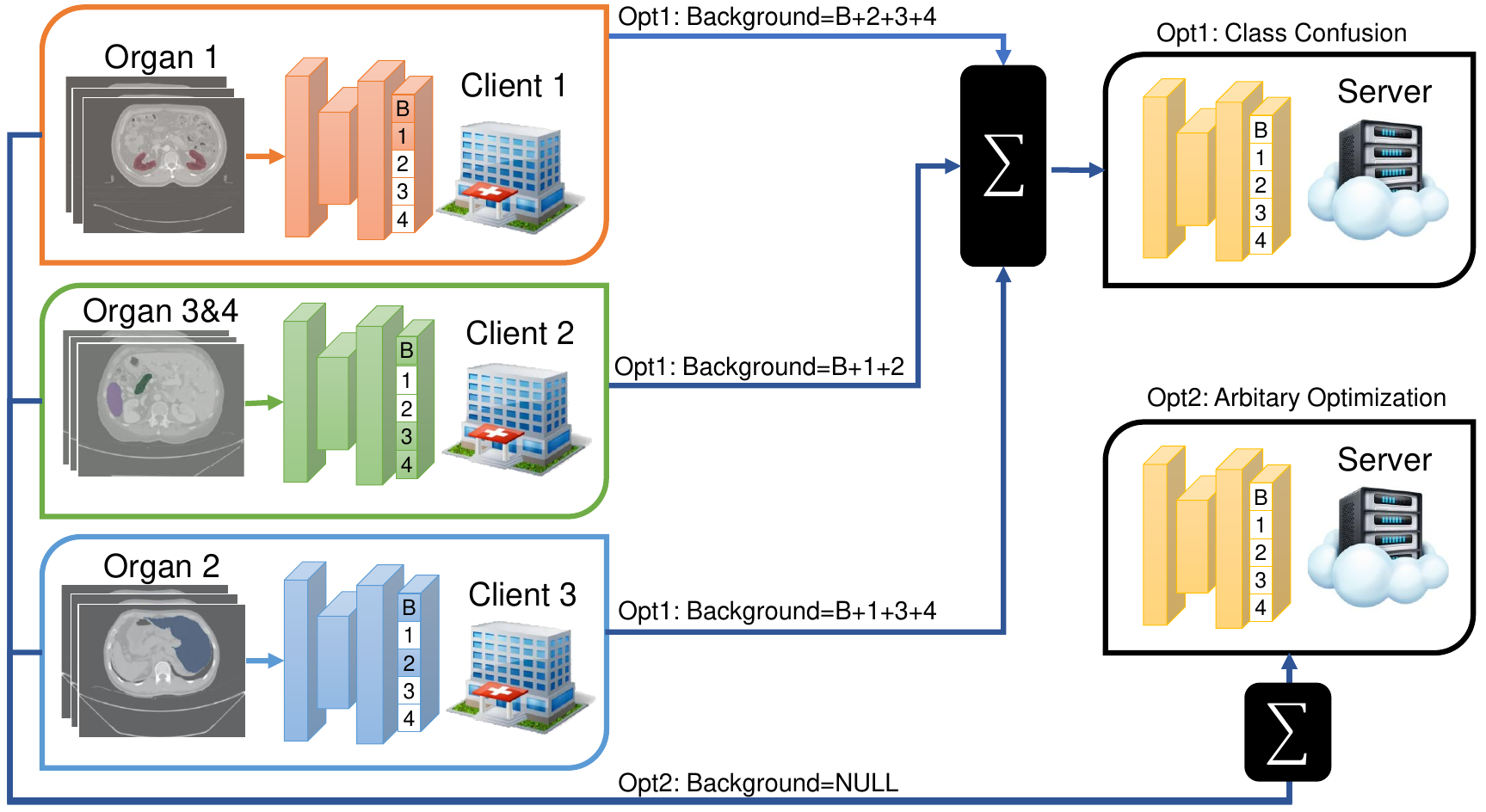}
	\caption{\textbf{Illustration of solutions in centralized learning to class heterogeneity problem for FPSS on CT images. }\\Each client has only two or three classes annotated, which are colored in the segmentation head. 'Opt 1' merges unannotated classes into the background class to calculate the loss. 'Opt 2' only uses the foreground class(es) to calculate the loss.}
	\label{Figure 1}
\end{figure}

\textit{The learning process for FPSS encounters two major challenges, namely, class heterogeneity and client drift.} The class heterogeneity problem is caused by inconsistency of annotated classes among clients. To give a straightforward illustration of solutions in centralized learning to the class heterogeneity problem, we present an example in \hyperref[Figure 1]{Figure 1}. When all unannotated classes are merged into the background class to calculate loss functions (Opt 1), the global model suffers severe class conflict. It comes to the fact that foreground annotations for each client are mistaken as background ones in other clients. When clients only use foreground classes to calculate loss (Opt 2), foreground channels without supervision can be optimized to any false direction. One simple but feasible solution for the class conflict issue is to aggregate part of the whole model globally and keep rest parts local, e.g., excluding the segmentation head from global model aggregation. However, some classes may be relevant in the medical field (e.g., relative organ position in CT image\cite{BALTER1998939, Zheng2017, chen2022magicnet}). This approach may hinder potential interaction between classes in the course of training and requires extra computational cost during evaluation.

In FL, client drift is caused by the assumption that data across clients are in non-independent and identically distributed distribution (non-IID)\cite{zhao2018federated, li2019convergence}. It can be ascribed to multiple factors in the medical field like differences in data collection protocols\cite{VANOMMEN201965} or devices\cite{tischenko2006new}, population diversity\cite{sharma2021improving}, etc. When weights or gradients with a huge divergence between clients are aggregated under the non-IID setting, the global model can be optimized to a suboptimal solution and the convergence speed can also be decelerated. Previous methods to tackle the problem in FL can be categorized into three major directions: local optimization rectification\cite{li2020federated, karimireddy2020scaffold, zhang2021federated, jiang2022harmofl, gao2022feddc, mendieta2022local, 10.1007/978-3-031-20050-2_38}, client selection\cite{balakrishnan2022diverse, tang2022fedcor}, and contrastive learning\cite{li2021model, 10.1007/978-3-031-20056-4_40, 10.1007/978-3-031-16443-9_25, yu2023multimodal, MU202393}. While client selection is mainly designed for full class supervised learning, contrastive learning is costly both in computational time and memory, especially for Unified Federated Partially-labeled Segmentation (UFPS) in the medical field. Although there are lots of works about local optimization rectification, most of them only emphasize on optimization, ignoring the importance of local data distribution.

To resolve the aforementioned two challenges, we propose a framework called Unified Federated Partially-labeled Segmentation (UFPS). It alleviates effects of class heterogeneity and client drift via Unified Label Learning (ULL) and sparse Unified Sharpness Aware Minimization (sUSAM) , respectively. 

In FPSS, only a subset of all classes are labeled for each client, and common PSS methods in centralized learning fail to generalize. Therefore, to exploit underlying class intersections without class conflict problem, ULL labels all classes in a unified manner based on pretrained class-specific teachers. Since the pretraining step is executed locally, clients can train their models at any time, which is free from communication burden and stability in FL\cite{posner2021federated}. Different from traditional pseudo labeling process, we filter the intersection part within pretrained teachers in the background channel as the ground truth to avert concept collision among clients. It is the first attempt to bring the pseudo labeling idea into FPSS. \textit{Thus, the class heterogeneity problem is translated to a noisy label learning issue.} 

We tackle the noisy label learning issue from both global and local perspectives. Since the global model benefits from overall data distribution and class interactions, it can serve as a more reliable source of pseudo labels. By increasing model aggregation weights of clients with high-quality data, the global model is more likely to concentrate on credible knowledge, thus better guiding local models. For local models, their common bottleneck is mainly caused by the coupling between noise and hard classes in pseudo labels. A loss weight scheduler is then proposed to alleviate side effects of noises while better fitting hard classes.

As a counterpart for the non-IID issue, Adaptive Sharpness Aware Minimization (ASAM), as an effective two-step approach has been proven in previous work FedASAM\cite{kwon2021asam}. Despite the good performance of ASAM, some directions may be far from global cliffy ways since these two steps are both based on the plain local dataset. When each local model is optimized towards the sharpest local direction, parts of these directions may be relevant to some client-specific attributes, thus restricting the generalization ability of the global model. Besides, the training time of FedASAM is doubled compared with FedAvg. For better generalization, our proposed sUSAM allows local models to approximate a unified optimization target for all clients through strong data augmentation. The effect of data augmentation is merely studied in the field of FL since underlying data information is banned from sharing. Besides, attempts to transfer traditional data augmentation to FL either promote the global model performance limitedly through slight augmentation or worsen it through strong augmentation \cite{kwon2021asam}. By decoupling training data for two steps in ASAM, local models can be free from training instability caused by strong data augmentation. To accelerate the ASAM based framework and avoid overfitting local-specific attributes, we only concentrate on the most essential ascent directions to optimize with and explore latent ones to enhance generalization ability of the global model. Through our experiments, we demonstrate our approach is capable of getting a large margin beyond previous methods. Our contributions can be concluded as:

\begin{itemize}
	\item We investigate challenges for federated partially-annotated segmentation (FPSS) systematically and propose a heterogeneous benchmark based on our solutions.
	\item We propose a Unified Federated Partially-annotated Segmentation (UFPS) framework for FPSS based on pseudo labeling technology for the first time, in which unified label learning (ULL) and sparse Unified Sharpness Aware Minimization (sUSAM) are designed to cope with class heterogeneity and client drift issues, respectively.
	\item The comprehensive experiments on the benchmark validate the effectiveness of our proposed method.
\end{itemize}

\section{RELATED WORK} \label{RELATED WORK}

\subsection{Partially Supervised Segmentation in Medical Domain}	\label{RELATED WORK Partially Supervised Segmentation in Medical Domain}
Many efforts have been made to conduct PSS in the medical domain. Deep Learning (DL) based approaches can be divided into three main branches, i.e., prior guided segmentation, index conditioned segmentation, and pseudo label based segmentation.

Priori guided segmentation methods like PaNN\cite{zhou2019prior} distills the volume ratio of target organs based on a fully labeled dataset, which may be hard to collect in real applications. PRIMP\cite{LIAN2023104339} proposes to use average masks for several groups as a priori, but it relies on manual preprocessing for start and end slices. These methods can be hard to implement in an FL scenario once such priori varies significantly between domains.

Cond-dec\cite{dmitriev2019learning} and DoDNet\cite{zhang2021dodnet} integrate organ indexes into the network. While the former encodes indexes to hash values and takes them as additional activations for each layer, the latter combines the bottleneck feature and an index vector to dynamically generate weights and bias for the segmentation head. Both of them require repetitive forward steps for all organs during the process of reference, which is time-consuming, especially in the medical domain.

CPS\cite{chen2021semi}, a pseudo label based segmentation method, proposes to use siamese networks supervising each other to correct potential noises in pseudo labels. Another way in this research field is MS-KD\cite{feng2021ms}, designing a framework that pretrains teacher models based on several datasets, each for one organ, to give pseudo labels. Features in all layers are distilled along with final logits in MS-KD to ease model training when the loss function is mere Kullback-Leibler (KL) loss\cite{hall1987kullback}. Other methods belong to none of these branches like PIPO\cite{fang2020multi} utilizing multi-scale inputs and features to capture details and global context. For all methods mentioned above, the domain gap issue is not taken into consideration which is common in the medical domain.

\subsection{Federated Learning}	\label{RELATED WORK Federated Learning}
One of the most serious challenges in FL is statistical heterogeneity of decentralized data. In order to surmount this barrier, numerous works are put forward. For instance, a regularization term between the global model and local models is proposed in FedProx\cite{li2020federated}. SCAFFOLD\cite{karimireddy2020scaffold} uses control variants to mitigate local gradient drift. These two methods are limited in highly non-IID scenarios. 

MOON\cite{li2021model} performs contrastive learning based on positive pairs between the local and global models, and on negative pairs between the current local model and the one in a previous round. FedCRLD\cite{10.1007/978-3-031-16443-9_25} reinforces the positive correlation in MOON and stability of local models via cross-attention and the local history distillation module, respectively. Nevertheless, they are costly both in computational time and memory.

Recently, several works have presented solutions based on high order information and managed to promote generalization ability of the global model to a great extent. FedAlign\cite{mendieta2022local} proposes to distillate Lipschitz constant between the original network block and a slimmed one. FedASAM\cite{10.1007/978-3-031-20050-2_38} combines Adaptive Sharpness Aware Minimization (ASAM)\cite{kwon2021asam} and Stochastic Weight Averaging (SWA)\cite{izmailov2018averaging} in FL, which is prolonged in training. 

Existing methods in FL mostly ameliorate the optimization process but seldom concentrate on data relevant techniques since data sharing among clients is prohibited. Direct data augmentation may even lead to model degradation in FL as proven in FedASAM.

\subsection{Federated Partially Supervised Segmentation}	\label{RELATED WORK Federated Partially Supervised Segmentation}
MENU-Net\cite{xu2022federated}, as the early work in FPSS, trains a model with multiple encoders and deep supervision layers via marginal and exclusive loss. However, some foreground channels for one client may be within a background one for other clients, which is also mentioned in \cite{10.1007/978-3-031-18523-6_6}. The global segmentation head for all organs with direct aggregation is doomed to incur class conflict problem, which is demonstrated in the \hyperref[empirical study]{empirical study} of our work. Besides, the global model with multiple encoders\cite{liang2020think} is inferior to the one with multiple decoders\cite{collins2021exploiting}. The same theory is verified in personalized federated learning (pFL)\cite{tan2022towards}. 

Naturally, a superior global model generalizes better than locally trained models to clients who may join FL in the future, which is common in the FL setting. Therefore, our main purpose is to train a global model instead of personalized ones for FPSS. Compared with MENU-Net, we represent comprehensive results based on originally fully labeled datasets in our experiment. To our best knowledge, it is also the first time that the global model is trained by clients with partially-annotated non-IID datasets but the benchmark performance for all organs in each client is reported. Our method only needs to forward the global model once for evaluation no matter how many organs are segmented and the global model in our approach generalizes better than local models on unseen clients.

\section{RESULTS}

\subsection{Empirical study}	\label{empirical study}
\begin{figure}
	\centering
		\includegraphics[width=\columnwidth]{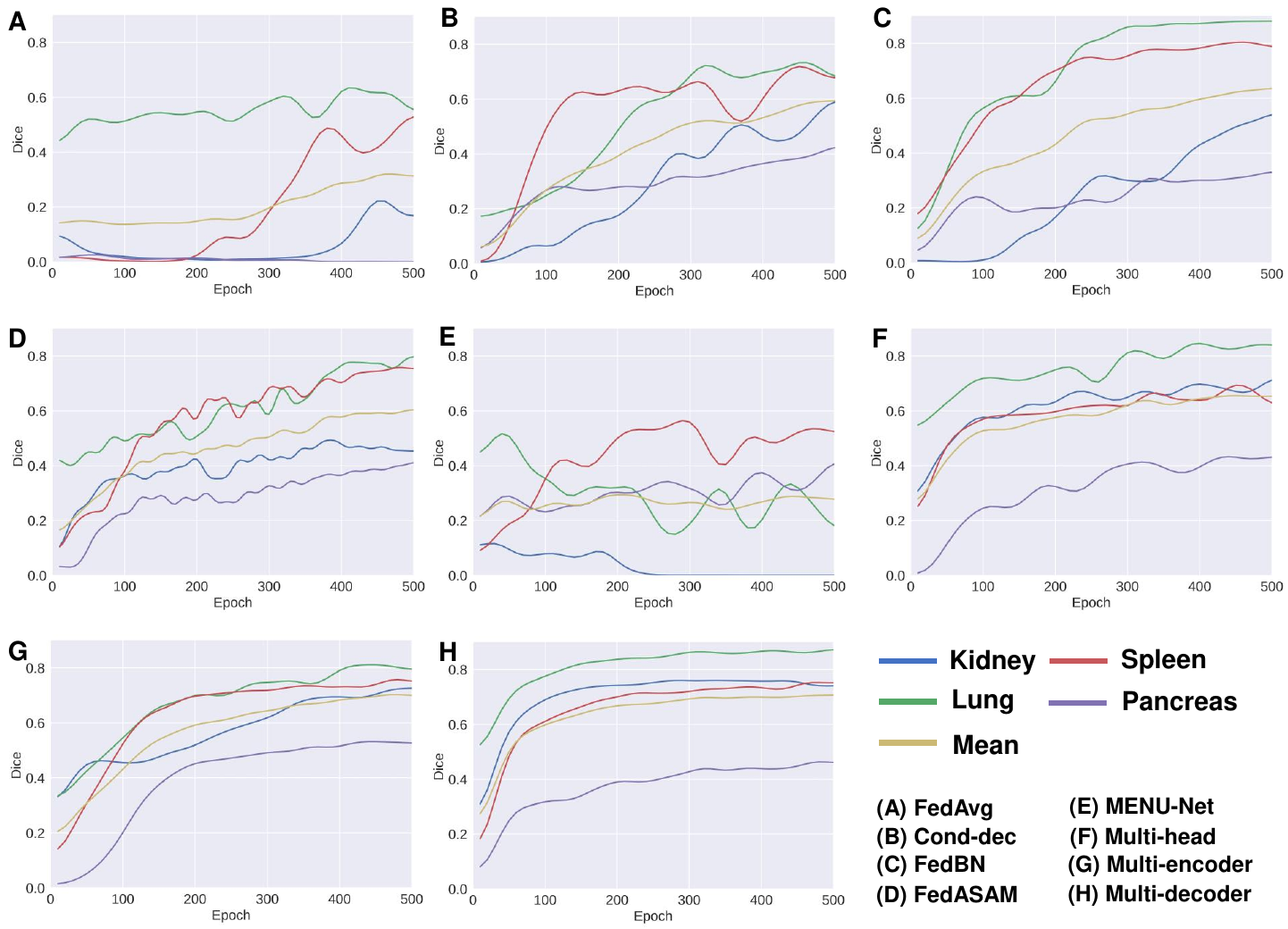}
	\caption{\textbf{Mean test dice curves over all clients for different combinations between PSS and FL.}}
	\label{Figure 2}
\end{figure}
Before formulating FPSS, we first study rationality of directly combining some methods for PSS with FL (\hyperref[Figure 1]{Figure 1}). Opt 1 in \hyperref[Figure 1]{Figure 1} corresponds to (E) in \hyperref[Figure 2]{Figure 2}, that is, MENU-Net. Opt 2 in \hyperref[Figure 1]{Figure 1} corresponds to (A), (B), (C), (D) in \hyperref[Figure 2]{Figure 2}. Implementation details can be found in \hyperref[SUPPLEMENTAL INFORMATION]{Note S6}.

To investigate the effect of solutions in centralized learning on the class heterogeneity issue, we start from the simplest case where locally trained segmentation networks are completely aggregated in each communication round. We observe that FedAvg (\hyperref[Figure 2]{Figure 2}A) fails to segment pancreas in CT images, which is the hardest class among all organs since its anatomical morphology is highly varied between domains. Besides, FedAvg suffers tremendous oscillation and low convergence speed compared with the rest. We put the blame on the fact that foreground channels without supervision can be optimized to any false direction. Even with client index input into each layer of the decoder (\hyperref[Figure 2]{Figure 2}B) as auxiliary information or with batch normalization layers separated for each client (\hyperref[Figure 2]{Figure 2}C) as feature decoupling to ease training, the severe oscillation can still not be alleviated. When the method designed for highly non-IID setting (\hyperref[Figure 2]{Figure 2}D) is used, successive fluctuation caused by class heterogeneity still exists. It proves that the class heterogeneity issue is a more critical challenge than the client drift in FPSS. 

When unannotated classes are merged into the background channel, MENU-Net (\hyperref[Figure 2]{Figure 2}E) suffers the most serious oscillation among all methods because foreground annotations for each client are mistaken as background ones in other clients when there exists no class intersections between clients. Thus, it can be concluded that both options in \hyperref[Figure 1]{Figure 1} are not suitable for FPSS since Opt 1 and Opt 2 cannot deal with class conflict for all classes and the background class, respectively.

As long as some certain part of the local model is separated from aggregation and the part is updated based on the loss for complete partial labels (e.g., one annotated class as foreground and its inverse set as background), the oscillation can be remarkably reduced. This phenomenon stresses the significance of dealing with class heterogeneity and client drift simultaneously in FPSS. Furthermore, personalizing decoder (\hyperref[Figure 2]{Figure 2}H) achieves better performance than personalizing encoder (\hyperref[Figure 2]{Figure 2}G). The reason for this phenomenon is that global representations are closely related to better generalization ability of models in FL, which has also been proven in previous work in personalized federated learning \cite{collins2021exploiting}. It is a priori that some classes are relevant in the medical field (e.g., relative organ position in CT image) and that layers around the bottleneck of the network usually extract high level information. Hence, personalizing such parts may hinder potential interactions between classes in the course of training. Despite the success of pFL based methods, they suffer from long inference time which is proportional to the number of classes.

Therefore, we aim to segment all classes without class conflict by filling up missing labels in a unified manner. Different from training personalized models, we also hope to train a generalized global model to forward once during testing, no matter how many target classes there are, by absorbing knowledge from all classes and all clients to learn organ interactions.

\subsection{Problem formulation}
In this subsection, we review objectives of PSS and FL and then define the formulation of FPSS based on \hyperref[empirical study]{empirical study}.

We first give a short review on PSS. Let $x \in X$ be the input and $y \in Y$ be its corresponding annotated label map. Suppose the entire dataset $D^p={\{D_i^p\}}_{i=1}^N$ can be divided into $N$ partially-annotated datasets where each subset $D_i^p$ includes $N_i$ data samples, i.e.,  $D_i^p=(X_i,Y_i^p)=\{{\{(x_{ij},y_{ij}^p)\}}_{j=1}^{N_i}\}^{N}_{i=1}$. We denote $C_i \subset C$ as the label set of $Y_i$, where $|C|$ is the total number of classes.  Here, we conclude some properties about classes in PSS.

\newtheorem{property}{Property}

\begin{property}	\label{Property 1}
The number of classes for the joint label space $Y$ is fixed to $N_c$: $ |\bigcup_{i=1}^N C_i |=N_c=|C| $.
\end{property}

\begin{property}	\label{Property 2}
The amount of partially-annotated classes for any subset $Y_i^p$ is usually limited:$ 0<|C_i\ |\ll N_c $.
\end{property}

\begin{property}	\label{Property 3}
The intersection of classes is restricted: $ \forall i,j\in[1,N], i\neq j, C_i\cap C_j=c_{i,j}$, where $0\leq |c_{i,j} |\ll N_c $.
\end{property}

Consider the FL setting with $N$ clients and the overall dataset $D$ with $N_c$ classes. Each client has the dataset $D_i$ separated from $D$.  Let $w_i, w_0$ denote the local model from client $i$ and the global model, respectively. In each round, all clients upload their trained local model to the server for aggregation, and the server distributes the global model to clients as the initial local model at the next round. The global objective is to minimize the average of local empirical risks:
\begin{equation}
$$
\mathop{min}\limits_{w_0\in W}f(w_0)=\frac{1}{N}\sum_{i=1}^N f_i (D_i,w_0 ),	 \tag{Equation 1}
$$
\label{Equation 1}
\end{equation}

\noindent where $f(\centerdot)$ is the loss function.

Now, we give the formulation of FPSS and list feasible solutions for it. Suppose each client has a partially-labeled dataset $D_i^p$ separated from $D^p$. The global objective for FPSS is almost same as the one for FL, i.e., \hyperref[Equation 1]{Equation 1}, but with more constraints (i.e., three properties listed in PSS setting):
\begin{equation}
\operatorname*{min}_{w_{0}\in W}f(w_{0})=\frac{1}{N}\sum_{i=1}^{N}f_{i}\bigl(D_{i}^{p},w_{0}\bigr),	 \tag{Equation 2}
\label{Equation 2}
\end{equation}

\noindent Without any preliminary step or part model aggregation, it is impossible to achieve global minimal in \hyperref[Equation 2]{Equation 2} based on partially-annotated datasets due to class heterogeneity problem. Thus, one simple but feasible way is to separate part of the whole model from aggregation:
\begin{equation}
\operatorname*{min}_{w_{0}^{G}\in W^{G},w_{i}^{L}\in W^{L}} f(w_{0}^G, w_{i}^{L})={\frac{1}{N}}\sum_{i=1}^{N}f_{i}\big (D_{i}^{p},w_{0}^{G},w_{i}^{L}\big).		 \tag{Equation 3}
\label{Equation 3}
\end{equation}
\noindent The whole model $w$ can be divided into $w_0^G,\{w_i^L\}_{i=1}^L$, denoting model parts aggregated globally and kept local, respectively.

\clearpage
\begin{figure*}
	\centering
	\begin{minipage}{\linewidth}
		\centering
		\subfloat{\includegraphics[width=\columnwidth]{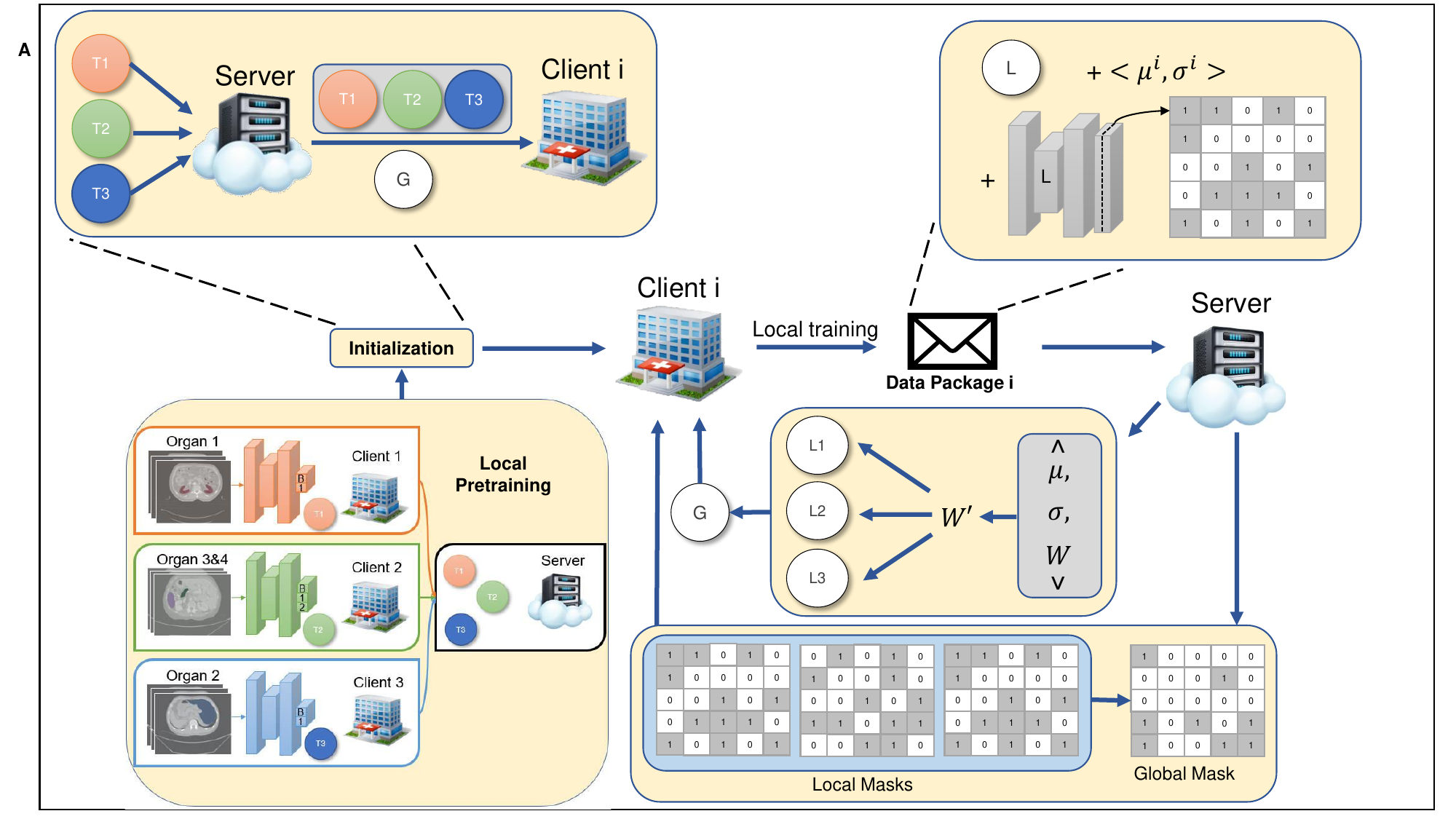}}
	\end{minipage}
	\begin{minipage}{\linewidth}
		\centering
		\subfloat{\includegraphics[width=\columnwidth]{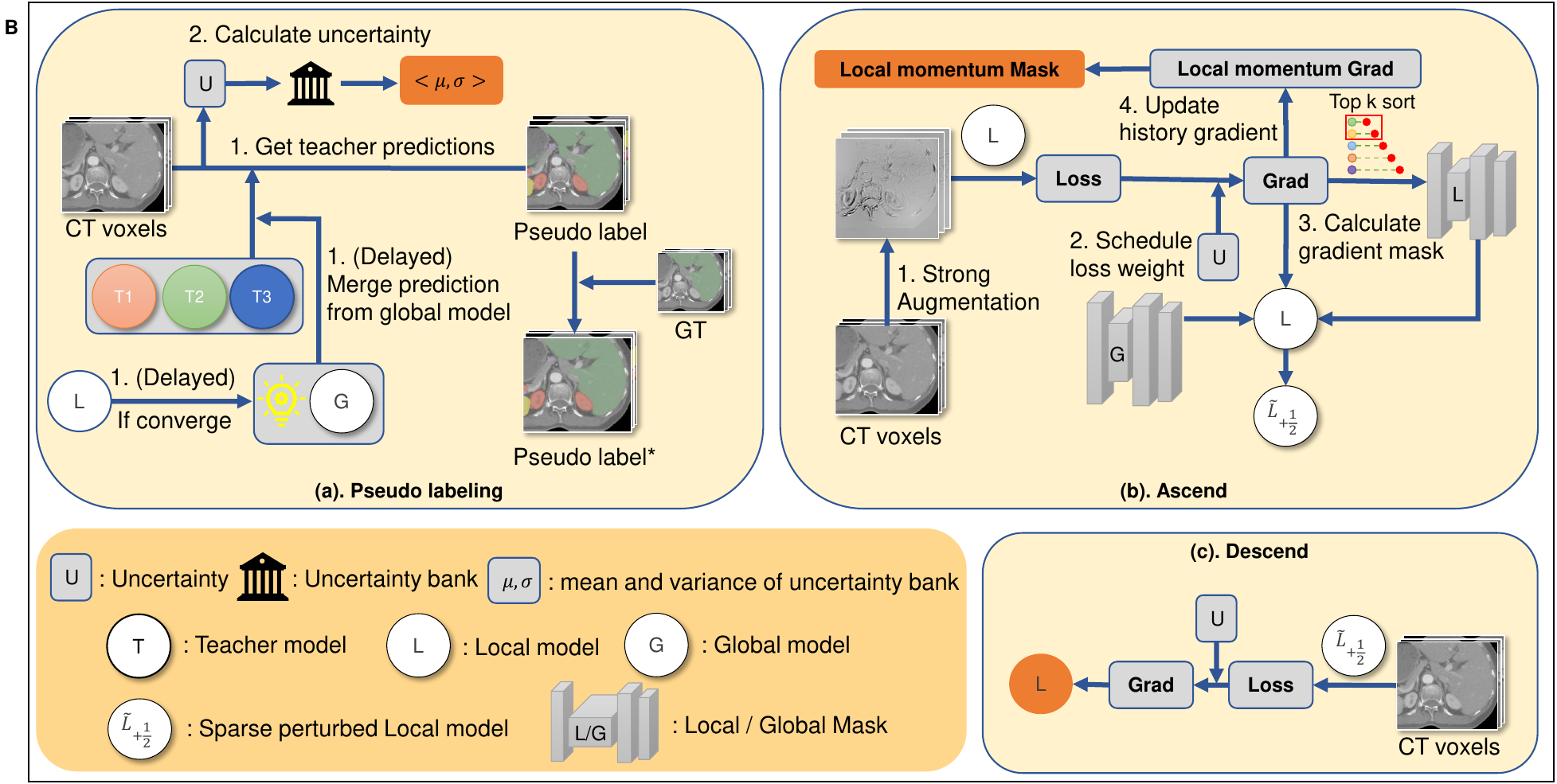}}
	\end{minipage}
	\caption{\textbf{The overall flow of our proposed UFPS framework.}\\ (A) Federated learning process. Operations in pseudo labeling are condensed in the 'initialization' box. The 'data package' refers to the local mask, the mean and variance of uncertainty bank for client $i$, and the local model. The low right part denotes taking the non-intersection part as the global mask. Aggregation weights are recomputed with statistics of the uncertainty bank and the original proportion weight. \\(B) Local training loop. Uncertainty score for each batch is deposited into the uncertainty bank and used to reweight loss. The local mask acquired in the ascent step is combined with part of the global mask to perturb the clean model. The local momentum mask is the mask sent to the server after local training. When local training converges, the global model replaces pretrained teachers as the main teacher.}
	\label{Figure 3}
	\label{fig3}
\end{figure*}
\clearpage

\subsection{Unified Label Learning} 	\label{METHOD Unified Label Learning}
In this subsection, we propose the pseudo labeling process for class heterogeneity problem and mechanisms for the noisy label learning issue from both global and local perspectives. The overall flow is depicted in \hyperref[Figure 3]{Figure 3}.

\textbf{Denoising pseudo label generation.} A better solution to the class heterogeneity issue should be free from class confliction and learn class interactions for better segmentation ability of the global model. In ULL, each client $i$ first pretrains a local model as a class-specific teacher for other clients with partially-annotated labels before federation. After sending the pretrained local teacher and receiving pretrained teachers $w^T$ from others, at each local round in FL, each client uses all pretrained teachers to get pseudo labels for all classes except ones with ground truth kept local. The background class for the pseudo label is the intersection of all teachers' background predictions, and foreground classes are merged in the predefined sequence. FL is then performed as:
\begin{equation}
\operatorname*{min}_{\begin{array}{l}{{w_{0}\in W}}\end{array}}f(w_{0})={\frac{1}{N}}\sum_{i=1}^{N}f_{i}\big(X_{i},Y_{i,gt},P L_{i}(X_{i},w^{T}),w_{0}\big),		 \tag{Equation 4}
\label{Equation 4}
\end{equation}
\noindent where $PL_i$ represents the operation to predict pseudo labels for $X_i$ by $w^T$. $Y_{i,gt}$ denotes the annotated ground-truth labels within foreground classes for each client $i$. However, even though the annotated class is replaced by the ground truth, noise in rest channels may still be severe as the process of pseudo labeling is hindered by domain gaps. For this reason, the class heterogeneity problem is naturally transformed into a noisy label learning issue.

As the direct source of pseudo labels, a noise-robust teacher model can play an important role in the noisy learning problem. Otherwise, local models for full-organ segmentation may overfit misguided information in noisy pseudo labels, thus stuck at a local minimum. In \cite{diao2022semifl}, the global model trained on a labeled public dataset serves as the teacher model to give pseudo labels as it is more reliable than local models. However, such a public dataset is not always available in the medical domain due to privacy concerns. Even though a fully annotated public dataset to enhance model reliability is not available in our setting, predictions from the global model may still become less noisy than ones from local teacher models at some point in time since the global model absorbs knowledge from multiple organs and global data distribution. Thus, we use the global model as the main teacher (global main teacher, GMT) in the training course at that time to better supervise local models. 

Although the ability of the global model to locate organs is promoted, its prediction for segmentation boundaries may be weakened. This phenomenon can be explained that the global model in FL is usually smoother than locally trained models since client drifts result in counteractions in some dimensions. Therefore, we use locally pretrained models as auxiliary teachers to refine boundary areas. Specifically, when the foreground prediction intersection of a patch between the global model and auxiliary teachers is greater than the volume percentage threshold $\mathscr{v}$, we use the intersection as the pseudo label. Otherwise, we only use forecast of the global model as convincing pseudo supervision:
\begin{gather*}
\tilde{q}=\left\{\begin{array}{l}
\tilde{q}^G,\left|\tilde{q}^{w^T}_{c\neq0} \cap \tilde{q}^G_{c\neq0}\right|<\mathscr{v} \cdot\left|\tilde{q}^G_{c\neq0}\right|, \\
\tilde{q}^{w^T} \cap \tilde{q}^G, \text { else },		 \tag{Equation 5}
\end{array}\right.
\label{Equation 5}
\end{gather*}
\noindent where $\tilde{q}^G$ denotes one-hot pseudo label from the global model and $|\cdot|$ here represents volume of the prediction.

\textbf{Enlarging impact of less noisy local models.} 
The denoising effect of GMT counts on a reliable global model, which is directly affected by the noise degree of local models. Considering that, we propose uncertainty-aware global aggregation (UA) to enhance reliability of the global model by enlarging aggregation weights of less noisy local models. 

It is a common occurrence in FL that aggregation weights $A^w$ only depend on the number of samples. However, clients with a large amount of data are not necessarily with high quality data. For FPSS, data quality can be reflected by confidence of pseudo labels. Using merged prediction from all teacher models, we first calculate data-wise uncertainty $U$ for each sample $j$:
\begin{equation}
\begin{aligned}
& U_j=\frac{1}{N_{c}}\sum_{c=0}^{N_{c}-1} \frac{\sum_{vox}E_{vox}\cdot\tilde{q}_{vox,c}^{w^{T}}}{\sum_{vox}\tilde{q}_{vox,c}^{w^{T}}+1},
\end{aligned}	 \tag{Equation 6}
\label{Equation 6}
\end{equation}
\noindent where $E_{vox}$ is average entropy across all classes and $vox$ denotes voxel. Uncertainty scores for each client are then deposited in their individual uncertainty bank.

Besides, since local teachers are pretrained at different sites and for different organs, using only uncertainty of pseudo labels may not correctly rectify the aggregation weights. Thus, we calculate both mean $\mu_i$ and variance $\sigma_i$ of the uncertainty bank for each site and combine them with the number of samples to decide the aggregation weight of each client $i$:
\begin{equation}
\widehat{A}^w_i=\frac{1}{3}\left(\frac{e^{-\frac{\mu_i}{\tau^{\mu}}}}{\sum_j^N e^{-\frac{\mu_j}{\tau^{\mu}}}}+\frac{e^{-\frac{\sigma_i}{\tau^{\sigma}}}}{\sum_j^N e^{-\frac{\sigma_j}{\tau^{\sigma}}}}+A^w_i\right),		 \tag{Equation 7}
\label{Equation 7}
\end{equation}
\noindent where ${\tau^{\mu}},{\tau^{\sigma}}$ are temperature hyper-parameters for mean and variance, respectively. By giving higher aggregation weights to these clients with reliable pseudo labels, the global model is less likely to be affected by these label noises. Direct assignment of $\widehat{A}^w_i$ can be improper because not all parts of the whole model are closely coupled with uncertainty. Consequently, we only apply this module to the decoder, which is most relevant to the final prediction. 

\textbf{Uncertainty-guided loss weight scheduler and noise robust loss.} 
Another key factor restricting reliability of the global model is the inadequate learning for hard classes, resulting in even more severe noise than one of head classes. Pseudo labels with low confidence are more likely to be noisy or hard to segment. To avoid underfitting hard classes and overfitting pure noise, we propose to use weight scheduler (WS) based on self-entropy for loss functions. The proposed scheduler, named tail shift (TS), is formulated as:
\begin{equation}
\begin{aligned}
&w(U_j)=\left\{\begin{array}{l}
2-e^{\operatorname{norm}(U_j)-\frac{r}{R}}, U_j>U_{\mathcal{T}}, \\
2-e^{\operatorname{norm}(U_j)}, \text { else },
\end{array}\right. \\
& \text{where} \quad \operatorname{norm}(U_j)=\frac{U_j-\mu}{U_{\max }-U_{\min }},
\end{aligned}		 \tag{Equation 8}
\label{Equation 8}
\end{equation}
\noindent where $U_{\mathcal{T}}$ corresponds to the uncertainty value at lowest $\mathcal{T}$ percentage. $\mu, U_{max}, U_{min}$ represent mean, maximal, minimal uncertainty in the uncertainty bank of the current client, respectively. $w(U_j)$ is then multiplied with the overall loss function to ensure enough fitting emphasis on hard classes. Other schedulers and their impact are introduced in \hyperref[SUPPLEMENTAL INFORMATION]{Note S8}.

Pseudo labels given by teacher models can be quite noisy under some circumstances (e.g., restricted amount of labeled data)\cite{zheng2021rectifying, 10.1007/978-3-030-87240-3_22}. As predictions from student models may become even more reliable than ones from teacher models during training, we use reverse cross entropy (RCE)\cite{wang2019symmetric} loss and reweight it based on current training epoch $r$ and total training epoch $R$ (adaptive RCE loss, aRCE):

\begin{equation}
	f_{aRCE}\ =\,e^{-20(1-\frac{r}{R})}\cdot (q(x) log\left(p(x)\right)),		 \tag{Equation 9}
\label{Equation 9}
\end{equation}

\noindent where $q(x)$ is model prediction and $p(x)$ is ground truth. 

\subsection{Sparse Unified Sharpness Aware Minimization} 	\label{METHOD Unified Sharpness Aware Minimization}

In this subsection, to alleviate the client drift problem, we introduce a unified ASAM (USAM) and its accelerating version, sparse USAM (sUSAM) , based on the ASAM framework.

\textbf{Optimizing towards global direction with strong data augmentation.} FedASAM has been proven as an effective method for the client drift problem. In the ascent step, the objective is to approximate the steepest optimization direction. By further optimizing from the sharpest direction for the original parameter in the descent step, the global model achieves flatter minima and smoother loss landscape at each iteration.

However, local models in FedASAM may overfit some local-specific attributes, thus failing to generalize on the non-IID global distribution. Therefore, we aim to find the steepest global direction in a unified manner while maintaining the modeling capacity of local datasets. Unlike previous methods dealing with data heterogeneity in FL, we alleviate the model drift issue by approximating the underlying global data distribution through data augmentation.  

From Theorem 1 in \hyperref[SUPPLEMENTAL INFORMATION]{Note S2}, it can be concluded that the gap between $D_{\text {global }}$ and $D_{\text {aug }}$ is mainly decided by constant $g$ and its increment, caused by excessive data augmentation. Local models thus incur larger error of the upper bound for generalization on $D_{\text {global }}$. \textit{Our key insights are that local models can be free from performance degradation when optimized on strongly augmented datasets indirectly and that the global model generalizes for unseen clients better when the local data distribution is extended to the global one through comprehensive augmentations in a privacy-preserving manner.}

Thus, we propose USAM to optimize local models towards the global direction. As our setting for FPSS concentrates on medical images, we apply CMIDG\cite{ouyang2022causality}, a causality-inspired data augmentation method designed for the single-source domain medical image segmentation (e.g., CT and MRI), to local datasets to imitate the underlying global data distribution. CMIDG integrates medical priors and simulates real-world data from various data centers by inflicting non-linear medical noises on data. Thus, we can naturally treat CMIDG as a strong and reasonable data augmentation method to explore global cliffy ways. When CMIDG is used for both steps in ASAM in our experiment, the global model is more unstable just as what is verified in Theorem 1. Unlike FedASAM inputs the same data for ascent and descend steps, we perform the ascent step of ASAM on the augmented data through CMIDG.

Since the time complexity of USAM is about two times of FedAvg in our experiment, we only use USAM when the global model converges in late communication rounds and find it works almost as well as used for more rounds.

\begin{figure}
	\centering
		\includegraphics[width=\columnwidth]{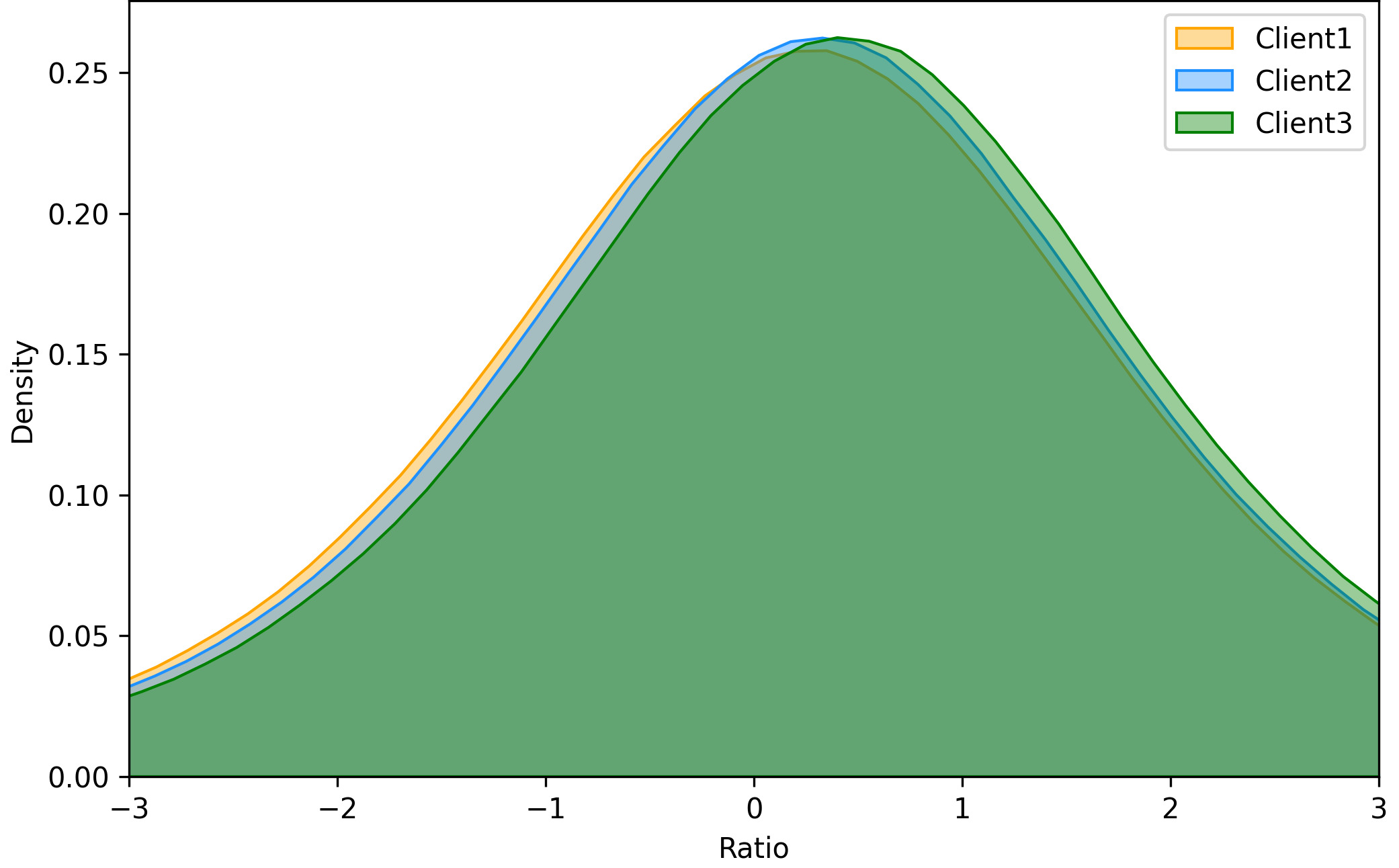}
	\caption{\textbf{Relative difference ratio of gradients between the pseudo label baseline and USAM.}}
	\label{Figure 4}
\end{figure}

\textbf{Accelerating and supplementing USAM wih gradient mask.} Although USAM has its potential to alleviate the effects of data heterogeneity, some sharp directions found in the ascent step may be relevant to attributes that only exist in a single local dataset. Besides, the time complexity of USAM is large even performed in limited rounds. We propose sUSAM focusing only on the most essential parts of perturbation to tackle both issues. Note that the principle of accelerating effect is discussed in the previous paper\cite{mi2022make}.

To illustrate whether all gradients deserve perturbation, we show the relative difference ratio of gradients between the pseudo label baseline and USAM:
\begin{equation}
r=\log \left|\frac{\nabla f_{USAM}-\nabla f_{\text {base }}}{\nabla f_{\text {base }}}\right|.		 \tag{Equation 10}
\label{Equation 10}
\end{equation}

\noindent As is demonstrated in \hyperref[Figure 4]{Figure 4}, about $60\%$ gradients are steep (ratio more than 0). Hence, we introduce a gradient mask $M_L$ to only reserve gradients with top $T_L\%$ absolute value in the ascent step for USAM. 

However, the sparse mask may erase some gradients accounting for vital global factors, which may be too hard to emphasize on for some clients but commonly stressed by others. To bridge this semantic gap, we propose to replace part of local masks with the nonintersecting global mask, which consists of three main steps, i.e., update of local masks and momentum gradients, communication for local and global masks, and mergence of local and global masks.

Local masks are always updated in the top-k manner. In the meantime, each client maintains local momentum gradients $G_{mo}$ on the ascent step, which is further used to calculate a momentum local mask $M_{L, mo}$. The local momentum gradients are updated at each iteration:
\begin{equation}
G_{m o}=\alpha_{m o} G_{m o}+\left(1-\alpha_{m o}\right) \nabla f,	 \tag{Equation 11}
\label{Equation 11}
\end{equation}

\noindent where $\alpha_{m o}$ is hyper-parameter empirically set to 0.9.

After finishing local training, each client sends $M_{L, mo}$ to the server to represent dominant positions of local features. When receiving momentum local masks from all clients, the server merges them as a global mask $M_G$ following the rule that nonintersecting parts of momentum local masks are set to 1 and the rest to 0:

\begin{equation}
\begin{aligned}
& \left(M_G\right)_0=\left(\sum_{i=1}^N M_{L, m o}\right)_0 \cup\left(\sum_{i=1}^N M_{L, m o}\right)_N, \\
& \left(M_G\right)_1=1-\left(M_G\right)_0,
\end{aligned}		 \tag{Equation 12}
\label{Equation 12}
\end{equation}

\noindent where $(\cdot)_i$ denotes positions with their values equal to $i$. The global mask ensures no redundant perturbation while exploring underlying global features in a unified manner. Compared with gradients of the float type, the global mask is of the bool type so the extra communication burden and privacy leakage can be almost negligible. 

For these gradients $G_N$ not in top $T_L\%$ of $M_L$ but in the nonintersecting part of $M_G$, we randomly choose part of them as content for the extra perturbation mask $M_E$ and the total length is:
\begin{equation}
\left|M_E\right|=\min \left(T_G|\nabla f|,\left|G_N\right|\right),		 \tag{Equation 13}
\label{Equation 13}
\end{equation}

\noindent where $T_G$ is hyper-parameter to decide the proportion of the extra mask. The final descent step at the $k$-th iteration based on sparse disturbance is formulated based on the mergence of masks:
\begin{equation}
w_{k+1} \leftarrow w_k-\left.\nabla_{w_k} f\left(D, w_k\right)\right|_{w_k+\hat{\epsilon}_k \cdot\left(M_L \cup M_E\right)}.	 \tag{Equation 14}
\label{Equation 14}
\end{equation}

To further save computational cost and stabilize training, the update for all masks is conducted every $r_{fre}$ rounds. Otherwise, $G_{mo}$ is not accumulated and a history local mask $M_L$ got in the last update is used. In our experiment, the average computational overhead of the local mask is only 5\% of the one for local global, which can be almost neglected. Next, we provide a summary convergence analysis for both full and part participating scenarios. Detailed assumption, proof, discussion are in \hyperref[SUPPLEMENTAL INFORMATION]{Note S3} and \hyperref[SUPPLEMENTAL INFORMATION]{Note S4}.

It can be concluded for Theorem 2 and Theorem 3 in \hyperref[SUPPLEMENTAL INFORMATION]{Note S4} that the sparse ratio for masks explicitly influences partial high order terms. Since the mask in sUSAM constrains sparse gradients, additional square and two-thirds terms are also negligible in magnitude. Besides, sUSAM is potential to generalize better by the dynamic mask, thus alleviating weight shifts in dominant terms for convergence.

\subsection{EXPERIMENTAL PROCEDURES}

\subsubsection{Resource availability}
\subsubsubsection{Lead contact}
Any further information, questions, or requests should be sent to Li Yan Ma (\href{liyanma@shu.edu.cn}{liyanma@shu.edu.cn}).

\subsubsubsection{Materials availability}
Our study did not generate any physical materials.

\subsubsubsection{Data and code availability}
This study uses previously published datasets. Our source code is available at GitHub (\href{https://github.com/tekap404/unified_federated_partially-labeled_segmentation}{\url{https://github.com/tekap404/unified_federated_partially-labeled_segmentation}}) and has been archived at Zenodo \cite{jiang4474880ufps}.

\subsubsection{Datasets}

\begin{table}
	\label{Table 1}
	\centering
	\rowcolors {3}{grey}{white}
	\renewcommand\arraystretch{1.2}
	\resizebox{\columnwidth}{!}{
		\begin{tabular}{lllll}
		\arrayrulecolor{red}
			\hline
			\multicolumn{5}{l}{\textcolor{red}{\textbf{Table 1. Statistics of Datasets}}}  \\
			\hline
			\textcolor{red}{Dataset}        & \textcolor{red}{WORD}   & \textcolor{red}{AbdomenCT-1K}       
			& \textcolor{red}{AMOS}  & \textcolor{red}{BTCV}  \\ \hline
			Total selected & 120    & 266                & 200   & 30    \\ 
			Partial Target & Kidney & Spleen \& Pancreas & Liver & All   \\ 
			in-FL/out-FL     & in-FL   & in-FL               & in-FL  & out-FL \\ 
			Client index   & 1      & 2                  & 3     & 4    \\
			\hline
		\end{tabular}
	}
\end{table}

Main information of datasets is listed in \hyperref[Table 1]{Table 1}. We conduct our experiments with four fully-annotated CT image datasets: WORD (\href{https://github.com/HiLab-git/WORD}{\url{https://github.com/HiLab-git/WORD}}), AbdomenCT-1K (\href{https://github.com/JunMa11/AbdomenCT-1K}{\url{https://github.com/JunMa11/AbdomenCT-1K}}), AMOS (\href{https://amos22.grand-challenge.org}{\url{https://amos22.grand-challenge.org}}) and BTCV (\href{https://www.synapse.org/\#!Synapse:syn3193805/wiki/217752}{\url{https://www.synapse.org/\#!Synapse:syn3193805/wiki/217752}}). Annotations for four organs are extracted from each dataset to serve as foreground classes, i.e., liver, kidney (left + right), spleen, pancreas. Whether a client is in the training process of FL is represented by 'in-FL' and 'out-FL'. Preprocessing detatails can be found in \hyperref[SUPPLEMENTAL INFORMATION]{Note S6}.

\subsection{Training} 
Only the partial target set and its inverse set (background) are used to pretrain partial teacher models. Dice and BCE losses are used as default loss functions. We train all methods for 500 communication rounds and 1 local round for each global one. We conduct 10 warmup rounds to increase the minimal learning rate and accumulate uncertainty values for loss weight scheduler. Unless specially remarked, post-processing is not used. Post-processing includes filling up holes and deleting small connected components. We only show dice and HD for the mean of each dataset. Please refer to \hyperref[SUPPLEMENTAL INFORMATION]{Note S6} for training details and complete results under more metrics.

\subsection{Main results}  
\subsubsection{Comparison with SOTAs}

\begin{figure}
	\centering
		\includegraphics[width=\columnwidth]{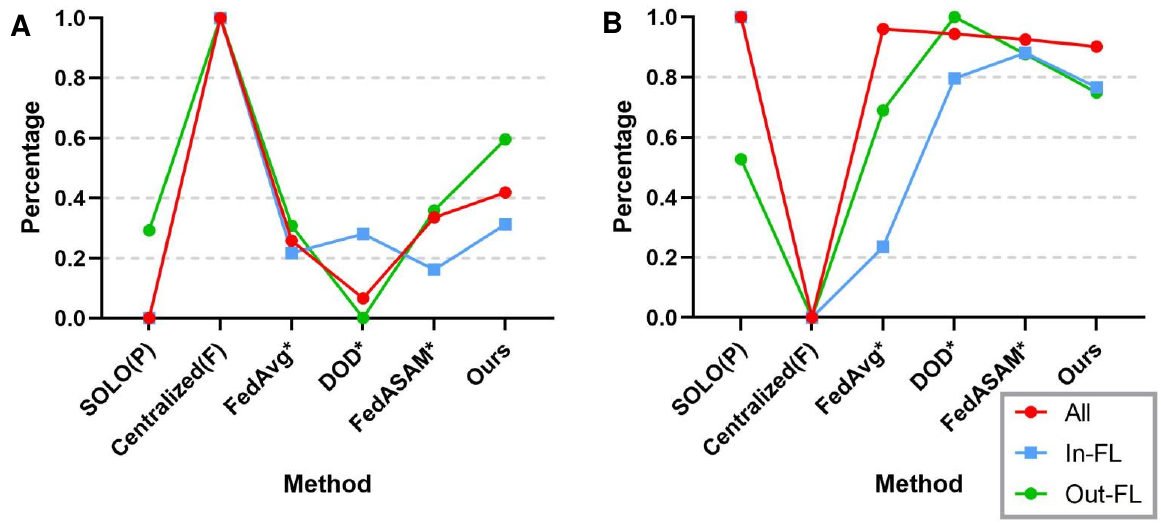}
	\caption{\textbf{Client-wise comparison between SOTAs after post-processing.}\\ (A) Normalized Dice ($\uparrow$).\\(B) Normalized HD ($\downarrow$).}
	\label{Figure 5}
\end{figure}

\begin{figure}
	\centering
		\includegraphics[width=\columnwidth]{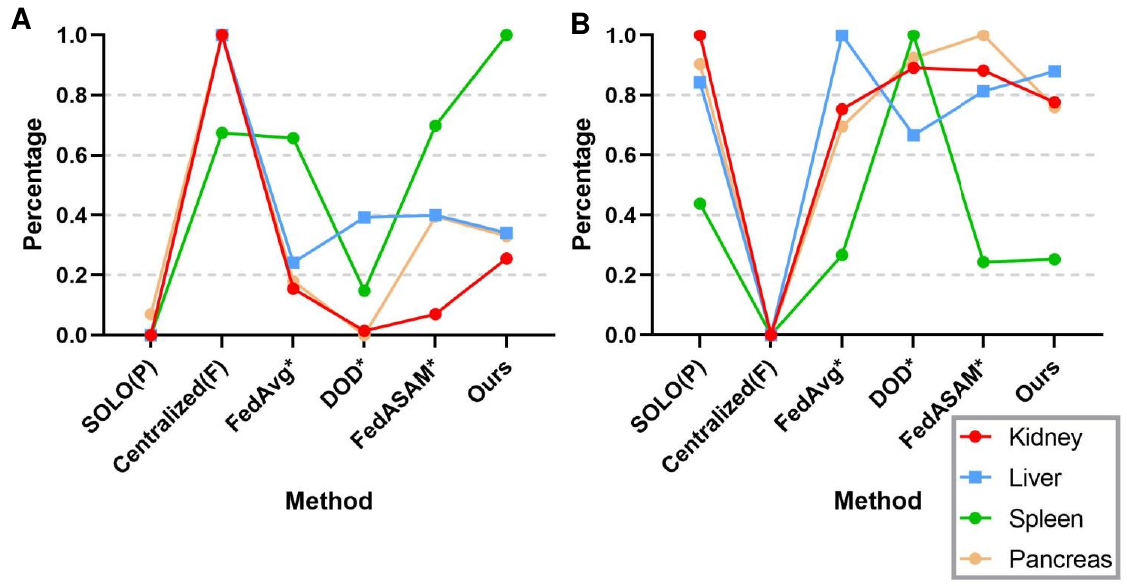}
	\caption{\textbf{Organ-wise comparison between SOTAs after post-processing.}\\ (A) Normalized Dice ($\uparrow$).\\(B) Normalized HD ($\downarrow$).}
	\label{Figure 6}
\end{figure}

To show the lower and upper bounds of benchmarks, we first conduct experiments under local training (SOLO) and centralized training (Centralized) based on partially-annotated datasets and fully annotated ones, respectively. We use models pretrained in SOLO as organ-specific teachers to give pseudo labels. To demonstrate the effectiveness of UFPS, we compare UFPS with a variety of SOTA methods. FedAvg* is a simple combination of FedAvg and our proposed pseudo labeling procedure. Note that DOD is originally a partially-annotated method based on part model aggregation for centralized learning and we modify it into the personalized federated learning setting. Besides, FedASAM is a method designed for heterogeneous data for fully annotated federated learning. Here we embed FedASAM into the pseudo labeling framework for partially-annotated segmentation as FedASAM*. Details for all methods can be found in \hyperref[SUPPLEMENTAL INFORMATION]{Note S6}. Experimental results on in-FL and out-FL datasets for some representative methods are shown in \hyperref[Table 2]{Table 2}, \hyperref[Figure 5]{Figure 5} and \hyperref[Figure 6]{Figure 6}.

The benchmark, i.e., SOLO (partial), is based on partially-annotated datasets within each client, thus incurring servere a domain gap for client 1 and client 3. For pFL based approaches, i.e., DOD* and FedCRLD, are essentially not compatible with FPSS since recognizing all classes largely depends on global universal features, as what is also demonstrated in \cite{jiang2022test} and that pFL models are likely to overfit to local biased distribution. Therefore, although DOD* is quite effective for client 2, owing to the largest aggregation weight based only on the amount of data, segmentation results for the rest are even worse than SOLO. The overall performance of FedCRLD is analogical to DOD* with only client 2 surviving the severe collapse. The over-fitting issue is exacerbated by the local momentum model. FedAvg*, FedASAM*, and UFPS (ours) are all based on the pseudo labeling framework, outperforming DOD*. Through unifying the class label space in FPSS, all these methods significantly benefite from class interactions, which proves the validity of using pseudo labels in FPSS for the first time. 

Among all methods originally designed for PSS, i.e., CPS*, MS-KD*, and DOD*, CPS* achieves the best performance (73.80 in Dice). By co-training, the noise degree of pseudo labels from teacher models is somehow alleviated in a local perspective, which justifies our basic idea that the class heterogeneity problem can be translated to a noisy label learning issue. 

Although the overall performance of MOON* is similar to FedAvg*, we notice that dice for client 4 is enhanced by 0.51 but in-FL results are not satisfying. We can conclude for the form of contrastive loss in \cite{li2021model} that forcing the local model to align with the global one and to keep away from its history version is potential to learn generalizable features for unseen domains, but feature extraction ability for native information may recede. Different from these contrastive learning based methods (FedCRLD and MOON*), UFPS not only justifies global distribution for all clients but it also takes some features which may be too hard to emphasize on for some clients but commonly stressed by others into account, leading to improvement for all clients compared with FedAvg*.

The tendency and principle of FedProx* are analogous to MOON* but with better out-FL performance and worse in-FL one. FedAlign refers to second-order calculation likewise. However, there is a large gap between it and sUSAM used alone, manifesting the significance of global distribution alignment under the highly non-iid setting. Since UFPS is universal for model type and regulation methods, appropriate combinations are probably beneficial.

As can be seen in \hyperref[Figure 5]{Figure 5}, thanks to ULL and sUSAM to denoise pseudo labels and to optimize towards the global direction, our method outperforms other methods (except upper bound Centralized Full) for both in-FL clients and out-FL clients. Specifically, our method increases by 5.35 for the baseline and 1.08 for FedASAM* in dice. Through simple post-processing, based on the accurate segmentation location and the intersection of teachers in GMT, HD of our method can also be reduced to a satisfying result, indicating its segmentation border is more refined than others. From \hyperref[Figure 6]{Figure 6}, it can be seen that our method even surpasses Centralized Full on the class 'Spleen', which proves the strong generalization ability of our method and its potential to save labor for labeling full annotations. Besides, our approach also gains a large margin for the class 'Kidney' compared with methods except for the upper bound and achieves approximate performance for other organs.

\begin{table*}[]
\label{Table 2}
	\centering
	\rowcolors {3}{grey}{white}
	\renewcommand\arraystretch{1.2}
	\resizebox{2\columnwidth}{!}{
		\begin{tabular}{lllllll}
			\arrayrulecolor{red}
			\hline
			\multicolumn{7}{l}{\textcolor{red}{\textbf{Table 2. Comparison with SOTAs. }}}  \\
			\hline
			\textcolor{red}{Method}   & \textcolor{red}{Client 1}     & \textcolor{red}{Client 2}    
			& \textcolor{red}{Client 3}     & \textcolor{red}{Client 4}     & \textcolor{red}{Mean}
			& \textcolor{red}{Post}           \\ \hline
			SOLO (partial, lower bound)     & 69.17 / 1.02   & 75.75 / 3.02   & 60.59 / 1.61   & 74.70 / 1.52   & 70.05 / 1.79   & 69.93 /   1.66 \\
			Centralized (full, upper bound) & 78.76 / 1.10   & 88.33 / 1.72   & 79.18 / 1.33   & 80.78 / 1.41   & 81.76 / 1.39   & 82.25 /   1.23 \\
			FedCRLD	& 67.82 / 2.21 & 77.77 / 3.06 & 58.11 / 2.06 & 72.30 / 1.84 & 69.00 / 2.29 & 68.14 / 1.76 \\
			DOD*     & 61.95 / 1.15 & 81.76 / 2.59 & 60.57 / 1.69 & 53.52 / 1.14 & 70.62 / 1.72 & 70.77 /   1.63 \\
			CPS*     & 75.78 / 0.99 & 78.05 / 2.91 & 65.75 / 1.70 & 75.60 / 1.65 & 73.80 / 1.81 & 73.78 / 1.62 \\
			MS-KD*	 & 74.35 / 1.00 & 76.68 / 2.96 & 63.33 / 1.66 & 73.10 / 1.64 & 71.86 / 1.81 & 71.77 / 1.82 \\
			FedAvg*                        & 74.94 /   1.29 & 78.10 /   2.83 & 64.53 /   1.83 & 74.74 /   1.69 & 73.07 /   1.91 & 73.13 /   1.64 \\
			FedProx*	& 74.41 / 1.39 & 77.33 / 2.88 & 64.56 / 1.75 & 75.60 / 1.64 & 72.97 / 1.92 & 73.06 / 1.64 \\
			MOON*		& 75.00 / 1.22 & 77.89 / 2.88 & 64.10 / 1.83 & 75.25 / 1.69 & 73.06 / 1.90 & 73.12 / 1.61 \\
			FedAlign*	& 75.18 / 1.20 & 77.07 / 2.89 & 64.31 / 1.77 & 76.22 / 1.62 & 73.20 / 1.87 & 73.28 / 1.61 \\
			FedASAM* & 77.02 / 1.39 & 78.15 / 2.91 & 65.60 / 1.75 & 75.14 / 1.71 & 73.97 / 1.94 & 74.20 /   1.63 \\
			UFPS (ours)                     & 76.22 /   1.45 & 79.56 /   2.82 & 66.82 /   2.04 & 77.22 /   1.72 & 74.95 /   2.01 & 75.28 /   1.62 \\
			\hline
			\rowcolor{white}\multicolumn{7}{l}{'Post' represents mean results after post-processing.} \\
			\rowcolor{white}\multicolumn{7}{l}{Here we only show Dice / HD (higher / lower numbers are better) for the mean of each dataset.} \\
			\rowcolor{white}\multicolumn{7}{l}{All methods marked asterisk are not FPSS methods originally and modified to fit the FPSS setting. DOD* is combined}  \\
			\rowcolor{white}\multicolumn{7}{l}{with the multi-decoder setting in \hyperref[empirical study]{empirical study}. Others with asterisk are combined with the pseudo labeling procedure.}  \\
			\rowcolor{white}\multicolumn{7}{l}{Please refer to \hyperref[SUPPLEMENTAL INFORMATION]{Note S6} for modification details and \hyperref[SUPPLEMENTAL INFORMATION]{Note S7} for complete results.}  \\
			\hline
		\end{tabular}
	}
\end{table*}

\subsubsection{Ablation study}
In this subsection, we prove validity of each module proposed in our paper and provide main ablation studies for all of them.

\textbf{Module validity.} \hyperref[Table 3]{Table 3} shows module validity for UFPS. We can conclude that the order of module importance is GMT > WS > sUSAM > UA > aRCE. Since local teachers pretrained at client 1 and client 3 are not generalized enough for the global distribution, which is dominant by the dataset from client 2, predictions from local models trained in the FL process can be gradually less noisy than these pseudo labels. Therefore, using aRCE as an extra loss gets a reasonable promotion. When WS is employed to force these models to concentrate on hard classes, all classes can be fitted simultaneously with similar emphasis. What is more important, class interactions are fully explored by local models in this situation, resulting in enhancement for all clients by a large margin. UA is basically designed for these clients with high data quality and with head class annotations. By correctly rectifying the aggregation weight, the model performance gains. When all classes are denoised through previous modules, GMT is able to use the global model to give more reliable pseudo labels than locally pretrained teachers. Due to another favorable factor that the global model is indirectly trained on global distribution, the overall performance becomes even better. Benefitting from optimization towards the global steepest direction at each site and guidance for latent global directions from other clients, ULL additional with sUSAM is profitable for most clients with few extra computational costs.

As for the relationship between modules, WS, as an essential part to ensure the training quality of the early training phase, mainly interacts with UA since a lot of uncertainty values are accumulated at this stage to model a reliable uncertainty distribution for each client. It also ensures that the noise degree of pseudo labels is not too large to affect basic parts in other modules implicitly. When GMT is invoked, it has mutual effects with sUSAM and UA considering the fact that sUSAM performs global alignment for the global model and UA rectifies the aggregation weights to guarantee the global model is dominated by local models trained with high quality data. Inversely, GMT offers more accurate pseudo labels for the two modules. Through analogy, aRCE relieves the influence of noisy labels as well, thus forming a virtuous cycle with GMT.

\begin{table*}[]
\label{Table 3}
	\centering
	\rowcolors {3}{grey}{white}
	\renewcommand\arraystretch{1.2}
	\resizebox{2\columnwidth}{!}{
		\begin{tabular}{p{0.7cm}<{\centering}p{0.7cm}<{\centering}p{0.7cm}<{\centering}p{0.7cm}<{\centering}p{0.7cm}<{\centering}p{0.9cm}<{\centering}lllll}
			\arrayrulecolor{red}
			\hline
			\multicolumn{11}{l}{\textcolor{red}{\textbf{Table 3. Ablation study on module validity.}}}  \\
			\hline
			\textcolor{red}{PL} &\textcolor{red}{aRCE} &\textcolor{red}{WS} &\textcolor{red}{UA} &\textcolor{red}{GMT} &\textcolor{red}{sUSAM} &\textcolor{red}{Client 1} &\textcolor{red}{Client 2}    
			&\textcolor{red}{Client 3}     &\textcolor{red}{Client 4}     &\textcolor{red}{Mean} \\ \hline
			 & & & & &                              & 69.17 / 1.02 & 75.75 / 3.02 & 60.59/ 1.61  & 74.70 / 1.52 & 70.05 / 1.79 \\
			\checkmark & & & & &                          & 74.94 / 1.29 & 78.10 / 2.83 & 64.53 / 1.83 & 74.74 / 1.69 & 73.07 / 1.91 \\
			\checkmark & \checkmark & & & &                     & 75.53 / 1.23 & 77.71 / 2.87 & 65.37 / 1.78 & 74.75 / 1.71 & 73.34 / 1.89 \\
			\checkmark &  & \checkmark & & &                     & 75.98 / 1.58 & 78.24 / 2.86 & 66.33 / 1.74 & 76.35 / 1.65 & 74.22 / 1.96 \\
			\checkmark &  & & \checkmark & &                     & 76.65 / 1.39 & 78.05 / 2.92 & 64.81 / 1.89 & 75.19 / 1.72 & 73.67 / 1.98 \\
			\checkmark &  & & & \checkmark &                     & 77.77 / 1.18 & 77.58 / 2.85 & 65.64 / 1.96 & 76.09 / 1.73 & 74.27 / 1.93 \\
			\checkmark &  & & & & \checkmark                     & 77.06 / 1.38 & 78.23 / 2.89 & 65.53 / 1.89 & 75.34 / 1.74 & 74.04 / 1.98 \\
			\checkmark & \checkmark & \checkmark & & &                 & 75.99 / 1.44 & 78.22 / 2.86 & 66.89 / 1.92 & 76.48 / 1.70 & 74.39 / 1.98 \\
			\checkmark & \checkmark & \checkmark & \checkmark & &             & 76.93 / 1.37 & 78.08 / 2.89 & 66.56 / 1.95 & 76.22 / 1.62 & 74.44 / 1.98 \\
			\checkmark & \checkmark & \checkmark & \checkmark & \checkmark &        & 76.12 / 1.51 & 78.83 / 2.86 & 67.30 / 1.95 & 77.07 / 1.70 & 74.83 / 2.00 \\
			\checkmark & \checkmark & \checkmark & \checkmark & \checkmark & \checkmark & 76.22 / 1.45 & 79.56 / 2.82 & 66.82 / 2.04 & 77.22 / 1.72 & 74.95 / 2.01  \\ 
			\hline
			\rowcolor{white}\multicolumn{11}{l}{Pseudo label, adaptive RCE loss, weight scheduler, uncertainty-aware global aggregation, global main teacher, sparse Unified}\\
			\rowcolor{white}\multicolumn{11}{l}{Sharpness Aware Minimization are denoted as PL, aRCE, WS, UA, GMT and sUSAM, respectively. }\\
			\rowcolor{white}\multicolumn{11}{l}{Here we only show dice / HD (higher / lower numbers are better) for the mean of each dataset. }\\
			\rowcolor{white}\multicolumn{11}{l}{Please refer to \hyperref[SUPPLEMENTAL INFORMATION]{Note S8} for more results.}\\
			\hline
		\end{tabular}
	}
\end{table*}

\begin{figure*}
	\centering
		\includegraphics[width=2\columnwidth]{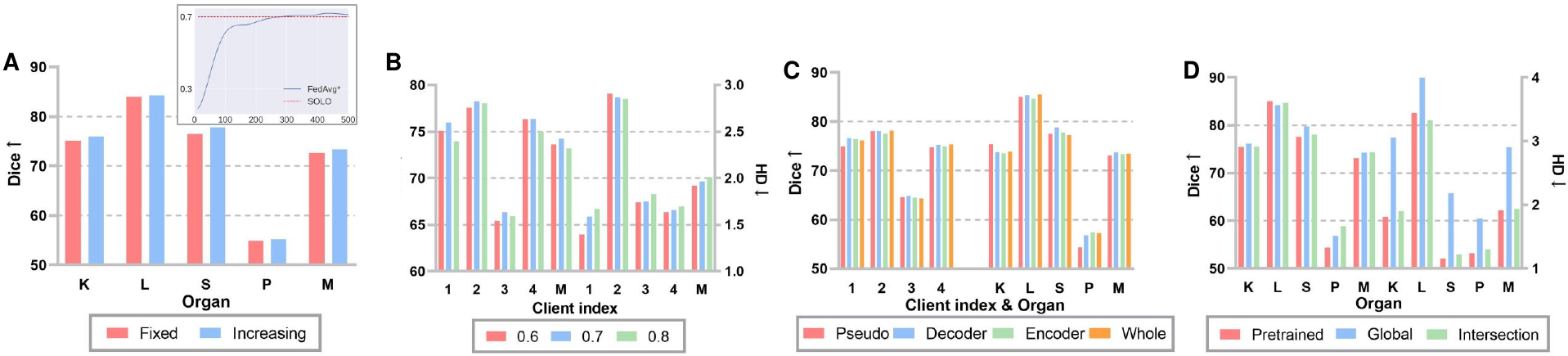}
	\caption{\textbf{Ablation for ULL.}\\K, L, S, P, M represent kidney, liver, spleen, pancreas, mean, respectively.\\(A) Organ-wise dice comparison for strategies of aRCE. The subfigure is a training dice curve for FedAvg*.\\(B) Client-wise dice and HD comparison for the uncertainty threshold in WS.\\(C) Client-wise and organ-wise dice comparison for module position of UA.\\(D) Organ-wise dice and HD comparison for strategies of GMT.}
	\label{Figure 7}
\end{figure*}

\textbf{Ablation for ULL.} Main ablations for ULL are all shown in \hyperref[Figure 7]{Figure 7}. In the subfigure of \hyperref[Figure 7]{Figure 7}A, it can be observed that the model performance of FedAvg* gets higher along with the training process and surpasses pretrained teachers, i.e., SOLO, at the 300-th epoch. Thus, it lays foundation to use RCE loss since predictions are better than pseudo labels due to organ interactions. The result in the bigger picture further proves our assumption that enlarging the coefficient for RCE loss, i.e., aRCE loss, is better than a fixed one for the increasing reliability of local models in the FL stage. 

For the threshold of shifting in \hyperref[Figure 7]{Figure 7}B, when it is set to a moderate value, the local model is neither greatly affected by label noises nor easily neglects tail classes, thus enhancing performance for all clients. For our future work, we intend to solve this problem by determining this hyper-parameter adaptively.

As demonstrated in \hyperref[Figure 7]{Figure 7}C, client 1 in this experimental setting only has the kidney class annotated before FL, whose aggregation weight is increased at most among all clients through UA. But the promotion is not from the class 'Kidney', which proves that the specific class(es) that a client has ground truth annotation is not necessarily related with the overall uncertainty of pseudo labels closely. What actually plays a key role is global model aggregation based on mean and variance of uncertainty. It rectifies model aggregation weights to those clients whose data is of high quality but of less amount. This operation sacrifices little or no fitting ability for others compared with the pseudo labeling baseline. Furthermore, conducting this module merely on the decoder is slightly better than on the whole model but much better than on the encoder plus deep supervision. The reason for this phenomenon is probably due to the closer relationship between prediction uncertainty and decoder. 

It is obvious from \hyperref[Figure 7]{Figure 7}D that when the global model is taken as the main teacher after a certain point, i.e., the 300-th epoch in our experiment, owing to its better generalization ability than that of pretrained teacher models, it achieves a huge performance gain on the client who is of worse performance. Furthermore, when we take intersection between global and pretrained teacher models, regions with high confidence are taken as our final prediction, so the ambiguity of border can be significantly alleviated, which is proven by HD.

\begin{figure*}
	\centering
		\includegraphics[width=2\columnwidth]{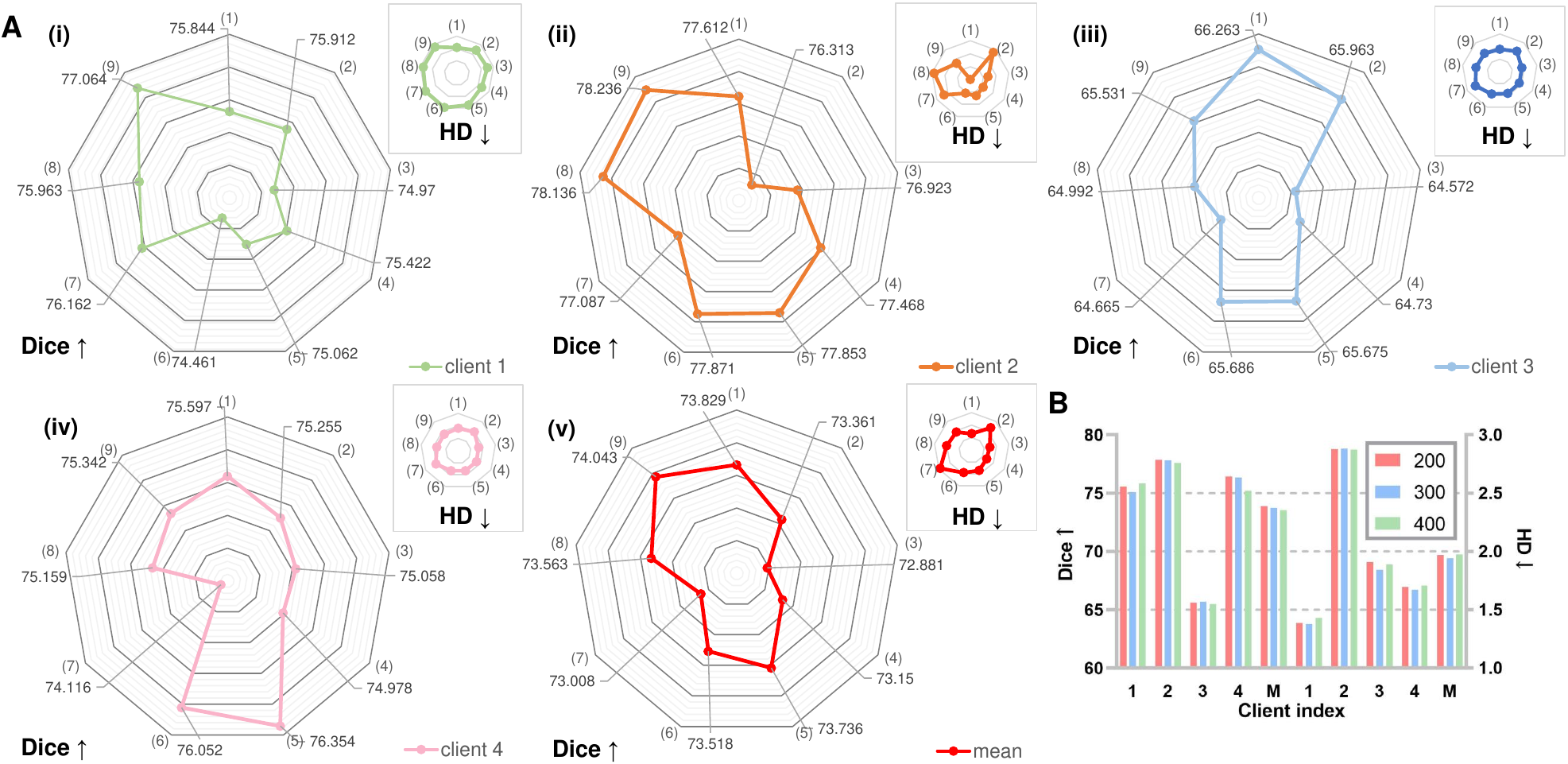}
	\caption{\textbf{Strategy ablation for sUSAM.}\\(A) The larger radar map corresponds to dice and the smaller one corresponds to HD.\\(1) CMIDG from beginning. (2) CMIDG from the 300th epoch. (3) Original data + ASAM. (4) Random perturbation + ASAM. (5) USAM. (6) USAM + (0.8 * original weight + 0.2 * perturbed weight for descent step). (7) CMIDG for both ascent and descent steps. (8) USAM + top k perturbation. (9) sUSAM.\\(B) The start epoch of USAM.}
	\label{Figure 8}
\end{figure*}

\textbf{Ablation for sUSAM.} We first display \hyperref[Figure 8]{Figure 8}A to comprehensively validate our motivations for sUSAM. The data augmentation method that we use, i.e., CMIDG, whose distribution density is in direct proportion to the number of epochs. The comparison between (1) and (2) indicates its sensitivity to training rounds, thus resulting in heavy computational costs since the augmentation is generated from a network. Since the local sharpest direction is not necessarily the global steepest direction, simply performing the ascent step of ASAM on original data in (3) intensifies client drifts, and it is even worse than random perturbation in (4). USAM, i.e., (5), with comparison to (2) and (7), demonstrates our insight that local models can be free from performance degradation when optimized on strongly augmented datasets indirectly. That is, conducting CMIDG in the ascent step and descending on the original data benefit models from abundant data distribution while keeping their fitting ability on local distribution even under excessive augmentation. What should be paid most attention to is that the out-FL client BTCV achieves the best result in this setting, which shows the great potential of USAM to generalize better on unseen data distribution. If original parameters and perturbed ones are combined to calculate loss, i.e., (6), model performance slightly drops, proving necessity of the perturbation in the ascent step. 

For partial gradient perturbation (8), it degrades the model performance for a little degree just as results in SSAM\cite {mi2022make}. Through the non-intersection global mask in sUSAM, underlying steep directions for global distribution are fully explored. Consequently, model performance for in-FL clients in (9) is greatly increased. 

When modifying the start epoch of USAM from 300 to 200 in \hyperref[Figure 8]{Figure 8}B, we find the gain from data density is limited, so we choose 300 as the start epoch to balance training speed and accuracy. Besides, benefiting from most essential perturbation directions in a global perspective, our method mere with sUSAM for 200 epochs surpasses the SOTA method FedASAM with 500 epochs for ASAM.

\begin{figure*}
	\centering
		\includegraphics[width=2\columnwidth]{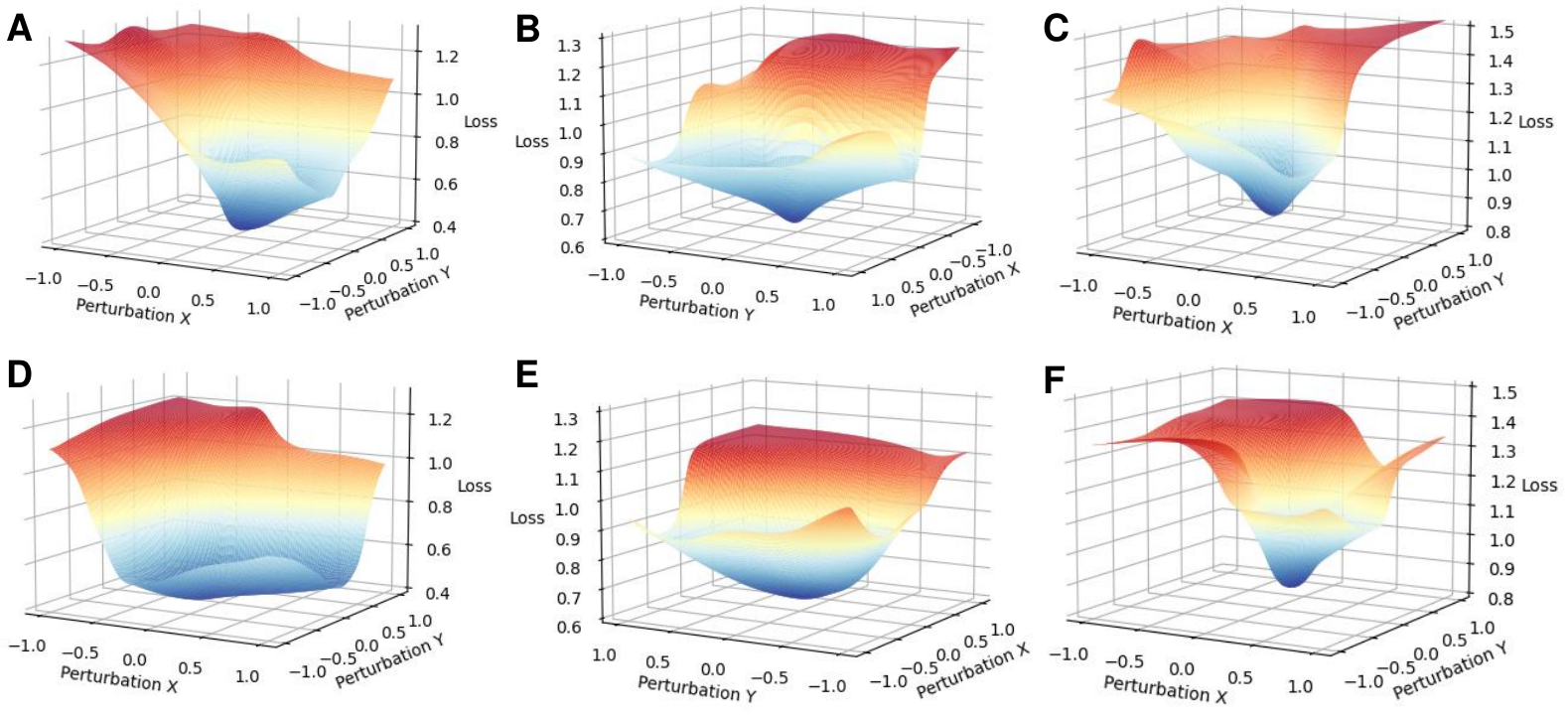}
	\caption{\textbf{Loss landscape on the training set.} \\ Model generalization is better when the overall loss landscape is flatter.\\(A, B and C) Loss landscapes from FedAvg* for client 1, client 2 and client 3, respectively.\\ (D, E and F) Loss landscapes from UFPS (ours) for client 1, client 2 and client 3, respectively.}
	\label{Figure 9}
\end{figure*}

\begin{figure*}
	\centering
		\includegraphics[width=2\columnwidth]{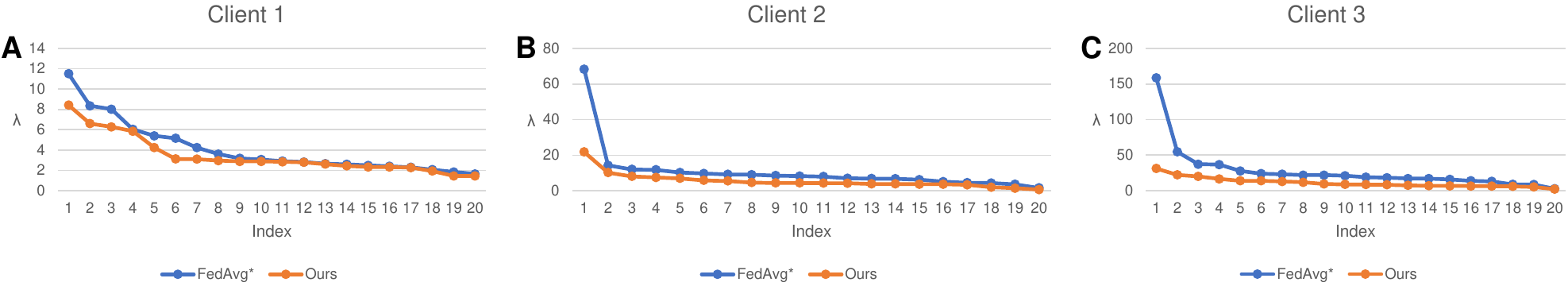}
	\caption{\textbf{Hessian eigenspectra of the global model.} \\(A, B and C) Statistics of Hessian for client 1, client 2, client 3, respectively. }
	\label{Figure 10}
\end{figure*}

To prove the generalization ability of sUSAM, we first plot loss landscape on the training set (\hyperref[Figure 9]{Figure 9}). Compared with FedAvg* achieving sharp minima for all clients, UFPS achieves lower loss for all clients. For client 1, the landscape under loss value 0.8 is overall flatter benefiting from the corrected model aggregation weights and gradient mask for the global descending direction. All statistics extracted from the Hessian for the global model (\hyperref[Table 4]{Table 4} and \hyperref[Figure 10]{Figure 10}) all demonstrate the generalization ability can be improved by seeking to flatter minima explicitly in a heterogeneous setting.

\begin{table*}[]
\label{Table 4}
\centering
\rowcolors {3}{grey}{white}
	\renewcommand\arraystretch{1.2}
	\resizebox{2\columnwidth}{!}{
		\begin{tabular}{lllllll}
		\arrayrulecolor{red}
		\hline
		\multicolumn{7}{l}{\textcolor{red}{\textbf{Table 4. Statistics related to model generalization.}}}  \\
		\hline
		\textcolor{red}{Client}   & \textcolor{red}{$\lambda_{max}$(pseudo)}     & \textcolor{red}{$\lambda_{max}$(ours)}    
		& \textcolor{red}{$\lambda_{max} / \lambda_{5}$(pseudo)}     & \textcolor{red}{$\lambda_{max} / \lambda_{5}$(ours)}     & \textcolor{red}{trace(pseudo)} 
		& \textcolor{red}{trace(ours)}\\ \hline
		1      & 11.512      & 8.422     & 2.125       & 1.956     & 170.1         & 60.8        \\
		2      & 68.286      & 21.84     & 6.643       & 3.132     & 393.7         & 269.6        \\
		3      & 158.768	 & 31.221    & 5.723       & 2.245     & 165.6         & 57.2       \\
		\hline
		\rowcolor{white}\multicolumn{7}{l}{Model generalization is better when all of these statics are lower. }\\
		\rowcolor{white}\multicolumn{7}{l}{$\lambda_{max}$ and $\lambda_{5}$ mean the top eigenvalue and the 5th top eigenvalue of the Hessian for the global model, respectively.}\\
		\hline
		\end{tabular}
	}
\end{table*}

\subsubsection{Visual Evaluation}	\label{EXPERIMENTS Main results Visual Evaluation}

\begin{figure*}
	\centering
		\includegraphics[width=2\columnwidth]{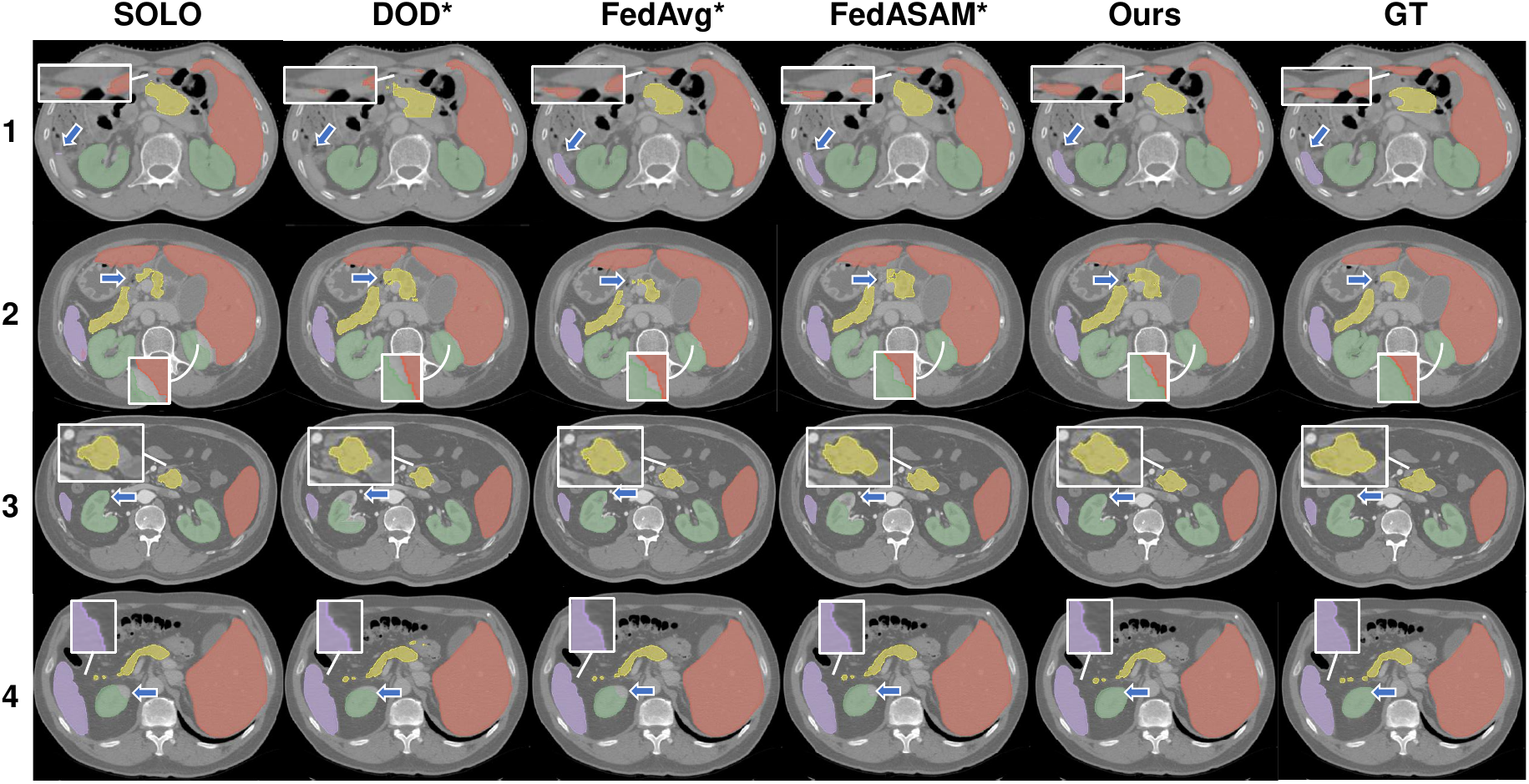}
	\caption{\textbf{2D segmentation result on the test set.}\\Numbers on the left side of images refer to client index. Green, red, purple, yellow regions represent kidney, liver, spleen and pancreas, respectively.}
	\label{Figure 11}
\end{figure*}

As the global model in our method is trained from multiple sites and organs, it is effective to reduce false negatives compared with other methods, e.g., spleen in client 1, kidney in client 2 and 3 in \hyperref[Figure 11]{Figure 11}. Besides, the overall contour predicted by our global model is obviously smoother, especially for pancreas and junctions between organs.

\begin{figure*}
	\centering
		\includegraphics[width=2\columnwidth]{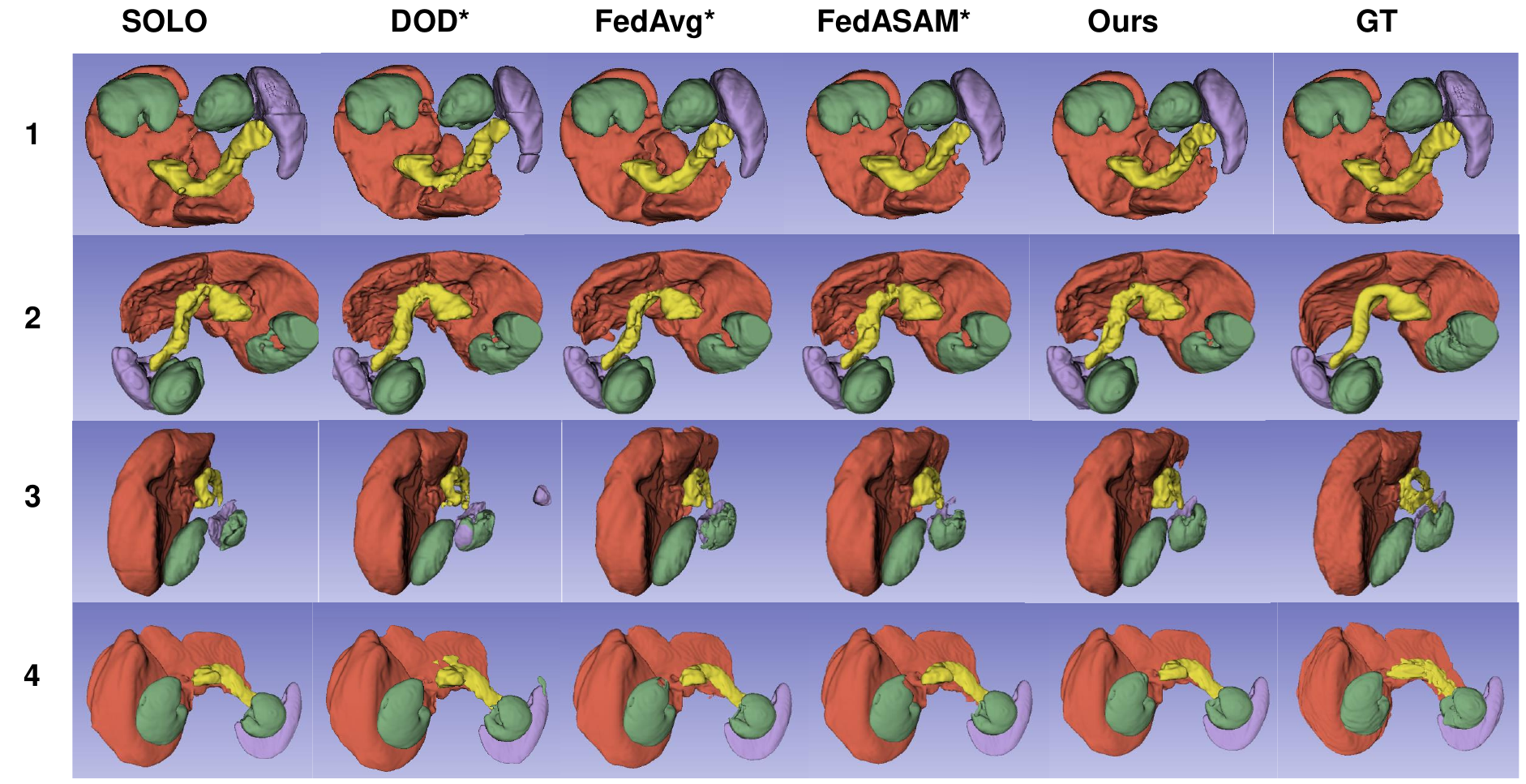}
	\caption{\textbf{3D segmentation result on the test set.}\\Numbers on the left side of images refer to client index. Green, red, purple, yellow regions represent kidney, liver, spleen and pancreas, respectively.}
	\label{Figure 12}
\end{figure*}

From the 3D segmentation results in \hyperref[Figure 12]{Figure 12}, DOD* generates more false positives for client 3 and client 4. It can explained by the facts that data distribution of client 3 is relatively biased from the global and that client 4 is not involved in training. This proves that personalized models do not have desirable generalization ability. By contrast, UFPS uses a single model to generalize well on all datasets and all classes. Furthermore, UFPS is also able to fix some unnatural segmentation in ground truth, e.g., spleen in client 1, which shows the great potential to apply our method in real-world applications.

\section{Discussion}  \label{Discussion}
In this work, we analyze challenges in FPSS for direct combinations between PSS and FL methods. Our proposed UFPS is able to segment all classes based on several partially-annotated datasets by a single global model. Our training process integrates ULL and sUSAM. While ULL denoises pseudo labels and explores underlying values in hard classes, sUSAM unifies local training in FL to a global direction. The overall framework is of low demand for computational resources and time-saving during test time compared with pFL based methods.

Our experiments demonstrate the strong generalization ability of UFPS since it absorbs knowledge from multiple sites and organ interactions. Effectiveness and sensitivity of hyper-parameters for each module in ULL are also comprehensively investigated. Through detailed module ablation studies of sUSAM, we verify our key insights on how to enhance ASAM based framework to a more generalized and faster version in FL. 

In terms of limitation, some key hyper-parameters, e.g. threshold in WS and perturbation radius of UFPS, rely on manual fine-tuning. This can be resolved by reinforced learning or other automatic parameter adjusting methods.

\section{SUPPLEMENTAL INFORMATION} \label{SUPPLEMENTAL INFORMATION}
\noindent Supplemental information can be found at Supplemental information.pdf.

\section{ACKNOWLEDGMENTS}
\noindent This work was supported in part by the National Key R\&D Program of China (No. 2021YFA1003004), in part by the Shanghai Municipal Natural Science Foundation under Grant 21ZR1423300. 

\section{AUTHOR CONTRIBUTIONS}
\noindent Conceptualization, Jiang, L.; methodology, Jiang, L.; formal analysis, Jiang, L. and Ma, L.-Y.; investigation, Jiang, L. and Ma, L.-Y.; writing – original draft, Jiang, L. and Ma, L.-Y.; writing – review \& editing, all authors; visualization, Jiang, L.; funding acquisition, Ma, L.-Y., Zeng, T.-Y. and Ying, S.-H.; resources, Ma, L.-Y. and Zeng, T.-Y.; supervision, Ma, L.-Y. and Zeng, T.-Y.

\section{DECLARATION OF INTERESTS}
\noindent The authors declare no competing interests.

\section{Reference}  \label{Reference}

\bibliographystyle{unsrt}
\bibliography{refs}

\begin{thebibliography}{10}

\bibitem{lecun2015deep}
Yann LeCun, Yoshua Bengio, and Geoffrey Hinton.
\newblock Deep learning.
\newblock {\em nature}, 521(7553):436--444, 2015.
\newblock \href{https://doi.org/10.1038/nature14539}{https://doi.org/10.1038/nature14539}.

\bibitem{DOI199997}
Kunio Doi, Heber MacMahon, Shigehiko Katsuragawa, Robert~M Nishikawa, and Yulei Jiang.
\newblock Computer-aided diagnosis in radiology: potential and pitfalls.
\newblock {\em European Journal of Radiology}, 31(2):97--109, 1999.
\newblock \href{https://doi.org/10.1016/S0720-048X(99)00016-9}{https://doi.org/10.1016/S0720-048X(99)00016-9}.

\bibitem{DOI2007198}
Kunio Doi.
\newblock Computer-aided diagnosis in medical imaging: Historical review, current status and future potential.
\newblock {\em Computerized Medical Imaging and Graphics}, 31(4):198--211, 2007.
\newblock \href{https://doi.org/10.1016/j.compmedimag.2007.02.002}{https://doi.org/10.1016/j.compmedimag.2007.02.002}.

\bibitem{greenwald2022whole}
Noah~F Greenwald, Geneva Miller, Erick Moen, Alex Kong, Adam Kagel, Thomas Dougherty, Christine~Camacho Fullaway, Brianna~J McIntosh, Ke~Xuan Leow, Morgan~Sarah Schwartz, et~al.
\newblock Whole-cell segmentation of tissue images with human-level performance using large-scale data annotation and deep learning.
\newblock {\em Nature biotechnology}, 40(4):555--565, 2022.
\newblock \href{https://doi.org/10.1038/s41587-021-01094-0}{https://doi.org/10.1038/s41587-021-01094-0}.

\bibitem{david2019applications}
Laurianne David, Josep Ar{\'u}s-Pous, Johan Karlsson, Ola Engkvist, Esben~Jannik Bjerrum, Thierry Kogej, Jan~M Kriegl, Bernd Beck, and Hongming Chen.
\newblock Applications of deep-learning in exploiting large-scale and heterogeneous compound data in industrial pharmaceutical research.
\newblock {\em Frontiers in pharmacology}, 10:1303, 2019.
\newblock \href{https://www.frontiersin.org/articles/10.3389/fphar.2019.01303}{https://www.frontiersin.org/articles/10.3389/fphar.2019.01303}.

\bibitem{zhou2019prior}
Yuyin Zhou, Zhe Li, Song Bai, Chong Wang, Xinlei Chen, Mei Han, Elliot Fishman, and Alan~L Yuille.
\newblock Prior-aware neural network for partially-supervised multi-organ segmentation.
\newblock In {\em Proceedings of the IEEE/CVF international conference on computer vision}, pages 10672--10681, 2019.
\newblock \href{https://doi.org/10.1109/ICCV.2019.01077}{https://doi.org/10.1109/ICCV.2019.01077}.

\bibitem{SHI2021101979}
Gonglei Shi, Li~Xiao, Yang Chen, and S.~Kevin Zhou.
\newblock Marginal loss and exclusion loss for partially supervised multi-organ segmentation.
\newblock {\em Medical Image Analysis}, 70:101979, 2021.
\newblock \href{https://doi.org/10.1016/j.media.2021.101979}{https://doi.org/10.1016/j.media.2021.101979}.

\bibitem{10.1007/978-3-030-58598-3_23}
Qi~Fan, Lei Ke, Wenjie Pei, and Yu-Wing Tang, Chi-Keungand~Tai.
\newblock Commonality-parsing network across shape and appearance for partially supervised instance segmentation.
\newblock In {\em Computer Vision -- ECCV 2020}, pages 379--396. Springer International Publishing, 2020.
\newblock \href{https://doi.org/10.1007/978-3-030-58598-3\_23}{https://doi.org/10.1007/978-3-030-58598-3\_23}.

\bibitem{fang2020multi}
Xi~Fang and Pingkun Yan.
\newblock Multi-organ segmentation over partially labeled datasets with multi-scale feature abstraction.
\newblock {\em IEEE Transactions on Medical Imaging}, 39(11):3619--3629, 2020.
\newblock \href{https://doi.org/10.1109/TMI.2020.3001036}{https://doi.org/10.1109/TMI.2020.3001036}.

\bibitem{annas2002medical}
George~J Annas.
\newblock Medical privacy and medical research: judging the new federal regulations.
\newblock {\em New England Journal of Medicine}, 346:216, 2002.
\newblock \href{https://www.nejm.org/doi/full/10.1056/NEJM200205233462118}{https://www.nejm.org/doi/full/10.1056/NEJM200205233462118}.

\bibitem{mcmahan2017communication}
Brendan McMahan, Eider Moore, Daniel Ramage, Seth Hampson, and Blaise~Aguera y~Arcas.
\newblock Communication-efficient learning of deep networks from decentralized data.
\newblock In {\em Artificial intelligence and statistics}, pages 1273--1282. PMLR, 2017.
\newblock \href{https://www.nejm.org/doi/full/10.1056/NEJM200205233462118}{https://www.nejm.org/doi/full/10.1056/NEJM200205233462118}.

\bibitem{BALTER1998939}
James~M. Balter, Kwok~L. Lam, Cornealeus~J. McGinn, Theodore~S. Lawrence, and Randall~K. {Ten Haken}.
\newblock Improvement of ct-based treatment-planning models of abdominal targets using static exhale imaging.
\newblock {\em International Journal of Radiation Oncology*Biology*Physics}, 41(4):939--943, 1998.
\newblock \href{https://doi.org/10.1016/S0360-3016(98)00130-8}{https://doi.org/10.1016/S0360-3016(98)00130-8}.

\bibitem{Zheng2017}
Yefeng Zheng, David Liu, Bogdan Georgescu, Daguang Xu, and Dorin Comaniciu.
\newblock {\em Deep Learning Based Automatic Segmentation of Pathological Kidney in CT: Local Versus Global Image Context}, pages 241--255.
\newblock Springer International Publishing, 2017.
\newblock \href{https://doi.org/10.1007/978-3-319-42999-1\_14}{https://doi.org/10.1007/978-3-319-42999-1\_14}.

\bibitem{chen2022magicnet}
Duowen Chen, Yunhao Bai, Wei Shen, Qingli Li, Lequan Yu, and Yan Wang.
\newblock Magicnet: Semi-supervised multi-organ segmentation via magic-cube partition and recovery.
\newblock {\em arXiv preprint arXiv:2212.14310}, 2022.
\newblock \href{https://arxiv.org/abs/2212.14310}{https://arxiv.org/abs/2212.14310}.

\bibitem{zhao2018federated}
Yue Zhao, Meng Li, Liangzhen Lai, Naveen Suda, Damon Civin, and Vikas Chandra.
\newblock Federated learning with non-iid data.
\newblock {\em arXiv preprint arXiv:1806.00582}, 2018.
\newblock \href{https://arxiv.org/abs/1806.00582}{https://arxiv.org/abs/1806.00582}.

\bibitem{li2019convergence}
Xiang Li, Kaixuan Huang, Wenhao Yang, Shusen Wang, and Zhihua Zhang.
\newblock On the convergence of fedavg on non-iid data.
\newblock {\em arXiv preprint arXiv:1907.02189}, 2019.
\newblock \href{https://arxiv.org/abs/1907.02189}{https://arxiv.org/abs/1907.02189}.

\bibitem{VANOMMEN201965}
F.~{van Ommen}, H.W.A.M. {de Jong}, J.W. Dankbaar, E.~Bennink, T.~Leiner, and A.M.R. Schilham.
\newblock Dose of ct protocols acquired in clinical routine using a dual-layer detector ct scanner: A preliminary report.
\newblock {\em European Journal of Radiology}, 112:65--71, 2019.
\newblock \href{https://doi.org/10.1016/j.ejrad.2019.01.011}{https://doi.org/10.1016/j.ejrad.2019.01.011}.

\bibitem{tischenko2006new}
Oleg Tischenko, Yuan Xu, and Christoph Hoeschen.
\newblock A new scanning device in ct with dose reduction potential.
\newblock In {\em Medical Imaging 2006: Physics of Medical Imaging}, volume 6142, pages 893--899. SPIE, 2006.
\newblock \href{https://doi.org/10.1117/12.654463}{https://doi.org/10.1117/12.654463}.

\bibitem{sharma2021improving}
Ashwarya Sharma and Latha Palaniappan.
\newblock Improving diversity in medical research.
\newblock {\em Nature Reviews Disease Primers}, 7(1):74, 2021.
\newblock \href{https://doi.org/10.1038/s41572-021-00316-8}{https://doi.org/10.1038/s41572-021-00316-8}.

\bibitem{li2020federated}
Tian Li, Anit~Kumar Sahu, Manzil Zaheer, Maziar Sanjabi, Ameet Talwalkar, and Virginia Smith.
\newblock Federated optimization in heterogeneous networks.
\newblock {\em Proceedings of Machine learning and systems}, 2:429--450, 2020.
\newblock \href{https://proceedings.mlsys.org/paper\_files/paper/2020/file/38af86134b65d0f10fe33d30dd76442e-Paper.pdf}{https://proceedings.mlsys.org/paper\_files/paper/2020/file/38af86134\\b65d0f10fe33d30dd76442e-Paper.pdf}.

\bibitem{karimireddy2020scaffold}
Sai~Praneeth Karimireddy, Satyen Kale, Mehryar Mohri, Sashank Reddi, Sebastian Stich, and Ananda~Theertha Suresh.
\newblock Scaffold: Stochastic controlled averaging for federated learning.
\newblock In {\em International Conference on Machine Learning}, pages 5132--5143. PMLR, 2020.
\newblock \href{https://proceedings.mlr.press/v119/karimireddy20a.html}{https://proceedings.mlr.press/v119/karimireddy20a.html}.

\bibitem{zhang2021federated}
Lin Zhang, Yong Luo, Yan Bai, Bo~Du, and Ling-Yu Duan.
\newblock Federated learning for non-iid data via unified feature learning and optimization objective alignment.
\newblock In {\em Proceedings of the IEEE/CVF international conference on computer vision}, pages 4420--4428, 2021.
\newblock \href{https://doi.org/10.1109/ICCV48922.2021.00438}{https://doi.org/10.1109/ICCV48922.2021.00438}.

\bibitem{jiang2022harmofl}
Meirui Jiang, Zirui Wang, and Qi~Dou.
\newblock Harmofl: Harmonizing local and global drifts in federated learning on heterogeneous medical images.
\newblock In {\em Proceedings of the AAAI Conference on Artificial Intelligence}, volume~36, pages 1087--1095, 2022.
\newblock \href{https://doi.org/10.1609/aaai.v36i1.19993}{https://doi.org/10.1609/aaai.v36i1.19993}.

\bibitem{gao2022feddc}
Liang Gao, Huazhu Fu, Li~Li, Yingwen Chen, Ming Xu, and Cheng-Zhong Xu.
\newblock Feddc: Federated learning with non-iid data via local drift decoupling and correction.
\newblock In {\em Proceedings of the IEEE/CVF Conference on Computer Vision and Pattern Recognition}, pages 10112--10121, 2022.
\newblock \href{https://doi.org/10.1109/CVPR52688.2022.00987}{https://doi.org/10.1109/CVPR52688.2022.00987}.

\bibitem{mendieta2022local}
Matias Mendieta, Taojiannan Yang, Pu~Wang, Minwoo Lee, Zhengming Ding, and Chen Chen.
\newblock Local learning matters: Rethinking data heterogeneity in federated learning.
\newblock In {\em Proceedings of the IEEE/CVF Conference on Computer Vision and Pattern Recognition}, pages 8397--8406, 2022.
\newblock \href{https://doi.org/10.1109/CVPR52688.2022.00821}{https://doi.org/10.1109/CVPR52688.2022.00821}.

\bibitem{10.1007/978-3-031-20050-2_38}
Debora Caldarola, Barbara Caputo, and Marco Ciccone.
\newblock Improving generalization in federated learning by seeking flat minima.
\newblock In Shai Avidan, Gabriel Brostow, Moustapha Ciss{\'e}, Giovanni~Maria Farinella, and Tal Hassner, editors, {\em Computer Vision -- ECCV 2022}, pages 654--672, Cham, 2022. Springer Nature Switzerland.
\newblock \href{https://doi.org/10.1007/978-3-031-20050-2\_38}{https://doi.org/10.1007/978-3-031-20050-2\_38}.

\bibitem{balakrishnan2022diverse}
Ravikumar Balakrishnan, Tian Li, Tianyi Zhou, Nageen Himayat, Virginia Smith, and Jeff Bilmes.
\newblock Diverse client selection for federated learning via submodular maximization.
\newblock In {\em International Conference on Learning Representations}, 2022.
\newblock \href{https://openreview.net/pdf?id=nwKXyFvaUm}{https://openreview.net/pdf?id=nwKXyFvaUm}.

\bibitem{tang2022fedcor}
Minxue Tang, Xuefei Ning, Yitu Wang, Jingwei Sun, Yu~Wang, Hai Li, and Yiran Chen.
\newblock Fedcor: Correlation-based active client selection strategy for heterogeneous federated learning.
\newblock In {\em Proceedings of the IEEE/CVF Conference on Computer Vision and Pattern Recognition}, pages 10102--10111, 2022.
\newblock \href{https://doi.org/10.1109/CVPR52688.2022.00986}{https://doi.org/10.1109/CVPR52688.2022.00986}.

\bibitem{li2021model}
Qinbin Li, Bingsheng He, and Dawn Song.
\newblock Model-contrastive federated learning.
\newblock In {\em Proceedings of the IEEE/CVF Conference on Computer Vision and Pattern Recognition}, pages 10713--10722, 2021.
\newblock \href{https://doi.org/10.1109/CVPR46437.2021.01057}{https://doi.org/10.1109/CVPR46437.2021.01057}.

\bibitem{10.1007/978-3-031-20056-4_40}
Sungwon Han, Sungwon Park, Fangzhao Wu, Sundong Kim, Chuhan Wu, Xing Xie, and Meeyoung Cha.
\newblock Fedx: Unsupervised federated learning with cross knowledge distillation.
\newblock In Shai Avidan, Gabriel Brostow, Moustapha Ciss{\'e}, Giovanni~Maria Farinella, and Tal Hassner, editors, {\em Computer Vision -- ECCV 2022}, pages 691--707, Cham, 2022. Springer Nature Switzerland.
\newblock \href{https://doi.org/10.1007/978-3-031-20056-4\_40}{https://doi.org/10.1007/978-3-031-20056-4\_40}.

\bibitem{10.1007/978-3-031-16443-9_25}
Xiaoming Qi, Guanyu Yang, Yuting He, Wangyan Liu, Ali Islam, and Shuo Li.
\newblock Contrastive re-localization and history distillation in federated cmr segmentation.
\newblock In Linwei Wang, Qi~Dou, P.~Thomas Fletcher, Stefanie Speidel, and Shuo Li, editors, {\em Medical Image Computing and Computer Assisted Intervention -- MICCAI 2022}, pages 256--265, Cham, 2022. Springer Nature Switzerland.
\newblock \href{https://doi.org/10.1007/978-3-031-16443-9\_25}{https://doi.org/10.1007/978-3-031-16443-9\_25}.

\bibitem{yu2023multimodal}
Qiying Yu, Yang Liu, Yimu Wang, Ke~Xu, and Jingjing Liu.
\newblock Multimodal federated learning via contrastive representation ensemble.
\newblock {\em arXiv preprint arXiv:2302.08888}, 2023.
\newblock \href{https://ar5iv.labs.arxiv.org/html/2302.08888}{https://ar5iv.labs.arxiv.org/html/2302.08888}.

\bibitem{MU202393}
Xutong Mu, Yulong Shen, Ke~Cheng, Xueli Geng, Jiaxuan Fu, Tao Zhang, and Zhiwei Zhang.
\newblock Fedproc: Prototypical contrastive federated learning on non-iid data.
\newblock {\em Future Generation Computer Systems}, 143:93--104, 2023.
\newblock \href{https://doi.org/10.1016/j.future.2023.01.019}{https://doi.org/10.1016/j.future.2023.01.019}.

\bibitem{posner2021federated}
Jason Posner, Lewis Tseng, Moayad Aloqaily, and Yaser Jararweh.
\newblock Federated learning in vehicular networks: Opportunities and solutions.
\newblock {\em IEEE Network}, 35(2):152--159, 2021.
\newblock \href{https://doi.org/10.1109/MNET.011.2000430}{https://doi.org/10.1109/MNET.011.2000430}.

\bibitem{kwon2021asam}
Jungmin Kwon, Jeongseop Kim, Hyunseo Park, and In~Kwon Choi.
\newblock Asam: Adaptive sharpness-aware minimization for scale-invariant learning of deep neural networks.
\newblock In {\em International Conference on Machine Learning}, pages 5905--5914. PMLR, 2021.
\newblock \href{https://proceedings.mlr.press/v139/kwon21b.html}{https://proceedings.mlr.press/v139/kwon21b.html}.

\bibitem{LIAN2023104339}
Sheng Lian, Lei Li, Zhiming Luo, Zhun Zhong, Beizhan Wang, and Shaozi Li.
\newblock Learning multi-organ segmentation via partial- and mutual-prior from single-organ datasets.
\newblock {\em Biomedical Signal Processing and Control}, 80:104339, 2023.
\newblock \href{https://doi.org/10.1016/j.bspc.2022.104339}{https://doi.org/10.1016/j.bspc.2022.104339}.

\bibitem{dmitriev2019learning}
Konstantin Dmitriev and Arie~E Kaufman.
\newblock Learning multi-class segmentations from single-class datasets.
\newblock In {\em Proceedings of the IEEE/CVF Conference on Computer Vision and Pattern Recognition}, pages 9501--9511, 2019.
\newblock \href{https://doi.org/10.1109/CVPR.2019.00973}{https://doi.org/10.1109/CVPR.2019.00973}.

\bibitem{zhang2021dodnet}
Jianpeng Zhang, Yutong Xie, Yong Xia, and Chunhua Shen.
\newblock Dodnet: Learning to segment multi-organ and tumors from multiple partially labeled datasets.
\newblock In {\em Proceedings of the IEEE/CVF conference on computer vision and pattern recognition}, pages 1195--1204, 2021.
\newblock \href{https://doi.org/10.1109/CVPR46437.2021.00125}{https://doi.org/10.1109/CVPR46437.2021.00125}.

\bibitem{chen2021semi}
Xiaokang Chen, Yuhui Yuan, Gang Zeng, and Jingdong Wang.
\newblock Semi-supervised semantic segmentation with cross pseudo supervision.
\newblock In {\em Proceedings of the IEEE/CVF Conference on Computer Vision and Pattern Recognition}, pages 2613--2622, 2021.
\newblock \href{https://doi.org/10.1109/CVPR46437.2021.00264}{https://doi.org/10.1109/CVPR46437.2021.00264}.

\bibitem{feng2021ms}
Shixiang Feng, Yuhang Zhou, Xiaoman Zhang, Ya~Zhang, and Yanfeng Wang.
\newblock Ms-kd: Multi-organ segmentation with multiple binary-labeled datasets.
\newblock {\em arXiv preprint arXiv:2108.02559}, 2021.
\newblock \href{https://arxiv.org/abs/2108.02559}{https://arxiv.org/abs/2108.02559}.

\bibitem{hall1987kullback}
Peter Hall.
\newblock On kullback-leibler loss and density estimation.
\newblock {\em The Annals of Statistics}, pages 1491--1519, 1987.
\newblock \href{https://doi.org/10.1214/aos/1176350606}{https://doi.org/10.1214/aos/1176350606}.

\bibitem{izmailov2018averaging}
Pavel Izmailov, Dmitrii Podoprikhin, Timur Garipov, Dmitry Vetrov, and Andrew~Gordon Wilson.
\newblock Averaging weights leads to wider optima and better generalization.
\newblock {\em arXiv preprint arXiv:1803.05407}, 2018.
\newblock \href{https://arxiv.org/abs/1803.05407}{https://arxiv.org/abs/1803.05407}.

\bibitem{xu2022federated}
Xuanang Xu and Pingkun Yan.
\newblock Federated multi-organ segmentation with partially labeled data.
\newblock {\em arXiv preprint arXiv:2206.07156}, 2022.
\newblock \href{https://arxiv.org/abs/2206.07156}{https://arxiv.org/abs/2206.07156}.

\bibitem{10.1007/978-3-031-18523-6_6}
Chen Shen, Pochuan Wang, Dong Yang, Daguang Xu, Masahiro Oda, Po-Ting Chen, Kao-Lang Liu, Wei-Chih Liao, Chiou-Shann Fuh, Kensaku Mori, Weichung Wang, and Holger~R. Roth.
\newblock Joint multi organ and tumor segmentation from partial labels using federated learning.
\newblock In Shadi Albarqouni, Spyridon Bakas, Sophia Bano, M.~Jorge Cardoso, Bishesh Khanal, Bennett Landman, Xiaoxiao Li, Chen Qin, Islem Rekik, Nicola Rieke, Holger Roth, Debdoot Sheet, and Daguang Xu, editors, {\em Distributed, Collaborative, and Federated Learning, and Affordable AI and Healthcare for Resource Diverse Global Health}, pages 58--67, Cham, 2022. Springer Nature Switzerland.
\newblock \href{https://doi.org/10.1007/978-3-031-18523-6\_6}{https://doi.org/10.1007/978-3-031-18523-6\_6}.

\bibitem{liang2020think}
Paul~Pu Liang, Terrance Liu, Liu Ziyin, Nicholas~B Allen, Randy~P Auerbach, David Brent, Ruslan Salakhutdinov, and Louis-Philippe Morency.
\newblock Think locally, act globally: Federated learning with local and global representations.
\newblock {\em arXiv preprint arXiv:2001.01523}, 2020.
\newblock \href{https://arxiv.org/abs/2001.01523}{https://arxiv.org/abs/2001.01523}.

\bibitem{collins2021exploiting}
Liam Collins, Hamed Hassani, Aryan Mokhtari, and Sanjay Shakkottai.
\newblock Exploiting shared representations for personalized federated learning.
\newblock In {\em International Conference on Machine Learning}, pages 2089--2099. PMLR, 2021.
\newblock \href{https://proceedings.mlr.press/v139/collins21a.html}{https://proceedings.mlr.press/v139/collins21a.html}.

\bibitem{tan2022towards}
Alysa~Ziying Tan, Han Yu, Lizhen Cui, and Qiang Yang.
\newblock Towards personalized federated learning.
\newblock {\em IEEE Transactions on Neural Networks and Learning Systems}, 2022.
\newblock \href{https://doi.org/10.1109/TNNLS.2022.3160699}{https://doi.org/10.1109/TNNLS.2022.3160699}.

\bibitem{diao2022semifl}
Enmao Diao, Jie Ding, and Vahid Tarokh.
\newblock Semifl: Semi-supervised federated learning for unlabeled clients with alternate training.
\newblock {\em Advances in Neural Information Processing Systems}, 35:17871--17884, 2022.
\newblock \href{https://proceedings.neurips.cc/paper\_files/paper/2022/file/71c3451f6cd6a4f82bb822db25cea4fd-Paper-Conference.pdf}{https://proceedings.neurips.cc/paper\_files/paper/2022/file/71c3451f6\\cd6a4f82bb822db25cea4fd-Paper-Conference.pdf}.

\bibitem{zheng2021rectifying}
Zhedong Zheng and Yi~Yang.
\newblock Rectifying pseudo label learning via uncertainty estimation for domain adaptive semantic segmentation.
\newblock {\em International Journal of Computer Vision}, 129(4):1106--1120, 2021.
\newblock \href{https://doi.org/10.1007/s11263-020-01395-y}{https://doi.org/10.1007/s11263-020-01395-y}.

\bibitem{10.1007/978-3-030-87240-3_22}
Cheng Chen, Quande Liu, Yueming Jin, Qi~Dou, and Pheng-Ann Heng.
\newblock Source-free domain adaptive fundus image segmentation with denoised pseudo-labeling.
\newblock In Marleen de~Bruijne, Philippe~C. Cattin, St{\'e}phane Cotin, Nicolas Padoy, Stefanie Speidel, Yefeng Zheng, and Caroline Essert, editors, {\em Medical Image Computing and Computer Assisted Intervention -- MICCAI 2021}, pages 225--235, Cham, 2021. Springer International Publishing.
\newblock \href{https://doi.org/10.1007/978-3-030-87240-3\_22}{https://doi.org/10.1007/978-3-030-87240-3\_22}.

\bibitem{wang2019symmetric}
Yisen Wang, Xingjun Ma, Zaiyi Chen, Yuan Luo, Jinfeng Yi, and James Bailey.
\newblock Symmetric cross entropy for robust learning with noisy labels.
\newblock In {\em Proceedings of the IEEE/CVF International Conference on Computer Vision}, pages 322--330, 2019.
\newblock \href{https://doi.org/10.1109/ICCV.2019.00041}{https://doi.org/10.1109/ICCV.2019.00041}.

\bibitem{ouyang2022causality}
Cheng Ouyang, Chen Chen, Surui Li, Zeju Li, Chen Qin, Wenjia Bai, and Daniel Rueckert.
\newblock Causality-inspired single-source domain generalization for medical image segmentation.
\newblock {\em IEEE Transactions on Medical Imaging}, 2022.
\newblock \href{https://doi.org/10.1109/TMI.2022.3224067}{https://doi.org/10.1109/TMI.2022.3224067}.

\bibitem{mi2022make}
Peng Mi, Li~Shen, Tianhe Ren, Yiyi Zhou, Xiaoshuai Sun, Rongrong Ji, and Dacheng Tao.
\newblock Make sharpness-aware minimization stronger: A sparsified perturbation approach.
\newblock {\em arXiv preprint arXiv:2210.05177}, 2022.
\newblock \href{https://proceedings.neurips.cc/paper\_files/paper/2022/file/c859b99b5d717c9035e79d43dfd69435-Paper-Conference.pdf}{https://proceedings.neurips.cc/paper\_files/paper/2022/file/c859b99b5\\d717c9035e79d43dfd69435-Paper-Conference.pdf}.

\bibitem{jiang4474880ufps}
Le~Jiang, Liyan Ma, Tieyong Zeng, and Shi~Hui Ying.
\newblock Code, datasets, and results for the paper "ufps: A unified framework for partially-annotated federated segmentation in heterogeneous data distribution".
\newblock 2023.
\newblock \href{https://zenodo.org/doi/10.5281/zenodo.10140361}{https://zenodo.org/doi/10.5281/zenodo.10140361}.

\bibitem{jiang2022test}
Liangze Jiang and Tao Lin.
\newblock Test-time robust personalization for federated learning.
\newblock {\em arXiv preprint arXiv:2205.10920}, 2022.
\newblock \href{https://arxiv.org/abs/2205.10920}{https://arxiv.org/abs/2205.10920}.

\end{thebibliography}


\begin{thebibliography}{10}

\bibitem{sinha2017certifiable}
Aman Sinha, Hongseok Namkoong, and John Duchi.
\newblock Certifiable distributional robustness with principled adversarial training.
\newblock {\em arXiv preprint arXiv:1710.10571}, 2, 2017.
\newblock \href{https://arxiv.org/abs/1710.10571}{https://arxiv.org/abs/1710.10571}.

\bibitem{bartlett2001rademacher}
Peter~L Bartlett and Shahar Mendelson.
\newblock Rademacher and gaussian complexities: Risk bounds and structural results.
\newblock In {\em Computational Learning Theory: 14th Annual Conference on Computational Learning Theory, COLT 2001 and 5th European Conference on Computational Learning Theory, EuroCOLT 2001 Amsterdam, The Netherlands, July 16--19, 2001 Proceedings 14}, pages 224--240. Springer, 2001.
\newblock \href{https://doi.org/10.1007/3-540-44581-1\_15}{https://doi.org/10.1007/3-540-44581-1\_15}.

\bibitem{wellner2013weak}
Thomas Mikosch, Aad van~der Vaart, and Jon~A. Wellner.
\newblock {\em Weak Convergence and Empirical Processes: With Applications to Statistics}.
\newblock Springer Science \& Business Media, 1996.
\newblock \href{https://link.springer.com/book/10.1007/978-1-4757-2545-2}{https://link.springer.com/book/10.1007/978-1-4757-2545-2}.

\bibitem{qu2022generalized}
Zhe Qu, Xingyu Li, Rui Duan, Yao Liu, Bo~Tang, and Zhuo Lu.
\newblock Generalized federated learning via sharpness aware minimization.
\newblock In {\em International Conference on Machine Learning}, pages 18250--18280. PMLR, 2022.
\newblock \href{https://proceedings.mlr.press/v162/qu22a.html}{https://proceedings.mlr.press/v162/qu22a.html}.

\bibitem{yang2021achieving}
Haibo Yang, Minghong Fang, and Jia Liu.
\newblock Achieving linear speedup with partial worker participation in non-iid federated learning.
\newblock {\em arXiv preprint arXiv:2101.11203}, 2021.
\newblock \href{https://arxiv.org/abs/2101.11203}{https://arxiv.org/abs/2101.11203}.

\bibitem{isensee2021nnu}
Fabian Isensee, Paul~F Jaeger, Simon~AA Kohl, Jens Petersen, and Klaus~H Maier-Hein.
\newblock nnu-net: a self-configuring method for deep learning-based biomedical image segmentation.
\newblock {\em Nature methods}, 18(2):203--211, 2021.
\newblock \href{https://doi.org/10.1038/s41592-020-01008-z}{https://doi.org/10.1038/s41592-020-01008-z}.

\bibitem{mcmahan2017communication}
Brendan McMahan, Eider Moore, Daniel Ramage, Seth Hampson, and Blaise~Aguera y~Arcas.
\newblock Communication-efficient learning of deep networks from decentralized data.
\newblock In {\em Artificial intelligence and statistics}, pages 1273--1282. PMLR, 2017.
\newblock \href{https://www.nejm.org/doi/full/10.1056/NEJM200205233462118}{https://www.nejm.org/doi/full/10.1056/NEJM200205233462118}.

\bibitem{cardoso2022monai}
M~Jorge Cardoso, Wenqi Li, Richard Brown, Nic Ma, Eric Kerfoot, Yiheng Wang, Benjamin Murrey, Andriy Myronenko, Can Zhao, Dong Yang, et~al.
\newblock Monai: An open-source framework for deep learning in healthcare.
\newblock {\em arXiv preprint arXiv:2211.02701}, 2022.
\newblock \href{https://arxiv.org/abs/2211.02701}{https://arxiv.org/abs/2211.02701}.

\bibitem{loshchilov2017decoupled}
Ilya Loshchilov and Frank Hutter.
\newblock Decoupled weight decay regularization.
\newblock {\em arXiv preprint arXiv:1711.05101}, 2017.
\newblock \href{https://arxiv.org/abs/1711.05101}{https://arxiv.org/abs/1711.05101}.

\bibitem{grill2020bootstrap}
Jean-Bastien Grill, Florian Strub, Florent Altch{\'e}, Corentin Tallec, Pierre Richemond, Elena Buchatskaya, Carl Doersch, Bernardo Avila~Pires, Zhaohan Guo, Mohammad Gheshlaghi~Azar, et~al.
\newblock Bootstrap your own latent-a new approach to self-supervised learning.
\newblock {\em Advances in neural information processing systems}, 33:21271--21284, 2020.
\newblock \href{https://proceedings.neurips.cc/paper\_files/paper/2020/file/f3ada80d5c4ee70142b17b8192b2958e-Paper.pdf}{https://proceedings.neurips.cc/paper\_files/paper/2020/file/f3ada80d5c4ee70142b17b8192b2958e-Paper.pdf}.

\bibitem{kornblith2019similarity}
Simon Kornblith, Mohammad Norouzi, Honglak Lee, and Geoffrey Hinton.
\newblock Similarity of neural network representations revisited.
\newblock In {\em International Conference on Machine Learning}, pages 3519--3529. PMLR, 2019.
\newblock \href{http://proceedings.mlr.press/v97/kornblith19a/kornblith19a.pdf}{http://proceedings.mlr.press/v97/kornblith19a/kornblith19a.pdf}.

\bibitem{shokri2017membership}
Reza Shokri, Marco Stronati, Congzheng Song, and Vitaly Shmatikov.
\newblock Membership inference attacks against machine learning models.
\newblock In {\em 2017 IEEE symposium on security and privacy (SP)}, pages 3--18. IEEE, 2017.
\newblock \href{https://doi.org/10.1109/SP.2017.41}{https://doi.org/10.1109/SP.2017.41}.

\bibitem{chobola2022membership}
Tomas Chobola, Dmitrii Usynin, and Georgios Kaissis.
\newblock Membership inference attacks against semantic segmentation models.
\newblock {\em arXiv preprint arXiv:2212.01082}, 2022.
\newblock \href{https://arxiv.org/abs/2212.01082}{https://arxiv.org/abs/2212.01082}.

\bibitem{dwork2006differential}
Cynthia Dwork.
\newblock Differential privacy.
\newblock In {\em International colloquium on automata, languages, and programming}, pages 1--12. Springer, 2006.
\newblock \href{https://link.springer.com/chapter/10.1007/11787006\_1}{https://link.springer.com/chapter/10.1007/11787006\_1}.

\end{thebibliography}

\end{sloppypar}
\end{document}


\begin{sloppypar}

\let\WriteBookmarks\relax
\def\floatpagepagefraction{1}
\def\textpagefraction{.001}

\newcommand{\upcite}[1]{\textsuperscript{\cite{#1}}}
\captionsetup[figure]{labelfont={bf},name={Figure S},labelsep=none}
\thispagestyle{empty}
\pagestyle{empty}
\makeatletter
\def\@biblabel#1{#1.~}
\makeatother






\section*{Supplementary Information}  \label{Supplementary Information}

\section{Notations}   \label{Notations}

\makeatletter

\def\@captype{table*}
\label{Table S1}
\rowcolors {3}{grey}{white}
	\renewcommand\arraystretch{1.2}
	\resizebox{0.95\columnwidth}{!}{
	\begin{tabular}{ll}
	\arrayrulecolor{red}
	\hline
		\multicolumn{2}{l}{\textcolor{red}{\textbf{Table S1. Description for notations occurred in this paper.}}}  \\
	\hline
	\textcolor{red}{Notation} & \textcolor{red}{Description}                                               \\
	\hline
	$(x,y/y^p)$ & Image and corresponding fully-annotated / partially-annotated label.   \\
	$\left(X, Y / Y^p\right)$ & Joint space for image and fully-annotated / partially-annotated label. \\
	$D_i^p / D^p$ & Partially-annotated dataset for $i$-th client. / Global partially-annotated dataset.       \\
	$N_i / N$ & Number of data samples in $D_i^p/D^p$.          \\
	${C}$ & Label set for all classes occurring in $Y$.                    \\
	$\left|C_i\right|$ & Number of classes in label space for $i$-th client. \\
	$N_{c}$ & Number of classes in $Y$.                    \\
	${c}_{i, j}$ & Intersection between $C_i$ and $C_j$.   \\
	$R$ & Total number of communication round.          \\
	$K$ & Total number of local rounds. \\
	$w_i^r / w_0^r$ & Local model for $i$-th client at communication round $r$. / Global model at communication round $r$.    \\
	${w}^{G / L}$ & Model part aggregated globally / kept local.   \\
	$w_i^T / w^T$ & Teacher model pretrained by $i$-th client. / All pretrained teacher models. \\
	$f(\cdot)$ & Loss function. \\
	$PL$ & Operation to predict pseudo labels.                    \\
	$Y_l$ & Annotated ground-truth label within foreground classes.   \\
	$p(x)$  & Ground truth.          \\
	$q(x) / \tilde{q}(x)$ & Model prediction. / One-hot model prediction. \\
	$E$ & Information entropy.                    \\
	$\mathcal{T}$ & Hyper-parameter of uncertainty threshold.   \\
	$U / {U}_{\mathcal{T}}$ & Data-wise uncertainty. / Uncertainty value at lowest $\mathcal{T}$ percentage.    \\
	${U}_{\text {bank }}$ & Uncertainty bank. \\
	$\mu / \sigma / {U}_{\text { max } / \text { min } }$ & Mean / variance / maximal / minimal uncertainty value for the uncertainty bank. \\
	${\alpha}$ & Hyper-parameter to determine the minimal base in BD.                    \\
	${\beta}$ & Hyper-parameter to balance terms in RG.   \\
	${\tau}^{{\mu} / {\sigma}}$ & Temperature hyper-parameter for mean / variance in uncertainty-based aggregation.          \\
	${A^w}$ & Aggregation weight only based on the number of local samples. \\
	${\mathscr{v}}$ & Hyper-parameter of volume threshold.                    \\
	${W}$ & Joint model parameter space.          \\
	${T}_{{w}}^{-1}$ & Normalization operator. \\
	${\eta}$ & Hyper-parameter to balance stability and universality in ASAM.                    \\
	${\rho}$ & Hyper-parameter of searching radius for disturbance in ASAM. \\
	${\nabla} {f}_{\text {USAM }} / {\nabla} {f}_{\text {base }}$ & Gradients of USAM / pseudo label baseline.     \\
	$T_{\text {sUSAM }}$ & Hyper-parameter of local mask percentage threshold in sUSAM.          \\
	$M_L / M_G$ & Local / Global mask in sUSAM. \\
	${T}_L$ & Hyper-parameter of non-masking percentage of local mask in sUSAM. \\
	${T}_{{G}}$ & Hyper-parameter of non-masking percentage of local mask according to the global mask in sUSAM.     \\     
	${M}_{L, {m} o}$ & Momentum local mask. \\
	${\alpha}_{{m} o}$ & Hyper-parameter for the update of local momentum gradients.                    \\
	${G}_{{N}}$ & Gradients not in top $T_L\%$ of $ M_L$ but in the nonintersecting part of $M_G$. \\
	${r}_{warmup}$ & Number of warmup rounds. \\
	${r}_{WS}$ & End global round for weight scheduler.                    \\
	${r}_{UA / GMT / sUSAM}$ & Start global round for uncertainty-based aggregation / global mean teacher / sUSAM.  	\\
	${r}_{{fre}}$ & Hyper-parameter of the updating frequency for local mask in sUSAM.          \\
	\hline
	\end{tabular}
}

Notations are described in \hyperref[Table S1]{Table S1}.

\section{Proof of Theorem 1}   \label{Proof of Theorem 1}

\newtheorem{defi}{Definition}  \label{Definition 1}
\begin{defi}
Given a bounded loss satisfying $|f(x , w)| \leq$ $f_{\text {upper }}$, the Wasserstein distance between two data distribution $A, B$ is:
$$
\mathbb{W}_{f_c}(A, B)=\inf _{f_{\text {upper }} \in \Pi(A, B)} \mathbb{E}_{f_{\text {upper }}}\left[f_c\left(X, X^{\prime}\right)\right],
$$
where $\Pi(A, B)$ represents couplings of $A, B, f_c$ denotes nonnegative, lower semi-continuous cost function.
\end{defi}

\noindent Here we cite the lemma from Sinha et.al \upcite{sinha2017certifiable}, which demonstrates the result on any distribution $A$ and $B$.

\newtheorem{lemd}{Lemma}[section] \label{Lemma 2.1}
\begin{lemd}
Let $f: X \times W \rightarrow \mathbb{R}$ and $f_c: X \times X \rightarrow \mathbb{R}_{+}$ be continuous. Let $\varphi_\gamma\left(x^{\prime} , w\right)=\sup _{x \in X}\left[f(x , w)-\gamma c\left(x, x^{\prime}\right)\right]$ be the surrogate objective function. For any distribution $B$ and any $g>0,\gamma>0$
$$
\sup _{\mathbb{W}_{f_c}(A, B) \leq g} \mathbb{E}_A[f(X , w)]=\inf _{\gamma \geq 0} \gamma g+\mathbb{E}_B\left[\varphi_\gamma\left(x^{\prime} , w\right)\right],
$$
and for any $\gamma \geq 0$, we have
$$
\sup _A\left\{\mathbb{E}_A[f(X , w)]-\gamma \mathbb{W}_{f_c}(A, B)\right\}=\mathbb{E}_B\left[\varphi_\gamma(X , w)\right].
$$
\end{lemd}

\newtheorem{theo}{Theorem}  \label{Theorem 1}
\begin{theo}
For any constant $t>0, g>0, \gamma>0, w \in W$, expected error of risk function on the global data distribution satisfies the following inequality at the probability of $1-e^{-t}$:
$$
\begin{gathered}
\sup_{\mathbb{W}_{f_c}(D_{global } , D_{aug }) \leq g} \mathbb{E}_{D_{\text {global }}}[f(X , w)] \leq 
\mathbb{E}_{\widehat{D}_{\text {aug }}}\left[\varphi_\gamma(X , w)\right]+\gamma g+\epsilon_n(t)~, \\
\text{where}~ \epsilon_n(t)=\gamma b_1 \sqrt{\frac{f_{\text {upper }}}{n}} \cdot  
\int_0^1 \sqrt{\log N\left(\mathcal{F}, f_{\text {upper }} \epsilon,\|\cdot\|_{L^{\infty}(X)}\right)} d \epsilon
+b_2 f_{\text {upper }} \sqrt{\frac{t}{n}}~,
\end{gathered}
$$
$D_{\text {global }}, D_{\text {aug }}, \widehat{D}_{\text {aug }}$ corresponds to global data distribution, augmented local data distribution and empiral augmented local data distribution, respectively, $\varphi_\gamma\left(x , w\right)=$ $\sup _{x \in X}\left[f\left(x^{\prime} , w\right)-\gamma f_c\left(x^{\prime}, x\right)\right]$ is a surrogate objective function with penalty $\gamma$ and augmentation $x^{\prime}$ of $x$, $n$ is number of samples of $D_{\text {aug }}$, $\mathcal{F}$ is the hypothesis class, $N(\cdot)$ is covering numbers of $\mathcal{F}$, $b_1, b_2$ are both constants.
\end{theo}

\noindent \textit{Proof.}

Leveraging \hyperref[Lemma 2.1]{Lemma 2.1} to our problem, for all $g>0, \gamma>0$, distribution $D_{\text {aug }}$, we have the deterministic result
$$
\begin{gathered}
\sup _{\mathbb{W}_{f_c}(D_{global} , D_{aug}) \leq g} \mathbb{E}_{D_{global}} [f(X , w)] \leq 
\mathbb{E}_{D_{\text {aug }}}\left[\varphi_\gamma(X , w)\right]+\gamma g.
\end{gathered}
$$
To get the complete form of \hyperref[Theorem 1]{Theorem 1}, we first show that for the empirical augmented local data distribution $\widehat{D}_{\text {aug }}$, $\mathbb{E}_{\widehat{D}_{\text {aug }}}\left[\varphi_\gamma(X , w)\right]$ concentrates around its population counterpart at the usual rate. Since $-f_{\text {upper }} \leq f(x , w) \leq$ $\varphi_\gamma(X , w) \leq \sup _x f(x , w) \leq f_{\text {upper }}$, the functional $w \rightarrow F_n(w)$ satisfies bounded difference. Substituting standard result on Rademacher complexity \upcite{bartlett2001rademacher} and entropy integrals \upcite{wellner2013weak} gives the result.

\section{Assumption}

\newtheorem{assumption}{Assumption}
\begin{assumption}[Smoothness]	 \label{Assumption 1}
$\forall i \in[N], w_a, w_b \in W, f_i$ is $L$-smooth:
$$
\left\|\nabla f_i\left(w_a\right)-\nabla f_i\left(w_b\right)\right\| \leq L\left\|w_a-w_b\right\|.
$$
\end{assumption}

\begin{assumption}[Bounded variance of global gradient]		 \label{Assumption 2}
$\forall i \in$ $[N], r \in[R]$, the variance between global and local gradient for the global model is bounded by $\sigma_g$ :
$$
\left\|\nabla f_i\left(w^r\right)-\nabla f\left(w^r\right)\right\|^2 \leq \sigma_g^2.
$$
\end{assumption}

\begin{assumption}[Bounded variance of stochastic gradient]		 \label{Assumption 3}
$\forall i \in$ $[N]$, the stochastic gradient $\nabla f_i\left(w, \xi_i\right)$, computed by the $i$-th client of $w$ using mini-batch $\xi_i$, is an unbiased estimator $\nabla f_i(w)$ with variance bounded by $\sigma_l$ :
$$
\mathbb{E}_{\xi_i}\left\|\frac{\nabla f_i\left(w, \xi_i\right)}{\left\|\nabla f_i\left(w, \xi_i\right)\right\|}-\frac{\nabla f_i(w)}{\left\|\nabla f_i(w)\right\|}\right\|^2 \leq \sigma_l^2,
$$
\end{assumption}
\noindent where the expectation is over all local datasets.

\begin{assumption}[Upper bound of stochastic gradient with sparse mask]		\label{Assumption 4}
$\forall i \in[N]$, the arbitrary stochastic gradient with sparse mask $m$ is bounded by $G_s$ :
$$
\mathbb{E}_{\xi_i}\left\|\frac{\nabla f_i\left(w, \xi_i\right)}{\left\|\nabla f_i\left(w, \xi_i\right)\right\|} \odot(1-m)\right\|^2 \leq G_s^2,
$$
\end{assumption}

\noindent where $G_s$ and $m$ depend on hyper-parameters $T_L,T_G$.

\section{Proof of Theorem 2 and 3}   \label{Proof of Theorem 2 and 3}
\noindent \textbf{Description of sUSAM Algorithm and Key Lemmas.} We list flow of sUSAM in \hyperref[Algorithm 2]{Algorithm 2}. For the part client participating circumstance, we randomly choose $S^r \subseteq[N]$ clients with $\left|S^r\right|=S$ in global round $r$ and conduct following updates:
the server first sends the global model for local initialization $w_{i, 0}^r=w^r$. Then each client updates the local model in $K$ local rounds:
$$
\begin{gathered}
\widetilde{w}_{i, k+\frac{1}{2}}^r=w_{i, k-1}^r+\rho \frac{g_{i, k-1}^r}{\left\|g_{i, k-1}^r\right\|} \odot m, \\
w_{i, k}^r=w_{i, k-1}^r+\eta_l \tilde{g}_{i, k+\frac{1}{2}}^r, \\
\Delta_i^r=w_{i, k}^r-w^r, \\
\Delta^r=\frac{1}{N} \sum_{i=1}^N \Delta_i^{r-1}, \\
w^{r+1}=w^r+\eta_g \Delta^r,
\end{gathered}
$$
\noindent where $g_{i, k-1}^r$ is the gradient at the $r$-th global round and $(k-1)$-th local round for client $i$, the fractional part and tilde signal in $\tilde{g}_{i, k+\frac{1}{2}}^r$ denotes 'already perturbed' and 'sparsely perturbed', respectively. $\eta_g$ means the global learning rate.

\newtheorem{leme}{Lemma}[section]	 \label{Lemma 4.1}
\begin{leme}
(Relaxed triangle inequality). Let $\left\{v_1, \ldots, v_n\right\}$ be n vectors in $\mathbb{R}^d$. The following folds: (1) $\forall a>0, \left\| v_i+v_j \right\|^2 \leq(1+a) \left\| v_i \right\|^2+\left(1+\frac{1}{a}\right) \left\| v_j \right\|^2,$ and (2) $\left\|\sum_{i=1}^n v_i\right\|^2 \leq$ $n \sum_{i=1}^n\left\|v_i\right\|^2.$
\end{leme}

\begin{leme}	 \label{Lemma 4.2}
For random variables $x_1, \ldots, x_n$, the following holds:
$$
\mathbb{E}\left[\left\|x_1+\cdots+x_n\right\|^2\right] \leq n \mathbb{E}\left[\left\|x_1\right\|^2+\cdots+\left\|x_n\right\|^2\right].
$$
\end{leme}

\begin{leme}	 \label{Lemma 4.3}
For independent, mean 0 variables $x_1, \ldots, x_n$, the following holds:
$$
\mathbb{E}\left[\left\|x_1+\cdots+x_n\right\|^2\right] \leq \mathbb{E}\left[\left\|x_1\right\|^2+\cdots+\left\|x_n\right\|^2\right].
$$
\end{leme}

\begin{leme}	 \label{Lemma 4.4}
(Separating mean and variance for sUSAM). The accumulated sparse gradient for the $i$-th client at the $k$-th local round and the $r$-th global round in sUSAM is bounded by:
$$
\mathbb{E}\left[\left\|\sum_{k=0}^{K-1} \tilde{g}_{i, k}^r\right\|^2\right] \leq K \sum_{k=0}^{K-1} \mathbb{E}\left[\left\|\tilde{g}_{i, k+\frac{1}{2}}^r\right\|^2\right]+\frac{K L^2 \rho^2}{N}\left(\sigma_l^2+G_s^2\right).
$$
\end{leme}
\noindent \textit{Proof.}
\begin{align*}
& \mathbb{E}\left[\left\|\sum_{k=0}^{K-1} \tilde{g}_{i, k+\frac{1}{2}}^r\right\|^2\right] \\
& \stackrel{(1)}{=}{\mathbb{E}}\left[\left\|\sum_{k=0}^{K-1} \tilde{g}_{i, k+\frac{1}{2}}^r\right\|^2\right] 
+\mathbb{E}\left[\left\|\sum_{k=0}^{K-1}\left(\tilde{g}_{i, k+\frac{1}{2}}^r-\nabla f\left(w_{i, k+\frac{1}{2}}^r\right)\right)\right\|^2\right] \\
& \stackrel{(2)}{\leq} \sum_{k=0}^{K-1} \mathbb{E}\left[\left\|\tilde{g}_{i, k+\frac{1}{2}}^r\right\|^2\right] +L^2 \sum_{k=0}^{K-1} \mathbb{E}\left[\left\|\frac{1}{N} \sum_{i \in[N]}\left(w_{i, k+\frac{1}{2}}^r \right.\right.\right.
\left.\left.\left.-\widehat{w}_{i, k+\frac{1}{2}}^r+ \widetilde{w}_{i, k+\frac{1}{2}}^r-w_{i, k+\frac{1}{2}}^r \right)\right\|^2\right] \\
& \stackrel{(3)}{\leq} K \sum_{k=0}^{K-1} \mathbb{E}\left[\left\|\tilde{g}_{i, k+\frac{1}{2}}^r\right\|^2\right] 
+L^2 K \mathbb{E}\left[\frac { 1 } { N ^ { 2 } } \cdot N \cdot 2 \left(\left\|w_{i, k+\frac{1}{2}}^r-\widehat{w}_{i, k+\frac{1}{2}}^r\right\|^2\right.\right. 
\left.\left.+\left\|\widetilde{w}_{i, k+\frac{1}{2}}^r-w_{i, k+\frac{1}{2}}^r\right\|^2\right)\right] \\
& \stackrel{(4)}{\leq} \sum_{k=0}^{K-1} \mathbb{E}\left[\left\|\tilde{g}_{i, k+\frac{1}{2}}^r\right\|^2\right] +\frac{2 L^2 K}{N} \mathbb{E}\left[\left\| w_{i, k-1}^r \right.\right.
\left. +\delta_{i, k}^r\left(w_{i, k-1}^r , \xi_{i, k-1}^r\right) -w_{i, k-1}^r+\delta_{i, k}^r\left(w_{i, k-1}^r\right)\right\|^2
\left. +\left\| \widetilde{w}_{i, k}^r-w_{i, k}^r \right\|^2\right]\\
& \stackrel{(5)}{\leq} K \sum_{k=0}^{K-1} \mathbb{E}\left[\left\|\tilde{g}_{i, k+\frac{1}{2}}^r\right\|^2\right] 
+\frac{2 L^2 K}{N} \mathbb{E}\left[\left\| \rho \frac{\nabla f_i\left(w_{i, k-1}^r, \xi_i\right)}{\left\|\nabla f_i\left(w_{i, k-1}^r, \xi_i\right)\right\|}\right.\right. 
\left.\left.-\rho \frac{\nabla f_i\left(w_{i, k-1}^r\right)}{\left\|\nabla f_i\left(w_{i, k-1}^r\right)\right\|}\right\|^2+\left\| \widetilde{w}_{i, k}^r-w_{i, k}^r \right\|^2\right]\\
& \stackrel{(6)}{\leq} K \sum_{k=0}^{K-1} \mathbb{E}\left[\left\|\tilde{g}_{i, k+\frac{1}{2}}^r\right\|^2\right]+\frac{2 L^2 K \rho^2 \sigma_l^2}{N} 
+\frac{2 L^2 K}{N} \mathbb{E}\left[\left\| \rho \frac{\nabla f_i\left(w_{i, k-1}^r, \xi_i\right)}{\left\|\nabla f_i\left(w_{i, k-1}^r, \xi_i\right)\right\|} \odot(1 -m) \right\|^2\right]\\
& \stackrel{(7)}{\leq} K \sum_{k=0}^{K-1} \mathbb{E}\left[\left\|\tilde{g}_{i, k+\frac{1}{2}}^r\right\|^2\right]+\frac{2 L^2 K \rho^2}{N}\left(\sigma_l^2+G_s^2\right)
\end{align*}
where $\widehat{w}_{i, k}^r$ denotes unbiased perturbed weight without mask, $w_{i, k}^r$ denotes perturbed weight without mask, $\widetilde{w}_{i, k}^r$ denotes perturbed weight with mask. (1) is from the assumption that the stochastic gradient $\nabla f_i\left(\widetilde{w}, \xi_i\right)$ computed by the $i$-th client of $w$ using mini-batch $\xi_i$ is an unbiased estimator $\nabla f_i(w)$ with variance bounded by $\sigma_l$. (2) is from \hyperref[Assumption 1]{Assumption 1}. (3) is from \hyperref[Lemma 4.1]{Lemma 4.1} with $a=1$ and \hyperref[Lemma 4.2]{Lemma 4.2}. (4) is from the definition of perturbed weight. (5) extends the formulation of perturbation. (6) is similar to (4) and (5). (7) is from \hyperref[Assumption 4]{Assumption 4}.

\begin{leme}	 \label{Lemma 4.5}
(Bounded variance of global perturbed gradient). The variance of local and global gradients with sparse perturbation $\tilde{\delta}$ can be bounded as:
$$
\left\|\nabla f_i\left(\widetilde{w}_{+\frac{1}{2}}\right)-\nabla f\left(\widetilde{w}_{+\frac{1}{2}}\right)\right\|^2 \leq 3 \sigma_g^2+6 L^2 \rho^2.
$$
\end{leme}
\noindent \textit{Proof.}
\begin{align*}
& \left\|\nabla f_i\left(\widetilde{w}_{+\frac{1}{2}}\right)-\nabla f\left(\widetilde{w}_{+\frac{1}{2}}\right)\right\|^2 \\
& \stackrel{(1)}{=}\left\|\nabla f_i\left(w+\tilde{\delta}_i\right)-\nabla f \left(w+\tilde{\delta} \right)\right\|^2 \\
& =\left\|\left(\nabla f_i\left(w+\tilde{\delta}_i\right)-\nabla f_i(w)\right)\right.
\left. +\left(\nabla f_i(w)-\nabla f(w)\right)+\left(\nabla f(w) -\nabla f(w+\tilde{\delta}) \right) \right\|^2 \\
& \stackrel{(2)}{\leq} 3\left\|\nabla f_i\left(w+\tilde{\delta}_i\right)-\nabla f_i(w)\right\|^2 
+3\left\|\nabla f_i(w)-\nabla f(w)\right\|^2 +3\left\|\nabla f(w)-\nabla f(w+\tilde{\delta}) \right\|^2\\
& \stackrel{(3)}{\leq} 3 \sigma_g^2  
+3 L^2{\left\|w+\rho \cdot \operatorname{sign}\left(\nabla f_i(w)\right) \frac{\nabla f_i(w)}{\left\|\nabla f_i(w)\right\|} \odot m-w \right\|}^2
+3 L^2 \left\| w+\rho \cdot \operatorname{sign}\left(\nabla f(w)\right) \frac{\nabla f(w)}{\|\nabla f(w)\|} \odot m-w \right\|^2  \\
& \stackrel{(4)}{\leq} 3 \sigma_g^2+6 L^2 \rho^2 
\end{align*}
(1) is from the definition of perturbed weight. (2) is from \hyperref[Lemma 4.2]{Lemma 4.2}. (3) is from \hyperref[Assumption 1]{Assumption 1}, \hyperref[Assumption 2]{Assumption 2} and formulation of perturbation. (4) is from the fact that $\left\|\frac{\nabla f(w)}{\|\nabla f(w)\|} \odot m\right\|^2<1$.

\begin{leme}	 \label{Lemma 4.6}
(Bounded $\varepsilon_\delta$ of sUSAM). Suppose $\exists \alpha_{g_s}$, s.t. $\frac{\left\|f_i\left(\widetilde{w}_{i, k+\frac{1}{2}}\right) \odot m\right\|^2}{\left\|\nabla f_i\left(w_{i, 0}\right)\right\|^2} \leq \alpha_{g_s},$ and local learning rates satisfy $\eta_l \leq \frac{1}{4 K L} . \quad$ Denote $\delta_{i, k}=\rho \frac{\nabla f_i\left(w_{i, k} \xi_i\right)}{\left\|\nabla f_i\left(w_{i, k}, \xi_i\right)\right\|}, \quad \tilde{\delta}=$ $\rho \frac{\nabla f(w)}{\|\nabla f(w)\|} \odot m, \tilde{\delta}_{i, k}=\rho \frac{\nabla f_i\left(w_{i, k}, \xi_i\right)}{\left\|\nabla f_i\left(w_{i, k}, \xi_i\right)\right\|} \odot m$. Under \hyperref[Assumption 1]{Assumption 1}, \hyperref[Assumption 2]{2}, \hyperref[Assumption 3]{3}, \hyperref[Assumption 4]{4}, the shift of perturbation of sUSAM can be bounded as follows:
$$
\varepsilon_\delta=\frac{1}{N} \sum_i \mathbb{E}\left[\left\|\tilde{\delta}_{i, k}-\tilde{\delta}\right\|^2\right] \leq 2 \rho^2 K^2 L^2 \eta_l^2 \alpha_{g_s}.
$$
\end{leme}
\noindent \textit{Proof.}

When the local learning rate is small, the gradient $\nabla f_i\left(w_{i, k}, \xi_i\right)$ is small too. Based on the first order Hessian approximation, the optimal gradient is:
$$
\begin{gathered}
\nabla f_i\left(w_{i, k}\right)=\nabla f_i\left(w_{i, k-1}-\eta_l \tilde{g}_{i, k+\frac{1}{2}}\right) \\
=\nabla f_i\left(w_{i, k-1}\right)-H \eta_l \tilde{g}_{i, k+\frac{1}{2}}-O\left(\left\|\eta_l \tilde{g}_{i, k+\frac{1}{2}}\right\|^2\right),
\end{gathered}
$$
where $H$ is the Hessian matrix at $w_{i, k-1}$. The shift of perturbation is then:
\begin{align*}
& \varepsilon_{\delta}=\mathbb{E}\left[\left\|\tilde{\delta}_{i, k}-\tilde{\delta}\right\|^2\right] \\
& =\rho^2 \mathbb{E}\left[\left\|\left(\frac{\nabla f_i\left(w_{i, k}\right)}{\left\|\nabla f_i\left(w_{i, k}\right)\right\|}-\frac{\nabla f(w)}{\|\nabla f(w)\|}\right) \odot m\right\|^2\right] \\
& \stackrel{(1)}{\leq} \rho^2 \frac{\left\|\left(\nabla f_i\left(w_{i, k}\right)-\nabla f_i\left(w_{i, 0}\right)\right) \odot m\right\|^2}{\left\|\nabla f_i\left(w_{i, 0}\right)\right\|^2}\\
& \stackrel{(2)}{\leq} \rho^2\left(\left(1+\frac{1}{k-1}\right) \cdot 
\frac{\left\|\left(\nabla f_i\left(w_{i, k-1}\right)-\nabla f_i\left(w_{i, 0}\right)\right) \odot m\right\|^2}{\left\|\nabla f_i\left(w_{i, 0}\right)\right\|^2} 
+K \frac{\left\|\left(H \eta_l \tilde{g}_{i, k+\frac{1}{2}}+\left(\left\|\eta_l \tilde{g}_{i, k+\frac{1}{2}}\right\|^2\right)\right) \odot m\right\|^2}{\left\|\nabla f_i\left(w_{i, 0}\right)\right\|^2}\right) \\
& \stackrel{(3)}{\leq} \rho^2\left(\left(1+\frac{1}{k-1}\right)   \cdot 
\frac{\left\|\left(\nabla f_i\left(w_{i, k-1}\right)-\nabla f_i\left(w_{i, 0}\right)\right) \odot m\right\|^2}{\left\|\nabla f_i\left(w_{i, 0}\right)\right\|^2} +K L^2 \eta_l^2 \alpha_{g_s}\right)\\
& \frac{1}{N} \sum_{i \in[N]} \mathbb{E}\left[\left\|\tilde{\delta}_{i, k}-\tilde{\delta}\right\|^2\right] \\
& \stackrel{(4)}{\leq} \rho^2\left(\sum_{\tau=0}^{k-1}\left(1+\frac{1}{k-1}\right)^\tau K L^2 \eta_l^2 \alpha_{g_s}\right) \\
& \leq 2 \rho^2 K^2 L^2 \eta_l^2 \alpha_{g_s} 
\end{align*}
(1) can be explained through spherical coordinate system (Please refer to FedSAM\upcite{qu2022generalized} for more details). (2) is from the first order Hessian approximation and \hyperref[Lemma 4.1]{Lemma 4.1} with $a=\frac{1}{K-1}$. (3) is due to top eigenvalue of $H$ is bounded by $L$ under \hyperref[Assumption 1]{Assumption 1}. Expanding the recursion between (1) and (3) results in (4).

\begin{leme}	 \label{Lemma 4.7}
(Bounded $\varepsilon_w$ of sUSAM). Suppose local learning rates satisfy $\eta_l \leq \frac{1}{10 K L}$. Under \hyperref[Assumption 1]{Assumption 1}, \hyperref[Assumption 2]{2}, \hyperref[Assumption 3]{3}, \hyperref[Assumption 4]{4}, the shift of model parameters can of sUSAM can be bounded as follows:
$$
\begin{gathered}
\varepsilon_w=\frac{1}{N} \sum_i \mathbb{E}\left[\left\|w_{i, k}-w\right\|^2\right] 
\leq 5 K^2 \eta_l^2\left(2 L^2 \rho^2 \sigma_l^2+6 K\left(3 \sigma_g^2+6 L^2 \rho^2\right)\right. 
\left.+6 K \left\|\nabla f(\widetilde{w}) \right\|^2+24 K^3 {\eta_l}^4 L^4 \rho^2\right).
\end{gathered}
$$
Proof of this Lemma is same as the one in FedSAM\upcite{qu2022generalized}.
\end{leme}

\begin{leme}	 \label{Lemma 4.8}
\begin{align*}
& \left\langle\nabla f\left(\widetilde{w}_{+\frac{1}{2}}^r\right), \mathbb{E}_r\left[\Delta^r\right.\right. \left.\left.+\eta_l K \nabla f\left(\widetilde{w}_{+\frac{1}{2}}^r\right)\right]\right\rangle \\
& \leq \frac{\eta_l K}{2}\left\|\nabla f\left(\widetilde{w}_{+\frac{1}{2}}^r\right)\right\|^2 +K \eta_l L^2 \varepsilon_w
+K \eta_l L^2 \varepsilon_\delta -\frac{\eta_l}{2 K N^2} \mathbb{E}_r\left\|\sum_{i, k} \nabla f_i\left(\widetilde{w}_{i, k+\frac{1}{2}}^r\right)\right\|^2.
\end{align*}
\end{leme}
\noindent \textit{Proof.}
\begin{align*}
& \left\langle \nabla f\left(\widetilde{w}_{+\frac{1}{2}}^r\right), \mathbb{E}_r\left[\Delta^r+\eta_l K \nabla f\left(\widetilde{w}_{+\frac{1}{2}}^r\right)\right] \right\rangle \\
& \stackrel{(1)}{=} \sqrt{K} \frac{1}{\sqrt{K}}\left\langle\nabla f\left(\widetilde{w}_{+\frac{1}{2}}^r\right), \mathbb{E}_r\left[-\frac{\eta_l}{N} \sum_{i, k} \tilde{g}_{i, k-1}^{r-1}\right.\right. 
\left.\left.+\frac{\eta_l}{N} \sum_{i, k} \nabla f_i\left(\widetilde{w}_{+\frac{1}{2}}^r\right)\right] \right\rangle \\
& \stackrel{(2)}{=} \frac{\eta_l K}{2}\left\|\nabla f\left(\widetilde{w}_{+\frac{1}{2}}^r\right)\right\|^2 
+\frac{\eta_l}{2 K N^2} \mathbb{E}_r \left\| \sum_{i, k}\left(\nabla f_i\left(\widetilde{w}_{i, k+\frac{1}{2}}^r\right)\right. \left.-\nabla f_i\left(\widetilde{w}_{+\frac{1}{2}}^r\right)\right) \right\|^2 
-\frac{\eta_l}{2 K N^2} \mathbb{E}_r\left\|\sum_{i, k} \nabla f_i\left(\widetilde{w}_{i, k+\frac{1}{2}}^r\right)\right\|^2 \\
& \stackrel{(3)}{\leq} \frac{\eta_l K}{2}\left\|\nabla f\left(\widetilde{w}_{+\frac{1}{2}}^r\right)\right\|^2 
+\frac{\eta_l}{2 N} \sum_{i, k} \mathbb{E}_r\left\|\nabla f_i\left(\widetilde{w}_{i, k+\frac{1}{2}}^r\right)-\nabla f_i\left(\widetilde{w}_{+\frac{1}{2}}^r\right)\right\|^2 
-\frac{\eta_l}{2 K N^2} \mathbb{E}_r\left\|\sum_{i, k} \nabla f_i\left(\widetilde{w}_{i, k+\frac{1}{2}}^r\right)\right\|^2 \\
& \stackrel{(4)}{\leq} \frac{\eta_l K}{2}\left\|\nabla f\left(\widetilde{w}_{+\frac{1}{2}}^r\right)\right\|^2 
+\frac{\eta_l L^2}{2 N} \sum_{i, k} \mathbb{E}_r\left\|\widetilde{w}_{i, k+\frac{1}{2}}^r-\widetilde{w}_{+\frac{1}{2}}^r\right\|^2 
-\frac{\eta_l}{2 K N^2} \mathbb{E}_r\left\|\sum_{i, k} \nabla f_i\left(\widetilde{w}_{i, k+\frac{1}{2}}^r\right)\right\|^2 \\
& \stackrel{(5)}{\leq} \frac{\eta_l K}{2}\left\|\nabla f\left(\widetilde{w}_{+\frac{1}{2}}^r\right)\right\|^2+\frac{\eta_l L^2}{N} \sum_{i, k} \mathbb{E}_r\left\|w_{i, k}^r-w^r\right\|^2 
+\frac{\eta_l L^2}{N} \sum_{i, k} \mathbb{E}_r\left\|\tilde{\delta}_{i, k}^r-\tilde{\delta}^r\right\|^2 
-\frac{\eta_l}{2 K N^2} \mathbb{E}_r\left\|\sum_{i, k} \nabla f_i\left(\widetilde{w}_{i, k+\frac{1}{2}}^r\right)\right\|^2 \\
& \stackrel{(6)}{\leq} \frac{\eta_l K}{2}\left\|\nabla f\left(\widetilde{w}_{+\frac{1}{2}}^r\right)\right\|^2+K \eta_l L^2 \varepsilon_w+K \eta_l L^2 \varepsilon_\delta 
-\frac{\eta_l}{2 K N^2} \mathbb{E}_r\left\|\sum_{i, k} \nabla f_i\left(\widetilde{w}_{i, k+\frac{1}{2}}^r\right)\right\|^2
\end{align*}
(1) can be derived from:
\begin{align*}
& \eta_l K \nabla f\left(\widetilde{w}_{+\frac{1}{2}}^r\right) \\
& =\eta_l K \frac{1}{N} \sum_{i \in[N]} \nabla f_i\left(\widetilde{w}_{+\frac{1}{2}}^r\right) \\
& =\eta_l K \frac{1}{N} \sum_{i \in[N]} \frac{1}{K} \sum_{k \in[K]} \nabla f_i\left(\widetilde{w}_{+\frac{1}{2}}^r\right) \\
& =\frac{\eta_l}{N} \sum_{i, k} \nabla f_i\left(\widetilde{w}_{+\frac{1}{2}}^r\right) \\
& \Delta^r=\frac{1}{N} \sum_{i=1}^N \Delta_i^{r-1} \\
& =\frac{1}{N} \sum_{i=1}^N\left(w_{i, 0}^{r-1}-\sum_k \eta_l \tilde{g}_{i, k-1}^{r-1}-w_i^{r-1}\right) \\
& =-\frac{\eta_l}{N} \sum_{i, k} \tilde{g}_{i, k-1}^{r-1}
\end{align*}
(2) is from the lemma that $\langle a, b\rangle=\frac{1}{2}\left(\|a\|^2+\|b\|^2-\| a-b \|^2\right) \quad$ with $\quad a=\sqrt{\eta_l K} \nabla f\left(\widetilde{w}_{+\frac{1}{2}}^r\right) \quad$ and $\quad b=-\frac{\sqrt{\eta_l}}{N \sqrt{K}} \sum_{i, k}\left(\nabla f_i\left(\widetilde{w}_{i, k+\frac{1}{2}}^r\right)-\nabla f_i\left(\widetilde{w}_{+\frac{1}{2}}^r\right)\right)$. (3) is from \hyperref[Lemma 4.2]{Lemma 4.2}. (4) is from \hyperref[Assumption 1]{Assumption 1}. (5) is from definition of sparse perturbation and \hyperref[Lemma 4.2]{Lemma 4.2}. (6) is from \hyperref[Lemma 4.6]{Lemma 4.6} and \hyperref[Lemma 4.7]{Lemma 4.7}.

\begin{leme}	 \label{Lemma 4.9}
For the full client participating scheme, the bound of $\mathbb{E}_r\left[\left\|\Delta^r\right\|^2\right]$ is:
$$
\begin{gathered}
\mathbb{E}_r\left[\left\|\Delta^r\right\|^2\right] \leq \frac{2 K \eta_l^2 L^2 \rho^2}{N}\left(\sigma_l^2+G_s^2\right) 
+\frac{\eta_l^2}{N^2}\left\|\sum_{i, k} \nabla f_i\left(\widetilde{w}_{i, k+\frac{1}{2}}^r\right)\right\|^2.
\end{gathered}
$$
\end{leme}
\noindent \textit{Proof.}
\begin{align*}
& \mathbb{E}_r\left[\left\|\Delta^r\right\|^2\right] \\
& \stackrel{(1)}{\leq} \frac{\eta_l^2}{N^2} \mathbb{E}_r\left\|\sum_{i, k} \tilde{g}_{i, k+\frac{1}{2}}^r\right\|^2 \\
& \stackrel{(2)}{=} \frac{\eta_l^2}{N^2} \mathbb{E}_r\left\|\sum_{i, k}\left(\tilde{g}_{i, k+\frac{1}{2}}^r-\nabla f_i\left(\widetilde{w}_{i, k+\frac{1}{2}}^r\right)\right)\right\|^2 
+\frac{\eta_l^2}{N^2} \mathbb{E}_r\left\|\sum_{i, k} \nabla f_i\left(\widetilde{w}_{i, k+\frac{1}{2}}^r\right)\right\|^2 \\
& \stackrel{(3)}{\leq} \frac{2 K \eta_l^2 L^2 \rho^2}{N}\left(\sigma_l^2+G_s^2\right)+\frac{\eta_l^2}{N^2} \mathbb{E}_r\left\|\sum_{i, k} \nabla f_i\left(\widetilde{w}_{i, k+\frac{1}{2}}^r\right)\right\|^2 
\end{align*}
(1) is from \hyperref[Lemma 4.2]{Lemma 4.2} and the fact that the optimal gradient is smaller than the empiral one. (2) is from \hyperref[Lemma 4.3]{Lemma 4.3} since the gradient is merely within one local epoch $k$ and the mean for $r$ is 0 in this situation. (3) is scaled from the second term in \hyperref[Lemma 4.4]{Lemma 4.4.(1)}.

\begin{leme}	 \label{Lemma 4.10}
For all $r \in[R-1]$, with proper choices of local and global learning rates, the iterates from sUSAM satisfy:
\begin{align*}
& \mathbb{E}_r\left[f\left(\widetilde{w}_{+\frac{1}{2}}^{r+1}\right)\right] \leq f\left(\widetilde{w}_{+\frac{1}{2}}^r\right)-K \eta_g \eta_L\left(\frac{1}{2}
-30 K^2 L^2 \eta_L^2\right)\left\|\nabla f\left(\widetilde{w}_{+\frac{1}{2}}^r\right)\right\|^2 
+K \eta_g \eta_L \left(10 K L^4 \eta_L^2 \rho^2 \sigma_l^2 \right.\\
& +90 K^2 L^2 \eta_L^2 \sigma_g^2+180 K^2 L^4 \eta_L^2 \rho^2 +24 K^4 L^6 \eta_L^6 \rho^2 \alpha_{g_s} 
\left.+\frac{\eta_g \eta_L L^3 \rho^2}{N}\left(\sigma_l^2+G_s^2\right)\right).
\end{align*}
\end{leme}
\noindent \textit{Proof.}
\begin{align*}
& \mathbb{E}_r\left[f\left(\widetilde{w}_{+\frac{1}{2}}^{r+1}\right)\right] \\
& \stackrel{(1)}{\leq} f\left(\widetilde{w}_{+\frac{1}{2}}^r\right)
+\mathbb{E}_r\left\langle\nabla f\left(\widetilde{w}_{+\frac{1}{2}}^r\right), \widetilde{w}_{+\frac{1}{2}}^{r+1}-\widetilde{w}_{+\frac{1}{2}}^r\right\rangle 
+\frac{L}{2} \mathbb{E}_r\left\|\widetilde{w}_{+\frac{1}{2}}^{r+1}-\widetilde{w}_{+\frac{1}{2}}^r\right\|^2 \\
& \stackrel{(2)}{=} f\left(\widetilde{w}_{+\frac{1}{2}}^r\right)+\mathbb{E}_r\left\langle\nabla f\left(\widetilde{w}_{+\frac{1}{2}}^r\right),-\eta_g \Delta^r \right. 
+K \eta_g \eta_L \nabla f\left(\widetilde{w}_{+\frac{1}{2}}^r\right) \left.-K \eta_g \eta_L \nabla f\left(\widetilde{w}_{+\frac{1}{2}}^r\right)\right\rangle 
+\frac{L \eta_g^2}{2} \mathbb{E}_r\left\|\Delta^r\right\|^2 \\
& \stackrel{(3)}{=} f\left(\widetilde{w}_{+\frac{1}{2}}^r\right)-K \eta_g \eta_L\left\|\nabla f\left(\widetilde{w}^r_{+\frac{1}{2}} \right)\right\|^2 
+\eta_g\left\langle\nabla f\left(\widetilde{w}^r_{+\frac{1}{2}} \right), \mathbb{E}_r\left[-\Delta^r\right.\right. 
\left.\left.+K \eta_L \nabla f\left(\widetilde{w}_{+\frac{1}{2}}^r\right)\right]\right\rangle \\
& +\frac{L \eta_g^2}{2} \mathbb{E}_r\left\|\Delta^r\right\|^2 \\
& \stackrel{(4)}{\leq} f\left(\widetilde{w}_{+\frac{1}{2}}^r\right)-\frac{K \eta_g \eta_L}{2}\left\|\nabla f\left(\widetilde{w}_{+\frac{1}{2}}^r\right)\right\|^2 
+K \eta_g \eta_L L^2 \varepsilon_w +K \eta_g \eta_L L^2 \varepsilon_\delta 
-\frac{\eta_g \eta_L}{2 K N} \mathbb{E}_r\left\|\sum_{i, k} \nabla f_i\left(\widetilde{w}_{i, k+\frac{1}{2}}^r\right)\right\|^2 \\
& +\frac{L}{2} \eta_g^2 \mathbb{E}_r\left\|\Delta^r\right\|^2 \\
& \stackrel{(5)}{\leq} f\left(\widetilde{w}_{+\frac{1}{2}}^r\right)-\frac{K \eta_g \eta_L}{2}\left\|\nabla f\left(\widetilde{w}_{+\frac{1}{2}}^r\right)\right\|^2 
+K \eta_g \eta_L L^2 \varepsilon_w +K \eta_g \eta_L L^2 \varepsilon_\delta 
+\frac{K \eta_g^2 \eta_l^2 L^3 \rho^2}{N}\left(\sigma_l^2+G_s^2\right) \\
& \stackrel{(6)}{\leq} f\left(\widetilde{w}_{+\frac{1}{2}}^r\right)
-K \eta_g \eta_L\left(\frac{1}{2}-30 K^2 L^2 \eta_L^2\right)\left\|\nabla f\left(\widetilde{w}_{+\frac{1}{2}}^r\right)\right\|^2 
+K \eta_g \eta_L\left(10 K L^4 \eta_L^2 \rho^2 \sigma_l^2+90 K^2 L^2 \eta_L^2 \sigma_g^2\right. \\
& +180 K^2 L^4 \eta_L^2 \rho^2+24 K^4 L^6 \eta_L^6 \rho^2 \alpha_{g_s} 
\left.+\frac{\eta_g \eta_L L^3 \rho^2}{N}\left(\sigma_l^2+G_s^2\right)\right) 
\end{align*}
(1) is from the lemma of L-smooth that $f(a) \leq f(b)+$ $\nabla f(b)(a-b)+\frac{L}{2}\|a-b\|^2$. (2) is from the description of sUSAM. (3) is from the unbiased estimators. (4) is from \hyperref[Lemma 4.3]{Lemma 4.3}. (5) is from \hyperref[Lemma 4.9]{Lemma 4.9} and neglection of the negative term. (6) is from \hyperref[Lemma 4.6]{Lemma 4.6}, \hyperref[Lemma 4.7]{Lemma 4.7} and the assumption that $\eta_l \leq \frac{1}{10 K L}$.

\begin{theo}	  \label{Theorem 2}
Let local and global learning rates be set as $\eta_l \leq$ $\frac{1}{10 K L}, \eta_l \eta_G \leq\frac{1}{K L}$. Under \hyperref[Assumption 1]{Assumption 1}, \hyperref[Assumption 2]{2}, \hyperref[Assumption 3]{3}, \hyperref[Assumption 4]{4} and full client participation, the sequence of iterates generated by sUSAM satisfies:
\begin{align*}
& \quad \min _{r \in[R]} \mathbb{E}\left[ \left\|\nabla f\left(w^r\right) \right\|^2 \right] \leq \frac{f^0-f^*}{C K \eta_g \eta_L}+\Phi~, 
\end{align*}
where $ \quad \Phi=\frac{1}{C}\left(10 K L^4 \eta_L^2 \rho^2 \sigma_l^2+90 K^2 L^2 \eta_L^2 \sigma_g^2\right. 
+180 K^2 L^4 \eta_L^2 \rho^2+24 K^4 L^6 \eta_L^6 \rho^2 \alpha_{g_s}
+16 K^3 L^6 \eta_L^4 \rho^2 \alpha_{g_s}+\left.\frac{\eta_g \eta_L L^3 \rho^2}{N}\left(\sigma_l^2+G_s^2\right)\right)$. If local and global learning rates are chosen as $\eta_l=\frac{1}{\sqrt{R} K L}$ and $\eta_G=\sqrt{K N}$, and perturbation radius is chosen as $\rho=$ $\frac{1}{\sqrt{R}}$, the convergence rates can be expressed as:
\begin{align*}
& \frac{1}{R} \sum_{r=1}^R \mathbb{E}\left[ \left\| f\left(w^{r+1}\right) \right\|^2\right] =O\left(\frac{F L}{\sqrt{R K N}}+\frac{\sigma_g^2}{R}\right. 
+\frac{L^2\left(\sigma_l^2+G_s^2\right)}{R^{\frac{3}{2}} \sqrt{K N}}+\frac{L^2}{R^2}\left.+\frac{L^2 \sigma_l^2}{R^2 K}\right).
\end{align*}
\end{theo}
\noindent \textit{Proof.}
\begin{align*}
& \frac{1}{R} \sum_{r=1}^R \mathbb{E}\left[\left\|f\left(w^{r+1}\right)\right\|^2\right] \\
& \stackrel{(1)}{=} \frac{1}{R} \sum_{r=1}^R \mathbb{E}\left[\left\|f\left(\widetilde{w}_{+\frac{1}{2}}^{r+1}\right)\right\|^2\right]\\
& \stackrel{(2)}{\leq} \frac{f\left(\widetilde{w}_{+\frac{1}{2}}^r\right)-f\left(\widetilde{w}_{+\frac{1}{2}}^{r+1}\right)}{C K \eta_g \eta_L R}+\frac{1}{C}\left(10 K L^4 \eta_L^2 \rho^2 \sigma_l^2
+90 K^2 L^2 \eta_L^2 \sigma_g^2+180 K^2 L^4 \eta_L^2 \rho^2 +24 K^4 L^6 \eta_L^6 \rho^2 \alpha_{g_s}\right.  \\
& +16 K^3 L^6 \eta_L^4 \rho^2 \alpha_{g_s} \left.+\frac{\eta_g \eta_L L^3 \rho^2}{N}\left(\sigma_l^2+G_s^2\right)\right) \\
& \stackrel{(3)}{\leq} \frac{f\left(\widetilde{w}_{+\frac{1}{2}}^0\right)-f^*}{C K \eta_g \eta_L R} 
+\frac{1}{C}\left(10 K L^4 \eta_L^2 \rho^2 \sigma_l^2+90 K^2 L^2 \eta_L^2 \sigma_g^2\right. 
+180 K^2 L^4 \eta_L^2 \rho^2+24 K^4 L^6 \eta_L^6 \rho^2 \alpha_{g_s} \\
& +16 K^3 L^6 \eta_L^4 \rho^2 \alpha_{g_s} \left.+\frac{\eta_g \eta_L L^3 \rho^2}{N}\left(\sigma_l^2+G_s^2\right)\right)
\end{align*}
(1) is due to \hyperref[Assumption 1]{Assumption 1} and taking expectation of $f\left(\widetilde{w}_{+\frac{1}{2}}^{r+1}\right)$ over randomness at global round $r$. (2) is from transposition of \hyperref[Lemma 4.10]{Lemma 4.10}, summing it for $r=[R]$, and multiplying both sides by $\frac{1}{C K \eta_g \eta_L R}$ with $0<C<\frac{1}{2}-$ $30 K^2 L^2 \eta_L^2$ if $\eta_L<\frac{1}{\sqrt{30} K L}$. (3) is from the relationship between current loss and optimal loss $f\left(\widetilde{w}_{+\frac{1}{2}}^{r+1}\right) \geq$ $f^*$, and the one between current loss and initial loss $f\left(\widetilde{w}_{+\frac{1}{2}}^0\right) \geq f\left(\widetilde{w}_{+\frac{1}{2}}^r\right)$. If we choose local and global learning rates as $\eta_l=\frac{1}{\sqrt{R} K L}$ and $\eta_G=\sqrt{K N}$, and choose perturbation radius as $\rho=\frac{1}{\sqrt{R}}$, we have:
\begin{align*}
& \frac{1}{R} \sum_{r=1}^R \mathbb{E}\left[\| f\left(w^{r+1}\right) \|^2\right] =O\left(\frac{F L}{\sqrt{R K N}}+\right. \frac{\sigma_g^2}{R}
+\frac{L^2\left(\sigma_l^2+G_s^2\right)}{R^{\frac{3}{2}} \sqrt{K N}}+\frac{L^2}{R^2} \left.+\frac{L^2 \sigma_l^2}{R^2 K}+\frac{L^2 \alpha_{g_s}}{R^3 K}+\frac{\alpha_{g_s}}{R^4 K^2}\right).
\end{align*}
After neglecting high order terms, we have:
$$
\begin{gathered}
\frac{1}{R} \sum_{r=1}^R \mathbb{E}\left[\left\|f\left(w^{r+1}\right)\right\|^2\right]=O\left(\frac{F L}{\sqrt{R K N}}+\frac{\sigma_g^2}{R}\right. 
\left.+\frac{L^2\left(\sigma_l^2+G_s^2\right)}{R^{\frac{3}{2}} \sqrt{K N}}+\frac{L^2}{R^2}+\frac{L^2 \sigma_l^2}{R^2 K}\right).
\end{gathered}
$$

\begin{leme}	 \label{Lemma 4.11}
For the circumstance of partial client participation without replacement, the upper bound of $\mathbb{E}_r\left[\left\|\Delta^r\right\|^2\right]$ is:
\begin{align*}
\mathbb{E}_r\left[\left\|\Delta^r\right\|^2\right] 
\leq \frac{2 K \eta_l^2 L^2 \rho^2}{N}\left(\sigma_l^2+G_S^2\right) 
+\frac{\eta_l^2}{N S} \sum_i\left\|\sum_{j=1}^{K-1} \nabla f_i\left(\widetilde{w}_{i, j+\frac{1}{2}}^r\right)\right\|^2 
+\frac{(S-1) \eta_L^2}{S N^2}\left\|\sum_{j=0}^{K-1} \nabla f_i\left(\widetilde{w}_{i, j+\frac{1}{2}}^r\right)\right\|^2.
\end{align*}
\end{leme}
\noindent \textit{Proof.}
\begin{align*}
& \mathbb{E}_r\left[\left\|\Delta^r\right\|^2\right] \\
& \stackrel{(1)}{\leq} \frac{\eta_l^2}{S^2} \mathbb{E}_r\left[\left\|\sum_{i \in S^r} \sum_k \tilde{g}_{i, k+\frac{1}{2}}\right\|^2\right] \\
& =\frac{\eta_l^2}{S^2} \mathbb{E}_r\left[\left\|\sum_i \mathbb{1}\left\{i \in S^r\right\} \sum_k \tilde{g}_{i, k+\frac{1}{2}}\right\|^2\right] \\
& \stackrel{(2)}{\leq} \frac{\eta_l^2}{N S} \mathbb{E}_r\left[\left\|\sum_i \sum_{j=0}^{K-1}\left(\tilde{g}_{i, j+\frac{1}{2}}-\nabla f_i\left(\widetilde{w}_{i, j+\frac{1}{2}}^r\right)\right)\right\|^2\right] 
+\frac{\eta_l^2}{S^2} \mathbb{E}_r\left[\left\| \sum_i \mathbb{I}\{i\right. \left.\left.\in S^r\right\} \sum_{j=0}^{K-1} \nabla f_i\left(\widetilde{w}_{i, j+\frac{1}{2}}^r\right) \right\|^2\right] \\
& \stackrel{(3)}{\leq} \frac{2 K \eta_l^2 L^2 \rho^2}{N}\left(\sigma_l^2+G_s^2\right) 
+\frac{\eta_l^2}{S^2} \mathbb{E}_r\left[\left\|\sum_{i=1}^s \sum_{j=0}^{K-1} \nabla f_i\left(\widetilde{w}_{i, j+\frac{1}{2}}^r\right)\right\|^2\right] \\
& \stackrel{(4)}{\leq} \frac{2 K \eta_l^2 L^2 \rho^2}{N}\left(\sigma_l^2+G_s^2\right) 
+\frac{\eta_l^2}{N S} \sum_i\left\|\sum_{j=1}^{K-1} \nabla f_i\left(\widetilde{w}_{i, j+\frac{1}{2}}^r\right)\right\| ^2  
+\frac{(S-1) \eta_L^2}{S N^2}\left\|\sum_{j=0}^{K-1} \nabla f_i\left(\widetilde{w}_{i, j+\frac{1}{2}}^r\right)\right\|^2
\end{align*}
(1) is from \hyperref[Lemma 4.2]{Lemma 4.2}. (2) is from \hyperref[Lemma 4.3]{Lemma 4.3}. (3) is from \hyperref[Lemma 4.4]{Lemma 4.4}. (4) is from Yang's proof \upcite{yang2021achieving} in 'For strategy 1' of their Theorem 2.

\begin{leme}	 \label{Lemma 4.12}
If we choose all $k \in[K]$ and $i \in[N]$ for sUSAM, the upper bound of $\sum_i \mathbb{E}\left[\left\|\sum_k \nabla f_i\left(\widetilde{w}_{i, k+\frac{1}{2}}\right)\right\|^2\right]$ is then:
\begin{align*}
\sum_i \mathbb{E}\left[\left\|\sum_k \nabla f_i\left(\widetilde{w}_{i, k+\frac{1}{2}}\right)\right\|^2\right] 
\leq 30 N K^2 L^2 \eta_L^2 \left(2 L^2 \rho^2 \sigma_l^2+6 K\left(3 \sigma_g^2+6 L^2 \rho^2\right)
+6 K\left\|\nabla f\left(\widetilde{w}_{+\frac{1}{2}}\right)\right\|^2\right. \\
+144 K^4 L^6 \eta_L^4 \rho^2 +12 N K^4 L^2 \eta_L^2 \rho^2 \alpha_{g_s} 
\left. +3 N K^2\left(3 \sigma_g^2+6 L^2 \rho^2\right) +3 N K^2\left\|\nabla f\left(\widetilde{w}_{+\frac{1}{2}}\right)\right\|^2\right).
\end{align*}
\end{leme}
\noindent \textit{Proof.}
\begin{align*}
& \sum_i \mathbb{E}\left[\left\|\sum_k \nabla f_i\left(\widetilde{w}_{i, k+\frac{1}{2}}\right)\right\|^2\right] \\
& =\sum_i \mathbb{E} \left[\left\| \sum_k \nabla f_i\left(\widetilde{w}_{i, k+\frac{1}{2}}\right)-\nabla f_i\left(\widetilde{w}_{+\frac{1}{2}}\right) 
+\nabla f_i\left(\widetilde{w}_{i, k+\frac{1}{2}}\right)-\nabla f\left(\widetilde{w}_{+\frac{1}{2}}\right) +\nabla f\left(\widetilde{w}_{+\frac{1}{2}}\right) \right\|^2\right] \\
& \stackrel{(1)}{\leq} 6KL^2 \sum_{i, k} \mathbb{E}\left[\left\|w_{i, k}-w\right\|^2\right] 
+6 K L^2 \sum_{i, k} \mathbb{E}\left[\left\|\tilde{\delta}_{i, k}-\tilde{\delta}\right\|^2\right]+3 N K^2\left(3 \sigma_g^2+6 L^2 \rho^2\right) 
+3 N K^2\left\|\nabla f\left(\widetilde{w}_{+\frac{1}{2}}\right)\right\|^2 \\
& \stackrel{(2)}{\leq} 30 N K^2 L^2 \eta_L^2 \left(2 L^2 \rho^2 \sigma_l^2+6 K\left(3 \sigma_g^2+6 L^2 \rho^2\right)\right.
+6 K\left\|\nabla f\left(\widetilde{w}_{+\frac{1}{2}}\right)\right\|^2 +144 K^4 L^6 \eta_L^4 \rho^2 +12 N K^4 L^2 \eta_L^2 \rho^2 \alpha_{g_s} \\
& \left. +3 N K^2\left(3 \sigma_g^2+6 L^2 \rho^2\right) +3 N K^2\left\|\nabla f\left(\widetilde{w}_{+\frac{1}{2}}\right)\right\|^2\right)
\end{align*}
(1) is from \hyperref[Lemma 4.2]{Lemma 4.2}, \hyperref[Assumption 1]{Assumption 1}, definition of perturbed model parameter, and \hyperref[Lemma 4.5]{Lemma 4.5}. (2) is from \hyperref[Lemma 4.6]{Lemma 4.6} and \hyperref[Lemma 4.7]{Lemma 4.7}.

\begin{theo}	  \label{Theorem 3}
Let local and global learning rates be set as $\eta_l \leq$ $\frac{1}{10 K L}, \eta_l \eta_G \leq\frac{1}{K L}$. Under \hyperref[Assumption 1]{Assumption 1}, \hyperref[Assumption 2]{2}, \hyperref[Assumption 3]{3}, \hyperref[Assumption 4]{4} and part client participation, the sequence of iterates generated by sUSAM satisfies:
\begin{align*}
\min _{r \in[R]} \mathbb{E}\left[\left\|\nabla f\left(w^r\right)\right\|^2\right] \leq \frac{f^0-f^*}{C K \eta_g \eta_L}+\Phi~,
\end{align*}
where $\Phi=\frac{1}{C}\left(10 K L^4 \eta_L^2 \rho^2 \sigma_l^2+90 K^2 L^2 \eta_L^2 \sigma_g^2\right. 
+180 K^2 L^4 \eta_L^2 \rho^2+24 K^4 L^6 \eta_L^6 \rho^2+2 K^2 L^4 \eta_L^2 \rho^2 \alpha_{g_s} 
+16 K^3 L^6 \eta_L^4 \rho^2+\frac{\eta_g \eta_L L^3 \rho^2}{2 S}\left(\sigma_l^2+G_s^2\right) 
+\frac{\eta_g \eta_L}{S}\left(30 K L^5 \eta_L^2 \rho^2 \sigma_l^2+270 K^2 L^3 \eta_L^2 \sigma_g^2\right.
+540 K^2 L^5 \eta_L^2 \rho^2+72 K^3 L^7 \eta_L^4 \rho^2 
\left.\left.+6 K^3 L^3 \eta_L^2 \rho^2 \alpha_{g_s}+\frac{9}{2} K L \sigma_g^2+9 K L^3 \rho^2\right)\right)$. 
If local and global learning rates are chosen as $\eta_l=\frac{1}{\sqrt{R} K L}$ and $\eta_G=$ $\sqrt{K S}$, and perturbation radius is chosen as $\rho=\frac{1}{\sqrt{R}}$, the convergence rate can be expressed as:
\begin{align*}
& \frac{1}{R} \sum_{r=1}^R \mathbb{E}\left[ \left\| f\left(w^{r+1}\right) \right\|^2\right] =O\left(\frac{F L}{\sqrt{R K S}}+\frac{\sqrt{K} \sigma_g^2}{\sqrt{R S}}
+\frac{\sigma_g^2}{R} +\frac{L^2\left(\sigma_l^2+G_s^2\right)+\sigma_g^2}{R^{\frac{3}{2}} \sqrt{K S}}+\frac{\sqrt{K} L^2}{R^{\frac{3}{2}}}
+\frac{L^2}{R^2}\left(1 +\alpha_{g_s}\right)\right).
\end{align*}
\end{theo}
\noindent \textit{Proof.}
\begin{align*}
& \mathbb{E}\left\|f\left(\widetilde{w}_{+\frac{1}{2}}^{r+1}\right)\right\| \\
& \stackrel{(1)}{\leq} f\left(\widetilde{w}^r_{+\frac{1}{2}}\right)-\frac{K \eta_g \eta_L}{2}\left\|\nabla f\left(\widetilde{w}^r_{+\frac{1}{2}}\right)\right\|^2 
+K \eta_g \eta_L L^2 \varepsilon_w+K \eta_g \eta_L L^2 \varepsilon_\delta 
-\frac{K \eta_g \eta_L}{2} \mathbb{E}_r\left\|\sum_{i, k} \nabla f_i\left(\widetilde{w}_{i, k+\frac{1}{2}}^r\right)\right\|^2 \\
& +\frac{L}{2} \eta_g^2 \mathbb{E}_r\left[\left\|\Delta^r\right\|^2\right]\\
& \stackrel{(2)}{\leq} f\left(\widetilde{w}_{+\frac{1}{2}}^r\right)-\frac{K \eta_g \eta_L}{2}\left\|\nabla f\left(\widetilde{w}_{+\frac{1}{2}}^r\right)\right\|^2 
+K \eta_g \eta_L L^2 \varepsilon_w+K \eta_g \eta_L L^2 \varepsilon_\delta 
-\frac{K \eta_g \eta_L}{2} \mathbb{E}_r\left\|\sum_{i, k} \nabla f_i\left(\widetilde{w}_{i, k+\frac{1}{2}}^r\right)\right\|^2 \\
& +\frac{K \eta_g^2 \eta_l^2 L^3 \rho^2}{2 S}\left(\sigma_l^2+G_s^2\right) 
+\frac{\eta_g^2 L S}{2 N} \sum_i\left\|\sum_{j=1}^{K-1} \nabla f_i\left(\widetilde{w}_{i, j+\frac{1}{2}}^r\right)\right\|^2 
+\frac{\eta_g^2 L S(S-1)}{2 N^2}\left\|\sum_{j=0}^{K-1} \nabla f_i\left(\widetilde{w}_{i, j+\frac{1}{2}}^r\right)\right\|^2 \\
& \stackrel{(3)}{\leq} f\left(\widetilde{w}_{+\frac{1}{2}}^r\right)-\frac{K \eta_g \eta_L}{2}\left\|\nabla f\left(\widetilde{w}_{+\frac{1}{2}}^r\right)\right\|^2 
+K \eta_g \eta_L L^2 \varepsilon_w+K \eta_g \eta_L L^2 \varepsilon_\delta 
+\frac{K \eta_g^2 \eta_l^2 L^3 \rho^2}{2 S}\left(\sigma_l^2+G_s^2\right)\\
& +\frac{L \eta_g^2 \eta_l^2}{2 N S} \sum_i\left\|\sum_k \nabla f_i\left(\widetilde{w}_{i, k+\frac{1}{2}}^r\right)\right\|^2 \\
& \stackrel{(4)}{\leq} f\left(\widetilde{w}_{+\frac{1}{2}}^r\right)-K \eta_g \eta_L\left(\frac{1}{2}-30 K^2 L^2 \eta_L^2
-\frac{K \eta_g \eta_L}{2 S}\left(3 K +180 K^2 L^4 \eta_L^2 \rho^2\right)\right)\left\|\nabla f\left(\widetilde{w}^r_{+\frac{1}{2}}\right)\right\|^2 \\
& +K \eta_g \eta_L\left(10 K L^4 \eta_L^2 \rho^2 \sigma_l^2+90 K^2 L^2 \eta_L^2 \sigma_g^2\right. 
+180 K^2 L^4 \eta_L^2 \rho^2+24 K^4 L^6 \eta_L^6 \rho^2 +2 K^2 L^4 \eta_L^2 \rho^2 \alpha_{g_s}+16 K^3 L^6 \eta_L^4 \rho^2 \\
& \left.+\frac{\eta_g \eta_L L^3 \rho^2}{2 S}\left(\sigma_l^2+G_s^2\right)\right) 
+\frac{K \eta_g^2 \eta_L^2}{S}\left(30 K L^5 \eta_L^2 \rho^2 \sigma_l^2 +270 K^2 L^3 \eta_L^2 \sigma_g^2
+540 K^2 L^5 \eta_L^2 \rho^2 +72 K^3 L^7 \eta_L^4 \rho^2 \right.\\
& \left.\left.+6 K^3 L^3 \eta_L^2 \rho^2 \alpha_{g_s}+\frac{9}{2} K L \sigma_g^2+9 K L^3 \rho^2\right)\right) \\
& \stackrel{(5)}{\leq} f\left(\widetilde{w}_{+\frac{1}{2}}^r\right)-C K \eta_g \eta_L\left\|\nabla f\left(\widetilde{w}_{+\frac{1}{r}}^r\right)\right\|^2 
+K \eta_g \eta_L\left(10 K L^4 \eta_L^2 \rho^2 \sigma_l^2+90 K^2 L^2 \eta_L^2 \sigma_g^2+180 K^2 L^4 \eta_L^2 \rho^2 \right. \\
& +24 K^4 L^6 \eta_L^6 \rho^2 +2 K^2 L^4 \eta_L^2 \rho^2 \alpha_{g_s}
\left.+16 K^3 L^6 \eta_L^4 \rho^2 +\frac{\eta_g \eta_L L^3 \rho^2}{2 S}\left(\sigma_l^2+G_s^2\right)\right) 
+\frac{K \eta_g^2 \eta_L^2}{S}\left(30 K L^5 \eta_L^2 \rho^2 \sigma_l^2\right. \\
& \left.+270 K^2 L^3 \eta_L^2 \sigma_g^2 +540 K^2 L^5 \eta_L^2 \rho^2 +72 K^3 L^7 \eta_L^4 \rho^2
+6 K^3 L^3 \eta_L^2 \rho^2 \alpha_{g_s} +\frac{9}{2} K L \sigma_g^2+9 K L^3 \rho^2\right)
\end{align*}
(1) is from \hyperref[Lemma 4.10]{Lemma 4.10}.(4). (2) is from \hyperref[Lemma 4.11]{Lemma 4.11}. (3) is due to taking the expectation of $r$-th round and under the assumption that learning rates satisfy $K L \eta_g \eta_L \leq \frac{S-1}{S}$. (4) is from \hyperref[Lemma 4.6]{Lemma 4.6}, \hyperref[Lemma 4.7]{Lemma 4.7}, \hyperref[Lemma 4.12]{Lemma 4.12}. (5) is because there exists $C> 0,~s.t.~0<C<\left(\frac{1}{2}-30 K^2 L^2 \eta_L^2-\frac{K \eta_g \eta_L}{2 S}(3 K+\left.180 K^2 L^4 \eta_L^2 \rho^2\right)\right)$.\\
After \hyperref[Lemma 4.10]{Lemma 4.10}.(6) for $r=[R]$ is summed and both sides are multiplied by $\frac{1}{C K \eta_g \eta_L R}$, the following holds:
\begin{align*}
& \frac{1}{R} \sum_{r=1}^R \mathbb{E}\left[\left\|f\left(w^{r+1}\right)\right\|^2\right] \\
& \stackrel{(6)}{\leq} \frac{f\left(\widetilde{w}^0_{+\frac{1}{2}}\right)-f^*}{C K \eta_g \eta_L R} 
+\frac{1}{C}\left(10 K L^4 \eta_L^2 \rho^2 \sigma_l^2+90 K^2 L^2 \eta_L^2 \sigma_g^2\right. 
+180 K^2 L^4 \eta_L^2 \rho^2+24 K^4 L^6 \eta_L^6 \rho^2 +2 K^2 L^4 \eta_L^2 \rho^2 \alpha_{g_s} \\
& +16 K^3 L^6 \eta_L^4 \rho^2 +\frac{\eta_g \eta_L L^3 \rho^2}{2 S}\left(\sigma_l^2+G_s^2\right) 
+\frac{\eta_g \eta_L}{S}\left(30 K L^5 \eta_L^2 \rho^2 \sigma_l^2+270 K^2 L^3 \eta_L^2 \sigma_g^2\right. 
+540 K^2 L^5 \eta_L^2 \rho^2+72 K^3 L^7 \eta_L^4 \rho^2 \\
& \left.\left.+6 K^3 L^3 \eta_L^2 \rho^2 \alpha_{g_s}+\frac{9}{2} K L \sigma_g^2+9 K L^3 \rho^2\right)\right). \\
\end{align*}
(6) is from the relationship between current loss and optimal loss $f\left(\widetilde{w}_{+\frac{1}{2}}^{r+1}\right) \geq f^*$, and the one between current loss and initial loss $f\left(\widetilde{w}_{+\frac{1}{2}}^0\right) \geq f\left(\widetilde{w}_{+\frac{1}{2}}^r\right)$. If we choose local and global learning rates as $\eta_l=$ $\frac{1}{\sqrt{R} K L}$ and $\eta_G=\sqrt{K S}$, and choose perturbation radius as $\rho=\frac{1}{\sqrt{R}}$ and $F=$ $f\left(\widetilde{w}_{+\frac{1}{2}}^0\right)-f^*$, we have:
\begin{align*}
& \frac{1}{R} \sum_{r=1}^R \mathbb{E}\left[\left\| f\left(w^{r+1}\right) \right\|^2\right] \\
& \leq \frac{F L}{C K \eta_g \eta_L R}+\frac{1}{C}\left(10 \frac{K L^2 \sigma_l^2}{R}\right. +90 \frac{\sigma_g^2}{R}
+180 \frac{L^2}{R^2}+24 \frac{1}{R^4 K^2} +2 \frac{L^2}{R^2} \alpha_{g_s}+16 \frac{L^2}{R^3 K} 
+\frac{L^2}{2 R^{\frac{3}{2}} \sqrt{K S}}\left(\sigma_l^2+G_s^2\right) \\
& +30 \frac{L^2 \sigma_l^2}{R^{\frac{5}{2}} K^{\frac{3}{2}} \sqrt{S}}
+270 \frac{\sigma_g^2}{R^{\frac{3}{2}} \sqrt{S K}} +540 \frac{L^2}{R^{\frac{5}{2}} \sqrt{K S}}+72 \frac{L^2}{R^{\frac{7}{2}} K^{\frac{3}{2}} \sqrt{S}} 
+6 \frac{\sqrt{K} \alpha_{g_s}}{R^{\frac{5}{2}} \sqrt{S}}\left.+\frac{9 \sqrt{K} \sigma_g^2}{2 \sqrt{S R}}+9 \frac{\sqrt{K} L^2}{R^{\frac{3}{2}}}\right).
\end{align*}
If the number of sampled clients is larger than the one of epochs and high order terms are neglected, the convergence rate of part client participation for sUSAM is:
\begin{align*}
& \frac{1}{R} \sum_{r=1}^R \mathbb{E}\left[\left\| f\left(w^{r+1}\right)\right\|^2\right] 
=O\left(\frac{F L}{\sqrt{R K S}}+\frac{\sqrt{K} \sigma_g^2}{\sqrt{R S}}+\frac{\sigma_g^2}{R}
+\frac{L^2\left(\sigma_l^2+G_s^2\right)+\sigma_g^2}{R^{\frac{3}{2}} \sqrt{K S}}+\frac{\sqrt{K} L^2}{R^{\frac{3}{2}}} +\frac{L^2}{R^2}\left(1+\alpha_{g_s} \right)\right).
\end{align*}

\noindent \textbf{Discussion.} It can be concluded for \hyperref[Theorem 2]{Theorem 2} and \hyperref[Theorem 3]{Theorem 3} that $\frac{\sigma_g^2}{R}$ and $\frac{\sqrt{K} \sigma_g^2}{\sqrt{R S}}$ are caused by heterogeneity between clients; $\frac{L^2 \sigma_l^2}{R^{\frac{3}{2}} \sqrt{K N}}, \frac{L^2 \sigma_l^2}{R^{\frac{3}{2}} \sqrt{K S}}, \frac{\sqrt{K} L^2}{R^{\frac{3}{2}}}, \frac{L^2}{R^2}$, $\frac{L^2 \sigma_l^2}{R^2 K}$ are relevant to local SGD; $\frac{L^2 G_S^2}{R^{\frac{3}{2}} \sqrt{K N}}, \frac{L^2 G_S^2}{R^{\frac{3}{2}} \sqrt{K S}}, \frac{L^2 \alpha_{g_S}}{R^2}$ depend on hyper-parameters of the sparse ratio. $\frac{F L}{\sqrt{R K N}}, \frac{F L}{\sqrt{R K S}}$ and $\frac{\sqrt{K} \sigma_g^2}{\sqrt{R S}}$ represent main terms for the convergence rate. If learning rates are set properly, the convergence rate for sUSAM can be compatible with existing non-convex FL works. Since the mask in sUSAM constrains sparse gradients, additional square and two-thirds terms are also negligible. Besides, sUSAM is potential to generalize better by the dynamic mask, thus alleviating weight shifts in $\frac{\sigma_g^2}{R}$ and $\frac{\sqrt{K} \sigma_g^2}{\sqrt{R S}}$.

\section{Complete Algorithms}   \label{Complete Algorithms}
\hyperref[Algorithm 1]{Algorithm 1} describes the local pretraining process, which can be conducted at any time before the federated training stage. Each client pretrains an organ(s)-specific teacher model based on the partially-annotated dataset. After that, all clients send locally pretrained teacher models to the server for the future pseudo labeling stage.

\begin{algorithm*} 
      \caption{UFPS(pretraining)}  
      \label{Algorithm 1}  
      \begin{algorithmic}[1]  
        \Require  
          {Number of clients $N$; number of local epochs $K$; initial local models $w^1,...,w^N$; partially-annotated local datasets $D_1^p,...,D_N^p$.}
        \For{each client $i \in \{1,...,N\}$ in parallel}  
		\For{$k \in {1,...,K}$ }  
			\State  $w_i^T$ = MiniBatchUpdate($D_i^p,w_i^T$)
		\EndFor   
        \EndFor  
        \State  return $\{ w_i^T \}_{i=1}^N$
      \end{algorithmic}  
    \end{algorithm*}

\hyperref[Algorithm 2]{Algorithm 2} introduces the federated learning flow. Gradient masks and statistics for uncertainty are communicated according to $r_{sUSAM}$ and $r_{UA}$, respectively. $r_{UA}$ is used to ensure statistics of uncertainty banks are sufficient to modify the aggregation weight. After uncertainty-aware global aggregation and calculation of the nonintersecting global gradient mask, the global mask is sent to clients along with the global model.
\renewcommand{\thealgorithm}{2}
\begin{algorithm*}
      \caption{UFPS(federated training)}  
      \label{Algorithm 2}  
      \begin{algorithmic}[1]  
        \Require{Number of local and global rounds $K$ and $R$; number of clients $N$; partially-annotated local datasets $D_1^p,...,D_N^p$;
			pretrained teacher models $w^T_1,...,w^T_N$; initial global model $w^0_0$; 
			uncertainty banks $U_{bank}^1,...,U_{bank}^N$; global mask $M_G$; local masks $M^1_L,...,w^N_L$;
			local momentum gradients $G_{mo}^1,...,G_{mo}^N$; start epoch for global mean teacher, uncertainty-aware aggregation, sUSAM $r_{GMT}$, $r_{UA}$, $r_{sUSAM}$.}
	 \State  server sends all teacher models $w^T$ to all clients
	 \For {each global round $r \in \{1,...,R\}$}  
        	\For{each client $i \in \{1,...,N\}$ in parallel}  
			\State  $w_{i,0}^r=w_0^r$
			\For {each local round $k \in \{1,...,K\} $ }  
				\If	{ $r < r_{sUSAM}$}
					\State  $w_{i,k}^r$ = LocalTraining($D_i^{PL},w^T$)
					\State  client $i$ sends $w_{i,k}^r$ to server
				\ElsIf { $r \geq r_{sUSAM}$ and $r < r_{UA}$}
					\State  $w_{i,k}^r, M^i_L$ = LocalTraining($D_i^{PL},w^T$)
					\State  client $i$ sends $w_{i,k}^r, M^i_L$ to server
				\Else
					\State  $w_{i,k}^r, \mu^i, \sigma^i, M^i_L$ = LocalTraining($D_i^{PL},w^T$)
					\State  client $i$ sends $w_{i,k}^r, \mu^i, \sigma^i, M^i_L$ to server
				\EndIf
			\EndFor  
        	\EndFor 
		\If{ $r < r_{UA}$}		
			\State  $w_0^r=\sum_{i\in[N]} A^w_i w_i^r$
		\Else
			\State  $\hat{A}^w$ $\leftarrow$ server calculates aggregation weights based on $\mu^i, \sigma^i$ according to Equation 7
			\State  $w_0^r=\sum_{i\in[N]} \hat{A}^w_i w_i^r$
		\EndIf
		\State  server sends $w_0^r$ to all clients
		\If{ $r \geq r_{sUSAM}$}		
			\State  $M_G$ $\leftarrow$ server updates global mask according to Equation 12
			\State  server sends global mask $M_G$ to all clients
		\EndIf
	 \EndFor \\  
	\Return $w^*_0$ 
      \end{algorithmic}  
    \end{algorithm*}

\hyperref[Algorithm 3]{Algorithm 3} is about the local training stage in FL. At each local round, each client first uses pretrained local teacher models to get the pseudo label and replaces classes with ground truth at hand. During this process, the uncertainty value is also calculated with logits from teachers. When local training converges, one can decide whether to use the global model to get the pseudo label and refine it by pretrained teachers. With data and pseudo labels, each client updates the local model either through vanilla SGD or sUSAM depending on the current training stage. After local training, each client uploads information according to the current global epoch.
\renewcommand{\thealgorithm}{3}
\begin{algorithm*}  
      \caption{UFPS(LocalTraining)}  
      \label{Algorithm 3}  
      \begin{algorithmic}[1]  
        \Require{Current local round $r$; batch size $B$; partially-annotated dataset $D^P$; local model $w$;
			random initialized blender node and image transform node for CMIDG; warmup epoch $r_{warmup}$; 
			end epoch for weight scheduler $r_{WS}$; start epoch for global mean teacher, uncertainty-aware aggregation, sUSAM $r_{GMT}$, $r_{UA}$, $r_{sUSAM}$;
			updating frequency for sUSAM $r_{fre}$; global mask $M_G$.}
	 \State  $\mu=0,\sigma=0$
	 \For{each batch $\{x,y\}_{i=1}^{B} \in D^P$}
	 	\State  $q^T$ $\leftarrow$  forward all teacher models $w^T_1,...,w^T_N$ on $\{x\}_{i=1}^{B}$
		\State  $U$ $\leftarrow$ calculate Equation 6 through $q^T$ 
		\State   ${\tilde{q}}^T$ $\leftarrow$ merge foreground classes and post-processing background class for $\hat{q}^T$        	
		\If {$r\geq r_{GMT}$}
			\State   ${\tilde{q}}^G$ $\leftarrow$ forward global model on $\{x\}_{i=1}^{B}$
			\State   $\tilde{q}$ $\leftarrow$ refine ${\tilde{q}}^G$ through ${\tilde{q}}^T$ according to Equation 5
			\State   $y'$ $\leftarrow$ merge foreground class in groud truth $y$ into $\tilde{q}$ 
	 	\EndIf
		\If {$r < r_{sUSAM}$}
			\State   $\hat{q}$ $\leftarrow$ forward $w$ on $\{x\}_{i=1}^{B}$ 
			\State   $f = CalculateLoss(\hat{q}, y', w(U))$
			\State $w = w - \eta\bigtriangledown f$
		\Else
			\State   $\{x_{aug}\}_{i=1}^{B}$ $\leftarrow$ perform CMIDG on $\{x\}_{i=1}^{B}$ 
			\State   $\hat{q}_{aug}$ $\leftarrow$ forward $w$ on $\{x_{aug}\}_{i=1}^{B}$ 
			\State   $f = CalculateLoss(\hat{q}_{aug}, y', w(U))$
			\If {$r = r_{sUSAM}$}
				\State   $M_L \leftarrow TopKSort_{T_L} \left|\bigtriangledown f \right|$
				\State   $G_{mo}$ $\leftarrow$ initialize local momentum gradient
			\ElsIf {$r>r_{sUSAM}$ and $r\%r_{fre}=1$}
				\State   $M_L \leftarrow TopKSort_{T_L} \left|\bigtriangledown f \right|$
				\State   $G_{mo}$ $\leftarrow$ update local momentum gradient according to Equation 11
			\Else
				\State   take history mask $M_E$ as current local mask
			\EndIf
			\State   $\tilde{w}_{+\frac{1}{2}}$ $\leftarrow$ perturb model parameters with sparse mask
			\State   $\hat{q}$ $\leftarrow$ forward $\tilde{w}_{+\frac{1}{2}}$  on $\{x\}_{i=1}^{B}$ 
			\State   $\tilde{f}_{+\frac{1}{2}} = CalculateLoss(\hat{q}, y', w(U))$
			\State $w = w - \eta\bigtriangledown \tilde{f}_{+\frac{1}{2}} $
		\EndIf
	 \EndFor
	 \If {$r \geq r_{UA}$}
	 	\State $\mu=\frac{1}{|U_{bank}|}\sum_{i=1}^{|U_{bank}|}U_{bank}^i$
		\State $\sigma=\frac{1}{|U_{bank}|}\sum_{i=1}^{|U_{bank}|}(U_{bank}^i-\mu)^2$
	 \EndIf
	 \If {$r \geq r_{sUSAM}$}
	 	\State $M_L \leftarrow argmax_{T_L} G_{mo}$
	 \EndIf
	 \If {$r < r_{sUSAM}$}
	 	\State \Return $w$
	 \ElsIf { $r \geq r_{sUSAM}$ and $r < r_{UA}$}
		\State \Return $w,M_L$
	 \Else
		\State \Return $w,\mu,\sigma, M_L$
	 \EndIf
      \end{algorithmic}  
    \end{algorithm*}

\hyperref[Algorithm 4]{Algorithm 4} mainly discriminates the calculation of losses. In the warmup global epochs, only basic losses are used. Later, the weight scheduler is used to enforce the local model to concentrate on classes with specific uncertainty values. When the local model converges, aRCE loss is then adopted to help escape from local minima caused by noisy pseudo labels. Note that RCE loss only works when predictions from local models are confirming, so we use it after the end epoch of WS.
\renewcommand{\thealgorithm}{4}
\begin{algorithm*}
      \caption{CalculateLoss}  
      \label{Algorithm 4}  
      \begin{algorithmic}[1]  
        \Require{Model prediction $\hat{q}$; pseudo label $y'$; model parameter $w$; uncertainty based weight scheduler $w(U)$.}
	 	\If {$r < r_{warmup}$}
			\State $w = w - \eta\bigtriangledown f_{Dice+BCE}(\hat{q},y')$
		\ElsIf{$r \geq r_{warmup}$ and $r < r_{WS}$ }
			\State $w = w - \eta\bigtriangledown (w(U)f_{Dice+BCE}(\hat{q},y'))$
		\Else
			\State $w = w - \eta\bigtriangledown f_{Dice+BCE+RCE}(\hat{q},y')$
		\EndIf\\
        \Return $f$
      \end{algorithmic}  
    \end{algorithm*}

\section{Detailed Experimental Setups}   \label{Detailed Experimental Setups}
\subsection{Datasets}  	\label{Detailed Experimental Setups Datasets}

We conduct our experiments with four originally fully-annotated (labels for liver, kidney (left + right), spleen, pancreas) CT image datasets: WORD, AbdomenCT-1K, AMOS (AMOS2022), and BTCV. In the preprocessing stage, we first crop the foreground region with Hounsfield units (HU) more than 0 with a bounding box. Image intensities are then truncated into [-500, 500] to filtrate non-organ volumes. We divide each dataset with a ratio of 7:1:2 for training, validation, and testing sets.

\noindent \textbf{WORD.} Whole abdominal ORgan Dataset (WORD) totally contains 150 abdominal CT volumes (30495 slices) from 150 patients before the radiation therapy. Each volume consists of 159 to 330 slices of 512 $\times$ 512 pixels with an in-plane resolution of 0.976 $mm$ $\times$ 0.976 $mm$ and slice spacing of 2.5-3.0 $mm$, indicating that the WORD is a very high-resolution dataset. The whole dataset has 16 organs with fine pixel-level annotations and scribble-based sparse annotation, including the liver, spleen, kidney (L), kidney (R), stomach, gallbladder, esophagus, duodenum, colon, intestine, adrenal, rectum, bladder, head of the femur (L), and head of the femur (R). For large and small organs, pixel distribution is extremely unbalanced (as shown in Fig2 in the paper of WORD). 120 volumes with annotation and 30 without annotation are published.

\noindent \textbf{AbdomenCT-1K.} AbdomenCT-1K totally contains 1112 abdominal CT volumes from 12 medical centers. Since the information about which center each data belongs to is privacy concerned, we do not further split this dataset into more datasets. Organs in this dataset are liver, kidney, spleen, and pancreas. Resolution of all CT scans is 512 $\times$ 512 pixels with varying pixel sizes and slice thickness between 1.25-5 $mm$. 1000 volumes with annotation and 63 without annotation are published. We pick 266 samples from 1000 labeled volumes with the smallest file size to accelerate the training process. 

\noindent \textbf{AMOS.} AMOS consists of 500 CT and 100 MRI with 15 organs, including spleen, right kidney, left kidney, gallbladder, esophagus, liver, stomach, aorta, inferior vena cava, pancreas, right adrenal gland, left adrenal gland, duodenum, bladder, prostate/uterus. The dataset is from five domains, whereas we do not split it since no domain information is provided. 200 annotated CT images and 40 annotated MRI images are published and we only use CT images.

\noindent \textbf{BTCV.} BTCV provides 50 CT volumes captured during portal venous contrast phase with variable volume sizes 512 $\times$ 512 $\times$ (85-198) and field of views 280 $\times$ 280 $\times$ 280 $mm^3$ - 500 $\times$ 500 $\times$ 650 $mm^3$. The in-plane resolution varies from 0.54 $\times$ 0.54 $mm^2$ to 0.98 $\times$ 0.98 $mm^2$, while the slice thickness ranges from 2.5-5.0 $mm$. 13 organs are in BTCV, i.e., spleen, right kidney, left kidney, gallbladder, esophagus, liver, stomach, aorta, inferior vena cava, portal vein and splenic vein, pancreas, right adrenal gland, left adrenal gland. We merge left kidney and right kidney when testing. We choose 30 training samples in divided BTCV as our out-FL client.

\subsection{Training} 

All methods are based on generic modular UNet from nnUNet \upcite{isensee2021nnu} with auto mixed precision to save GPU memory and accelerate training. We modify deep supervision brunches after decoder to ones after encoder to extract more informative features. The attribute 'tracking\_running\_states' of batch normalization layers is set to False, which is same as FedAvg\upcite{mcmahan2017communication}. The output channel number of the segmentation head is set according to the specific method.

The best global model evaluated on validation sets of clients is used to perform testing for the final result. Only the partial target set and its inverse set (background) are used to train local teacher models. Training transform is conducted by MONAI\upcite{cardoso2022monai}, including normalize to [-1,1], spatial pad, random crop patch (pos:neg=1:1, pos=HU>0) with size (80,192,192), random rotate and flip in x and y axis, affine, grid distortion, cutout, random scale and shift intensity. Dice and BCE losses are used as default loss functions. Transformations for validation and testing only include normalization and spatial pad. Sliding window inference from MONAI is used with ROI size (80,192,192), sw batch size 2 and overlap 0.5. We train all methods for 500 communication rounds. In each communication round, all clients in FL are selected for one epoch of local training. We choose AdamW\upcite{loshchilov2017decoupled} as our optimizer with weight decay $10^{-6}$. We conduct 10 warmup rounds from minimal learning rate $10^{-6}$ to the initiate learning rate $10^{-4}$ and change the learning rate by warmup restart scheduler in rest rounds. All methods are implemented in PyTorch on two Nvidia GeForce RTX 3090s.

\subsection{Metric}  	\label{Detailed Experimental Setups Metric}
Here, we use TP, TN, FP, FN as abbreviation for true positive, true negative, false positive and false negative. Besides, $x_{GT},x_{pred}$ represent ground truth and model prediction.

\noindent \textbf{Dice.} The ratio between the intersection part of two objects and the total area:
$$
\text { Dice }=\frac{2\left|x_{G T} \cap x_{pred }\right|}{\left|x_{G T}\right|+\left|x_{pred }\right|}=\frac{2 T P}{2 T P+F P+F N}
$$

\noindent \textbf{Hausdorff Distance (HD).} HD measures the similarity between the point sets $\{X,Y\}$ for segmentation border:
\begin{align*}
& HD=\max \left\{d_{X Y},d_{Y X}\right\} \\
& =\max \left\{\max _{x \in X} \min _{y \in Y} d(x, y), \min _{y \in Y} \max _{x \in X} d(x, y)\right\}
\end{align*}

\noindent \textbf{Jaccard Coefficient (JC).} The ratio between intersection and union:
$$
JC=\frac{\left|x_{G T} \cap x_{pred }\right|}{\left|x_{G T} \cup x_{pred }\right|}=\frac{T P}{T P+F P+F N}
$$

\noindent \textbf{Sensitivity (Sen).} Sensitivity is also called recall, which measures the model ability to segment interested regions: 
$$
Sen=\frac{\left|x_{G T} \cap x_{p r e d}\right|}{\left|x_{G T}\right|}=\frac{T P}{T P+F N}
$$

\noindent \textbf{Specificity (Spe).} Specificity measures the model ability to judge uninterested regions correctly:
$$
Spe=\frac{T N}{T N+F P}
$$

\noindent \textbf{Relative Volume Error (RVE).} Ratio between absolute error and measurement being taken:
$$
RVE=\frac{abs\left(\left|x_{G T}\right|-\left|x_{p r e d}\right|\right)}{\left|x_{G T}\right|}
$$

\subsection{Complete model aggregation}  	\label{Detailed Experimental Setups Complete model aggregation}
Datasets, i.e., WORD, AbdomenCT-1K, AMOS, with partially-annotated labels are used to train the model(s) by Dice loss and binary cross entropy (BCE) loss only for annotated foreground class(es). Except for Multi-head, Multi-encoder, and Multi-decoder, we only use foreground label(s) to calculate dice and BCE losses.


\noindent \textbf{FedAvg.} Above three datasets are used to train three local models. The aggregated global model is used for evaluation.

\noindent \textbf{Cond-dec.} Above three datasets are used to train three local models. Client index is additionally input into the decoder after each BN layer as a hash value matching magnitude for BN output. Other procedures are same with FedAvg. During validation, same operation as the one in training is performed to get foreground classes corresponding to client indexes. For the background class, three logits for corresponding indexes are averaged in the $0$-th channel.

\noindent \textbf{FedBN.} Above three datasets are used to train three local models. All procedures are same as FedAvg except that batch normalization layers are personalized. During validation, same operations as ones in Cond-dec are performed for foreground classes and the background class.

\noindent \textbf{FedASAM.} Above three datasets are used to train three local models. All procedures are same as FedAvg except methods in FedASAM are additionally used.

\noindent \textbf{MENU-Net.} Above three datasets are used to train local models with multi-encoders by ME loss.  The total number of encoders is equal to the organ amount in our experiment, i.e., four encoders. During the forward pass of training, all features from the last layer of four encoders are concatenated as input to the decoder and all features from each layer of four encoders are concatenated as input to deep supervision layers. Moreover, only the encoder corresponding to the specific organ, which is the current training target,  is updated (thus we update Abdomen-1K twice). It should be noted that the whole patch training strategy in MENU-Net is changed to one random patch chosen by the percentage of pos:neg = 1:1 since the time gap can be tens of times in our experiment. When the server aggregates local models, for the multi-encoder part(s), only the encoder(s) from the client owning the annotated organ label is(are) used to update the part(s) in the global model. When testing, all clients forward all encoders to get four features of the bottleneck layer as the input to the decoder.

\subsection{Part model aggregation}  	\label{Detailed Experimental Setups Part model aggregation}
Datasets, i.e., WORD, AbdomenCT-1K, AMOS, with partially-annotated labels are used to train models by dice and BCE losses only for annotated foreground class(es). Validation in this category is performed before model aggregation. For Multi-encoder and Multi-decoder, the same operation is performed for foreground classes and the background class as in Cond-dec since the output channel of the segmentation head is five (four for the foreground and one for the background).

\noindent \textbf{Multi-head.} All procedures are same with FedAvg except for what is mentioned above and that segmentation heads (the final convolution layer) are personalized. 

\noindent \textbf{Multi-encoder.} All procedures are same with FedAvg except for what is mentioned above and that encoders are personalized.

\noindent \textbf{Multi-decoder.} All procedures are same with FedAvg except for what is mentioned above and that decoders are personalized.

\subsection{Baseline and SOTAs}  	\label{Detailed Experimental Setups Baseline and SOTAs}

We modify existing methods to better suit our experiment and tune hyper-parameters under each method. For all methods except SOLO, Centralized, FedCRLD, and DOD*, model aggregation is conduct before validation and we evaluate on test sets only by the global model. Unless specified, dice and BCE losses are used to train the model(s).

\noindent \textbf{SOLO (partial).} Each client trains a model on the partially-annotated local dataset by dice and BCE losses. During test time, we use three models to get three predictions and merge metrics from all of them. Result models for this method are used as pretrained teacher models for all SOTAs except FedCRLD and DOD*.

\noindent \textbf{Centralized (full).} Only a centralized model is trained on WORD, AbdomenCT-1K, and AMOS with fully annotated labels.

\noindent \textbf{FedCRLD.} FedCRLD is a pFL method with encoders personalized so we perform evaluation before aggregation. This method is based on contrastive learning like BYOL \upcite{grill2020bootstrap}. Cross-attention and self-attention of features are added to the network. Besides segmentation loss, there are three additional losses: MSE loss for predictions between the current local model and the momentum local model; MSE loss for features between the current local model and the momentum local model, and between the global model and the momentum local model; knowledge distillation (KD) loss between the current local model and the global model.

\noindent \textbf{DOD*.} DOD is originally a PSS method designed for centralized learning. A controller is proposed to generate dynamic weights for the segmentation head. The input of the controller is the organ index and the bottleneck feature. The encoder and deep supervision are globally aggregated and the rest remains local. We perform evaluation before model aggregation for this method just as other pFL methods.

\noindent \textbf{CPS*.} CPS is originally a PSS method designed for centralized learning. A co-training method in which two local models with different initializations offer extra pseudo labels for cooperative learning. We modify it into our pseudo labeling framework.

\noindent \textbf{MS-KD*.} MS-KD is originally a PSS method designed for centralized learning. Several losses are added: multi-scale knowledge distillation loss for original logits between local model and teacher models, and the one for features between local model and teacher models. Both of them use one-hot pseudo labels predicted from teacher models to serve as foreground masks.

\noindent \textbf{FedAvg*.} FedAvg is originally an FL method for datasets with all class annotated. We modify it into our pseudo labeling framework. After all clients send their pretained teacher models to the server, the server sends all of them and the initialized global model to clients. Other procedures are same with FedAvg except that each client uses pretrained teacher models to get pseudo labels in each round.

\noindent \textbf{FedProx*.} FedProx is originally an FL method for datasets with all classes annotated. We modify it into our pseudo labeling framework. MSE loss for model parameters between local and global models is added.

\noindent \textbf{MOON*.} MOON is originally an FL method for datasets with all classes annotated. We modify it into our pseudo labeling framework. The contrastive loss of MOON includes features between current local and global models as positive pairs and ones between the current local model and the local model in the last round as negative pairs.

\noindent \textbf{FedAlign*.} FedAlign is originally an FL method for datasets with all classes annotated. A regulation loss of Lipschitz constant for features between the complete local model and the slimmed local model is added. 

\noindent \textbf{FedASAM*.} The two-step minimization method of FedASAM is most similar with the one in our work. Original data are used in the ascent step. Stochastic weight averaging (SWA) is used on the server-side after model aggregation. FedASAM is modified into the pseudo label framework as FedASAM*. For ASAM, we set the start epoch as 0, perturbation radius $\rho$ as 0.7, and balancing coefficient $\eta$ as 0.001. For SWA, we tune the start epoch percentage from {0.25, 0.5, 0.75} and choose 0.75 as the best one, we also tune the cycle from {2,4,6} and choose 2 as the best one.

\noindent \textbf{UFPS.} Here we list all hyper-parameters for all modules in our method. The coefficient for aRCE loss is set to 0.01 and the start epoch is set to 200. For weight scheduler based on uncertainty score, we choose tail shift with 0.7 as the tail percentage threshold and set the end epoch for weight scheduler as 200. For uncertainty-aware global aggregation, we choose decoder as the target part and set $\tau^m,\tau^v$ as 0.05, 0.001, respectively. The corresponding start epoch $r_{UA}$ is set as 300 to accumulate enough statistics in the uncertainty bank. For global mean teacher, we choose 300 as the start epoch for this module and the intersection percentage is set to 0.8. For sUSAM, we set the start epoch $r_{sUSAM}$ as 300, perturbation radius $\rho$ as 0.7, balancing coefficient $\eta$ as 0.001, non-masked percentage for local mask $T_L$ as 0.4, non-masked percentage for global mask $T_L$ as 0.1, gradient momentum coefficient $\alpha_{mo}$ as 0.9, local mask update frequency $r_{fre}$ as 5.

\section{Complete comparison between SOTAs}
Unless 'Post' is written as the header of a table or post-processing is marked in the title of the table, the default setting does not include any post-processing. Post-processing methods consist of filling up binary holes and removing small connected components which are below 20\% percentage of the largest non-background connected component.

Original results and ones with post-processing are demonstrated in \hyperref[Table S2]{Table S2} and \hyperref[Table S3]{Table S3}, separately. Dice is considered as the most important metric in our experiment as it directly reflects the overlap between ground truth and prediction. No matter whether post-processing is used, our model is able to predict accurate results for whole organs, thus getting a great result for dice and other supplementary metrics, i.e., JC and Sen. Just getting a large overlap is not enough since medical diagnosis requires an elaborate segmentation. The ability to refine segmentation border is closely related to HD and RVE. We note that FL based methods all suffer from the excess of smoothness, thus directly not satisfying for the two metrics. However, they can be significantly reduced by simple post-processing in practice. For Specificity, all methods achieve a great score close to 1, meaning all methods are capable of filtering out the most uninterested regions, so the lowest percentage of our method makes little difference compared with other methods in real medical applications.

\begin{table*}[]
	\label{Table S2}
	\centering
	\rowcolors {4}{grey}{white}
	\renewcommand\arraystretch{1.2}
	\resizebox{\columnwidth}{!}{
		\begin{tabular}{llllllllllllllllllllll}
		\arrayrulecolor{red}

		\hline
		\multicolumn{22}{l}{\textcolor{red}{\textbf{Table S2. Complete comparison between SOTAs.}}}  \\
		\hline
		
		 & \multicolumn{21}{l}{\textcolor{red}{Client index (the next line) \& Organ index   (after name of method)}} \\ 
		\cmidrule[0.5mm]{2 -21}
		
		\multirow{-2}{*}{\textcolor{red}{Method \& Metric}} &
		\multicolumn{5}{l}{\textcolor{red}{1}} &
		\multicolumn{5}{l}{\textcolor{red}{2}} &
		\multicolumn{5}{l}{\textcolor{red}{3}} &
		\multicolumn{5}{l}{\textcolor{red}{4}} &
		\textcolor{red}{Mean}  \\ 
		\hline
		
		SOLO partial 
		& 1 & 2 & 3 & 4 & mean 
		& 1 & 2 & 3 & 4 & mean 
		& 1 & 2 & 3 & 4 & mean 
		& 1 & 2 & 3 & 4 & mean &   \\
		Dice & 
		91.480 & 85.928 & 53.943 & 45.359 & 69.178 & 68.785 & 72.387 & 91.847 & 70.010 & 75.757 & 53.098 & 90.202 & 64.664 & 34.435 & 60.600 & 77.485 & 86.167 & 75.398 & 59.763 & 74.703 & 70.059 \\
		HD &
		0.803 & 1.219 & 1.229 & 0.852 & 1.026 & 3.204 & 6.348 & 1.050 & 1.530 & 3.033 & 2.113 & 1.745 & 1.270 & 1.352 & 1.620 & 1.575 & 2.282 & 1.129 & 1.116 & 1.526 & \textcolor{red}{1.801} \\
		JC &
		0.168 & 0.205 & 0.070 & 0.047 & 0.123 & 0.275 & 0.405 & 0.282 & 0.181 & 0.286 & 0.107 & 0.250 & 0.100 & 0.042 & 0.125 & 0.193 & 0.327 & 0.159 & 0.094 & 0.193 & 0.182 \\
		RVE &
		0.061 & 0.095 & 0.398 & 0.351 & 0.226 & 0.345 & 2.045 & 0.092 & 0.220 & 0.676 & 0.471 & 0.063 & 0.265 & 0.421 & 0.305 & 0.210 & 0.080 & 0.120 & 0.226 & 0.159 & 0.341 \\
		Sen &
		0.910 & 0.824 & 0.472 & 0.404 & 0.653 & 0.648 & 0.719 & 0.897 & 0.699 & 0.741 & 0.513 & 0.887 & 0.604 & 0.307 & 0.578 & 0.774 & 0.863 & 0.729 & 0.604 & 0.743 & 0.678 \\
		Spe &
		1.000 & 0.999 & 1.000 & 1.000 & 1.000 & 0.999 & 0.929 & 1.000 & 0.999 & 0.982 & 1.000 & 0.999 & 1.000 & 1.000 & 1.000 & 0.999 & 0.997 & 1.000 & 1.000 & 0.999 & 0.995  \\
		Centralized full 
		& 1 & 2 & 3 & 4 & mean 
		& 1 & 2 & 3 & 4 & mean 
		& 1 & 2 & 3 & 4 & mean 
		& 1 & 2 & 3 & 4 & mean & \\
		Dice &
		87.979 &		83.599 &		70.348 &		73.130 &		78.764 &		94.994 &		95.440 &		91.348 &		71.520 &		88.325 &		88.422 &		89.558 &		72.715 &		66.019 &		79.179 &		88.684 &		89.583 &		77.146 &		67.706 &		80.780 &		81.762 \\
		HD &
		0.825 &		2.238 &		0.684 &		0.678 &		1.106 &		1.681 &		2.733 &		1.055 &		1.428 &		1.724 &		1.211 &		2.314 &		0.838 &		0.958 &		1.330 &		1.383 &		2.286 &		1.011 &		0.985 &		1.416 &		1.394 \\
		JC &
		0.159 &		0.196 &		0.097 &		0.086 &		0.134 &		0.415 &		0.557 &		0.279 &		0.187 &		0.360 &		0.194 &		0.246 &		0.116 &		0.089 &		0.161 &		0.232 &		0.343 &		0.164 &		0.109 &		0.212 &		0.217 \\
		RVE &
		0.090 &
		0.222 &
		0.278 &
		0.120 &
		0.178 &
		0.044 &
		0.032 &
		0.076 &
		0.216 &
		0.092 &
		0.078 &
		0.114 &
		0.166 &
		0.186 &
		0.136 &
		0.416 &
		0.085 &
		0.114 &
		0.167 &
		0.196 &
		0.151 \\
		Sen &
		0.850 &
		0.889 &
		0.642 &
		0.719 &
		0.775 &
		0.957 &
		0.969 &
		0.896 &
		0.700 &
		0.880 &
		0.892 &
		0.919 &
		0.697 &
		0.641 &
		0.787 &
		0.921 &
		0.913 &
		0.762 &
		0.672 &
		0.817 &
		0.815 \\
		Spe &
		1.000 &
		0.996 &
		1.000 &
		1.000 &
		0.999 &
		0.999 &
		0.998 &
		1.000 &
		1.000 &
		0.999 &
		1.000 &
		0.997 &
		1.000 &
		1.000 &
		0.999 &
		0.999 &
		0.996 &
		1.000 &
		1.000 &
		0.999 &
		0.999 \\
		FedCRLD &
		1 &
		2 &
		3 &
		4 &
		mean &
		1 &
		2 &
		3 &
		4 &
		mean &
		1 &
		2 &
		3 &
		4 &
		mean &
		1 &
		2 &
		3 &
		4 &
		mean &
		\\
		Dice &
		89.638 &
		79.277 &
		62.261 &
		40.113 &
		67.822 &
		71.392 &
		79.237 &
		93.171 &
		67.283 &
		77.771 &
		56.759 &
		84.789 &
		61.519 &
		29.407 &
		58.119 &
		74.840 &
		84.667 &
		71.439 &
		58.277 &
		72.306 &
		69.004 \\
		
		HD &
		1.155 &
		5.394 &
		1.283 &
		1.020 &
		2.213 &
		3.402 &
		6.175 &
		1.092 &
		1.580 &
		3.062 &
		2.560 &
		2.795 &
		1.645 &
		1.253 &
		2.063 &
		2.021 &
		2.721 &
		1.448 &
		1.201 &
		1.848 &
		2.297 \\
		JC &
		0.162 &
		0.182 &
		0.084 &
		0.042 &
		0.118 &
		0.284 &
		0.441 &
		0.288 &
		0.172 &
		0.296 &
		0.111 &
		0.228 &
		0.095 &
		0.035 &
		0.117 &
		0.181 &
		0.317 &
		0.149 &
		0.091 &
		0.185 &
		0.179 \\
		
		RVE &
		0.090 &
		0.302 &
		0.287 &
		0.344 &
		0.256 &
		0.340 &
		1.634 &
		0.067 &
		0.268 &
		0.577 &
		0.345 &
		0.182 &
		0.345 &
		0.520 &
		0.348 &
		0.188 &
		0.088 &
		0.166 &
		0.242 &
		0.171 &
		0.338 \\
		Sen &
		0.893 &
		0.852 &
		0.575 &
		0.359 &
		0.670 &
		0.681 &
		0.784 &
		0.933 &
		0.688 &
		0.772 &
		0.529 &
		0.874 &
		0.589 &
		0.259 &
		0.563 &
		0.746 &
		0.846 &
		0.700 &
		0.596 &
		0.722 &
		0.682 \\
		
		Spe &
		1.000 &
		0.995 &
		1.000 &
		1.000 &
		0.999 &
		0.998 &
		0.945 &
		1.000 &
		0.999 &
		0.986 &
		0.999 &
		0.995 &
		1.000 &
		1.000 &
		0.999 &
		0.999 &
		0.996 &
		1.000 &
		1.000 &
		0.999 &
		0.995\\
		DOD* &
		1 &
		2 &
		3 &
		4 &
		mean &
		1 &
		2 &
		3 &
		4 &
		mean &
		1 &
		2 &
		3 &
		4 &
		mean &
		1 &
		2 &
		3 &
		4 &
		mean &
		\\
		Dice &
		86.762 &
		83.903 &
		54.178 &
		22.977 &
		61.955 &
		79.231 &
		89.113 &
		92.128 &
		66.570 &
		81.760 &
		51.877 &
		89.853 &
		64.434 &
		36.129 &
		60.573 &
		73.214 &
		85.997 &
		69.744 &
		53.529 &
		70.621 &
		70.832 \\
		HD &
		0.996 &
		1.328 &
		1.454 &
		0.829 &
		1.152 &
		3.117 &
		4.607 &
		1.117 &
		1.534 &
		2.594 &
		2.005 &
		1.789 &
		1.673 &
		1.315 &
		1.695 &
		1.916 &
		2.423 &
		1.437 &
		1.142 &
		1.729 &
		1.857 \\
		JC &
		0.154 &
		0.199 &
		0.067 &
		0.023 &
		0.111 &
		0.320 &
		0.503 &
		0.281 &
		0.171 &
		0.319 &
		0.104 &
		0.247 &
		0.096 &
		0.044 &
		0.123 &
		0.176 &
		0.325 &
		0.142 &
		0.083 &
		0.182 &
		0.186 \\
		RVE &
		0.101 &
		0.097 &
		0.653 &
		0.713 &
		0.391 &
		0.214 &
		0.433 &
		0.080 &
		0.242 &
		0.242 &
		0.472 &
		0.066 &
		0.333 &
		0.451 &
		0.331 &
		0.287 &
		0.083 &
		0.193 &
		0.289 &
		0.213 &
		\textcolor{red}{0.271} \\
		Sen &
		0.831 &
		0.799 &
		0.551 &
		0.179 &
		0.590 &
		0.759 &
		0.890 &
		0.902 &
		0.645 &
		0.799 &
		0.481 &
		0.880 &
		0.612 &
		0.348 &
		0.580 &
		0.710 &
		0.851 &
		0.677 &
		0.522 &
		0.690 &
		0.693 \\
		Spe &
		1.000 &
		0.999 &
		0.999 &
		1.000 &
		0.999 &
		0.999 &
		0.987 &
		1.000 &
		1.000 &
		0.996 &
		1.000 &
		0.999 &
		0.999 &
		1.000 &
		0.999 &
		0.999 &
		0.997 &
		0.999 &
		1.000 &
		0.999 &
		\textcolor{red}{0.997} \\
		CPS* &
		1 &
		2 &
		3 &
		4 &
		mean &
		1 &
		2 &
		3 &
		4 &
		mean &
		1 &
		2 &
		3 &
		4 &
		mean &
		1 &
		2 &
		3 &
		4 &
		mean &
		\\
		Dice &
		89.739 &
		89.348 &
		75.950 &
		48.116 &
		75.788 &
		73.501 &
		77.192 &
		93.272 &
		68.266 &
		78.058 &
		59.186 &
		89.972 &
		74.797 &
		39.061 &
		65.754 &
		80.089 &
		87.951 &
		75.219 &
		59.170 &
		75.607 &
		73.802 \\
		
		HD &
		1.008 &
		1.193 &
		0.966 &
		0.806 &
		0.993 &
		3.028 &
		6.065 &
		1.050 &
		1.525 &
		2.917 &
		2.185 &
		1.931 &
		1.376 &
		1.317 &
		1.702 &
		1.849 &
		2.293 &
		1.336 &
		1.146 &
		1.656 &
		1.817 \\
		JC &
		0.163 &
		0.217 &
		0.106 &
		0.051 &
		0.134 &
		0.302 &
		0.437 &
		0.288 &
		0.176 &
		0.301 &
		0.121 &
		0.249 &
		0.117 &
		0.048 &
		0.134 &
		0.200 &
		0.335 &
		0.158 &
		0.092 &
		0.196 &
		0.191 \\
		
		RVE &
		0.068 &
		0.055 &
		0.174 &
		0.293 &
		0.147 &
		0.290 &
		2.112 &
		0.089 &
		0.221 &
		0.678 &
		0.411 &
		0.098 &
		0.246 &
		0.391 &
		0.287 &
		0.270 &
		0.075 &
		0.153 &
		0.197 &
		0.174 &
		0.322 \\
		Sen &
		0.884 &
		0.883 &
		0.698 &
		0.432 &
		0.724 &
		0.702 &
		0.779 &
		0.933 &
		0.677 &
		0.773 &
		0.578 &
		0.901 &
		0.711 &
		0.359 &
		0.637 &
		0.820 &
		0.891 &
		0.734 &
		0.595 &
		0.760 &
		0.724 \\
		
		Spe &
		1.000 &
		0.999 &
		1.000 &
		1.000 &
		1.000 &
		0.999 &
		0.927 &
		1.000 &
		0.999 &
		0.981 &
		0.999 &
		0.997 &
		1.000 &
		1.000 &
		0.999 &
		0.999 &
		0.996 &
		1.000 &
		1.000 &
		0.999 &
		0.995\\
		MS-KD* &
		1 &
		2 &
		3 &
		4 &
		mean &
		1 &
		2 &
		3 &
		4 &
		mean &
		1 &
		2 &
		3 &
		4 &
		mean &
		1 &
		2 &
		3 &
		4 &
		mean &
		\\
		Dice &
		89.920 &
		88.511 &
		71.978 &
		46.995 &
		74.351 &
		72.861 &
		75.769 &
		90.539 &
		66.100 &
		76.317 &
		55.820 &
		89.298 &
		69.066 &
		39.118 &
		63.326 &
		77.914 &
		86.683 &
		71.644 &
		56.173 &
		73.104 &
		71.775 \\
		
		HD &
		0.899 &
		1.282 &
		1.020 &
		0.813 &
		1.004 &
		2.978 &
		6.159 &
		1.209 &
		1.585 &
		2.983 &
		2.088 &
		1.880 &
		1.360 &
		1.336 &
		1.666 &
		1.777 &
		2.332 &
		1.323 &
		1.132 &
		1.641 &
		1.824 \\
		JC &
		0.163 &
		0.214 &
		0.098 &
		0.050 &
		0.131 &
		0.298 &
		0.428 &
		0.277 &
		0.168 &
		0.293 &
		0.114 &
		0.246 &
		0.106 &
		0.048 &
		0.128 &
		0.193 &
		0.328 &
		0.149 &
		0.086 &
		0.189 &
		0.185 \\
		
		RVE &
		0.067 &
		0.071 &
		0.191 &
		0.268 &
		0.149 &
		0.319 &
		1.955 &
		0.141 &
		0.240 &
		0.664 &
		0.462 &
		0.094 &
		0.239 &
		0.396 &
		0.298 &
		0.210 &
		0.081 &
		0.145 &
		0.236 &
		0.168 &
		0.320 \\
		Sen &
		0.895 &
		0.860 &
		0.656 &
		0.442 &
		0.713 &
		0.690 &
		0.748 &
		0.903 &
		0.665 &
		0.752 &
		0.543 &
		0.877 &
		0.645 &
		0.374 &
		0.610 &
		0.783 &
		0.858 &
		0.698 &
		0.580 &
		0.730 &
		0.701 \\
		
		Spe &
		1.000 &
		0.999 &
		1.000 &
		1.000 &
		1.000 &
		0.999 &
		0.932 &
		0.999 &
		0.999 &
		0.983 &
		1.000 &
		0.998 &
		1.000 &
		1.000 &
		0.999 &
		0.999 &
		0.997 &
		1.000 &
		1.000 &
		0.999 &
		0.995\\
		FedAvg* &
		1 &
		2 &
		3 &
		4 &
		mean &
		1 &
		2 &
		3 &
		4 &
		mean &
		1 &
		2 &
		3 &
		4 &
		mean &
		1 &
		2 &
		3 &
		4 &
		mean &
		\\
		Dice &
		90.114 &
		84.690 &
		72.959 &
		51.983 &
		74.937 &
		72.890 &
		79.384 &
		93.662 &
		66.448 &
		78.096 &
		59.062 &
		88.036 &
		68.126 &
		42.890 &
		64.528 &
		79.517 &
		87.911 &
		75.399 &
		56.123 &
		74.738 &
		73.075 \\
		HD &
		0.857 &
		2.593 &
		0.810 &
		0.910 &
		1.292 &
		2.770 &
		6.033 &
		0.993 &
		1.527 &
		2.831 &
		1.869 &
		2.590 &
		1.552 &
		1.310 &
		1.830 &
		1.730 &
		2.548 &
		1.291 &
		1.206 &
		1.694 &
		1.912 \\
		JC &
		0.164 &
		0.200 &
		0.101 &
		0.056 &
		0.130 &
		0.296 &
		0.449 &
		0.290 &
		0.169 &
		0.301 &
		0.119 &
		0.239 &
		0.107 &
		0.054 &
		0.130 &
		0.197 &
		0.333 &
		0.158 &
		0.086 &
		0.194 &
		0.189 \\
		RVE &
		0.065 &
		0.176 &
		0.223 &
		0.196 &
		0.165 &
		0.310 &
		2.211 &
		0.063 &
		0.240 &
		0.706 &
		0.409 &
		0.221 &
		0.343 &
		0.348 &
		0.330 &
		0.249 &
		0.093 &
		0.125 &
		0.210 &
		0.169 &
		0.343 \\
		Sen &
		0.897 &
		0.892 &
		0.677 &
		0.485 &
		0.738 &
		0.692 &
		0.803 &
		0.926 &
		0.645 &
		0.766 &
		0.571 &
		0.920 &
		0.640 &
		0.402 &
		0.633 &
		0.807 &
		0.904 &
		0.734 &
		0.561 &
		0.752 &
		0.722 \\
		Spe &
		1.000 &
		0.997 &
		1.000 &
		1.000 &
		0.999 &
		0.999 &
		0.925 &
		1.000 &
		0.999 &
		0.981 &
		1.000 &
		0.994 &
		0.999 &
		1.000 &
		0.998 &
		0.999 &
		0.995 &
		1.000 &
		1.000 &
		0.998 &
		0.994 \\
		FedProx* &
		1 &
		2 &
		3 &
		4 &
		mean &
		1 &
		2 &
		3 &
		4 &
		mean &
		1 &
		2 &
		3 &
		4 &
		mean &
		1 &
		2 &
		3 &
		4 &
		mean &
		\\
		Dice &
		89.394 &
		83.151 &
		73.447 &
		51.684 &
		74.419 &
		70.491 &
		76.653 &
		94.049 &
		68.145 &
		77.334 &
		55.708 &
		88.826 &
		72.188 &
		41.519 &
		64.560 &
		79.392 &
		86.664 &
		76.453 &
		59.897 &
		75.601 &
		72.979 \\
		
		HD &
		0.915 &
		2.924 &
		0.917 &
		0.816 &
		1.393 &
		2.880 &
		6.108 &
		1.051 &
		1.509 &
		2.887 &
		1.906 &
		2.284 &
		1.663 &
		1.183 &
		1.759 &
		1.699 &
		2.416 &
		1.327 &
		1.151 &
		1.648 &
		1.922 \\
		JC &
		0.162 &
		0.194 &
		0.101 &
		0.056 &
		0.128 &
		0.285 &
		0.433 &
		0.291 &
		0.175 &
		0.296 &
		0.113 &
		0.243 &
		0.112 &
		0.052 &
		0.130 &
		0.198 &
		0.328 &
		0.161 &
		0.094 &
		0.195 &
		0.187 \\
		
		RVE &
		0.076 &
		0.233 &
		0.220 &
		0.247 &
		0.194 &
		0.363 &
		2.060 &
		0.057 &
		0.239 &
		0.680 &
		0.459 &
		0.148 &
		0.276 &
		0.378 &
		0.315 &
		0.228 &
		0.075 &
		0.129 &
		0.214 &
		0.162 &
		0.338 \\
		Sen &
		0.876 &
		0.885 &
		0.673 &
		0.479 &
		0.728 &
		0.661 &
		0.767 &
		0.939 &
		0.680 &
		0.762 &
		0.534 &
		0.900 &
		0.685 &
		0.388 &
		0.627 &
		0.799 &
		0.872 &
		0.756 &
		0.618 &
		0.761 &
		0.720 \\
		
		Spe &
		1.000 &
		0.996 &
		1.000 &
		1.000 &
		0.999 &
		0.999 &
		0.929 &
		1.000 &
		0.999 &
		0.982 &
		1.000 &
		0.996 &
		0.999 &
		1.000 &
		0.999 &
		0.999 &
		0.996 &
		1.000 &
		1.000 &
		0.999 &
		0.995\\
		MOON* &
		1 &
		2 &
		3 &
		4 &
		mean &
		1 &
		2 &
		3 &
		4 &
		mean &
		1 &
		2 &
		3 &
		4 &
		mean &
		1 &
		2 &
		3 &
		4 &
		mean &
		\\
		Dice &
		90.051 &
		87.758 &
		78.271 &
		43.931 &
		75.003 &
		73.214 &
		76.661 &
		93.128 &
		68.591 &
		77.899 &
		58.463 &
		88.662 &
		72.081 &
		37.228 &
		64.109 &
		79.703 &
		85.711 &
		75.293 &
		60.303 &
		75.252 &
		73.066 \\
		
		HD &
		1.078 &
		1.997 &
		1.004 &
		0.819 &
		1.225 &
		2.781 &
		6.167 &
		1.046 &
		1.527 &
		2.880 &
		2.255 &
		2.240 &
		1.703 &
		1.143 &
		1.836 &
		1.904 &
		2.426 &
		1.388 &
		1.069 &
		1.697 &
		1.910 \\
		JC &
		0.164 &
		0.211 &
		0.109 &
		0.046 &
		0.132 &
		0.299 &
		0.432 &
		0.287 &
		0.176 &
		0.299 &
		0.119 &
		0.243 &
		0.112 &
		0.046 &
		0.130 &
		0.198 &
		0.324 &
		0.158 &
		0.095 &
		0.194 &
		0.189 \\
		
		RVE &
		0.092 &
		0.060 &
		0.137 &
		0.331 &
		0.155 &
		0.305 &
		2.022 &
		0.073 &
		0.234 &
		0.658 &
		0.440 &
		0.125 &
		0.290 &
		0.381 &
		0.309 &
		0.311 &
		0.077 &
		0.169 &
		0.229 &
		0.196 &
		0.330 \\
		Sen &
		0.898 &
		0.869 &
		0.734 &
		0.395 &
		0.724 &
		0.704 &
		0.764 &
		0.924 &
		0.686 &
		0.769 &
		0.574 &
		0.891 &
		0.684 &
		0.343 &
		0.623 &
		0.828 &
		0.859 &
		0.743 &
		0.608 &
		0.760 &
		0.719 \\
		
		Spe &
		1.000 &
		0.999 &
		1.000 &
		1.000 &
		0.999 &
		0.999 &
		0.930 &
		1.000 &
		0.999 &
		0.982 &
		0.999 &
		0.997 &
		0.999 &
		1.000 &
		0.999 &
		0.999 &
		0.996 &
		1.000 &
		1.000 &
		0.999 &
		0.995\\
		FedAlign* &
		1 &
		2 &
		3 &
		4 &
		mean &
		1 &
		2 &
		3 &
		4 &
		mean &
		1 &
		2 &
		3 &
		4 &
		mean &
		1 &
		2 &
		3 &
		4 &
		mean &
		\\
		Dice &
		89.215 &
		85.110 &
		73.804 &
		52.606 &
		75.184 &
		70.796 &
		76.502 &
		94.095 &
		66.913 &
		77.077 &
		56.253 &
		88.256 &
		73.643 &
		39.114 &
		64.317 &
		81.152 &
		87.635 &
		76.489 &
		59.623 &
		76.225 &
		73.201 \\
		
		HD &
		0.825 &
		2.248 &
		0.968 &
		0.782 &
		1.206 &
		2.804 &
		6.204 &
		0.997 &
		1.559 &
		2.891 &
		1.840 &
		2.453 &
		1.613 &
		1.205 &
		1.778 &
		1.617 &
		2.475 &
		1.281 &
		1.116 &
		1.622 &
		1.874 \\
		JC &
		0.161 &
		0.202 &
		0.101 &
		0.057 &
		0.130 &
		0.282 &
		0.430 &
		0.292 &
		0.170 &
		0.294 &
		0.113 &
		0.240 &
		0.115 &
		0.048 &
		0.129 &
		0.202 &
		0.332 &
		0.161 &
		0.093 &
		0.197 &
		0.188 \\
		
		RVE &
		0.073 &
		0.099 &
		0.207 &
		0.272 &
		0.163 &
		0.387 &
		2.124 &
		0.068 &
		0.257 &
		0.709 &
		0.451 &
		0.142 &
		0.281 &
		0.426 &
		0.325 &
		0.169 &
		0.069 &
		0.149 &
		0.226 &
		0.153 &
		0.338 \\
		Sen &
		0.866 &
		0.863 &
		0.689 &
		0.482 &
		0.725 &
		0.652 &
		0.759 &
		0.941 &
		0.670 &
		0.755 &
		0.532 &
		0.897 &
		0.705 &
		0.370 &
		0.626 &
		0.810 &
		0.884 &
		0.756 &
		0.611 &
		0.765 &
		0.718 \\
		
		Spe &
		1.000 &
		0.998 &
		1.000 &
		1.000 &
		0.999 &
		1.000 &
		0.927 &
		1.000 &
		0.999 &
		0.981 &
		1.000 &
		0.996 &
		0.999 &
		1.000 &
		0.999 &
		0.999 &
		0.996 &
		1.000 &
		1.000 &
		0.999 &
		0.995\\
		FedASAM* &
		1 &
		2 &
		3 &
		4 &
		mean &
		1 &
		2 &
		3 &
		4 &
		mean &
		1 &
		2 &
		3 &
		4 &
		mean &
		1 &
		2 &
		3 &
		4 &
		mean &
		\\
		Dice &
		89.823 &
		84.635 &
		76.024 &
		57.602 &
		77.021 &
		71.257 &
		80.009 &
		93.295 &
		68.073 &
		78.159 &
		56.630 &
		89.340 &
		69.946 &
		46.494 &
		65.603 &
		78.792 &
		87.480 &
		73.454 &
		60.859 &
		75.146 &
		73.982 \\
		HD &
		1.059 &
		2.882 &
		0.791 &
		0.849 &
		1.395 &
		2.998 &
		5.976 &
		1.020 &
		1.658 &
		2.913 &
		1.854 &
		2.628 &
		1.269 &
		1.258 &
		1.752 &
		1.686 &
		2.603 &
		1.301 &
		1.260 &
		1.712 &
		1.943 \\
		JC &
		0.163 &
		0.200 &
		0.105 &
		0.063 &
		0.133 &
		0.288 &
		0.451 &
		0.288 &
		0.173 &
		0.300 &
		0.115 &
		0.245 &
		0.109 &
		0.058 &
		0.132 &
		0.196 &
		0.331 &
		0.153 &
		0.094 &
		0.194 &
		0.190 \\
		RVE &
		0.090 &
		0.200 &
		0.193 &
		0.180 &
		0.166 &
		0.347 &
		2.012 &
		0.085 &
		0.302 &
		0.686 &
		0.451 &
		0.161 &
		0.269 &
		0.332 &
		0.303 &
		0.237 &
		0.086 &
		0.143 &
		0.248 &
		0.178 &
		0.333 \\
		Sen &
		0.885 &
		0.885 &
		0.703 &
		0.564 &
		0.759 &
		0.672 &
		0.802 &
		0.921 &
		0.738 &
		0.783 &
		0.547 &
		0.919 &
		0.654 &
		0.473 &
		0.648 &
		0.794 &
		0.894 &
		0.716 &
		0.668 &
		0.768 &
		0.740 \\
		Spe &
		1.000 &
		0.997 &
		1.000 &
		1.000 &
		0.999 &
		0.999 &
		0.932 &
		1.000 &
		0.999 &
		0.982 &
		1.000 &
		0.996 &
		1.000 &
		1.000 &
		0.999 &
		0.999 &
		0.995 &
		1.000 &
		0.999 &
		0.998 &
		0.995 \\
		UFPS(Ours) &		1 &		2 &		3 &		4 &
		mean &
		1 &		2 &		3 &		4 &		mean &
		1 &		2 &		3 &		4 &		mean &
		1 &		2 &		3 &		4 &		mean &
		\\
		Dice &
		90.186 &		85.345 &		76.248 &		53.126 &		76.226 &		76.428 &		79.529 &		93.512 &		68.788 &		79.564 &		60.284 &		88.419 &		74.441 &		44.140 &		66.821 &		81.082 &		88.691 &		76.677 &		62.456 &		77.226 &		\textcolor{red}{74.959} \\
		HD &
		0.954 &		2.534 &		1.530 &		0.798 &		1.454 &		2.626 &		6.015 &		1.027 &		1.615 &		2.821 &		2.279 &		2.847 &		1.864 &		1.182 &		2.043 &		1.847 &		2.518 &		1.364 &		1.176 &		1.726 &		2.011 \\
		JC &
		0.164 &		0.202 &		0.106 &		0.057 &		0.132 &		0.315 &		0.449 &		0.289 &		0.176 &		0.307 &		0.124 &		0.242 &		0.115 &		0.055 &		0.134 &		0.204 &		0.338 &		0.162 &		0.097 &		0.200 &		\textcolor{red}{0.193} \\
		RVE &
		0.081 &		0.128 &		0.198 &		0.209 &		0.154 &		0.264 &		2.133 &		0.068 &		0.261 &		0.681 &		0.417 &		0.180 &		0.254 &		0.365 &		0.304 &		0.340 &		0.082 &		0.129 &		0.182 &		0.183 &		0.331 \\
		Sen &
		0.911 &		0.882 &		0.742 &		0.518 &		0.763 &		0.739 &		0.803 &		0.931 &		0.713 &		0.796 &		0.595 &		0.914 &		0.726 &		0.429 &		0.666 &		0.842 &		0.903 &		0.758 &		0.657 &		0.790 &		\textcolor{red}{0.754} \\
		Spe &
		1.000 &		0.997 &		1.000 &		1.000 &		0.999 &		0.999 &		0.927 &		1.000 &		0.999 &		0.981 &		0.999 &		0.995 &		0.999 &		1.000 &		0.998 &		0.999 &		0.996 &		1.000 &		0.999 &		0.998 &		0.994 \\
		\hline 
		\rowcolor{white}\multicolumn{22}{l}{Organ indexes are numbers on the straight right side of the name of method.} \\
		\rowcolor{white}\multicolumn{22}{l}{Model performance is better when metrics are higher, except for HD and RVE.} \\
		\rowcolor{white}\multicolumn{22}{l}{The best method (except Centralized full) for each metric is marked in red.} \\
		\hline

		\end{tabular}
	}
	\end{table*}

\begin{table*}[]
	\label{Table S3}
	\centering
	\rowcolors {4}{grey}{white}
	\renewcommand\arraystretch{1.2}
	\resizebox{\columnwidth}{!}{
		\begin{tabular}{llllllllllllllllllllll}
		\arrayrulecolor{red}

		\hline
		\multicolumn{22}{l}{\textcolor{red}{\textbf{Table S3. Complete comparison between SOTAs with post-processing.}}}  \\
		\hline
		
			& \multicolumn{21}{l}{\textcolor{red}{Client index (the next line) \& Organ index   (after name of method)}} \\ 
		\cmidrule[0.5mm]{2 -21}
		
		\multirow{-2}{*}{\textcolor{red}{Method \& Metric}} &
		\multicolumn{5}{l}{\textcolor{red}{1}} &
		\multicolumn{5}{l}{\textcolor{red}{2}} &
		\multicolumn{5}{l}{\textcolor{red}{3}} &
		\multicolumn{5}{l}{\textcolor{red}{4}} &
		\textcolor{red}{Mean}  \\ 
		\hline
		
		SOLO partial &
		1 &
		2 &
		3 &
		4 &
		mean &
		1 &
		2 &
		3 &
		4 &
		mean &
		1 &
		2 &
		3 &
		4 &
		mean &
		1 &
		2 &
		3 &
		4 &
		mean &
		\\
		Dice &
		91.486 &
		86.121 &
		54.106 &
		44.041 &
		68.939 &
		68.976 &
		70.662 &
		91.953 &
		70.088 &
		75.420 &
		52.805 &
		90.450 &
		65.057 &
		33.989 &
		60.575 &
		77.569 &
		86.495 &
		75.585 &
		59.494 &
		74.786 &
		69.930 \\
		HD &
		0.693 &
		1.192 &
		0.912 &
		0.743 &
		0.885 &
		3.037 &
		5.934 &
		0.979 &
		1.522 &
		2.868 &
		1.887 &
		1.583 &
		1.014 &
		1.228 &
		1.428 &
		1.521 &
		2.204 &
		1.041 &
		1.073 &
		1.460 &
		1.660 \\
		JC &
		0.168 &
		0.206 &
		0.071 &
		0.046 &
		0.123 &
		0.277 &
		0.398 &
		0.283 &
		0.181 &
		0.285 &
		0.107 &
		0.251 &
		0.101 &
		0.041 &
		0.125 &
		0.193 &
		0.329 &
		0.160 &
		0.094 &
		0.194 &
		0.182 \\
		RVE &
		0.059 &
		0.092 &
		0.436 &
		0.382 &
		0.242 &
		0.357 &
		1.860 &
		0.090 &
		0.224 &
		0.633 &
		0.477 &
		0.054 &
		0.273 &
		0.440 &
		0.311 &
		0.216 &
		0.076 &
		0.119 &
		0.227 &
		0.160 &
		0.336 \\
		Sen &
		0.909 &
		0.825 &
		0.470 &
		0.392 &
		0.649 &
		0.646 &
		0.705 &
		0.897 &
		0.697 &
		0.736 &
		0.510 &
		0.888 &
		0.604 &
		0.302 &
		0.576 &
		0.773 &
		0.863 &
		0.729 &
		0.601 &
		0.742 &
		0.676 \\
		Spe &
		1.000 &
		0.999 &
		1.000 &
		1.000 &
		1.000 &
		0.999 &
		0.936 &
		1.000 &
		0.999 &
		0.984 &
		1.000 &
		0.999 &
		1.000 &
		1.000 &
		1.000 &
		0.999 &
		0.997 &
		1.000 &
		1.000 &
		0.999 &
		0.996 \\
		Centralized full &
		1 &
		2 &
		3 &
		4 &
		mean &
		1 &
		2 &
		3 &
		4 &
		mean &
		1 &
		2 &
		3 &
		4 &
		mean &
		1 &
		2 &
		3 &
		4 &
		mean &
		\\
		Dice &
		88.096 &
		87.610 &
		70.339 &
		73.258 &
		79.826 &
		95.015 &
		95.699 &
		91.432 &
		71.640 &
		88.446 &
		88.620 &
		91.840 &
		72.749 &
		65.846 &
		79.764 &
		88.743 &
		90.148 &
		77.260 &
		67.745 &
		80.974 &
		82.253 \\
		HD &
		0.743 &
		1.365 &
		0.684 &
		0.628 &
		0.855 &
		1.669 &
		2.410 &
		1.036 &
		1.424 &
		1.635 &
		1.024 &
		1.625 &
		0.788 &
		0.893 &
		1.082 &
		1.325 &
		2.120 &
		0.998 &
		0.968 &
		1.353 &
		1.231 \\
		JC &
		0.159 &
		0.211 &
		0.097 &
		0.086 &
		0.138 &
		0.415 &
		0.560 &
		0.280 &
		0.187 &
		0.361 &
		0.195 &
		0.256 &
		0.116 &
		0.089 &
		0.164 &
		0.232 &
		0.347 &
		0.165 &
		0.110 &
		0.213 &
		0.219 \\
		RVE &
		0.084 &
		0.132 &
		0.278 &
		0.125 &
		0.155 &
		0.043 &
		0.022 &
		0.076 &
		0.212 &
		0.088 &
		0.071 &
		0.059 &
		0.170 &
		0.197 &
		0.124 &
		0.415 &
		0.079 &
		0.112 &
		0.160 &
		0.191 &
		0.140 \\
		Sen &
		0.851 &
		0.890 &
		0.642 &
		0.719 &
		0.775 &
		0.957 &
		0.969 &
		0.896 &
		0.699 &
		0.880 &
		0.891 &
		0.919 &
		0.696 &
		0.639 &
		0.787 &
		0.921 &
		0.913 &
		0.762 &
		0.670 &
		0.816 &
		0.815 \\
		Spe &
		1.000 &
		0.998 &
		1.000 &
		1.000 &
		0.999 &
		0.999 &
		0.998 &
		1.000 &
		1.000 &
		0.999 &
		1.000 &
		0.998 &
		1.000 &
		1.000 &
		0.999 &
		0.999 &
		0.996 &
		1.000 &
		1.000 &
		0.999 &
		0.999 \\
		FedCRLD &
		1 &
		2 &
		3 &
		4 &
		mean &
		1 &
		2 &
		3 &
		4 &
		mean &
		1 &
		2 &
		3 &
		4 &
		mean &
		1 &
		2 &
		3 &
		4 &
		mean &
		\\
		Dice &
		84.112 &
		85.878 &
		62.329 &
		38.752 &
		67.768 &
		67.770 &
		74.864 &
		93.433 &
		67.373 &
		75.860 &
		45.890 &
		85.537 &
		62.287 &
		28.543 &
		55.565 &
		77.017 &
		86.740 &
		72.047 &
		57.701 &
		73.376 &
		68.142 \\
		
		HD &
		0.821 &
		1.638 &
		0.905 &
		0.821 &
		1.046 &
		2.793 &
		5.796 &
		1.017 &
		1.563 &
		2.793 &
		1.829 &
		2.512 &
		1.116 &
		1.159 &
		1.654 &
		1.556 &
		2.402 &
		1.163 &
		1.081 &
		1.550 &
		1.761 \\
		JC &
		0.150 &
		0.204 &
		0.084 &
		0.041 &
		0.120 &
		0.271 &
		0.422 &
		0.289 &
		0.172 &
		0.288 &
		0.092 &
		0.229 &
		0.097 &
		0.034 &
		0.113 &
		0.190 &
		0.326 &
		0.151 &
		0.090 &
		0.189 &
		0.178 \\
		
		RVE &
		0.126 &
		0.139 &
		0.319 &
		0.388 &
		0.243 &
		0.390 &
		1.887 &
		0.066 &
		0.271 &
		0.653 &
		0.567 &
		0.231 &
		0.328 &
		0.545 &
		0.418 &
		0.258 &
		0.084 &
		0.157 &
		0.257 &
		0.189 &
		0.376 \\
		Sen &
		0.804 &
		0.875 &
		0.575 &
		0.347 &
		0.650 &
		0.624 &
		0.748 &
		0.933 &
		0.686 &
		0.748 &
		0.428 &
		0.893 &
		0.589 &
		0.250 &
		0.540 &
		0.750 &
		0.871 &
		0.700 &
		0.586 &
		0.727 &
		0.666 \\
		
		Spe &
		1.000 &
		0.998 &
		1.000 &
		1.000 &
		0.999 &
		1.000 &
		0.935 &
		1.000 &
		0.999 &
		0.983 &
		1.000 &
		0.994 &
		1.000 &
		1.000 &
		0.998 &
		0.999 &
		0.996 &
		1.000 &
		1.000 &
		0.999 &
		0.995\\

		DOD* &
		1 &
		2 &
		3 &
		4 &
		mean &
		1 &
		2 &
		3 &
		4 &
		mean &
		1 &
		2 &
		3 &
		4 &
		mean &
		1 &
		2 &
		3 &
		4 &
		mean &
		\\
		Dice &
		87.269 &
		84.003 &
		65.498 &
		41.000 &
		69.443 &
		79.903 &
		86.098 &
		92.608 &
		67.929 &
		81.635 &
		51.056 &
		89.947 &
		62.968 &
		35.182 &
		59.788 &
		73.609 &
		86.049 &
		71.120 &
		58.136 &
		72.229 &
		70.773 \\
		HD &
		0.789 &
		1.247 &
		0.969 &
		0.769 &
		0.944 &
		2.687 &
		4.955 &
		1.078 &
		1.590 &
		2.578 &
		1.767 &
		1.744 &
		1.209 &
		1.147 &
		1.467 &
		1.634 &
		2.259 &
		1.255 &
		1.075 &
		1.556 &
		1.636 \\
		JC &
		0.156 &
		0.199 &
		0.087 &
		0.044 &
		0.122 &
		0.324 &
		0.486 &
		0.284 &
		0.174 &
		0.317 &
		0.103 &
		0.249 &
		0.095 &
		0.043 &
		0.123 &
		0.178 &
		0.327 &
		0.147 &
		0.091 &
		0.186 &
		0.187 \\
		RVE &
		0.102 &
		0.098 &
		0.216 &
		0.326 &
		0.186 &
		0.220 &
		0.852 &
		0.078 &
		0.272 &
		0.356 &
		0.489 &
		0.072 &
		0.334 &
		0.462 &
		0.339 &
		0.306 &
		0.077 &
		0.185 &
		0.208 &
		0.194 &
		\textcolor{red}{0.269} \\
		Sen &
		0.834 &
		0.799 &
		0.593 &
		0.402 &
		0.657 &
		0.757 &
		0.864 &
		0.924 &
		0.713 &
		0.815 &
		0.471 &
		0.884 &
		0.593 &
		0.337 &
		0.571 &
		0.699 &
		0.850 &
		0.697 &
		0.607 &
		0.713 &
		0.689 \\
		Spe &
		1.000 &
		0.999 &
		1.000 &
		1.000 &
		1.000 &
		0.999 &
		0.973 &
		1.000 &
		0.999 &
		0.993 &
		1.000 &
		0.998 &
		1.000 &
		1.000 &
		1.000 &
		0.999 &
		0.997 &
		1.000 &
		1.000 &
		0.999 &
		\textcolor{red}{0.998} \\
		CPS* &
		1 &
		2 &
		3 &
		4 &
		mean &
		1 &
		2 &
		3 &
		4 &
		mean &
		1 &
		2 &
		3 &
		4 &
		mean &
		1 &
		2 &
		3 &
		4 &
		mean &
		\\
		Dice &
		89.827 &
		89.562 &
		76.114 &
		47.121 &
		75.656 &
		73.555 &
		76.436 &
		93.895 &
		68.271 &
		78.039 &
		59.407 &
		90.512 &
		75.327 &
		37.740 &
		65.747 &
		80.476 &
		88.199 &
		75.757 &
		58.339 &
		75.693 &
		73.784 \\
		
		HD &
		0.805 &
		1.158 &
		0.660 &
		0.720 &
		0.836 &
		2.894 &
		5.609 &
		0.958 &
		1.511 &
		2.743 &
		1.710 &
		1.783 &
		0.917 &
		1.208 &
		1.404 &
		1.564 &
		2.185 &
		1.148 &
		1.092 &
		1.498 &
		1.620 \\
		JC &
		0.164 &
		0.218 &
		0.106 &
		0.050 &
		0.135 &
		0.303 &
		0.434 &
		0.291 &
		0.176 &
		0.301 &
		0.122 &
		0.252 &
		0.118 &
		0.047 &
		0.135 &
		0.201 &
		0.336 &
		0.160 &
		0.091 &
		0.197 &
		0.192 \\
		
		RVE &
		0.068 &
		0.056 &
		0.208 &
		0.313 &
		0.161 &
		0.306 &
		1.917 &
		0.078 &
		0.221 &
		0.630 &
		0.419 &
		0.090 &
		0.221 &
		0.421 &
		0.288 &
		0.268 &
		0.073 &
		0.135 &
		0.201 &
		0.169 &
		0.312 \\
		Sen &
		0.884 &
		0.883 &
		0.698 &
		0.423 &
		0.722 &
		0.699 &
		0.773 &
		0.933 &
		0.675 &
		0.770 &
		0.576 &
		0.901 &
		0.710 &
		0.346 &
		0.633 &
		0.820 &
		0.891 &
		0.734 &
		0.587 &
		0.758 &
		0.721 \\
		
		Spe &
		1.000 &
		0.999 &
		1.000 &
		1.000 &
		1.000 &
		0.999 &
		0.934 &
		1.000 &
		0.999 &
		0.983 &
		1.000 &
		0.997 &
		1.000 &
		1.000 &
		0.999 &
		0.999 &
		0.997 &
		1.000 &
		1.000 &
		0.999 &
		0.995 \\
		MS-KD* &
		1 &
		2 &
		3 &
		4 &
		mean &
		1 &
		2 &
		3 &
		4 &
		mean &
		1 &
		2 &
		3 &
		4 &
		mean &
		1 &
		2 &
		3 &
		4 &
		mean &
		\\
		Dice &
		90.098 &
		88.623 &
		72.650 &
		46.607 &
		74.494 &
		72.711 &
		72.907 &
		90.868 &
		66.194 &
		75.670 &
		55.131 &
		89.829 &
		69.580 &
		38.031 &
		63.143 &
		77.887 &
		86.950 &
		71.788 &
		55.553 &
		73.044 &
		71.588 \\
		
		HD &
		0.738 &
		1.182 &
		0.705 &
		0.750 &
		0.844 &
		2.868 &
		5.687 &
		1.140 &
		1.555 &
		2.812 &
		1.748 &
		1.743 &
		0.984 &
		1.249 &
		1.431 &
		1.562 &
		2.213 &
		1.171 &
		1.106 &
		1.513 &
		1.650 \\
		JC &
		0.164 &
		0.215 &
		0.100 &
		0.050 &
		0.132 &
		0.298 &
		0.416 &
		0.278 &
		0.169 &
		0.290 &
		0.113 &
		0.249 &
		0.108 &
		0.047 &
		0.129 &
		0.193 &
		0.330 &
		0.151 &
		0.085 &
		0.190 &
		0.185 \\
		
		RVE &
		0.069 &
		0.072 &
		0.247 &
		0.289 &
		0.169 &
		0.333 &
		1.743 &
		0.147 &
		0.240 &
		0.616 &
		0.468 &
		0.087 &
		0.238 &
		0.397 &
		0.298 &
		0.218 &
		0.081 &
		0.153 &
		0.239 &
		0.173 &
		0.314 \\
		Sen &
		0.895 &
		0.860 &
		0.655 &
		0.437 &
		0.712 &
		0.686 &
		0.725 &
		0.903 &
		0.663 &
		0.744 &
		0.535 &
		0.877 &
		0.644 &
		0.362 &
		0.604 &
		0.780 &
		0.858 &
		0.696 &
		0.571 &
		0.726 &
		0.697 \\
		
		Spe &
		1.000 &
		0.999 &
		1.000 &
		1.000 &
		1.000 &
		0.999 &
		0.939 &
		1.000 &
		0.999 &
		0.984 &
		1.000 &
		0.998 &
		1.000 &
		1.000 &
		0.999 &
		0.999 &
		0.997 &
		1.000 &
		1.000 &
		0.999 &
		0.996\\
		FedAvg* &
		1 &
		2 &
		3 &
		4 &
		mean &
		1 &
		2 &
		3 &
		4 &
		mean &
		1 &
		2 &
		3 &
		4 &
		mean &
		1 &
		2 &
		3 &
		4 &
		mean &
		\\
		Dice &
		90.278 &
		86.680 &
		73.132 &
		51.525 &
		75.404 &
		73.066 &
		77.390 &
		94.021 &
		66.579 &
		77.764 &
		58.541 &
		88.700 &
		68.305 &
		42.171 &
		64.429 &
		79.725 &
		88.599 &
		75.677 &
		55.678 &
		74.920 &
		73.129 \\
		HD &
		0.727 &
		1.443 &
		0.677 &
		0.730 &
		0.894 &
		2.652 &
		5.732 &
		0.942 &
		1.500 &
		2.707 &
		1.613 &
		2.122 &
		1.090 &
		1.088 &
		1.478 &
		1.558 &
		2.252 &
		1.065 &
		1.097 &
		1.493 &
		1.643 \\
		JC &
		0.165 &
		0.207 &
		0.101 &
		0.056 &
		0.132 &
		0.298 &
		0.440 &
		0.292 &
		0.170 &
		0.300 &
		0.119 &
		0.242 &
		0.107 &
		0.053 &
		0.130 &
		0.198 &
		0.337 &
		0.159 &
		0.086 &
		0.195 &
		0.189 \\
		RVE &
		0.060 &
		0.126 &
		0.230 &
		0.212 &
		0.157 &
		0.314 &
		1.992 &
		0.063 &
		0.247 &
		0.654 &
		0.419 &
		0.176 &
		0.333 &
		0.374 &
		0.326 &
		0.251 &
		0.081 &
		0.119 &
		0.215 &
		0.167 &
		0.326 \\
		Sen &
		0.897 &
		0.892 &
		0.677 &
		0.479 &
		0.736 &
		0.690 &
		0.787 &
		0.926 &
		0.642 &
		0.761 &
		0.565 &
		0.920 &
		0.639 &
		0.393 &
		0.629 &
		0.806 &
		0.905 &
		0.734 &
		0.555 &
		0.750 &
		0.719 \\
		Spe &
		1.000 &
		0.998 &
		1.000 &
		1.000 &
		0.999 &
		0.999 &
		0.931 &
		1.000 &
		1.000 &
		0.983 &
		1.000 &
		0.995 &
		1.000 &
		1.000 &
		0.998 &
		0.999 &
		0.996 &
		1.000 &
		1.000 &
		0.999 &
		0.995 \\
		FedProx* &
		1 &
		2 &
		3 &
		4 &
		mean &
		1 &
		2 &
		3 &
		4 &
		mean &
		1 &
		2 &
		3 &
		4 &
		mean &
		1 &
		2 &
		3 &
		4 &
		mean &
		\\
		Dice &
		89.571 &
		86.139 &
		73.563 &
		51.262 &
		75.134 &
		70.363 &
		74.443 &
		94.310 &
		68.048 &
		76.791 &
		55.753 &
		89.735 &
		72.597 &
		40.929 &
		64.754 &
		79.412 &
		86.976 &
		76.844 &
		59.107 &
		75.585 &
		73.066 \\
		
		HD &
		0.787 &
		1.550 &
		0.672 &
		0.715 &
		0.931 &
		2.781 &
		5.726 &
		0.973 &
		1.508 &
		2.747 &
		1.644 &
		1.871 &
		1.081 &
		1.076 &
		1.418 &
		1.513 &
		2.240 &
		1.145 &
		1.065 &
		1.491 &
		1.647 \\
		JC &
		0.163 &
		0.205 &
		0.101 &
		0.056 &
		0.131 &
		0.285 &
		0.424 &
		0.293 &
		0.175 &
		0.294 &
		0.113 &
		0.247 &
		0.114 &
		0.051 &
		0.131 &
		0.199 &
		0.330 &
		0.163 &
		0.093 &
		0.196 &
		0.188 \\
		
		RVE &
		0.073 &
		0.151 &
		0.236 &
		0.260 &
		0.180 &
		0.375 &
		1.826 &
		0.055 &
		0.252 &
		0.627 &
		0.466 &
		0.108 &
		0.247 &
		0.399 &
		0.305 &
		0.233 &
		0.075 &
		0.115 &
		0.220 &
		0.161 &
		0.318 \\
		Sen &
		0.876 &
		0.886 &
		0.672 &
		0.475 &
		0.727 &
		0.657 &
		0.751 &
		0.939 &
		0.677 &
		0.756 &
		0.533 &
		0.900 &
		0.683 &
		0.382 &
		0.625 &
		0.796 &
		0.872 &
		0.756 &
		0.608 &
		0.758 &
		0.717 \\
		
		Spe &
		1.000 &
		0.998 &
		1.000 &
		1.000 &
		0.999 &
		1.000 &
		0.937 &
		1.000 &
		0.999 &
		0.984 &
		1.000 &
		0.997 &
		1.000 &
		1.000 &
		0.999 &
		0.999 &
		0.997 &
		1.000 &
		1.000 &
		0.999 &
		0.995\\
		MOON* &
		1 &
		2 &
		3 &
		4 &
		mean &
		1 &
		2 &
		3 &
		4 &
		mean &
		1 &
		2 &
		3 &
		4 &
		mean &
		1 &
		2 &
		3 &
		4 &
		mean &
		\\
		Dice &
		90.157 &
		88.650 &
		79.007 &
		42.719 &
		75.133 &
		73.336 &
		75.883 &
		93.420 &
		68.633 &
		77.818 &
		58.531 &
		89.054 &
		72.496 &
		37.069 &
		64.288 &
		80.056 &
		86.124 &
		75.664 &
		59.187 &
		75.258 &
		73.124 \\
		
		HD &
		0.824 &
		1.198 &
		0.606 &
		0.760 &
		0.847 &
		2.665 &
		5.586 &
		0.954 &
		1.520 &
		2.681 &
		1.705 &
		1.868 &
		1.011 &
		1.105 &
		1.422 &
		1.616 &
		2.256 &
		1.112 &
		1.056 &
		1.510 &
		1.615 \\
		JC &
		0.164 &
		0.215 &
		0.110 &
		0.045 &
		0.134 &
		0.301 &
		0.430 &
		0.289 &
		0.177 &
		0.299 &
		0.120 &
		0.244 &
		0.113 &
		0.046 &
		0.131 &
		0.200 &
		0.326 &
		0.159 &
		0.093 &
		0.195 &
		0.190 \\
		
		RVE &
		0.088 &
		0.062 &
		0.168 &
		0.348 &
		0.166 &
		0.312 &
		1.798 &
		0.074 &
		0.241 &
		0.606 &
		0.442 &
		0.106 &
		0.264 &
		0.410 &
		0.306 &
		0.304 &
		0.075 &
		0.153 &
		0.237 &
		0.192 &
		0.318 \\
		Sen &
		0.897 &
		0.869 &
		0.734 &
		0.385 &
		0.721 &
		0.702 &
		0.759 &
		0.924 &
		0.684 &
		0.767 &
		0.572 &
		0.891 &
		0.683 &
		0.337 &
		0.621 &
		0.825 &
		0.859 &
		0.743 &
		0.595 &
		0.756 &
		0.716 \\
		
		Spe &
		1.000 &
		0.999 &
		1.000 &
		1.000 &
		1.000 &
		0.999 &
		0.938 &
		1.000 &
		0.999 &
		0.984 &
		1.000 &
		0.997 &
		1.000 &
		1.000 &
		0.999 &
		0.999 &
		0.997 &
		1.000 &
		1.000 &
		0.999 &
		0.996\\
		FedAlign* &
		1 &
		2 &
		3 &
		4 &
		mean &
		1 &
		2 &
		3 &
		4 &
		mean &
		1 &
		2 &
		3 &
		4 &
		mean &
		1 &
		2 &
		3 &
		4 &
		mean &
		\\
		Dice &
		89.234 &
		88.484 &
		74.082 &
		52.271 &
		76.018 &
		71.218 &
		74.609 &
		94.943 &
		66.882 &
		76.913 &
		55.168 &
		89.198 &
		73.160 &
		38.166 &
		63.923 &
		80.834 &
		88.516 &
		76.812 &
		58.904 &
		76.267 &
		73.280 \\
		
		HD &
		0.737 &
		1.224 &
		0.660 &
		0.709 &
		0.833 &
		2.728 &
		5.815 &
		0.941 &
		1.560 &
		2.761 &
		1.660 &
		1.931 &
		0.950 &
		1.092 &
		1.408 &
		1.479 &
		2.191 &
		1.060 &
		1.059 &
		1.447 &
		\textcolor{red}{1.612} \\
		JC &
		0.161 &
		0.213 &
		0.102 &
		0.057 &
		0.133 &
		0.281 &
		0.419 &
		0.292 &
		0.168 &
		0.290 &
		0.111 &
		0.244 &
		0.115 &
		0.047 &
		0.129 &
		0.202 &
		0.338 &
		0.162 &
		0.092 &
		0.198 &
		0.188 \\
		
		RVE &
		0.078 &
		0.072 &
		0.221 &
		0.280 &
		0.163 &
		0.395 &
		1.875 &
		0.071 &
		0.265 &
		0.652 &
		0.464 &
		0.115 &
		0.215 &
		0.448 &
		0.311 &
		0.185 &
		0.065 &
		0.124 &
		0.235 &
		0.152 &
		0.320 \\
		Sen &
		0.866 &
		0.873 &
		0.677 &
		0.479 &
		0.724 &
		0.671 &
		0.763 &
		0.959 &
		0.683 &
		0.769 &
		0.521 &
		0.901 &
		0.694 &
		0.356 &
		0.618 &
		0.803 &
		0.888 &
		0.752 &
		0.601 &
		0.761 &
		0.718 \\
		
		Spe &
		1.000 &
		0.999 &
		1.000 &
		1.000 &
		1.000 &
		1.000 &
		0.935 &
		1.000 &
		0.999 &
		0.983 &
		1.000 &
		0.997 &
		1.000 &
		1.000 &
		0.999 &
		0.999 &
		0.997 &
		1.000 &
		1.000 &
		0.999 &
		0.995\\
		FedASAM* &
		1 &
		2 &
		3 &
		4 &
		mean &
		1 &
		2 &
		3 &
		4 &
		mean &
		1 &
		2 &
		3 &
		4 &
		mean &
		1 &
		2 &
		3 &
		4 &
		mean &
		\\
		Dice &
		89.817 &
		88.664 &
		76.178 &
		57.691 &
		78.087 &
		71.096 &
		78.939 &
		93.439 &
		68.168 &
		77.910 &
		56.006 &
		90.685 &
		69.427 &
		45.598 &
		65.429 &
		78.840 &
		88.049 &
		73.625 &
		60.937 &
		75.363 &
		74.197 \\
		HD &
		0.830 &
		1.230 &
		0.654 &
		0.721 &
		0.859 &
		2.864 &
		5.473 &
		0.976 &
		1.638 &
		2.738 &
		1.625 &
		1.840 &
		0.930 &
		1.134 &
		1.382 &
		1.536 &
		2.252 &
		1.191 &
		1.143 &
		1.531 &
		1.628 \\
		JC &
		0.163 &
		0.215 &
		0.106 &
		0.063 &
		0.136 &
		0.288 &
		0.447 &
		0.289 &
		0.174 &
		0.300 &
		0.114 &
		0.251 &
		0.109 &
		0.057 &
		0.133 &
		0.196 &
		0.335 &
		0.154 &
		0.095 &
		0.195 &
		0.191 \\
		RVE &
		0.087 &
		0.074 &
		0.214 &
		0.157 &
		0.133 &
		0.359 &
		1.809 &
		0.089 &
		0.292 &
		0.637 &
		0.459 &
		0.120 &
		0.280 &
		0.342 &
		0.300 &
		0.241 &
		0.076 &
		0.148 &
		0.243 &
		0.177 &
		0.312 \\
		Sen &
		0.885 &
		0.886 &
		0.703 &
		0.562 &
		0.759 &
		0.669 &
		0.794 &
		0.921 &
		0.737 &
		0.780 &
		0.539 &
		0.919 &
		0.649 &
		0.463 &
		0.643 &
		0.792 &
		0.894 &
		0.714 &
		0.664 &
		0.766 &
		0.737 \\
		Spe &
		1.000 &
		0.999 &
		1.000 &
		1.000 &
		0.999 &
		1.000 &
		0.938 &
		1.000 &
		0.999 &
		0.984 &
		1.000 &
		0.997 &
		1.000 &
		1.000 &
		0.999 &
		0.999 &
		0.996 &
		1.000 &
		0.999 &
		0.999 &
		0.995 \\
		UFPS(Ours) &
		1 &
		2 &
		3 &
		4 &
		mean &
		1 &
		2 &
		3 &
		4 &
		mean &
		1 &
		2 &
		3 &
		4 &
		mean &
		1 &
		2 &
		3 &
		4 &
		mean &
		\\
		Dice &
		90.203 &
		87.908 &
		76.880 &
		53.169 &
		77.040 &
		76.509 &
		78.394 &
		93.868 &
		68.789 &
		79.390 &
		60.630 &
		89.088 &
		75.939 &
		43.346 &
		67.251 &
		81.274 &
		89.099 &
		77.232 &
		62.164 &
		77.442 &
		\textcolor{red}{75.281} \\
		HD &
		0.794 &
		1.236 &
		0.678 &
		0.692 &
		0.850 &
		2.527 &
		5.545 &
		0.967 &
		1.587 &
		2.657 &
		1.726 &
		2.057 &
		0.980 &
		1.082 &
		1.461 &
		1.560 &
		2.223 &
		1.135 &
		1.101 &
		1.505 &
		1.618 \\
		JC &
		0.164 &
		0.212 &
		0.107 &
		0.057 &
		0.135 &
		0.316 &
		0.445 &
		0.291 &
		0.176 &
		0.307 &
		0.125 &
		0.245 &
		0.119 &
		0.055 &
		0.136 &
		0.204 &
		0.340 &
		0.164 &
		0.097 &
		0.201 &
		\textcolor{red}{0.195} \\
		RVE &
		0.071 &
		0.084 &
		0.212 &
		0.207 &
		0.143 &
		0.273 &
		1.901 &
		0.064 &
		0.255 &
		0.623 &
		0.421 &
		0.141 &
		0.207 &
		0.384 &
		0.288 &
		0.312 &
		0.077 &
		0.100 &
		0.168 &
		0.164 &
		0.305 \\
		Sen &
		0.911 &
		0.882 &
		0.741 &
		0.517 &
		0.763 &
		0.737 &
		0.795 &
		0.931 &
		0.711 &
		0.793 &
		0.595 &
		0.914 &
		0.725 &
		0.418 &
		0.663 &
		0.841 &
		0.903 &
		0.758 &
		0.650 &
		0.788 &
		\textcolor{red}{0.752} \\
		Spe &
		1.000 &
		0.998 &
		1.000 &
		1.000 &
		0.999 &
		0.999 &
		0.935 &
		1.000 &
		0.999 &
		0.983 &
		1.000 &
		0.996 &
		1.000 &
		1.000 &
		0.999 &
		0.999 &
		0.996 &
		1.000 &
		1.000 &
		0.999 &
		0.995 \\
		\hline 
		\rowcolor{white}\multicolumn{22}{l}{Organ indexes are numbers on the straight right side of the name of method.} \\
		\rowcolor{white}\multicolumn{22}{l}{Model performance is better when metrics are higher, except for HD and RVE.} \\
		\rowcolor{white}\multicolumn{22}{l}{The best method (except Centralized full) for each metric is marked in red.} \\
		\hline

		\end{tabular}
	}
\end{table*}

\section{More ablation study}

\textbf{Ablation for adaptive RCE.} From \hyperref[Table S4]{Table S4}, it can be seen that model performance is better when the coefficient for aRCE loss is set to a moderate value. When it is set too large, model prediction can be dominant when it is not reliable enough compared to pseudo labels. Otherwise, noises in pseudo labels are not sufficiently alleviated.

\begin{table*}[]
\label{Table S4}
	
	\centering
	\rowcolors {3}{grey}{white}
	\renewcommand\arraystretch{1.2}
	\resizebox{\columnwidth}{!}{
		\begin{tabular}{lllllllllllllllllllllll}
		\arrayrulecolor{red}
	
		\hline
		\multicolumn{22}{l}{\textcolor{red}{\textbf{Table S4. Ablation study on aRCE loss.}}}  \\
		\hline
		
			& \multicolumn{21}{l}{\textcolor{red}{Client index (the next line) \& Organ index   (after name of method)}} \\ 
		\cmidrule[0.5mm]{2 -21}
		
		\multirow{-2}{*}{\textcolor{red}{Strategy \& Metric}} &
		\multicolumn{5}{l}{\textcolor{red}{1}} &
		\multicolumn{5}{l}{\textcolor{red}{2}} &
		\multicolumn{5}{l}{\textcolor{red}{3}} &
		\multicolumn{5}{l}{\textcolor{red}{4}} &
		\textcolor{red}{Mean}  \\ 
		\hline
		
		Fixed &
		1 &
		2 &
		3 &
		4 &
		mean &
		1 &
		2 &
		3 &
		4 &
		mean &
		1 &
		2 &
		3 &
		4 &
		mean &
		1 &
		2 &
		3 &
		4 &
		mean &
		&
		\\
		Dice &
		89.303 &
		87.873 &
		73.221 &
		50.823 &
		75.305 &
		71.233 &
		74.049 &
		91.461 &
		66.335 &
		75.769 &
		59.378 &
		88.175 &
		66.823 &
		42.846 &
		64.306 &
		80.671 &
		85.775 &
		74.408 &
		59.726 &
		75.145 &
		72.631 &
		\\
		HD &
		0.998 &
		1.610 &
		0.874 &
		0.782 &
		1.066 &
		2.928 &
		6.232 &
		1.001 &
		1.556 &
		2.929 &
		2.082 &
		2.217 &
		1.272 &
		1.190 &
		1.690 &
		1.714 &
		2.413 &
		1.196 &
		1.161 &
		1.621 &
		\textcolor{red}{1.827} &
		\\
		JC &
		0.162 &
		0.211 &
		0.100 &
		0.054 &
		0.132 &
		0.289 &
		0.416 &
		0.281 &
		0.169 &
		0.289 &
		0.120 &
		0.241 &
		0.104 &
		0.054 &
		0.130 &
		0.202 &
		0.324 &
		0.157 &
		0.093 &
		0.194 &
		0.186 &
		\\
		RVE &
		0.090 &
		0.075 &
		0.224 &
		0.257 &
		0.161 &
		0.396 &
		1.999 &
		0.118 &
		0.265 &
		0.694 &
		0.404 &
		0.148 &
		0.266 &
		0.360 &
		0.294 &
		0.242 &
		0.082 &
		0.130 &
		0.198 &
		0.163 &
		\textcolor{red}{0.328} &
		\\
		Sen &
		0.887 &
		0.878 &
		0.666 &
		0.470 &
		0.725 &
		0.670 &
		0.732 &
		0.888 &
		0.642 &
		0.733 &
		0.575 &
		0.884 &
		0.620 &
		0.403 &
		0.620 &
		0.816 &
		0.857 &
		0.714 &
		0.613 &
		0.750 &
		0.707 &
		\\
		Spe &
		1.000 &
		0.998 &
		1.000 &
		1.000 &
		0.999 &
		0.999 &
		0.930 &
		1.000 &
		1.000 &
		0.982 &
		0.999 &
		0.996 &
		1.000 &
		1.000 &
		0.999 &
		0.999 &
		0.996 &
		1.000 &
		1.000 &
		0.999 &
		\textcolor{red}{0.995} &
		\\
		Increasing &
		1 &
		2 &
		3 &
		4 &
		mean &
		1 &
		2 &
		3 &
		4 &
		mean &
		1 &
		2 &
		3 &
		4 &
		mean &
		1 &
		2 &
		3 &
		4 &
		mean &
		 &
		\\
		Dice &
		90.001 &
		84.213 &
		75.411 &
		52.485 &
		75.527 &
		73.361 &
		78.513 &
		93.009 &
		65.937 &
		77.705 &
		59.755 &
		87.164 &
		69.408 &
		45.136 &
		65.366 &
		80.774 &
		87.242 &
		73.534 &
		57.458 &
		74.752 &
		\textcolor{red}{73.338} &
		\\
		HD &
		0.893 &
		2.347 &
		0.786 &
		0.895 &
		1.230 &
		2.839 &
		6.107 &
		0.994 &
		1.559 &
		2.875 &
		1.974 &
		2.504 &
		1.400 &
		1.261 &
		1.785 &
		1.782 &
		2.536 &
		1.330 &
		1.224 &
		1.718 &
		1.902 &
		\\
		JC &
		0.164 &
		0.200 &
		0.104 &
		0.057 &
		0.131 &
		0.299 &
		0.443 &
		0.287 &
		0.168 &
		0.299 &
		0.122 &
		0.238 &
		0.109 &
		0.057 &
		0.131 &
		0.202 &
		0.330 &
		0.154 &
		0.089 &
		0.194 &
		\textcolor{red}{0.189} &
		\\
		RVE &
		0.070 &
		0.146 &
		0.201 &
		0.194 &
		0.153 &
		0.291 &
		2.238 &
		0.076 &
		0.238 &
		0.711 &
		0.395 &
		0.193 &
		0.277 &
		0.311 &
		0.294 &
		0.286 &
		0.088 &
		0.154 &
		0.205 &
		0.183 &
		0.335 &
		\\
		Sen &
		0.898 &
		0.874 &
		0.691 &
		0.494 &
		0.739 &
		0.697 &
		0.791 &
		0.910 &
		0.642 &
		0.760 &
		0.580 &
		0.904 &
		0.652 &
		0.428 &
		0.641 &
		0.822 &
		0.889 &
		0.709 &
		0.586 &
		0.752 &
		\textcolor{red}{0.723} &
		\\
		Spe &
		1.000 &
		0.997 &
		1.000 &
		1.000 &
		0.999 &
		0.999 &
		0.924 &
		1.000 &
		0.999 &
		0.981 &
		1.000 &
		0.995 &
		1.000 &
		1.000 &
		0.998 &
		0.999 &
		0.995 &
		1.000 &
		1.000 &
		0.998 &
		0.994 & \setcounter{rownum}{0}
		\\
		\hline

		& \multicolumn{21}{l}{\textcolor{red}{Client index (the next line) \& Organ index   (after name of method)}} \setcounter{rownum}{0}\\ 
		\cmidrule[0.5mm]{2 -21}
		
		\multirow{-2}{*}{\textcolor{red}{Coefficient \& Metric}} &
		\multicolumn{5}{l}{\textcolor{red}{1}} &
		\multicolumn{5}{l}{\textcolor{red}{2}} &
		\multicolumn{5}{l}{\textcolor{red}{3}} &
		\multicolumn{5}{l}{\textcolor{red}{4}} &
		\textcolor{red}{Mean}  \\ 
		\hline

		0.1 &
		1 &
		2 &
		3 &
		4 &
		mean &
		1 &
		2 &
		3 &
		4 &
		mean &
		1 &
		2 &
		3 &
		4 &
		mean &
		1 &
		2 &
		3 &
		4 &
		mean &
		&
		\\
		Dice &
		88.956 &
		87.513 &
		73.106 &
		47.670 &
		74.311 &
		70.003 &
		77.757 &
		91.819 &
		64.715 &
		76.074 &
		55.920 &
		88.767 &
		73.197 &
		36.014 &
		63.474 &
		79.886 &
		87.630 &
		75.264 &
		57.077 &
		74.964 &
		72.206 &
		\\
		HD &
		0.925 &
		2.081 &
		0.912 &
		0.757 &
		1.169 &
		2.885 &
		6.096 &
		1.054 &
		1.549 &
		2.896 &
		1.946 &
		2.575 &
		1.737 &
		1.114 &
		1.843 &
		1.724 &
		2.447 &
		1.334 &
		1.118 &
		1.656 &
		1.891 &
		\\
		JC &
		0.161 &
		0.210 &
		0.100 &
		0.051 &
		0.130 &
		0.282 &
		0.438 &
		0.282 &
		0.164 &
		0.292 &
		0.113 &
		0.243 &
		0.114 &
		0.044 &
		0.128 &
		0.200 &
		0.332 &
		0.157 &
		0.088 &
		0.194 &
		0.186 &
		\\
		RVE &
		0.076 &
		0.093 &
		0.223 &
		0.360 &
		0.188 &
		0.376 &
		2.047 &
		0.093 &
		0.290 &
		0.702 &
		0.452 &
		0.153 &
		0.331 &
		0.460 &
		0.349 &
		0.237 &
		0.082 &
		0.159 &
		0.226 &
		0.176 &
		0.354 &
		\\
		Sen &
		0.872 &
		0.890 &
		0.660 &
		0.421 &
		0.711 &
		0.649 &
		0.777 &
		0.891 &
		0.617 &
		0.733 &
		0.533 &
		0.903 &
		0.683 &
		0.318 &
		0.609 &
		0.796 &
		0.886 &
		0.719 &
		0.557 &
		0.739 &
		0.698 &
		\\
		Spe &
		1.000 &
		0.998 &
		1.000 &
		1.000 &
		0.999 &
		0.999 &
		0.929 &
		1.000 &
		1.000 &
		0.982 &
		1.000 &
		0.996 &
		0.999 &
		1.000 &
		0.999 &
		0.999 &
		0.996 &
		1.000 &
		1.000 &
		0.999 &
		\textcolor{red}{0.995} &
		\\
		0.01 &
		1 &
		2 &
		3 &
		4 &
		mean &
		1 &
		2 &
		3 &
		4 &
		mean &
		1 &
		2 &
		3 &
		4 &
		mean &
		1 &
		2 &
		3 &
		4 &
		mean &
		 &
		\\
		Dice &
		90.001 &
		84.213 &
		75.411 &
		52.485 &
		75.527 &
		73.361 &
		78.513 &
		93.009 &
		65.937 &
		77.705 &
		59.755 &
		87.164 &
		69.408 &
		45.136 &
		65.366 &
		80.774 &
		87.242 &
		73.534 &
		57.458 &
		74.752 &
		\textcolor{red}{73.338} &
		\\
		HD &
		0.893 &
		2.347 &
		0.786 &
		0.895 &
		1.230 &
		2.839 &
		6.107 &
		0.994 &
		1.559 &
		2.875 &
		1.974 &
		2.504 &
		1.400 &
		1.261 &
		1.785 &
		1.782 &
		2.536 &
		1.330 &
		1.224 &
		1.718 &
		1.902 &
		\\
		JC &
		0.164 &
		0.200 &
		0.104 &
		0.057 &
		0.131 &
		0.299 &
		0.443 &
		0.287 &
		0.168 &
		0.299 &
		0.122 &
		0.238 &
		0.109 &
		0.057 &
		0.131 &
		0.202 &
		0.330 &
		0.154 &
		0.089 &
		0.194 &
		\textcolor{red}{0.189} &
		\\
		RVE &
		0.070 &
		0.146 &
		0.201 &
		0.194 &
		0.153 &
		0.291 &
		2.238 &
		0.076 &
		0.238 &
		0.711 &
		0.395 &
		0.193 &
		0.277 &
		0.311 &
		0.294 &
		0.286 &
		0.088 &
		0.154 &
		0.205 &
		0.183 &
		0.335 &
		\\
		Sen &
		0.898 &
		0.874 &
		0.691 &
		0.494 &
		0.739 &
		0.697 &
		0.791 &
		0.910 &
		0.642 &
		0.760 &
		0.580 &
		0.904 &
		0.652 &
		0.428 &
		0.641 &
		0.822 &
		0.889 &
		0.709 &
		0.586 &
		0.752 &
		\textcolor{red}{0.723} &
		\\
		Spe &
		1.000 &
		0.997 &
		1.000 &
		1.000 &
		0.999 &
		0.999 &
		0.924 &
		1.000 &
		0.999 &
		0.981 &
		1.000 &
		0.995 &
		1.000 &
		1.000 &
		0.998 &
		0.999 &
		0.995 &
		1.000 &
		1.000 &
		0.998 &
		0.994 &
		\\
		0.001 &
		1 &
		2 &
		3 &
		4 &
		mean &
		1 &
		2 &
		3 &
		4 &
		mean &
		1 &
		2 &
		3 &
		4 &
		mean &
		1 &
		2 &
		3 &
		4 &
		mean &
		&
		\\
		Dice &
		89.363 &
		86.046 &
		75.599 &
		50.741 &
		75.437 &
		71.975 &
		74.055 &
		91.625 &
		66.485 &
		76.035 &
		60.252 &
		87.589 &
		69.150 &
		41.460 &
		64.613 &
		80.935 &
		85.855 &
		75.241 &
		60.220 &
		75.563 &
		72.912 &
		\\
		HD &
		0.985 &
		1.913 &
		0.917 &
		0.810 &
		1.157 &
		2.996 &
		6.236 &
		1.035 &
		1.554 &
		2.955 &
		2.038 &
		2.228 &
		1.369 &
		1.245 &
		1.720 &
		1.775 &
		2.398 &
		1.279 &
		1.163 &
		1.654 &
		\textcolor{red}{1.872} &
		\\
		JC &
		0.162 &
		0.205 &
		0.105 &
		0.054 &
		0.132 &
		0.293 &
		0.416 &
		0.281 &
		0.170 &
		0.290 &
		0.122 &
		0.239 &
		0.107 &
		0.052 &
		0.130 &
		0.202 &
		0.324 &
		0.159 &
		0.094 &
		0.195 &
		0.187 &
		\\
		RVE &
		0.077 &
		0.117 &
		0.182 &
		0.249 &
		0.157 &
		0.352 &
		2.030 &
		0.108 &
		0.248 &
		0.685 &
		0.402 &
		0.149 &
		0.259 &
		0.367 &
		0.294 &
		0.260 &
		0.078 &
		0.114 &
		0.207 &
		0.165 &
		\textcolor{red}{0.325} &
		\\
		Sen &
		0.891 &
		0.877 &
		0.700 &
		0.466 &
		0.734 &
		0.683 &
		0.732 &
		0.893 &
		0.651 &
		0.740 &
		0.584 &
		0.881 &
		0.644 &
		0.391 &
		0.625 &
		0.820 &
		0.855 &
		0.731 &
		0.617 &
		0.756 &
		0.714 &
		\\
		Spe &
		1.000 &
		0.998 &
		1.000 &
		1.000 &
		0.999 &
		0.999 &
		0.930 &
		1.000 &
		0.999 &
		0.982 &
		0.999 &
		0.996 &
		1.000 &
		1.000 &
		0.999 &
		0.999 &
		0.997 &
		1.000 &
		1.000 &
		0.999 &
		\textcolor{red}{0.995} &
		\\
		\hline 
			\rowcolor{white}\multicolumn{22}{l}{Organ indexes are numbers on the straight right side of the name of method.} \\
			\rowcolor{white}\multicolumn{22}{l}{Model performance is better when metrics are higher, except for HD and RVE.} \\
		\hline
		
		\end{tabular}
	}
\end{table*}

\begin{figure}
	\centering
		\includegraphics[width=\columnwidth]{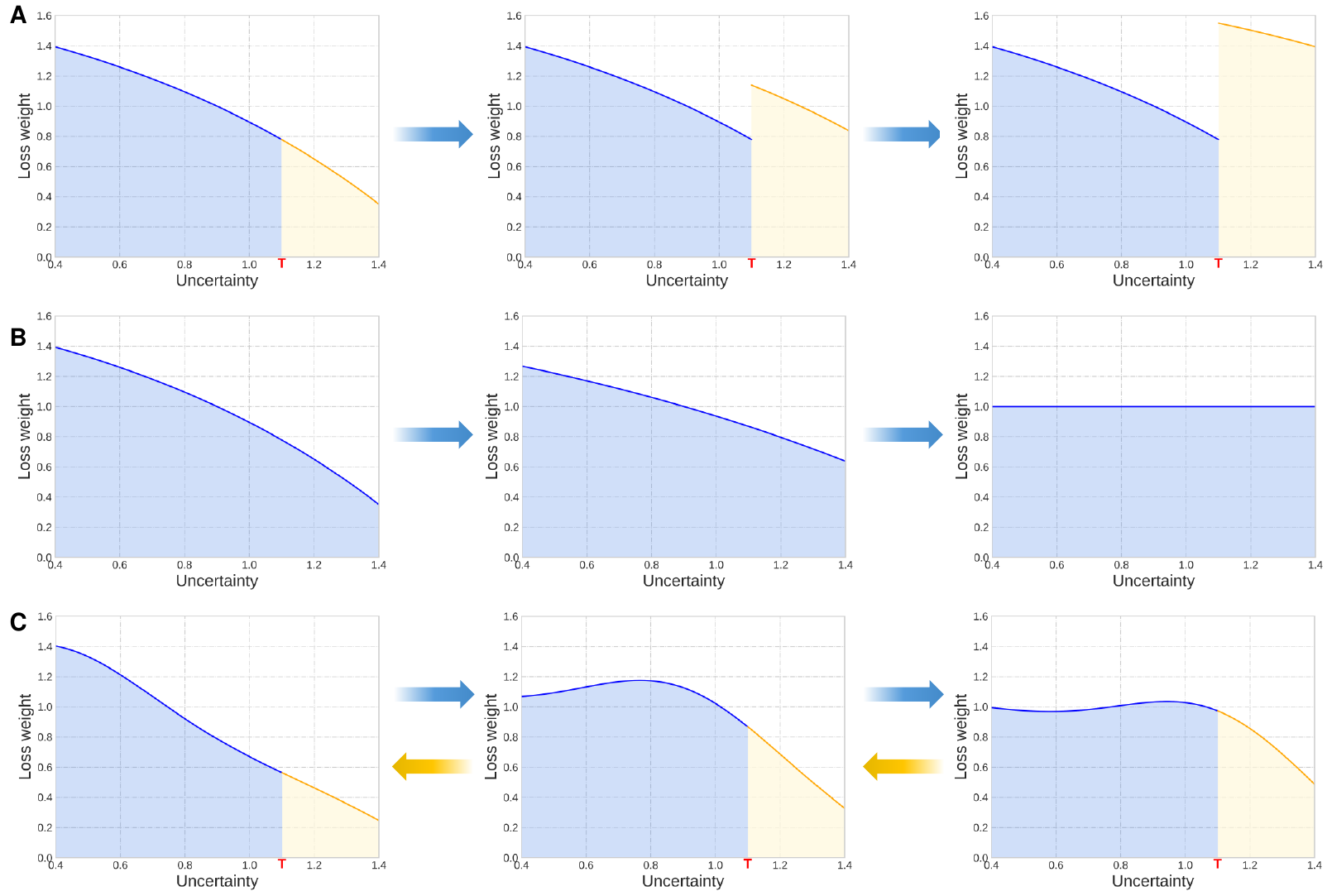}
	\caption{\textbf{Explanatory chart for three schedulers.}\\(A) Tail shift. \\(B) Base decrement. \\(C) Round-trip gaussian.}
	\label{Figure S1}
\end{figure}

\textbf{Ablation for Weight schedulers.} Here we propose three schedulers focusing on different ranges of uncertainty for various situations as shown in \hyperref[Figure S1]{Figure S1}.

\noindent \textbf{Tail shift (TS).}
$$
\begin{gathered}
w(U_j)=\left\{\begin{array}{c}
2-e^{\operatorname{norm}(U_j)-\frac{r}{R}}, U>U_{\mathcal{T}} \\
2-e^{\operatorname{norm}(U_j)}, \text { else }
\end{array}\right. ,~
\text{where} \quad \operatorname{norm}(U_j)=\frac{U_j-\mu}{U_{\max }-U_{\min }}
\label{eqS1}
\end{gathered}
$$

\noindent where $U_{\mathcal{T}}$ corresponds to the uncertainty value at the lowest $\mathcal{T}$ percentage. $\mu, U_{max}, U_{min}$ represent mean, maximal, minimal uncertainty in the uncertainty bank, respectively.

\noindent \textbf{Base Decrement (BD).}
$$
\begin{gathered}
w(U_j)=2-\left(\frac{\alpha-e}{R} r+e\right)^{n o r m(U_j)} ,~
\text{where} \quad \operatorname{norm}(U_j)=\frac{U_j-\mu}{U_{\max }-U_{\min }}
\label{eqS2}
\end{gathered}
$$

\noindent $\alpha$ is a hyper-parameter determining the minimal base, and we empirically set it to 1.

\noindent \textbf{Round-trip Gaussian (RG)}
$$
\begin{gathered}
w(U_j)=(1-\beta) \cdot G(U_j)+\beta \cdot\left(2-e^{\text {norm }(U_j)}\right) ,~ \\
\text{where} \quad G(U_j)=\frac{1}{\sqrt{2 \pi} \varrho} \cdot e^{-\frac{\left(\text { norm }(U_j)-\operatorname{norm}\left(\mu\right)-\text { Range }(r)\right)^2}{2 \varrho^2}}  ,~\\
\text { Range }(r)=\left\{\begin{array}{c}
\frac{U_{\text {range }} \cdot r}{\left\lfloor\frac{R}{2}\right\rfloor}, r \leq\left\lfloor\frac{R}{2}\right\rfloor \\
U_{\text {range }}-\frac{U_{\text {range }} \cdot\left(r-\left\lfloor\frac{R}{2}\right\rfloor\right)}{\left\lfloor\frac{R}{2}\right\rfloor}, \text { else }
\end{array}\right.  ,~\\
U_{\text {range }}=\operatorname{norm}\left(U_{\mathcal{T}}\right)-\operatorname{norm}\left(U_{\min }\right) ,~\\
\quad \operatorname{norm}(x)=\frac{x-\mu}{U_{\max }-U_{\min }},
\label{eqS3}
\end{gathered}
$$

\noindent where $\beta$ is a balancing factor, $\varrho$ is used to calibrate amplitude with other schedulers, $\lfloor\cdot\rfloor$ is a rounding operator. 

At the beginning of training, less uncertain samples are endowed with higher loss weight compared to dubious ones under any weight scheduler. Afterwards, TS puts more emphasis on least confident patches, which mainly consist of noisy pseudo labels and ones for hard classes. BD equally treats each part, which is equivalent to weight decreasement for the head part and increasement for the middle and tail parts. RG pays less attention to the tail part since the shift in the uncertainty axis is blocked before the threshold. In our experiment, the loss weight scheduler for any client is disabled during warm-up epochs $r_{warmup}$ until distribution for uncertainty can be approximately estimated. 

\begin{figure}
	\centering
		\includegraphics[width=\columnwidth]{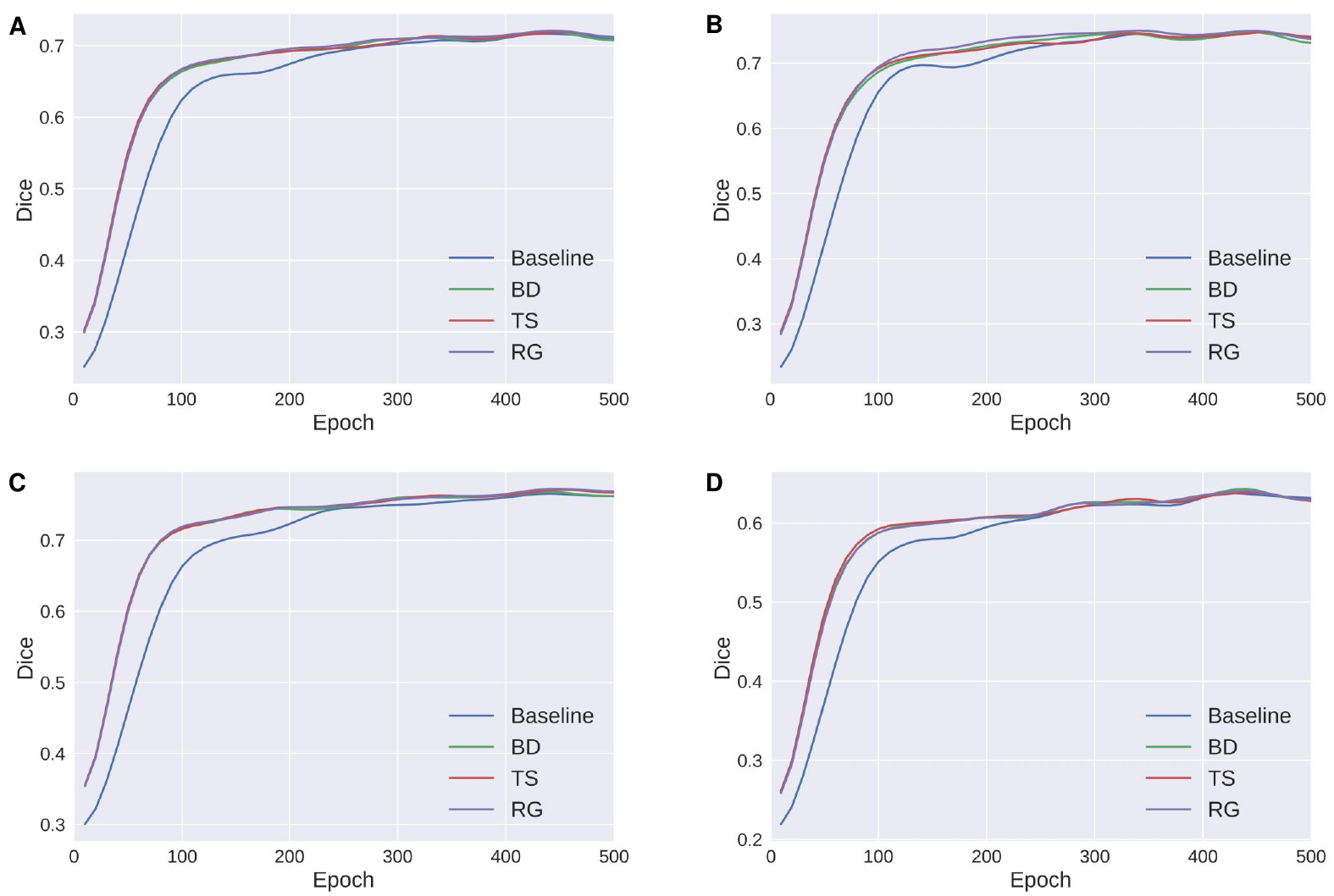}
	\caption{\textbf{Training curves for loss weight schedulers.}\\ (A) Testing dice during training for the mean of all clients.\\(B) Testing dice during training for the mean of client 1.\\(C) Testing dice during training for the mean of client 2.\\ (D) Testing dice during training for the mean of client 3.}
	\label{Figure S2}
\end{figure}

From \hyperref[Table S5]{Table S5}, it can be seen that TS, focusing on head and tail, achieves the best result in our experimental setting. The final epoch for weight scheduler should be decided depending on the fitting degree for different classes, which can be implicitly reflected from model performance for each class. Let us take TS as an example. As shown in \hyperref[Figure S2]{Figure S2}, model performance on the tail class, i.e., pancreas, is relatively stable after 150 epochs. Then, the larger weight on the tail class than head and middle classes should be controlled after this point. Through the results, we can find that prolonging the final epoch makes little difference as long as the noisy degree of tail classes is low. By using any one of these loss weight schedulers, the convergence speed is obviously accelerated due to concentration on uncertainty samples in previous rounds in \hyperref[Figure S2]{Figure S2} and exploration of hard but valuable ones later.

\begin{table*}[]
\label{Table S5}
	\centering
	\rowcolors {3}{grey}{white}
	\renewcommand\arraystretch{1.2}
	\resizebox{\columnwidth}{!}{
		\centering
		\begin{tabular}{llllllllllllllllllllll}
		\arrayrulecolor{red}

		\hline
		\multicolumn{22}{l}{\textcolor{red}{\textbf{Table S5. Ablation study on weight scheduler.}}}  \\
		\hline
		
			& \multicolumn{21}{l}{\textcolor{red}{Client index (the next line) \& Organ index   (after name of method)}} \\ 
		\cmidrule[0.5mm]{2 -21}
		
		\multirow{-2}{*}{\textcolor{red}{Strategy \& Metric}} &
		\multicolumn{5}{l}{\textcolor{red}{1}} &
		\multicolumn{5}{l}{\textcolor{red}{2}} &
		\multicolumn{5}{l}{\textcolor{red}{3}} &
		\multicolumn{5}{l}{\textcolor{red}{4}} &
		\textcolor{red}{Mean}  \\ 
		\hline
		
		TS &
		1 &
		2 &
		3 &
		4 &
		mean &
		1 &
		2 &
		3 &
		4 &
		mean &
		1 &
		2 &
		3 &
		4 &
		mean &
		1 &
		2 &
		3 &
		4 &
		mean &
		\\
		Dice &
		90.162 &
		87.442 &
		76.888 &
		49.430 &
		75.980 &
		72.955 &
		76.875 &
		94.070 &
		69.066 &
		78.242 &
		58.440 &
		89.017 &
		75.613 &
		42.264 &
		66.334 &
		80.019 &
		86.902 &
		76.442 &
		62.072 &
		76.359 &
		\textcolor{red}{74.229} \\
		HD &
		0.936 &
		2.275 &
		2.318 &
		0.817 &
		1.587 &
		2.889 &
		6.066 &
		0.989 &
		1.526 &
		2.867 &
		2.111 &
		2.070 &
		1.629 &
		1.175 &
		1.746 &
		1.768 &
		2.366 &
		1.389 &
		1.107 &
		1.658 &
		\textcolor{red}{1.965} \\
		JC &
		0.164 &
		0.210 &
		0.106 &
		0.053 &
		0.133 &
		0.298 &
		0.433 &
		0.292 &
		0.177 &
		0.300 &
		0.119 &
		0.245 &
		0.118 &
		0.053 &
		0.134 &
		0.200 &
		0.329 &
		0.161 &
		0.098 &
		0.197 &
		\textcolor{red}{0.191} \\
		RVE &
		0.069 &
		0.084 &
		0.156 &
		0.288 &
		0.149 &
		0.320 &
		1.990 &
		0.053 &
		0.231 &
		0.649 &
		0.431 &
		0.103 &
		0.251 &
		0.390 &
		0.294 &
		0.270 &
		0.079 &
		0.133 &
		0.175 &
		0.164 &
		\textcolor{red}{0.314} \\
		Sen &
		0.903 &
		0.877 &
		0.725 &
		0.451 &
		0.739 &
		0.694 &
		0.767 &
		0.937 &
		0.685 &
		0.771 &
		0.566 &
		0.891 &
		0.719 &
		0.397 &
		0.643 &
		0.817 &
		0.871 &
		0.753 &
		0.635 &
		0.769 &
		0.731 \\
		Spe &
		1.000 &
		0.998 &
		1.000 &
		1.000 &
		0.999 &
		0.999 &
		0.932 &
		1.000 &
		0.999 &
		0.983 &
		1.000 &
		0.997 &
		0.999 &
		1.000 &
		0.999 &
		0.999 &
		0.997 &
		1.000 &
		1.000 &
		0.999 &
		\textcolor{red}{0.995} \\
		BD &
		1 &
		2 &
		3 &
		4 &
		mean &
		1 &
		2 &
		3 &
		4 &
		mean &
		1 &
		2 &
		3 &
		4 &
		mean &
		1 &
		2 &
		3 &
		4 &
		mean &
		\\
		Dice &
		90.054 &
		85.203 &
		73.070 &
		52.624 &
		75.238 &
		73.605 &
		79.252 &
		93.431 &
		67.569 &
		78.464 &
		58.398 &
		87.818 &
		72.080 &
		45.090 &
		65.846 &
		79.404 &
		87.661 &
		75.007 &
		58.221 &
		75.073 &
		73.655 \\
		HD &
		0.884 &
		2.983 &
		1.917 &
		0.901 &
		1.671 &
		2.827 &
		6.017 &
		1.026 &
		1.545 &
		2.854 &
		1.954 &
		2.450 &
		1.491 &
		1.283 &
		1.794 &
		1.755 &
		2.464 &
		1.423 &
		1.205 &
		1.712 &
		2.008 \\
		JC &
		0.164 &
		0.203 &
		0.099 &
		0.057 &
		0.131 &
		0.301 &
		0.447 &
		0.289 &
		0.173 &
		0.302 &
		0.119 &
		0.240 &
		0.113 &
		0.057 &
		0.132 &
		0.198 &
		0.332 &
		0.157 &
		0.090 &
		0.194 &
		0.190 \\
		RVE &
		0.069 &
		0.118 &
		0.207 &
		0.194 &
		0.147 &
		0.295 &
		2.220 &
		0.072 &
		0.244 &
		0.708 &
		0.426 &
		0.173 &
		0.279 &
		0.343 &
		0.305 &
		0.282 &
		0.081 &
		0.135 &
		0.223 &
		0.180 &
		0.335 \\
		Sen &
		0.898 &
		0.871 &
		0.676 &
		0.499 &
		0.736 &
		0.702 &
		0.803 &
		0.923 &
		0.671 &
		0.775 &
		0.563 &
		0.904 &
		0.689 &
		0.426 &
		0.646 &
		0.809 &
		0.894 &
		0.733 &
		0.600 &
		0.759 &
		0.729 \\
		Spe &
		1.000 &
		0.998 &
		1.000 &
		1.000 &
		0.999 &
		0.999 &
		0.925 &
		1.000 &
		0.999 &
		0.981 &
		1.000 &
		0.995 &
		1.000 &
		1.000 &
		0.998 &
		0.999 &
		0.996 &
		1.000 &
		1.000 &
		0.998 &
		0.994 \\
		RG &
		1 &
		2 &
		3 &
		4 &
		mean &
		1 &
		2 &
		3 &
		4 &
		mean &
		1 &
		2 &
		3 &
		4 &
		mean &
		1 &
		2 &
		3 &
		4 &
		mean &
		\\
		Dice &
		89.983 &
		83.416 &
		74.119 &
		54.278 &
		75.449 &
		73.284 &
		77.649 &
		94.167 &
		67.751 &
		78.213 &
		59.423 &
		87.513 &
		70.165 &
		45.833 &
		65.734 &
		80.209 &
		87.297 &
		74.953 &
		58.421 &
		75.220 &
		73.654 \\
		HD &
		0.944 &
		3.106 &
		2.341 &
		0.971 &
		1.840 &
		2.872 &
		6.073 &
		1.071 &
		1.555 &
		2.893 &
		1.995 &
		2.415 &
		1.944 &
		1.309 &
		1.916 &
		1.742 &
		2.485 &
		1.513 &
		1.220 &
		1.740 &
		2.097 \\
		JC &
		0.164 &
		0.198 &
		0.101 &
		0.059 &
		0.130 &
		0.300 &
		0.438 &
		0.292 &
		0.173 &
		0.301 &
		0.121 &
		0.239 &
		0.108 &
		0.058 &
		0.131 &
		0.200 &
		0.331 &
		0.157 &
		0.090 &
		0.195 &
		0.189 \\
		RVE &
		0.083 &
		0.111 &
		0.250 &
		0.194 &
		0.159 &
		0.293 &
		2.209 &
		0.069 &
		0.244 &
		0.704 &
		0.414 &
		0.177 &
		0.387 &
		0.339 &
		0.329 &
		0.290 &
		0.078 &
		0.211 &
		0.236 &
		0.204 &
		0.349 \\
		Sen &
		0.904 &
		0.849 &
		0.717 &
		0.519 &
		0.747 &
		0.702 &
		0.780 &
		0.943 &
		0.682 &
		0.777 &
		0.577 &
		0.895 &
		0.683 &
		0.441 &
		0.649 &
		0.823 &
		0.881 &
		0.752 &
		0.606 &
		0.765 &
		\textcolor{red}{0.735} \\
		Spe &
		1.000 &
		0.998 &
		0.999 &
		1.000 &
		0.999 &
		0.999 &
		0.925 &
		1.000 &
		0.999 &
		0.981 &
		1.000 &
		0.995 &
		0.999 &
		1.000 &
		0.998 &
		0.999 &
		0.996 &
		0.999 &
		0.999 &
		0.998 &
		0.994 \\

		\hline
		
			& \multicolumn{21}{l}{\textcolor{red}{Client index (the next line) \& Organ index   (after name of method)}} \\ 
		\cmidrule[0.5mm]{2 -21}
		
		\multirow{-2}{*}{\textcolor{red}{Final epoch \& Metric}} &
		\multicolumn{5}{l}{\textcolor{red}{1}} &
		\multicolumn{5}{l}{\textcolor{red}{2}} &
		\multicolumn{5}{l}{\textcolor{red}{3}} &
		\multicolumn{5}{l}{\textcolor{red}{4}} &
		\textcolor{red}{Mean}  \\ 
		\hline

		150 &
		1 &
		2 &
		3 &
		4 &
		mean &
		1 &
		2 &
		3 &
		4 &
		mean &
		1 &
		2 &
		3 &
		4 &
		mean &
		1 &
		2 &
		3 &
		4 &
		mean &
		\\
		Dice &
		89.037 &
		83.958 &
		74.620 &
		51.831 &
		74.861 &
		71.923 &
		74.138 &
		92.544 &
		66.905 &
		76.378 &
		62.249 &
		87.084 &
		71.308 &
		42.420 &
		65.765 &
		81.640 &
		85.712 &
		74.988 &
		60.166 &
		75.626 &
		73.158 \\
		HD &
		1.204 &
		3.000 &
		2.408 &
		0.817 &
		1.857 &
		2.960 &
		6.235 &
		1.072 &
		1.553 &
		2.955 &
		2.133 &
		2.189 &
		1.629 &
		1.244 &
		1.799 &
		1.697 &
		2.361 &
		1.408 &
		1.158 &
		1.656 &
		2.067 \\
		JC &
		0.161 &
		0.197 &
		0.102 &
		0.056 &
		0.129 &
		0.292 &
		0.417 &
		0.285 &
		0.171 &
		0.291 &
		0.125 &
		0.238 &
		0.110 &
		0.053 &
		0.132 &
		0.204 &
		0.323 &
		0.158 &
		0.094 &
		0.195 &
		0.187 \\
		RVE &
		0.108 &
		0.157 &
		0.188 &
		0.230 &
		0.171 &
		0.392 &
		1.982 &
		0.089 &
		0.247 &
		0.677 &
		0.369 &
		0.131 &
		0.254 &
		0.391 &
		0.286 &
		0.245 &
		0.084 &
		0.154 &
		0.219 &
		0.175 &
		0.327 \\
		Sen &
		0.892 &
		0.866 &
		0.698 &
		0.484 &
		0.735 &
		0.679 &
		0.729 &
		0.914 &
		0.666 &
		0.747 &
		0.601 &
		0.865 &
		0.674 &
		0.403 &
		0.636 &
		0.827 &
		0.849 &
		0.737 &
		0.623 &
		0.759 &
		0.719 \\
		Spe &
		1.000 &
		0.997 &
		1.000 &
		1.000 &
		0.999 &
		0.999 &
		0.931 &
		1.000 &
		0.999 &
		0.982 &
		0.999 &
		0.997 &
		1.000 &
		1.000 &
		0.999 &
		0.999 &
		0.997 &
		1.000 &
		1.000 &
		0.999 &
		\textcolor{red}{0.995} \\
		200 &
		1 &
		2 &
		3 &
		4 &
		mean &
		1 &
		2 &
		3 &
		4 &
		mean &
		1 &
		2 &
		3 &
		4 &
		mean &
		1 &
		2 &
		3 &
		4 &
		mean &
		\\
		Dice &
		90.162 &
		87.442 &
		76.888 &
		49.430 &
		75.980 &
		72.955 &
		76.875 &
		94.070 &
		69.066 &
		78.242 &
		58.440 &
		89.017 &
		75.613 &
		42.264 &
		66.334 &
		80.019 &
		86.902 &
		76.442 &
		62.072 &
		76.359 &
		\textcolor{red}{74.229} \\
		HD &
		0.936 &
		2.275 &
		2.318 &
		0.817 &
		1.587 &
		2.889 &
		6.066 &
		0.989 &
		1.526 &
		2.867 &
		2.111 &
		2.070 &
		1.629 &
		1.175 &
		1.746 &
		1.768 &
		2.366 &
		1.389 &
		1.107 &
		1.658 &
		\textcolor{red}{1.965} \\
		JC &
		0.164 &
		0.210 &
		0.106 &
		0.053 &
		0.133 &
		0.298 &
		0.433 &
		0.292 &
		0.177 &
		0.300 &
		0.119 &
		0.245 &
		0.118 &
		0.053 &
		0.134 &
		0.200 &
		0.329 &
		0.161 &
		0.098 &
		0.197 &
		\textcolor{red}{0.191} \\
		RVE &
		0.069 &
		0.084 &
		0.156 &
		0.288 &
		0.149 &
		0.320 &
		1.990 &
		0.053 &
		0.231 &
		0.649 &
		0.431 &
		0.103 &
		0.251 &
		0.390 &
		0.294 &
		0.270 &
		0.079 &
		0.133 &
		0.175 &
		0.164 &
		\textcolor{red}{0.314} \\
		Sen &
		0.903 &
		0.877 &
		0.725 &
		0.451 &
		0.739 &
		0.694 &
		0.767 &
		0.937 &
		0.685 &
		0.771 &
		0.566 &
		0.891 &
		0.719 &
		0.397 &
		0.643 &
		0.817 &
		0.871 &
		0.753 &
		0.635 &
		0.769 &
		0.731 \\
		Spe &
		1.000 &
		0.998 &
		1.000 &
		1.000 &
		0.999 &
		0.999 &
		0.932 &
		1.000 &
		0.999 &
		0.983 &
		1.000 &
		0.997 &
		0.999 &
		1.000 &
		0.999 &
		0.999 &
		0.997 &
		1.000 &
		1.000 &
		0.999 &
		\textcolor{red}{0.995} \\
		250 &
		1 &
		2 &
		3 &
		4 &
		mean &
		1 &
		2 &
		3 &
		4 &
		mean &
		1 &
		2 &
		3 &
		4 &
		mean &
		1 &
		2 &
		3 &
		4 &
		mean &
		\\
		Dice &
		90.046 &
		83.004 &
		77.278 &
		52.935 &
		75.816 &
		72.447 &
		79.672 &
		93.434 &
		67.472 &
		78.256 &
		58.184 &
		86.879 &
		74.119 &
		46.158 &
		66.335 &
		79.815 &
		87.472 &
		77.062 &
		58.387 &
		75.684 &
		74.023 \\
		HD &
		0.905 &
		2.335 &
		1.494 &
		0.945 &
		1.420 &
		2.858 &
		6.040 &
		1.101 &
		1.562 &
		2.890 &
		1.908 &
		2.517 &
		1.866 &
		1.315 &
		1.902 &
		1.724 &
		2.545 &
		1.511 &
		1.231 &
		1.753 &
		1.991 \\
		JC &
		0.164 &
		0.196 &
		0.105 &
		0.057 &
		0.130 &
		0.296 &
		0.450 &
		0.288 &
		0.172 &
		0.302 &
		0.118 &
		0.237 &
		0.115 &
		0.058 &
		0.132 &
		0.199 &
		0.331 &
		0.160 &
		0.090 &
		0.195 &
		0.190 \\
		RVE &
		0.070 &
		0.133 &
		0.203 &
		0.197 &
		0.151 &
		0.342 &
		2.268 &
		0.066 &
		0.251 &
		0.732 &
		0.427 &
		0.182 &
		0.392 &
		0.337 &
		0.334 &
		0.263 &
		0.086 &
		0.215 &
		0.246 &
		0.202 &
		0.355 \\
		Sen &
		0.896 &
		0.854 &
		0.722 &
		0.510 &
		0.746 &
		0.691 &
		0.807 &
		0.928 &
		0.679 &
		0.776 &
		0.561 &
		0.896 &
		0.705 &
		0.448 &
		0.653 &
		0.813 &
		0.886 &
		0.753 &
		0.610 &
		0.766 &
		\textcolor{red}{0.735} \\
		Spe &
		1.000 &
		0.997 &
		1.000 &
		1.000 &
		0.999 &
		0.999 &
		0.923 &
		1.000 &
		0.999 &
		0.980 &
		1.000 &
		0.995 &
		0.999 &
		1.000 &
		0.998 &
		0.999 &
		0.995 &
		0.999 &
		0.999 &
		0.998 &
		0.994 \\

		\hline
		
			& \multicolumn{21}{l}{\textcolor{red}{Client index (the next line) \& Organ index   (after name of method)}} \\ 
		\cmidrule[0.5mm]{2 -21}
		
		\multirow{-2}{*}{\textcolor{red}{$U_{\mathcal{T}}$ \& Metric}} &
		\multicolumn{5}{l}{\textcolor{red}{1}} &
		\multicolumn{5}{l}{\textcolor{red}{2}} &
		\multicolumn{5}{l}{\textcolor{red}{3}} &
		\multicolumn{5}{l}{\textcolor{red}{4}} &
		\textcolor{red}{Mean}  \\ 
		\hline

		0.6 &
		1 &
		2 &
		3 &
		4 &
		mean &
		1 &
		2 &
		3 &
		4 &
		mean &
		1 &
		2 &
		3 &
		4 &
		mean &
		1 &
		2 &
		3 &
		4 &
		mean &
		\\
		Dice &
		90.087 &
		86.578 &
		75.214 &
		48.588 &
		75.117 &
		72.938 &
		77.279 &
		91.823 &
		68.181 &
		77.555 &
		58.373 &
		89.477 &
		72.041 &
		41.848 &
		65.435 &
		80.097 &
		86.761 &
		76.434 &
		62.076 &
		76.342 &
		73.612 \\
		HD &
		0.976 &
		2.313 &
		1.540 &
		0.756 &
		1.396 &
		2.916 &
		6.124 &
		1.070 &
		1.525 &
		2.908 &
		2.193 &
		2.227 &
		1.394 &
		1.153 &
		1.742 &
		1.814 &
		2.397 &
		1.218 &
		1.104 &
		1.633 &
		\textcolor{red}{1.920} \\
		JC &
		0.164 &
		0.207 &
		0.104 &
		0.051 &
		0.131 &
		0.298 &
		0.435 &
		0.281 &
		0.175 &
		0.297 &
		0.119 &
		0.246 &
		0.111 &
		0.052 &
		0.132 &
		0.201 &
		0.328 &
		0.161 &
		0.098 &
		0.197 &
		0.189 \\
		RVE &
		0.067 &
		0.113 &
		0.176 &
		0.330 &
		0.172 &
		0.313 &
		2.030 &
		0.095 &
		0.240 &
		0.670 &
		0.429 &
		0.126 &
		0.226 &
		0.395 &
		0.294 &
		0.269 &
		0.083 &
		0.121 &
		0.201 &
		0.168 &
		0.326 \\
		Sen &
		0.899 &
		0.883 &
		0.702 &
		0.432 &
		0.729 &
		0.695 &
		0.774 &
		0.896 &
		0.666 &
		0.758 &
		0.568 &
		0.900 &
		0.673 &
		0.386 &
		0.632 &
		0.823 &
		0.874 &
		0.741 &
		0.629 &
		0.767 &
		0.722 \\
		Spe &
		1.000 &
		0.998 &
		1.000 &
		1.000 &
		0.999 &
		0.999 &
		0.930 &
		1.000 &
		1.000 &
		0.982 &
		1.000 &
		0.997 &
		1.000 &
		1.000 &
		0.999 &
		0.999 &
		0.996 &
		1.000 &
		1.000 &
		0.999 &
		\textcolor{red}{0.995} \\
		0.7 &
		1 &
		&
		&
		&
		&
		2 &
		&
		&
		&
		&
		3 &
		&
		&
		&
		&
		4 &
		&
		&
		&
		&
		\\
		Dice &
		90.162 &
		87.442 &
		76.888 &
		49.430 &
		75.980 &
		72.955 &
		76.875 &
		94.070 &
		69.066 &
		78.242 &
		58.440 &
		89.017 &
		75.613 &
		42.264 &
		66.334 &
		80.019 &
		86.902 &
		76.442 &
		62.072 &
		76.359 &
		\textcolor{red}{74.229} \\
		HD &
		0.936 &
		2.275 &
		2.318 &
		0.817 &
		1.587 &
		2.889 &
		6.066 &
		0.989 &
		1.526 &
		2.867 &
		2.111 &
		2.070 &
		1.629 &
		1.175 &
		1.746 &
		1.768 &
		2.366 &
		1.389 &
		1.107 &
		1.658 &
		1.965 \\
		JC &
		0.164 &
		0.210 &
		0.106 &
		0.053 &
		0.133 &
		0.298 &
		0.433 &
		0.292 &
		0.177 &
		0.300 &
		0.119 &
		0.245 &
		0.118 &
		0.053 &
		0.134 &
		0.200 &
		0.329 &
		0.161 &
		0.098 &
		0.197 &
		\textcolor{red}{0.191} \\
		RVE &
		0.069 &
		0.084 &
		0.156 &
		0.288 &
		0.149 &
		0.320 &
		1.990 &
		0.053 &
		0.231 &
		0.649 &
		0.431 &
		0.103 &
		0.251 &
		0.390 &
		0.294 &
		0.270 &
		0.079 &
		0.133 &
		0.175 &
		0.164 &
		\textcolor{red}{0.314} \\
		Sen &
		0.903 &
		0.877 &
		0.725 &
		0.451 &
		0.739 &
		0.694 &
		0.767 &
		0.937 &
		0.685 &
		0.771 &
		0.566 &
		0.891 &
		0.719 &
		0.397 &
		0.643 &
		0.817 &
		0.871 &
		0.753 &
		0.635 &
		0.769 &
		\textcolor{red}{0.731} \\
		Spe &
		1.000 &
		0.998 &
		1.000 &
		1.000 &
		0.999 &
		0.999 &
		0.932 &
		1.000 &
		0.999 &
		0.983 &
		1.000 &
		0.997 &
		0.999 &
		1.000 &
		0.999 &
		0.999 &
		0.997 &
		1.000 &
		1.000 &
		0.999 &
		\textcolor{red}{0.995} \\
		0.8 &
		1 &
		&
		&
		&
		&
		2 &
		&
		&
		&
		&
		3 &
		&
		&
		&
		&
		4 &
		&
		&
		&
		&
		\\
		Dice &
		89.937 &
		83.021 &
		69.056 &
		53.694 &
		73.927 &
		72.708 &
		78.784 &
		93.477 &
		67.148 &
		78.029 &
		58.746 &
		87.359 &
		70.843 &
		46.870 &
		65.954 &
		79.690 &
		87.442 &
		75.228 &
		57.643 &
		75.001 &
		73.228 \\
		HD &
		0.885 &
		2.836 &
		2.019 &
		0.929 &
		1.667 &
		2.786 &
		6.049 &
		1.010 &
		1.553 &
		2.849 &
		1.884 &
		2.503 &
		1.611 &
		1.310 &
		1.827 &
		1.714 &
		2.486 &
		1.354 &
		1.228 &
		1.695 &
		2.010 \\
		JC &
		0.164 &
		0.197 &
		0.093 &
		0.058 &
		0.128 &
		0.297 &
		0.445 &
		0.289 &
		0.172 &
		0.300 &
		0.119 &
		0.239 &
		0.110 &
		0.059 &
		0.132 &
		0.199 &
		0.331 &
		0.158 &
		0.089 &
		0.194 &
		0.189 \\
		RVE &
		0.070 &
		0.150 &
		0.219 &
		0.209 &
		0.162 &
		0.312 &
		2.215 &
		0.068 &
		0.243 &
		0.710 &
		0.420 &
		0.185 &
		0.287 &
		0.327 &
		0.305 &
		0.271 &
		0.082 &
		0.162 &
		0.224 &
		0.185 &
		0.341 \\
		Sen &
		0.892 &
		0.860 &
		0.622 &
		0.517 &
		0.723 &
		0.691 &
		0.795 &
		0.916 &
		0.667 &
		0.767 &
		0.563 &
		0.900 &
		0.664 &
		0.454 &
		0.645 &
		0.808 &
		0.885 &
		0.732 &
		0.596 &
		0.755 &
		0.723 \\
		Spe &
		1.000 &
		0.997 &
		1.000 &
		1.000 &
		0.999 &
		0.999 &
		0.925 &
		1.000 &
		0.999 &
		0.981 &
		1.000 &
		0.995 &
		0.999 &
		1.000 &
		0.998 &
		0.999 &
		0.996 &
		1.000 &
		0.999 &
		0.998 &
		0.994
		\\
		\hline 
			\rowcolor{white}\multicolumn{22}{l}{Organ indexes are numbers on the straight right side of the name of method.} \\
			\rowcolor{white}\multicolumn{22}{l}{Model performance is better when metrics are higher, except for HD and RVE.} \\
		\hline

		\end{tabular}
	}

\end{table*}

\textbf{Ablation for UA.} \hyperref[Table S6]{Table S6} demonstrates a complete result for each organ and each client under more metrics. The conclusion for the usage of different model parts is basically same as what is discussed in our main paper.

\begin{table*}[]
	\label{Table S6}
	\rowcolors {3}{grey}{white}
	\renewcommand\arraystretch{1.2}
	\resizebox{\columnwidth}{!}{
	\centering
		\begin{tabular}{llllllllllllllllllllll}

		\arrayrulecolor{red}

		\hline
		\multicolumn{22}{l}{\textcolor{red}{\textbf{Table S6. Ablation study on uncertainty-based aggregation.}}}  \\
		\hline
		
			& \multicolumn{21}{l}{\textcolor{red}{Client index (the next line) \& Organ index   (after name of method)}} \\ 
		\cmidrule[0.5mm]{2 -21}
		
		\multirow{-2}{*}{\textcolor{red}{Part \& Metric}} &
		\multicolumn{5}{l}{\textcolor{red}{1}} &
		\multicolumn{5}{l}{\textcolor{red}{2}} &
		\multicolumn{5}{l}{\textcolor{red}{3}} &
		\multicolumn{5}{l}{\textcolor{red}{4}} &
		\textcolor{red}{Mean}  \\ 
		\hline

		Whole &
		  1 &
		  2 &
		  3 &
		  4 &
		  mean &
		  1 &
		  2 &
		  3 &
		  4 &
		  mean &
		  1 &
		  2 &
		  3 &
		  4 &
		  mean &
		  1 &
		  2 &
		  3 &
		  4 &
		  mean &
		   \\
		Dice &
		  89.152 &
		  87.143 &
		  72.443 &
		  55.928 &
		  76.167 &
		  72.077 &
		  80.032 &
		  92.170 &
		  68.436 &
		  78.179 &
		  55.467 &
		  88.943 &
		  68.425 &
		  44.227 &
		  64.266 &
		  78.831 &
		  86.120 &
		  75.791 &
		  60.651 &
		  75.348 &
		  73.490 \\
		HD &
		  1.019 &
		  2.313 &
		  1.171 &
		  0.927 &
		  1.357 &
		  2.959 &
		  6.003 &
		  1.079 &
		  1.642 &
		  2.921 &
		  1.933 &
		  2.425 &
		  1.845 &
		  1.370 &
		  1.893 &
		  1.734 &
		  2.569 &
		  1.306 &
		  1.260 &
		  1.717 &
		  1.972 \\
		JC &
		  0.162 &
		  0.209 &
		  0.100 &
		  0.060 &
		  0.133 &
		  0.294 &
		  0.451 &
		  0.284 &
		  0.174 &
		  0.301 &
		  0.113 &
		  0.244 &
		  0.107 &
		  0.055 &
		  0.130 &
		  0.196 &
		  0.325 &
		  0.159 &
		  0.094 &
		  0.194 &
		  \textcolor{red}{0.190} \\
		RVE &
		  0.095 &
		  0.147 &
		  0.203 &
		  0.206 &
		  0.163 &
		  0.335 &
		  2.010 &
		  0.085 &
		  0.299 &
		  0.682 &
		  0.459 &
		  0.142 &
		  0.323 &
		  0.342 &
		  0.317 &
		  0.242 &
		  0.087 &
		  0.135 &
		  0.260 &
		  0.181 &
		  0.336 \\
		Sen &
		  0.875 &
		  0.886 &
		  0.668 &
		  0.550 &
		  0.745 &
		  0.680 &
		  0.802 &
		  0.911 &
		  0.732 &
		  0.781 &
		  0.533 &
		  0.912 &
		  0.655 &
		  0.448 &
		  0.637 &
		  0.788 &
		  0.877 &
		  0.741 &
		  0.648 &
		  0.764 &
		  \textcolor{red}{0.732} \\
		Spe &
		  1.000 &
		  0.997 &
		  1.000 &
		  1.000 &
		  0.999 &
		  0.999 &
		  0.932 &
		  1.000 &
		  0.999 &
		  0.983 &
		  1.000 &
		  0.996 &
		  0.999 &
		  1.000 &
		  0.999 &
		  0.999 &
		  0.995 &
		  1.000 &
		  0.999 &
		  0.998 &
		  \textcolor{red}{0.995} \\
		Decoder &
		  1 &
		  2 &
		  3 &
		  4 &
		  mean &
		  1 &
		  2 &
		  3 &
		  4 &
		  mean &
		  1 &
		  2 &
		  3 &
		  4 &
		  mean &
		  1 &
		  2 &
		  3 &
		  4 &
		  mean &
		   \\
		Dice &
		  89.189 &
		  87.566 &
		  74.843 &
		  55.024 &
		  76.655 &
		  71.651 &
		  79.422 &
		  92.767 &
		  68.377 &
		  78.054 &
		  55.613 &
		  88.663 &
		  71.228 &
		  43.742 &
		  64.812 &
		  78.643 &
		  85.612 &
		  76.312 &
		  60.216 &
		  75.196 &
		  \textcolor{red}{73.679} \\
		HD &
		  0.988 &
		  2.413 &
		  1.342 &
		  0.849 &
		  1.398 &
		  2.940 &
		  6.004 &
		  1.133 &
		  1.629 &
		  2.927 &
		  1.973 &
		  2.468 &
		  1.906 &
		  1.249 &
		  1.899 &
		  1.750 &
		  2.574 &
		  1.356 &
		  1.226 &
		  1.726 &
		  1.988 \\
		JC &
		  0.162 &
		  0.211 &
		  0.103 &
		  0.059 &
		  0.134 &
		  0.290 &
		  0.448 &
		  0.285 &
		  0.174 &
		  0.300 &
		  0.114 &
		  0.243 &
		  0.111 &
		  0.054 &
		  0.130 &
		  0.196 &
		  0.323 &
		  0.161 &
		  0.093 &
		  0.193 &
		  0.189 \\
		RVE &
		  0.101 &
		  0.112 &
		  0.170 &
		  0.199 &
		  0.145 &
		  0.339 &
		  1.971 &
		  0.080 &
		  0.301 &
		  0.673 &
		  0.458 &
		  0.118 &
		  0.323 &
		  0.368 &
		  0.317 &
		  0.275 &
		  0.083 &
		  0.137 &
		  0.260 &
		  0.189 &
		  \textcolor{red}{0.331} \\
		Sen &
		  0.872 &
		  0.884 &
		  0.706 &
		  0.529 &
		  0.748 &
		  0.673 &
		  0.796 &
		  0.926 &
		  0.729 &
		  0.781 &
		  0.535 &
		  0.900 &
		  0.679 &
		  0.438 &
		  0.638 &
		  0.787 &
		  0.867 &
		  0.752 &
		  0.640 &
		  0.761 &
		  \textcolor{red}{0.732} \\
		Spe &
		  1.000 &
		  0.998 &
		  1.000 &
		  1.000 &
		  0.999 &
		  0.999 &
		  0.933 &
		  1.000 &
		  0.999 &
		  0.983 &
		  1.000 &
		  0.997 &
		  0.999 &
		  1.000 &
		  0.999 &
		  0.999 &
		  0.996 &
		  1.000 &
		  0.999 &
		  0.999 &
		  \textcolor{red}{0.995} \\
		Encoder &
		  1 &
		  2 &
		  3 &
		  4 &
		  mean &
		  1 &
		  2 &
		  3 &
		  4 &
		  mean &
		  1 &
		  2 &
		  3 &
		  4 &
		  mean &
		  1 &
		  2 &
		  3 &
		  4 &
		  mean &
		   \\
		Dice &
		  89.160 &
		  86.205 &
		  73.917 &
		  56.563 &
		  76.461 &
		  71.797 &
		  77.977 &
		  92.006 &
		  68.408 &
		  77.547 &
		  54.908 &
		  88.393 &
		  70.014 &
		  44.558 &
		  64.468 &
		  78.398 &
		  86.128 &
		  75.169 &
		  60.078 &
		  74.943 &
		  73.355 \\
		HD &
		  0.988 &
		  2.688 &
		  0.829 &
		  0.966 &
		  1.368 &
		  2.876 &
		  6.055 &
		  1.053 &
		  1.626 &
		  2.903 &
		  1.993 &
		  2.466 &
		  1.514 &
		  1.278 &
		  1.813 &
		  1.791 &
		  2.597 &
		  1.357 &
		  1.288 &
		  1.758 &
		  \textcolor{red}{1.961} \\
		JC &
		  0.162 &
		  0.205 &
		  0.102 &
		  0.061 &
		  0.132 &
		  0.292 &
		  0.439 &
		  0.283 &
		  0.174 &
		  0.297 &
		  0.112 &
		  0.241 &
		  0.110 &
		  0.056 &
		  0.130 &
		  0.195 &
		  0.325 &
		  0.158 &
		  0.093 &
		  0.193 &
		  0.188 \\
		RVE &
		  0.092 &
		  0.177 &
		  0.197 &
		  0.211 &
		  0.169 &
		  0.331 &
		  1.993 &
		  0.095 &
		  0.303 &
		  0.681 &
		  0.468 &
		  0.144 &
		  0.298 &
		  0.353 &
		  0.316 &
		  0.277 &
		  0.087 &
		  0.153 &
		  0.246 &
		  0.191 &
		  0.339 \\
		Sen &
		  0.878 &
		  0.887 &
		  0.679 &
		  0.551 &
		  0.749 &
		  0.678 &
		  0.784 &
		  0.907 &
		  0.730 &
		  0.775 &
		  0.531 &
		  0.905 &
		  0.666 &
		  0.447 &
		  0.637 &
		  0.787 &
		  0.874 &
		  0.737 &
		  0.641 &
		  0.760 &
		  0.730 \\
		Spe &
		  1.000 &
		  0.997 &
		  1.000 &
		  1.000 &
		  0.999 &
		  0.999 &
		  0.932 &
		  1.000 &
		  0.999 &
		  0.983 &
		  1.000 &
		  0.997 &
		  1.000 &
		  1.000 &
		  0.999 &
		  0.999 &
		  0.996 &
		  1.000 &
		  0.999 &
		  0.998 &
		  \textcolor{red}{0.995}
		  \\
		\hline 
			\rowcolor{white}\multicolumn{22}{l}{Organ indexes are numbers on the straight right side of the name of method.} \\
			\rowcolor{white}\multicolumn{22}{l}{Model performance is better when metrics are higher, except for HD and RVE.} \\
		\hline

		\end{tabular}
	}
\end{table*}

\begin{figure*}
	\centering
		\includegraphics[width=\columnwidth]{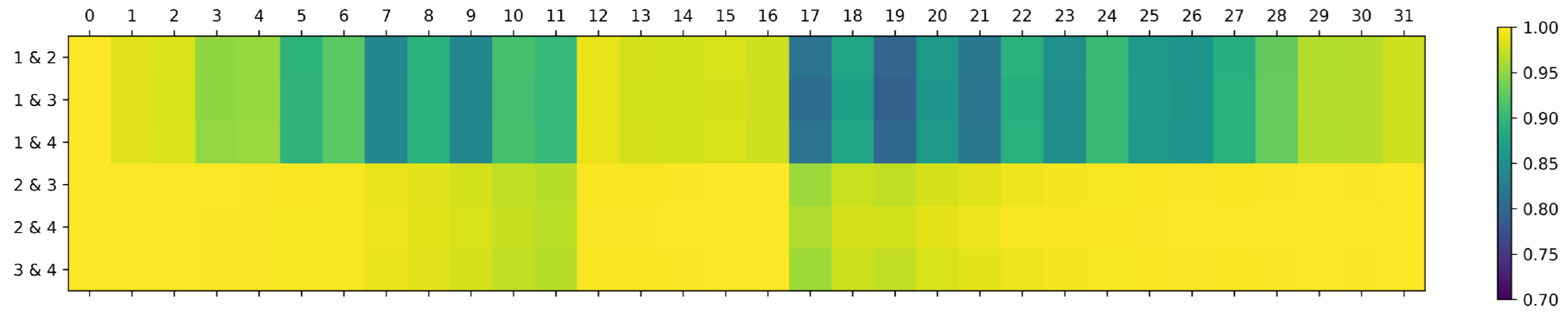}
	\caption{\textbf{Linear centered kernel alignment.}\\Numbers on the left side of the bar refer to method index, where FedAvg*, Whole, Encoder, Decoder are in the number sequence. Numbers above the bar are layer indexes, where encoder corresponds to 0 to 11, transpose convolution layers in decoder are from 12 to 16, vanilla convolution layers in decoder are from 17 to 26, deep supervision layers are from 27 to 30, The last one is segmentation head.}
	\label{Figure S3}
\end{figure*}

To investigate the effect of changing aggregation weights on different parts of models, we use linear centralized kernel alignment \upcite{kornblith2019similarity} (\hyperref[Figure S3]{Figure S3}) as a measurement for the layer-wise similarity of network parameters. It can be observed that when global aggregation weights are linked up with uncertainty, similarity between deep layers of the network is greatly changed, indicating the high level information related to noise level may be learned. When UA is performed on different layers, the main difference falls around the bottleneck layer, which is related to the deepest global context. 

\textbf{Ablation for GMT.} \hyperref[Table S7]{Table S7} demonstrates a complete result for each organ and each client under more metrics. The conclusion for the strategies is basically same as what is discussed in our main paper.

\begin{figure*}
	\centering
		\includegraphics[width=\columnwidth]{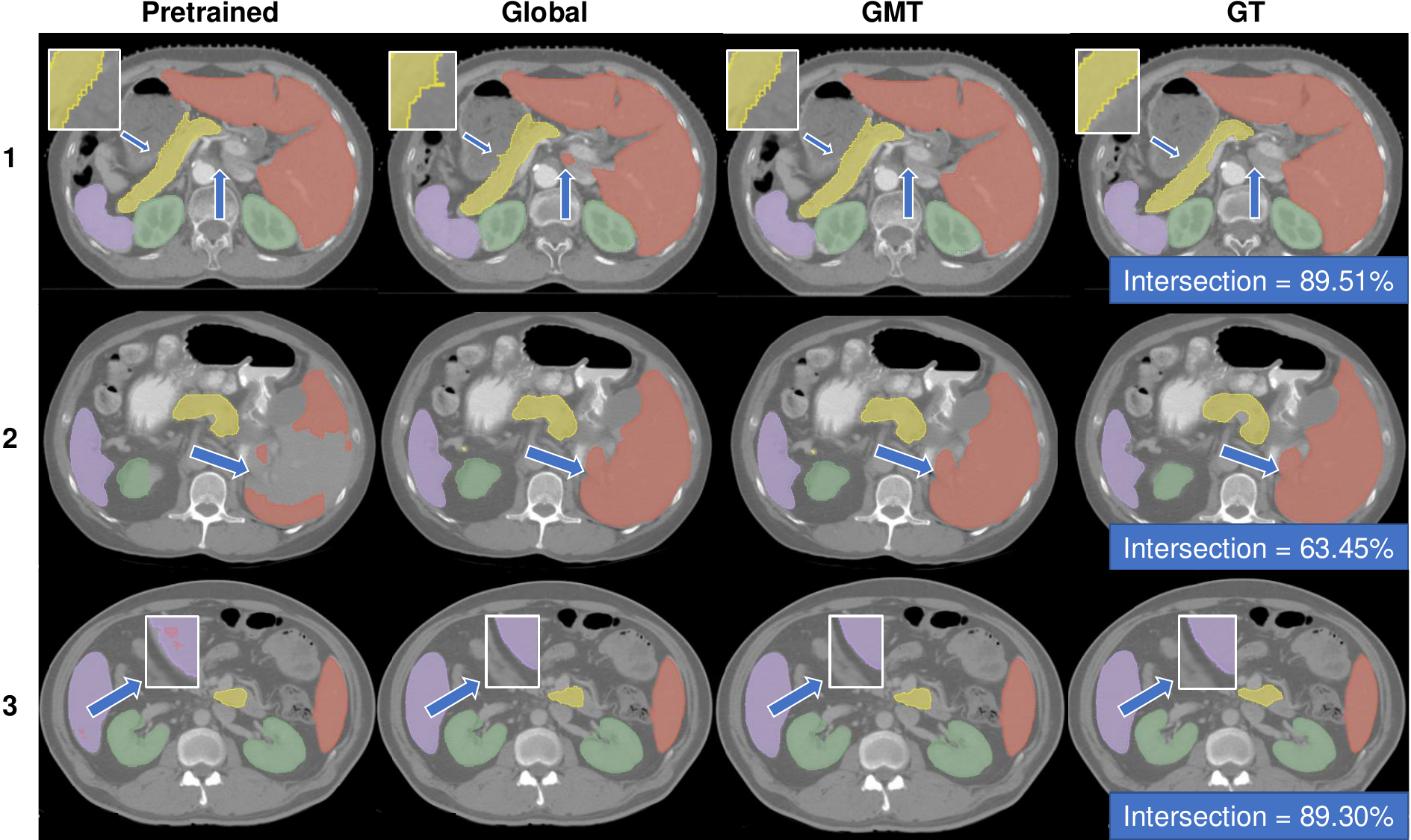}
	\caption{\textbf{Effect of global main teacher.}\\Numbers on the left side of images refer to client index. Green, red, purple, yellow regions represent kidney, liver, spleen and pancreas, respectively. Numbers in the blue box denotes intersection percentage between pretrained teachers and the global one.}
	\label{Figure S4}
\end{figure*}

At the first row of \hyperref[Figure S4]{Figure S4}, the intersectant percentage is over 80\%, so the intersection of predictions between pretrained teachers and the global one is taken. Although the global model over-segments some scraps and part of segmentation borders, the over-segmented result is offset with assistance from pretrained teacher models. In the third row, the circumstances happen to reverse. Thus, two kinds of main teachers give mutual aid when their segmentations are similar by eliminating noisy points. As to the second row, the intersectant percentage is below 80\%, so the accurate result from the global teacher is kept. This suggests that even when locally pretrained teachers may be hindered by domain gaps sometimes, the pseudo label from the GMT module can still be convincing owing to the generalization ability of the global model.

\begin{table*}[]
\label{Table S7}
	\rowcolors {3}{grey}{white}
	\renewcommand\arraystretch{1.2}
	\resizebox{\columnwidth}{!}{
	\centering
		\begin{tabular}{llllllllllllllllllllll}
		
		\arrayrulecolor{red}

		\hline
		\multicolumn{22}{l}{\textcolor{red}{\textbf{Table S7. Ablation study on global main teacher.}}}  \\
		\hline
		
			& \multicolumn{21}{l}{\textcolor{red}{Client index (the next line) \& Organ index   (after name of method)}} \\ 
		\cmidrule[0.5mm]{2 -21}
		
		\multirow{-2}{*}{\textcolor{red}{Strategy \& Metric}} &
		\multicolumn{5}{l}{\textcolor{red}{1}} &
		\multicolumn{5}{l}{\textcolor{red}{2}} &
		\multicolumn{5}{l}{\textcolor{red}{3}} &
		\multicolumn{5}{l}{\textcolor{red}{4}} &
		\textcolor{red}{Mean}  \\ 
		\hline
	
		Pretrained &
		1 &
		2 &
		3 &
		4 &
		mean &
		1 &
		2 &
		3 &
		4 &
		mean &
		1 &
		2 &
		3 &
		4 &
		mean &
		1 &
		2 &
		3 &
		4 &
		mean &
		\\
		Dice &
		90.114 &
		84.690 &
		72.959 &
		51.983 &
		74.937 &
		72.890 &
		79.384 &
		93.662 &
		66.448 &
		78.096 &
		59.062 &
		88.036 &
		68.126 &
		42.890 &
		64.528 &
		79.517 &
		87.911 &
		75.399 &
		56.123 &
		74.738 &
		73.075 \\
		HD &
		0.857 &
		2.593 &
		0.810 &
		0.910 &
		1.292 &
		2.770 &
		6.033 &
		0.993 &
		1.527 &
		2.831 &
		1.869 &
		2.590 &
		1.552 &
		1.310 &
		1.830 &
		1.730 &
		2.548 &
		1.291 &
		1.206 &
		1.694 &
		\textcolor{red}{1.912} \\
		JC &
		0.164 &
		0.200 &
		0.101 &
		0.056 &
		0.130 &
		0.296 &
		0.449 &
		0.290 &
		0.169 &
		0.301 &
		0.119 &
		0.239 &
		0.107 &
		0.054 &
		0.130 &
		0.197 &
		0.333 &
		0.158 &
		0.086 &
		0.194 &
		0.189 \\
		RVE &
		0.065 &
		0.176 &
		0.223 &
		0.196 &
		0.165 &
		0.310 &
		2.211 &
		0.063 &
		0.240 &
		0.706 &
		0.409 &
		0.221 &
		0.343 &
		0.348 &
		0.330 &
		0.249 &
		0.093 &
		0.125 &
		0.210 &
		0.169 &
		0.343 \\
		Sen &
		0.897 &
		0.892 &
		0.677 &
		0.485 &
		0.738 &
		0.692 &
		0.803 &
		0.926 &
		0.645 &
		0.766 &
		0.571 &
		0.920 &
		0.640 &
		0.402 &
		0.633 &
		0.807 &
		0.904 &
		0.734 &
		0.561 &
		0.752 &
		0.722 \\
		Spe &
		1.000 &
		0.997 &
		1.000 &
		1.000 &
		0.999 &
		0.999 &
		0.925 &
		1.000 &
		0.999 &
		0.981 &
		1.000 &
		0.994 &
		0.999 &
		1.000 &
		0.998 &
		0.999 &
		0.995 &
		1.000 &
		1.000 &
		0.998 &
		0.994 \\
		Global &
		1 &
		2 &
		3 &
		4 &
		mean &
		1 &
		2 &
		3 &
		4 &
		mean &
		1 &
		2 &
		3 &
		4 &
		mean &
		1 &
		2 &
		3 &
		4 &
		mean &
		\\
		Dice &
		85.067 &
		82.719 &
		76.878 &
		53.277 &
		74.485 &
		69.884 &
		79.873 &
		92.888 &
		66.808 &
		77.363 &
		67.640 &
		88.265 &
		73.222 &
		47.373 &
		69.125 &
		81.954 &
		85.995 &
		75.902 &
		59.939 &
		75.947 &
		74.230 \\
		HD &
		2.812 &
		6.117 &
		2.720 &
		1.854 &
		3.376 &
		2.858 &
		6.063 &
		1.290 &
		1.598 &
		2.952 &
		3.772 &
		3.205 &
		2.803 &
		2.008 &
		2.947 &
		2.755 &
		2.984 &
		1.914 &
		1.686 &
		2.335 &
		2.903 \\
		JC &
		0.151 &
		0.193 &
		0.107 &
		0.057 &
		0.127 &
		0.273 &
		0.447 &
		0.285 &
		0.170 &
		0.294 &
		0.134 &
		0.240 &
		0.113 &
		0.059 &
		0.136 &
		0.204 &
		0.323 &
		0.159 &
		0.093 &
		0.195 &
		0.188 \\
		RVE &
		0.321 &
		0.234 &
		0.753 &
		0.435 &
		0.436 &
		0.394 &
		1.598 &
		0.072 &
		0.298 &
		0.591 &
		0.229 &
		0.165 &
		0.614 &
		0.478 &
		0.371 &
		0.414 &
		0.082 &
		0.248 &
		0.248 &
		0.248 &
		0.412 \\
		Sen &
		0.847 &
		0.878 &
		0.766 &
		0.522 &
		0.753 &
		0.634 &
		0.787 &
		0.934 &
		0.642 &
		0.749 &
		0.673 &
		0.912 &
		0.735 &
		0.507 &
		0.707 &
		0.841 &
		0.865 &
		0.774 &
		0.622 &
		0.776 &
		\textcolor{red}{0.746} \\
		Spe &
		0.999 &
		0.996 &
		0.998 &
		0.999 &
		0.998 &
		0.999 &
		0.945 &
		1.000 &
		0.999 &
		0.986 &
		0.998 &
		0.995 &
		0.998 &
		0.999 &
		0.998 &
		0.998 &
		0.996 &
		0.999 &
		0.999 &
		0.998 &
		\textcolor{red}{0.995} \\
		Intersection &
		1 &
		2 &
		3 &
		4 &
		mean &
		1 &
		2 &
		3 &
		4 &
		mean &
		1 &
		2 &
		3 &
		4 &
		mean &
		1 &
		2 &
		3 &
		4 &
		mean &
		\\
		Dice &
		89.913 &
		87.246 &
		75.142 &
		58.777 &
		77.770 &
		71.843 &
		77.264 &
		93.326 &
		67.922 &
		77.589 &
		59.374 &
		87.560 &
		68.836 &
		46.827 &
		65.649 &
		80.884 &
		86.757 &
		74.798 &
		61.954 &
		76.098 &
		\textcolor{red}{74.277} \\
		HD &
		0.923 &
		1.928 &
		0.900 &
		1.001 &
		1.188 &
		2.771 &
		6.104 &
		1.003 &
		1.546 &
		2.856 &
		2.123 &
		2.729 &
		1.629 &
		1.379 &
		1.965 &
		1.805 &
		2.536 &
		1.338 &
		1.274 &
		1.738 &
		1.937 \\
		JC &
		0.163 &
		0.209 &
		0.104 &
		0.064 &
		0.135 &
		0.291 &
		0.436 &
		0.288 &
		0.174 &
		0.298 &
		0.121 &
		0.238 &
		0.107 &
		0.059 &
		0.131 &
		0.203 &
		0.328 &
		0.158 &
		0.097 &
		0.196 &
		\textcolor{red}{0.190} \\
		RVE &
		0.068 &
		0.095 &
		0.185 &
		0.191 &
		0.135 &
		0.341 &
		2.054 &
		0.072 &
		0.251 &
		0.680 &
		0.418 &
		0.206 &
		0.259 &
		0.347 &
		0.308 &
		0.262 &
		0.082 &
		0.132 &
		0.216 &
		0.173 &
		\textcolor{red}{0.324} \\
		Sen &
		0.900 &
		0.885 &
		0.698 &
		0.566 &
		0.762 &
		0.674 &
		0.773 &
		0.921 &
		0.675 &
		0.761 &
		0.573 &
		0.901 &
		0.652 &
		0.462 &
		0.647 &
		0.825 &
		0.876 &
		0.731 &
		0.649 &
		0.770 &
		0.735 \\
		Spe &
		1.000 &
		0.998 &
		1.000 &
		1.000 &
		0.999 &
		0.999 &
		0.929 &
		1.000 &
		0.999 &
		0.982 &
		1.000 &
		0.995 &
		1.000 &
		1.000 &
		0.998 &
		0.999 &
		0.996 &
		1.000 &
		1.000 &
		0.999 &
		\textcolor{red}{0.995}
		\\
		\hline 
			\rowcolor{white}\multicolumn{22}{l}{Organ indexes are numbers on the straight right side of the name of method.} \\
			\rowcolor{white}\multicolumn{22}{l}{Model performance is better when metrics are higher, except for HD and RVE.} \\
		\hline

		\end{tabular}
	}
\end{table*}

\textbf{Complete results for sUSAM.} \hyperref[Table S8]{Table S8} shows a complete result for each organ and each client under more metrics. (1) is 'CMIDG from 0 epoch'. We apply CMIDG to the data of the pseudo label baseline in the whole training process, i.e., 500 epochs, in which only descent steps are performed. (2) is 'CMIDG from 300 epoch', the mere difference between (1) and (2) is the start epoch, which is closely related to the generated data distribution. (3) is 'Original data + ASAM' in which ascent and descend steps are all performed on the original data without CMIDG. (4) is 'Random perturbation + ASAM' in which ascent steps are only performed to get a reference about perturbation amplitude and the perturbation follows a normal distribution. (5) is our proposed USAM in which CMIDG is only applied to the data in the ascent step. (6) is 'USAM + (0.8 * original weight + 0.2 * perturbed weight for descent step)' in which original gradients without perturbation and perturbed gradients by CMIDG are fused according to weights to update the model in the descent step. (7) is 'CMIDG for both ascent and descent step', in which CMIDG is applied to data in both steps. (8) is 'USAM + top k perturbation' based on (5) in which local perturbation is set to 50\% up to sorted gradients and the global gradient mask is not used. (9) is our proposed sUSAM. Except (1), we use these modules only from the 300-th epoch to the 500-th epoch. For 'ASAM for 500 epochs', please refer to FedASAM in our experiment for SOTA comparison, in which extra modules are used in the same way as what is mentioned in the original paper. The ablation of perturbation radius hints that a moderate perturbation is better for the trade-off between generalization and stability. The conclusion for the strategies is basically same as discussed in our main paper.

\hyperref[Table S9]{Table S9} displays hyper-parameter ablation for sUSAM. The conclusion for the start epoch is basically same as what is discussed in our main paper. For the perturbation radius, USAM is sensitive to this hyper-parameter and 0.7 is the best suiting one for this experimental setting.

\begin{table*}[]
	\label{Table S8}
	\rowcolors {3}{grey}{white}
	\renewcommand\arraystretch{1.2}
	\resizebox{\columnwidth}{!}{
	\centering
		\begin{tabular}{llllllllllllllllllllll}
		\arrayrulecolor{red}

		\hline
		\multicolumn{22}{l}{\textcolor{red}{\textbf{Table S8. Complete result for strategies of sUSAM.}}}  \\
		\hline
		
			& \multicolumn{21}{l}{\textcolor{red}{Client index (the next line) \& Organ index   (after name of method)}} \\ 
		\cmidrule[0.5mm]{2 -21}
		
		\multirow{-2}{*}{\textcolor{red}{Strategy \& Metric}} &
		\multicolumn{5}{l}{\textcolor{red}{1}} &
		\multicolumn{5}{l}{\textcolor{red}{2}} &
		\multicolumn{5}{l}{\textcolor{red}{3}} &
		\multicolumn{5}{l}{\textcolor{red}{4}} &
		\textcolor{red}{Mean}  \\ 
		\hline
		(1) &
			1 &
			2 &
			3 &
			4 &
			mean &
			1 &
			2 &
			3 &
			4 &
			mean &
			1 &
			2 &
			3 &
			4 &
			mean &
			1 &
			2 &
			3 &
			4 &
			mean &
			\\
		Dice &
			89.056 &
			87.866 &
			78.359 &
			48.093 &
			75.844 &
			83.081 &
			79.221 &
			90.457 &
			57.690 &
			77.612 &
			68.886 &
			88.081 &
			69.481 &
			38.603 &
			66.263 &
			85.400 &
			87.724 &
			74.219 &
			55.043 &
			75.597 &
			73.829 \\
		HD &
			1.085 &
			1.561 &
			0.794 &
			0.718 &
			1.039 &
			2.408 &
			6.069 &
			1.098 &
			1.632 &
			2.801 &
			2.019 &
			2.631 &
			1.401 &
			1.127 &
			1.795 &
			1.857 &
			2.932 &
			1.321 &
			1.174 &
			1.821 &
			\textcolor{red}{1.864} \\
		JC &
			0.161 &
			0.212 &
			0.110 &
			0.052 &
			0.134 &
			0.340 &
			0.447 &
			0.275 &
			0.143 &
			0.301 &
			0.140 &
			0.239 &
			0.106 &
			0.047 &
			0.133 &
			0.216 &
			0.330 &
			0.155 &
			0.084 &
			0.196 &
			\textcolor{red}{0.191} \\
		RVE &
			0.073 &
			0.068 &
			0.124 &
			0.360 &
			0.156 &
			0.280 &
			2.057 &
			0.095 &
			0.343 &
			0.694 &
			0.307 &
			0.146 &
			0.254 &
			0.374 &
			0.270 &
			0.226 &
			0.098 &
			0.157 &
			0.264 &
			0.186 &
			0.327 \\
		Sen &
			0.887 &
			0.881 &
			0.759 &
			0.430 &
			0.739 &
			0.787 &
			0.802 &
			0.884 &
			0.525 &
			0.750 &
			0.670 &
			0.902 &
			0.656 &
			0.360 &
			0.647 &
			0.861 &
			0.907 &
			0.721 &
			0.516 &
			0.751 &
			0.722 \\
		Spe &
			1.000 &
			0.998 &
			1.000 &
			1.000 &
			0.999 &
			1.000 &
			0.929 &
			1.000 &
			1.000 &
			0.982 &
			0.999 &
			0.996 &
			1.000 &
			1.000 &
			0.999 &
			0.999 &
			0.995 &
			1.000 &
			1.000 &
			0.998 &
			\textcolor{red}{0.995} \\
		(2) &
			1 &
			2 &
			3 &
			4 &
			mean &
			1 &
			2 &
			3 &
			4 &
			mean &
			1 &
			2 &
			3 &
			4 &
			mean &
			1 &
			2 &
			3 &
			4 &
			mean &
			\\
		Dice &
			88.112 &
			85.961 &
			76.718 &
			52.858 &
			75.912 &
			78.615 &
			76.530 &
			89.125 &
			60.981 &
			76.313 &
			67.829 &
			85.355 &
			69.328 &
			41.341 &
			65.963 &
			83.853 &
			86.615 &
			72.927 &
			57.625 &
			75.255 &
			73.361 \\
		HD &
			1.357 &
			1.553 &
			1.117 &
			0.801 &
			1.207 &
			2.581 &
			6.415 &
			1.259 &
			1.607 &
			2.965 &
			2.231 &
			3.059 &
			1.951 &
			1.231 &
			2.118 &
			1.942 &
			3.231 &
			1.596 &
			1.170 &
			1.985 &
			2.069 \\
		JC &
			0.159 &
			0.205 &
			0.107 &
			0.057 &
			0.132 &
			0.316 &
			0.430 &
			0.269 &
			0.152 &
			0.292 &
			0.136 &
			0.231 &
			0.106 &
			0.051 &
			0.131 &
			0.210 &
			0.326 &
			0.151 &
			0.089 &
			0.194 &
			0.187 \\
		RVE &
			0.097 &
			0.087 &
			0.231 &
			0.194 &
			0.152 &
			0.329 &
			2.096 &
			0.107 &
			0.307 &
			0.710 &
			0.300 &
			0.146 &
			0.338 &
			0.325 &
			0.277 &
			0.191 &
			0.080 &
			0.180 &
			0.200 &
			0.162 &
			\textcolor{red}{0.325} \\
		Sen &
			0.866 &
			0.838 &
			0.757 &
			0.495 &
			0.739 &
			0.725 &
			0.753 &
			0.882 &
			0.572 &
			0.733 &
			0.643 &
			0.843 &
			0.672 &
			0.399 &
			0.639 &
			0.826 &
			0.866 &
			0.719 &
			0.561 &
			0.743 &
			0.714 \\
		Spe &
			0.999 &
			0.999 &
			1.000 &
			1.000 &
			0.999 &
			1.000 &
			0.927 &
			0.999 &
			1.000 &
			0.982 &
			0.999 &
			0.996 &
			0.999 &
			1.000 &
			0.999 &
			0.999 &
			0.996 &
			0.999 &
			1.000 &
			0.999 &
			\textcolor{red}{0.995} \\
		(3) &
			1 &
			2 &
			3 &
			4 &
			mean &
			1 &
			2 &
			3 &
			4 &
			mean &
			1 &
			2 &
			3 &
			4 &
			mean &
			1 &
			2 &
			3 &
			4 &
			mean &
			\\
		Dice &
			89.315 &
			84.561 &
			75.661 &
			50.342 &
			74.970 &
			71.804 &
			76.503 &
			93.050 &
			66.334 &
			76.923 &
			58.716 &
			87.325 &
			70.761 &
			41.486 &
			64.572 &
			81.227 &
			85.450 &
			74.157 &
			59.398 &
			75.058 &
			72.881 \\
		HD &
			0.941 &
			2.400 &
			0.917 &
			0.778 &
			1.259 &
			2.791 &
			6.141 &
			1.040 &
			1.558 &
			2.883 &
			1.896 &
			2.523 &
			1.548 &
			1.170 &
			1.784 &
			1.661 &
			2.438 &
			1.320 &
			1.157 &
			1.644 &
			1.893 \\
		JC &
			0.162 &
			0.200 &
			0.105 &
			0.054 &
			0.130 &
			0.290 &
			0.432 &
			0.287 &
			0.169 &
			0.295 &
			0.119 &
			0.237 &
			0.109 &
			0.051 &
			0.129 &
			0.204 &
			0.322 &
			0.156 &
			0.093 &
			0.193 &
			0.187 \\
		RVE &
			0.070 &
			0.127 &
			0.176 &
			0.272 &
			0.161 &
			0.349 &
			2.033 &
			0.074 &
			0.252 &
			0.677 &
			0.420 &
			0.159 &
			0.274 &
			0.368 &
			0.305 &
			0.227 &
			0.080 &
			0.165 &
			0.203 &
			0.169 &
			0.328 \\
		Sen &
			0.879 &
			0.865 &
			0.711 &
			0.458 &
			0.728 &
			0.669 &
			0.757 &
			0.923 &
			0.653 &
			0.751 &
			0.559 &
			0.877 &
			0.678 &
			0.393 &
			0.627 &
			0.813 &
			0.853 &
			0.734 &
			0.605 &
			0.751 &
			0.714 \\
		Spe &
			1.000 &
			0.998 &
			1.000 &
			1.000 &
			0.999 &
			1.000 &
			0.930 &
			1.000 &
			0.999 &
			0.982 &
			1.000 &
			0.996 &
			1.000 &
			1.000 &
			0.999 &
			0.999 &
			0.996 &
			1.000 &
			1.000 &
			0.999 &
			\textcolor{red}{0.995} \\
		(4) &
			1 &
			2 &
			3 &
			4 &
			mean &
			1 &
			2 &
			3 &
			4 &
			mean &
			1 &
			2 &
			3 &
			4 &
			mean &
			1 &
			2 &
			3 &
			4 &
			mean &
			\\
		Dice &
			89.898 &
			85.934 &
			73.955 &
			51.902 &
			75.422 &
			72.841 &
			78.147 &
			92.656 &
			66.228 &
			77.468 &
			59.050 &
			88.062 &
			68.214 &
			43.595 &
			64.730 &
			79.909 &
			87.912 &
			75.289 &
			56.800 &
			74.978 &
			73.150 \\
		HD &
			0.886 &
			2.052 &
			0.793 &
			0.891 &
			1.155 &
			2.840 &
			6.088 &
			1.001 &
			1.541 &
			2.868 &
			1.969 &
			2.508 &
			1.400 &
			1.315 &
			1.798 &
			1.763 &
			2.508 &
			1.233 &
			1.225 &
			1.682 &
			1.876 \\
		JC &
			0.164 &
			0.205 &
			0.102 &
			0.056 &
			0.132 &
			0.296 &
			0.441 &
			0.286 &
			0.169 &
			0.298 &
			0.120 &
			0.240 &
			0.107 &
			0.055 &
			0.130 &
			0.200 &
			0.333 &
			0.158 &
			0.088 &
			0.195 &
			0.189 \\
		RVE &
			0.066 &
			0.128 &
			0.214 &
			0.204 &
			0.153 &
			0.306 &
			2.188 &
			0.076 &
			0.229 &
			0.700 &
			0.406 &
			0.205 &
			0.280 &
			0.322 &
			0.303 &
			0.283 &
			0.087 &
			0.135 &
			0.214 &
			0.180 &
			0.334 \\
		Sen &
			0.890 &
			0.893 &
			0.675 &
			0.484 &
			0.735 &
			0.688 &
			0.788 &
			0.909 &
			0.642 &
			0.757 &
			0.571 &
			0.915 &
			0.639 &
			0.406 &
			0.633 &
			0.808 &
			0.901 &
			0.729 &
			0.570 &
			0.752 &
			0.719 \\
		Spe &
			1.000 &
			0.998 &
			1.000 &
			1.000 &
			0.999 &
			0.999 &
			0.925 &
			1.000 &
			0.999 &
			0.981 &
			1.000 &
			0.995 &
			1.000 &
			1.000 &
			0.998 &
			0.999 &
			0.995 &
			1.000 &
			1.000 &
			0.998 &
			0.994 \\
		(5) &
			1 &
			2 &
			3 &
			4 &
			mean &
			1 &
			2 &
			3 &
			4 &
			mean &
			1 &
			2 &
			3 &
			4 &
			mean &
			1 &
			2 &
			3 &
			4 &
			mean &
			\\
		Dice &
			89.529 &
			84.727 &
			76.069 &
			49.922 &
			75.062 &
			73.099 &
			76.691 &
			93.864 &
			67.759 &
			77.853 &
			58.920 &
			88.507 &
			72.711 &
			42.560 &
			65.675 &
			79.970 &
			87.155 &
			76.630 &
			61.661 &
			76.354 &
			73.736 \\
		HD &
			1.033 &
			2.749 &
			0.948 &
			0.785 &
			1.379 &
			2.841 &
			6.141 &
			1.002 &
			1.530 &
			2.879 &
			2.256 &
			2.312 &
			1.641 &
			1.158 &
			1.842 &
			1.853 &
			2.404 &
			1.296 &
			1.133 &
			1.671 &
			1.943 \\
		JC &
			0.162 &
			0.200 &
			0.106 &
			0.053 &
			0.130 &
			0.298 &
			0.433 &
			0.290 &
			0.174 &
			0.299 &
			0.120 &
			0.243 &
			0.114 &
			0.053 &
			0.133 &
			0.200 &
			0.330 &
			0.162 &
			0.097 &
			0.197 &
			0.190 \\
		RVE &
			0.093 &
			0.143 &
			0.177 &
			0.285 &
			0.174 &
			0.335 &
			2.056 &
			0.060 &
			0.236 &
			0.672 &
			0.404 &
			0.154 &
			0.279 &
			0.373 &
			0.302 &
			0.278 &
			0.077 &
			0.122 &
			0.194 &
			0.168 &
			0.329 \\
		Sen &
			0.904 &
			0.883 &
			0.711 &
			0.450 &
			0.737 &
			0.697 &
			0.765 &
			0.934 &
			0.663 &
			0.764 &
			0.576 &
			0.896 &
			0.698 &
			0.400 &
			0.643 &
			0.822 &
			0.879 &
			0.753 &
			0.624 &
			0.770 &
			0.729 \\
		Spe &
			1.000 &
			0.997 &
			1.000 &
			1.000 &
			0.999 &
			0.999 &
			0.929 &
			1.000 &
			1.000 &
			0.982 &
			0.999 &
			0.996 &
			0.999 &
			1.000 &
			0.999 &
			0.999 &
			0.996 &
			1.000 &
			1.000 &
			0.999 &
			\textcolor{red}{0.995} \\
		(6) &
			1 &
			2 &
			3 &
			4 &
			mean &
			1 &
			2 &
			3 &
			4 &
			mean &
			1 &
			2 &
			3 &
			4 &
			mean &
			1 &
			2 &
			3 &
			4 &
			mean &
			\\
		Dice &
			89.846 &
			82.857 &
			75.490 &
			49.649 &
			74.461 &
			73.496 &
			76.523 &
			93.876 &
			67.588 &
			77.871 &
			59.113 &
			88.522 &
			73.592 &
			41.519 &
			65.686 &
			80.467 &
			86.866 &
			76.002 &
			60.874 &
			76.052 &
			73.518 \\
		HD &
			1.057 &
			3.105 &
			1.007 &
			0.766 &
			1.484 &
			2.807 &
			6.119 &
			1.020 &
			1.518 &
			2.866 &
			2.250 &
			2.408 &
			1.759 &
			1.150 &
			1.892 &
			1.795 &
			2.364 &
			1.430 &
			1.108 &
			1.674 &
			1.979 \\
		JC &
			0.163 &
			0.194 &
			0.105 &
			0.053 &
			0.129 &
			0.300 &
			0.432 &
			0.290 &
			0.173 &
			0.299 &
			0.120 &
			0.243 &
			0.114 &
			0.052 &
			0.132 &
			0.201 &
			0.329 &
			0.160 &
			0.096 &
			0.197 &
			0.189 \\
		RVE &
			0.091 &
			0.195 &
			0.194 &
			0.299 &
			0.195 &
			0.317 &
			2.013 &
			0.065 &
			0.242 &
			0.659 &
			0.411 &
			0.141 &
			0.313 &
			0.392 &
			0.314 &
			0.296 &
			0.078 &
			0.153 &
			0.186 &
			0.178 &
			0.337 \\
		Sen &
			0.902 &
			0.878 &
			0.712 &
			0.447 &
			0.735 &
			0.703 &
			0.764 &
			0.939 &
			0.652 &
			0.765 &
			0.578 &
			0.892 &
			0.703 &
			0.384 &
			0.639 &
			0.832 &
			0.872 &
			0.755 &
			0.606 &
			0.766 &
			0.726 \\
		Spe &
			1.000 &
			0.997 &
			1.000 &
			1.000 &
			0.999 &
			0.999 &
			0.931 &
			1.000 &
			1.000 &
			0.982 &
			0.999 &
			0.996 &
			0.999 &
			1.000 &
			0.999 &
			0.999 &
			0.996 &
			1.000 &
			1.000 &
			0.999 &
			\textcolor{red}{0.995} \\
		(7) &
			1 &
			2 &
			3 &
			4 &
			mean &
			1 &
			2 &
			3 &
			4 &
			mean &
			1 &
			2 &
			3 &
			4 &
			mean &
			1 &
			2 &
			3 &
			4 &
			mean &
			\\
		Dice &
			88.075 &
			85.074 &
			76.997 &
			54.502 &
			76.162 &
			80.481 &
			77.577 &
			88.778 &
			61.513 &
			77.087 &
			64.175 &
			83.632 &
			67.529 &
			43.325 &
			64.665 &
			82.431 &
			85.277 &
			71.514 &
			57.243 &
			74.116 &
			73.008 \\
		HD &
			1.079 &
			3.088 &
			0.962 &
			0.753 &
			1.471 &
			2.496 &
			6.427 &
			1.222 &
			1.615 &
			2.940 &
			2.153 &
			3.847 &
			1.758 &
			1.206 &
			2.241 &
			1.871 &
			3.743 &
			1.430 &
			1.187 &
			2.058 &
			2.178 \\
		JC &
			0.159 &
			0.202 &
			0.107 &
			0.059 &
			0.132 &
			0.324 &
			0.437 &
			0.266 &
			0.153 &
			0.295 &
			0.129 &
			0.223 &
			0.103 &
			0.054 &
			0.127 &
			0.205 &
			0.318 &
			0.147 &
			0.088 &
			0.190 &
			0.186 \\
		RVE &
			0.069 &
			0.124 &
			0.139 &
			0.202 &
			0.133 &
			0.319 &
			2.028 &
			0.122 &
			0.302 &
			0.693 &
			0.346 &
			0.242 &
			0.291 &
			0.381 &
			0.315 &
			0.141 &
			0.119 &
			0.172 &
			0.198 &
			0.158 &
			\textcolor{red}{0.325} \\
		Sen &
			0.857 &
			0.858 &
			0.750 &
			0.534 &
			0.750 &
			0.740 &
			0.777 &
			0.880 &
			0.595 &
			0.748 &
			0.601 &
			0.860 &
			0.658 &
			0.435 &
			0.639 &
			0.802 &
			0.873 &
			0.705 &
			0.575 &
			0.738 &
			0.719 \\
		Spe &
			1.000 &
			0.997 &
			1.000 &
			1.000 &
			0.999 &
			1.000 &
			0.929 &
			0.999 &
			0.999 &
			0.982 &
			0.999 &
			0.994 &
			0.999 &
			1.000 &
			0.998 &
			0.999 &
			0.994 &
			0.999 &
			1.000 &
			0.998 &
			0.994 \\
		(8) &
			1 &
			2 &
			3 &
			4 &
			mean &
			1 &
			2 &
			3 &
			4 &
			mean &
			1 &
			2 &
			3 &
			4 &
			mean &
			1 &
			2 &
			3 &
			4 &
			mean &
			\\
		Dice &
			89.468 &
			86.396 &
			72.558 &
			55.431 &
			75.963 &
			72.648 &
			78.573 &
			93.068 &
			68.255 &
			78.136 &
			55.325 &
			88.577 &
			70.853 &
			45.212 &
			64.992 &
			78.437 &
			86.464 &
			74.982 &
			60.753 &
			75.159 &
			73.563 \\
		HD &
			1.017 &
			2.511 &
			1.138 &
			0.887 &
			1.388 &
			3.120 &
			6.008 &
			1.110 &
			1.641 &
			2.970 &
			1.965 &
			2.464 &
			2.028 &
			1.253 &
			1.927 &
			1.777 &
			2.488 &
			1.454 &
			1.236 &
			1.739 &
			2.006 \\
		JC &
			0.162 &
			0.205 &
			0.100 &
			0.060 &
			0.132 &
			0.297 &
			0.443 &
			0.287 &
			0.174 &
			0.300 &
			0.113 &
			0.242 &
			0.111 &
			0.056 &
			0.130 &
			0.195 &
			0.327 &
			0.157 &
			0.094 &
			0.193 &
			0.189 \\
		RVE &
			0.107 &
			0.149 &
			0.223 &
			0.214 &
			0.173 &
			0.314 &
			1.988 &
			0.083 &
			0.328 &
			0.678 &
			0.467 &
			0.135 &
			0.355 &
			0.338 &
			0.324 &
			0.291 &
			0.083 &
			0.169 &
			0.277 &
			0.205 &
			0.345 \\
		Sen &
			0.883 &
			0.888 &
			0.679 &
			0.542 &
			0.748 &
			0.694 &
			0.790 &
			0.928 &
			0.731 &
			0.786 &
			0.535 &
			0.906 &
			0.687 &
			0.460 &
			0.647 &
			0.792 &
			0.880 &
			0.747 &
			0.656 &
			0.769 &
			0.738 \\
		Spe &
			1.000 &
			0.997 &
			1.000 &
			1.000 &
			0.999 &
			0.999 &
			0.932 &
			1.000 &
			0.999 &
			0.983 &
			1.000 &
			0.996 &
			0.999 &
			1.000 &
			0.999 &
			0.999 &
			0.996 &
			1.000 &
			0.999 &
			0.998 &
			\textcolor{red}{0.995} \\
		(9) &
			1 &
			2 &
			3 &
			4 &
			mean &
			1 &
			2 &
			3 &
			4 &
			mean &
			1 &
			2 &
			3 &
			4 &
			mean &
			1 &
			2 &
			3 &
			4 &
			mean &
			\\
		Dice &
			89.607 &
			86.188 &
			75.868 &
			56.591 &
			77.064 &
			72.590 &
			78.745 &
			93.043 &
			68.564 &
			78.236 &
			56.931 &
			88.331 &
			70.639 &
			46.224 &
			65.531 &
			78.799 &
			86.347 &
			75.005 &
			61.218 &
			75.342 &
			\textcolor{red}{74.043} \\
		HD &
			1.058 &
			2.354 &
			1.184 &
			0.933 &
			1.382 &
			2.856 &
			6.023 &
			1.083 &
			1.636 &
			2.899 &
			1.922 &
			2.545 &
			1.874 &
			1.239 &
			1.895 &
			1.758 &
			2.562 &
			1.427 &
			1.221 &
			1.742 &
			1.980 \\
		JC &
			0.163 &
			0.205 &
			0.105 &
			0.061 &
			0.133 &
			0.297 &
			0.445 &
			0.287 &
			0.174 &
			0.301 &
			0.116 &
			0.241 &
			0.111 &
			0.058 &
			0.132 &
			0.196 &
			0.326 &
			0.157 &
			0.095 &
			0.194 &
			0.190 \\
		RVE &
			0.104 &
			0.149 &
			0.175 &
			0.216 &
			0.161 &
			0.339 &
			1.993 &
			0.079 &
			0.325 &
			0.684 &
			0.452 &
			0.145 &
			0.313 &
			0.336 &
			0.311 &
			0.274 &
			0.081 &
			0.153 &
			0.259 &
			0.192 &
			0.337 \\
		Sen &
			0.886 &
			0.886 &
			0.706 &
			0.560 &
			0.759 &
			0.691 &
			0.790 &
			0.922 &
			0.739 &
			0.785 &
			0.550 &
			0.905 &
			0.673 &
			0.470 &
			0.650 &
			0.797 &
			0.876 &
			0.736 &
			0.660 &
			0.767 &
			\textcolor{red}{0.740} \\
		Spe &
			1.000 &
			0.997 &
			1.000 &
			1.000 &
			0.999 &
			0.999 &
			0.932 &
			1.000 &
			0.999 &
			0.983 &
			1.000 &
			0.996 &
			0.999 &
			1.000 &
			0.999 &
			0.999 &
			0.996 &
			1.000 &
			0.999 &
			0.999 &
			\textcolor{red}{0.995}
			\\
			\hline 
				\rowcolor{white}\multicolumn{22}{l}{Organ indexes are numbers on the straight right side of the name of method.} \\
				\rowcolor{white}\multicolumn{22}{l}{Model performance is better when metrics are higher, except for HD and RVE.} \\
			\hline
		\end{tabular}
		}
	\end{table*}

\begin{table*}[]
	\label{Table S9}
	\rowcolors {3}{grey}{white}
	\renewcommand\arraystretch{1.2}
	\resizebox{\columnwidth}{!}{
	\centering
	\begin{tabular}{llllllllllllllllllllll}
		\arrayrulecolor{red}

		\hline
		\multicolumn{22}{l}{\textcolor{red}{\textbf{Table S9. Hyper-parameter ablation for sUSAM.}}}  \\
		\hline
		
		& \multicolumn{21}{l}{\textcolor{red}{Client index (the next line) \& Organ index   (after name of method)}} \\ 
		\cmidrule[0.5mm]{2 -21}
		
		\multirow{-2}{*}{\textcolor{red}{Perturbation radius \&   Metric}} &
		\multicolumn{5}{l}{\textcolor{red}{1}} &
		\multicolumn{5}{l}{\textcolor{red}{2}} &
		\multicolumn{5}{l}{\textcolor{red}{3}} &
		\multicolumn{5}{l}{\textcolor{red}{4}} &
		\textcolor{red}{Mean}  \\ 
		\hline
		0.6 &
		1 &
		2 &
		3 &
		4 &
		mean &
		1 &
		2 &
		3 &
		4 &
		mean &
		1 &
		2 &
		3 &
		4 &
		mean &
		1 &
		2 &
		3 &
		4 &
		mean &
		 \\
		 Dice &
		 88.789 &
		 84.057 &
		 73.915 &
		 52.409 &
		 74.793 &
		 72.391 &
		 72.412 &
		 91.661 &
		 67.626 &
		 76.022 &
		 62.499 &
		 86.911 &
		 70.869 &
		 42.980 &
		 65.815 &
		 81.889 &
		 85.269 &
		 74.171 &
		 60.902 &
		 75.558 &
		 73.047 \\
	   HD &
		 1.279 &
		 2.428 &
		 0.964 &
		 0.831 &
		 1.375 &
		 2.989 &
		 6.251 &
		 1.065 &
		 1.584 &
		 2.972 &
		 2.433 &
		 2.203 &
		 1.562 &
		 1.281 &
		 1.870 &
		 1.978 &
		 2.341 &
		 1.307 &
		 1.205 &
		 1.707 &
		 1.981 \\
	   JC &
		 0.160 &
		 0.198 &
		 0.101 &
		 0.056 &
		 0.129 &
		 0.295 &
		 0.407 &
		 0.281 &
		 0.173 &
		 0.289 &
		 0.126 &
		 0.238 &
		 0.109 &
		 0.053 &
		 0.131 &
		 0.204 &
		 0.321 &
		 0.155 &
		 0.095 &
		 0.194 &
		 0.186 \\
	   RVE &
		 0.147 &
		 0.148 &
		 0.192 &
		 0.211 &
		 0.174 &
		 0.353 &
		 1.980 &
		 0.104 &
		 0.233 &
		 0.668 &
		 0.357 &
		 0.129 &
		 0.258 &
		 0.360 &
		 0.276 &
		 0.309 &
		 0.086 &
		 0.133 &
		 0.236 &
		 0.191 &
		 \textcolor{red}{0.327} \\
	   Sen &
		 0.905 &
		 0.865 &
		 0.680 &
		 0.500 &
		 0.737 &
		 0.699 &
		 0.713 &
		 0.903 &
		 0.684 &
		 0.750 &
		 0.620 &
		 0.859 &
		 0.669 &
		 0.422 &
		 0.643 &
		 0.849 &
		 0.843 &
		 0.725 &
		 0.643 &
		 0.765 &
		 0.724 \\
	   Spe &
		 0.999 &
		 0.997 &
		 1.000 &
		 1.000 &
		 0.999 &
		 0.999 &
		 0.931 &
		 1.000 &
		 0.999 &
		 0.982 &
		 0.999 &
		 0.997 &
		 1.000 &
		 1.000 &
		 0.999 &
		 0.999 &
		 0.997 &
		 1.000 &
		 1.000 &
		 0.999 &
		 \textcolor{red}{0.995} \\
	   0.7 &
		 1 &
		 2 &
		 3 &
		 4 &
		 mean &
		 1 &
		 2 &
		 3 &
		 4 &
		 mean &
		 1 &
		 2 &
		 3 &
		 4 &
		 mean &
		 1 &
		 2 &
		 3 &
		 4 &
		 mean &
		  \\
	   Dice &
		 89.529 &
		 84.727 &
		 76.069 &
		 49.922 &
		 75.062 &
		 73.099 &
		 76.691 &
		 93.864 &
		 67.759 &
		 77.853 &
		 58.920 &
		 88.507 &
		 72.711 &
		 42.560 &
		 65.675 &
		 79.970 &
		 87.155 &
		 76.630 &
		 61.661 &
		 76.354 &
		 \textcolor{red}{73.736} \\
	   HD &
		 1.033 &
		 2.749 &
		 0.948 &
		 0.785 &
		 1.379 &
		 2.841 &
		 6.141 &
		 1.002 &
		 1.530 &
		 2.879 &
		 2.256 &
		 2.312 &
		 1.641 &
		 1.158 &
		 1.842 &
		 1.853 &
		 2.404 &
		 1.296 &
		 1.133 &
		 1.671 &
		 \textcolor{red}{1.943} \\
	   JC &
		 0.162 &
		 0.200 &
		 0.106 &
		 0.053 &
		 0.130 &
		 0.298 &
		 0.433 &
		 0.290 &
		 0.174 &
		 0.299 &
		 0.120 &
		 0.243 &
		 0.114 &
		 0.053 &
		 0.133 &
		 0.200 &
		 0.330 &
		 0.162 &
		 0.097 &
		 0.197 &
		 \textcolor{red}{0.190} \\
	   RVE &
		 0.093 &
		 0.143 &
		 0.177 &
		 0.285 &
		 0.174 &
		 0.335 &
		 2.056 &
		 0.060 &
		 0.236 &
		 0.672 &
		 0.404 &
		 0.154 &
		 0.279 &
		 0.373 &
		 0.302 &
		 0.278 &
		 0.077 &
		 0.122 &
		 0.194 &
		 0.168 &
		 0.329 \\
	   Sen &
		 0.904 &
		 0.883 &
		 0.711 &
		 0.450 &
		 0.737 &
		 0.697 &
		 0.765 &
		 0.934 &
		 0.663 &
		 0.764 &
		 0.576 &
		 0.896 &
		 0.698 &
		 0.400 &
		 0.643 &
		 0.822 &
		 0.879 &
		 0.753 &
		 0.624 &
		 0.770 &
		 \textcolor{red}{0.729} \\
	   Spe &
		 1.000 &
		 0.997 &
		 1.000 &
		 1.000 &
		 0.999 &
		 0.999 &
		 0.929 &
		 1.000 &
		 1.000 &
		 0.982 &
		 0.999 &
		 0.996 &
		 0.999 &
		 1.000 &
		 0.999 &
		 0.999 &
		 0.996 &
		 1.000 &
		 1.000 &
		 0.999 &
		 \textcolor{red}{0.995} \\
	   0.8 &
		 1 &
		 2 &
		 3 &
		 4 &
		 mean &
		 1 &
		 2 &
		 3 &
		 4 &
		 mean &
		 1 &
		 2 &
		 3 &
		 4 &
		 mean &
		 1 &
		 2 &
		 3 &
		 4 &
		 mean &
		  \\
	   Dice &
		 89.563 &
		 81.818 &
		 73.000 &
		 54.332 &
		 74.678 &
		 71.880 &
		 77.954 &
		 92.785 &
		 67.677 &
		 77.574 &
		 58.873 &
		 87.737 &
		 70.434 &
		 44.521 &
		 65.391 &
		 80.880 &
		 87.264 &
		 75.637 &
		 60.457 &
		 76.060 &
		 73.426 \\
	   HD &
		 0.881 &
		 3.826 &
		 0.817 &
		 0.859 &
		 1.596 &
		 2.858 &
		 6.104 &
		 1.022 &
		 1.528 &
		 2.878 &
		 1.953 &
		 2.862 &
		 1.442 &
		 1.212 &
		 1.867 &
		 1.713 &
		 2.496 &
		 1.228 &
		 1.175 &
		 1.653 &
		 1.999 \\
	   JC &
		 0.163 &
		 0.191 &
		 0.101 &
		 0.059 &
		 0.128 &
		 0.292 &
		 0.441 &
		 0.286 &
		 0.173 &
		 0.298 &
		 0.120 &
		 0.239 &
		 0.109 &
		 0.056 &
		 0.131 &
		 0.203 &
		 0.330 &
		 0.159 &
		 0.095 &
		 0.197 &
		 0.189 \\
	   RVE &
		 0.071 &
		 0.278 &
		 0.220 &
		 0.207 &
		 0.194 &
		 0.380 &
		 2.094 &
		 0.075 &
		 0.247 &
		 0.699 &
		 0.416 &
		 0.195 &
		 0.260 &
		 0.366 &
		 0.309 &
		 0.224 &
		 0.081 &
		 0.141 &
		 0.173 &
		 0.155 &
		 0.339 \\
	   Sen &
		 0.880 &
		 0.881 &
		 0.677 &
		 0.508 &
		 0.737 &
		 0.678 &
		 0.779 &
		 0.914 &
		 0.660 &
		 0.758 &
		 0.567 &
		 0.895 &
		 0.670 &
		 0.424 &
		 0.639 &
		 0.822 &
		 0.885 &
		 0.737 &
		 0.620 &
		 0.766 &
		 0.725 \\
	   Spe &
		 1.000 &
		 0.996 &
		 1.000 &
		 1.000 &
		 0.999 &
		 0.999 &
		 0.928 &
		 1.000 &
		 0.999 &
		 0.982 &
		 1.000 &
		 0.995 &
		 1.000 &
		 1.000 &
		 0.998 &
		 0.999 &
		 0.996 &
		 1.000 &
		 1.000 &
		 0.999 &
		 \textcolor{red}{0.995} \\
	   0.9 &
		 1 &
		 2 &
		 3 &
		 4 &
		 mean &
		 1 &
		 2 &
		 3 &
		 4 &
		 mean &
		 1 &
		 2 &
		 3 &
		 4 &
		 mean &
		 1 &
		 2 &
		 3 &
		 4 &
		 mean &
		  \\
	   Dice &
		 89.705 &
		 82.774 &
		 72.666 &
		 53.104 &
		 74.562 &
		 71.986 &
		 77.983 &
		 93.409 &
		 66.928 &
		 77.576 &
		 57.547 &
		 86.915 &
		 69.881 &
		 45.579 &
		 64.980 &
		 80.307 &
		 87.796 &
		 76.619 &
		 58.002 &
		 75.681 &
		 73.200 \\
	   HD &
		 0.910 &
		 2.553 &
		 0.971 &
		 0.898 &
		 1.333 &
		 2.815 &
		 6.092 &
		 1.023 &
		 1.570 &
		 2.875 &
		 1.937 &
		 2.840 &
		 1.652 &
		 1.274 &
		 1.926 &
		 1.698 &
		 2.559 &
		 1.250 &
		 1.215 &
		 1.680 &
		 1.954 \\
	   JC &
		 0.163 &
		 0.196 &
		 0.099 &
		 0.058 &
		 0.129 &
		 0.292 &
		 0.440 &
		 0.288 &
		 0.171 &
		 0.298 &
		 0.117 &
		 0.237 &
		 0.109 &
		 0.058 &
		 0.130 &
		 0.201 &
		 0.333 &
		 0.162 &
		 0.090 &
		 0.196 &
		 0.188 \\
	   RVE &
		 0.073 &
		 0.165 &
		 0.197 &
		 0.176 &
		 0.153 &
		 0.329 &
		 2.264 &
		 0.073 &
		 0.241 &
		 0.727 &
		 0.436 &
		 0.226 &
		 0.318 &
		 0.325 &
		 0.326 &
		 0.268 &
		 0.088 &
		 0.134 &
		 0.205 &
		 0.174 &
		 0.345 \\
	   Sen &
		 0.887 &
		 0.863 &
		 0.673 &
		 0.502 &
		 0.731 &
		 0.680 &
		 0.784 &
		 0.925 &
		 0.665 &
		 0.763 &
		 0.553 &
		 0.905 &
		 0.667 &
		 0.439 &
		 0.641 &
		 0.812 &
		 0.895 &
		 0.750 &
		 0.598 &
		 0.764 &
		 0.725 \\
	   Spe &
		 1.000 &
		 0.997 &
		 1.000 &
		 1.000 &
		 0.999 &
		 0.999 &
		 0.922 &
		 1.000 &
		 0.999 &
		 0.980 &
		 1.000 &
		 0.994 &
		 0.999 &
		 1.000 &
		 0.998 &
		 0.999 &
		 0.995 &
		 1.000 &
		 1.000 &
		 0.998 &
		 0.994 \\

		\hline
		\rowcolor{white}
			& \multicolumn{21}{l}{\textcolor{red}{Client index (the next line) \& Organ index   (after name of method)}} \\ 
		\cmidrule[0.5mm]{2 -21}
		\rowcolor{white}
		\multirow{-2}{*}{\textcolor{red}{Start epoch \& Metric}} &
		\multicolumn{5}{l}{\textcolor{red}{1}} &
		\multicolumn{5}{l}{\textcolor{red}{2}} &
		\multicolumn{5}{l}{\textcolor{red}{3}} &
		\multicolumn{5}{l}{\textcolor{red}{4}} &
		\textcolor{red}{Mean}  \\ 
		\hline
		 
	   200 &
		 1 &
		 2 &
		 3 &
		 4 &
		 mean &
		 1 &
		 2 &
		 3 &
		 4 &
		 mean &
		 1 &
		 2 &
		 3 &
		 4 &
		 mean &
		 1 &
		 2 &
		 3 &
		 4 &
		 mean &
		  \\
	   Dice &
		89.849 &
		85.316 &
		75.887 &
		51.235 &
		75.572 &
		73.545 &
		77.112 &
		93.514 &
		67.290 &
		77.865 &
		59.906 &
		88.837 &
		71.160 &
		42.631 &
		65.634 &
		80.654 &
		87.517 &
		75.827 &
		61.718 &
		76.429 &
		\textcolor{red}{73.875} \\
		HD &
		1.013 &
		2.652 &
		1.073 &
		0.810 &
		1.387 &
		2.795 &
		6.135 &
		1.041 &
		1.543 &
		2.878 &
		2.216 &
		2.409 &
		1.781 &
		1.224 &
		1.907 &
		1.814 &
		2.443 &
		1.357 &
		1.164 &
		1.694 &
		1.967 \\
		JC &
		0.163 &
		0.202 &
		0.106 &
		0.054 &
		0.131 &
		0.301 &
		0.435 &
		0.289 &
		0.172 &
		0.299 &
		0.121 &
		0.244 &
		0.111 &
		0.053 &
		0.132 &
		0.201 &
		0.331 &
		0.159 &
		0.097 &
		0.197 &
		\textcolor{red}{0.190} \\
		RVE &
		0.102 &
		0.145 &
		0.168 &
		0.278 &
		0.173 &
		0.312 &
		2.092 &
		0.064 &
		0.243 &
		0.678 &
		0.401 &
		0.145 &
		0.293 &
		0.377 &
		0.304 &
		0.284 &
		0.080 &
		0.152 &
		0.194 &
		0.177 &
		0.333 \\
		Sen &
		0.911 &
		0.888 &
		0.713 &
		0.468 &
		0.745 &
		0.707 &
		0.773 &
		0.932 &
		0.653 &
		0.766 &
		0.587 &
		0.907 &
		0.682 &
		0.403 &
		0.645 &
		0.835 &
		0.887 &
		0.753 &
		0.626 &
		0.775 &
		\textcolor{red}{0.733} \\
		Spe &
		1.000 &
		0.997 &
		1.000 &
		1.000 &
		0.999 &
		0.999 &
		0.928 &
		1.000 &
		1.000 &
		0.982 &
		0.999 &
		0.996 &
		0.999 &
		1.000 &
		0.999 &
		0.999 &
		0.996 &
		1.000 &
		1.000 &
		0.999 &
		\textcolor{red}{0.995} \\
		300 &
		1 &
		2 &
		3 &
		4 &
		mean &
		1 &
		2 &
		3 &
		4 &
		mean &
		1 &
		2 &
		3 &
		4 &
		mean &
		1 &
		2 &
		3 &
		4 &
		mean &
		\\
		Dice &
		89.529 &
		84.727 &
		76.069 &
		49.922 &
		75.062 &
		73.099 &
		76.691 &
		93.864 &
		67.759 &
		77.853 &
		58.920 &
		88.507 &
		72.711 &
		42.560 &
		65.675 &
		79.970 &
		87.155 &
		76.630 &
		61.661 &
		76.354 &
		73.736 \\
		HD &
		1.033 &
		2.749 &
		0.948 &
		0.785 &
		1.379 &
		2.841 &
		6.141 &
		1.002 &
		1.530 &
		2.879 &
		2.256 &
		2.312 &
		1.641 &
		1.158 &
		1.842 &
		1.853 &
		2.404 &
		1.296 &
		1.133 &
		1.671 &
		\textcolor{red}{1.943} \\
		JC &
		0.162 &
		0.200 &
		0.106 &
		0.053 &
		0.130 &
		0.298 &
		0.433 &
		0.290 &
		0.174 &
		0.299 &
		0.120 &
		0.243 &
		0.114 &
		0.053 &
		0.133 &
		0.200 &
		0.330 &
		0.162 &
		0.097 &
		0.197 &
		\textcolor{red}{0.190} \\
		RVE &
		0.093 &
		0.143 &
		0.177 &
		0.285 &
		0.174 &
		0.335 &
		2.056 &
		0.060 &
		0.236 &
		0.672 &
		0.404 &
		0.154 &
		0.279 &
		0.373 &
		0.302 &
		0.278 &
		0.077 &
		0.122 &
		0.194 &
		0.168 &
		\textcolor{red}{0.329} \\
		Sen &
		0.904 &
		0.883 &
		0.711 &
		0.450 &
		0.737 &
		0.697 &
		0.765 &
		0.934 &
		0.663 &
		0.764 &
		0.576 &
		0.896 &
		0.698 &
		0.400 &
		0.643 &
		0.822 &
		0.879 &
		0.753 &
		0.624 &
		0.770 &
		0.729 \\
		Spe &
		1.000 &
		0.997 &
		1.000 &
		1.000 &
		0.999 &
		0.999 &
		0.929 &
		1.000 &
		1.000 &
		0.982 &
		0.999 &
		0.996 &
		0.999 &
		1.000 &
		0.999 &
		0.999 &
		0.996 &
		1.000 &
		1.000 &
		0.999 &
		\textcolor{red}{0.995} \\
		400 &
		1 &
		2 &
		3 &
		4 &
		mean &
		1 &
		2 &
		3 &
		4 &
		mean &
		1 &
		2 &
		3 &
		4 &
		mean &
		1 &
		2 &
		3 &
		4 &
		mean &
		\\
		Dice &
		90.077 &
		84.171 &
		75.436 &
		53.716 &
		75.850 &
		72.514 &
		77.703 &
		93.665 &
		66.522 &
		77.601 &
		59.985 &
		87.820 &
		69.560 &
		44.576 &
		65.485 &
		80.053 &
		88.187 &
		75.371 &
		57.203 &
		75.203 &
		73.535 \\
		HD &
		0.859 &
		3.096 &
		0.835 &
		0.920 &
		1.428 &
		2.822 &
		6.115 &
		1.004 &
		1.546 &
		2.872 &
		1.893 &
		2.686 &
		1.691 &
		1.289 &
		1.890 &
		1.724 &
		2.586 &
		1.320 &
		1.205 &
		1.708 &
		1.975 \\
		JC &
		0.164 &
		0.198 &
		0.104 &
		0.058 &
		0.131 &
		0.295 &
		0.439 &
		0.290 &
		0.170 &
		0.298 &
		0.121 &
		0.240 &
		0.109 &
		0.056 &
		0.132 &
		0.199 &
		0.334 &
		0.159 &
		0.088 &
		0.195 &
		0.189 \\
		RVE &
		0.065 &
		0.191 &
		0.196 &
		0.169 &
		0.155 &
		0.323 &
		2.252 &
		0.063 &
		0.237 &
		0.719 &
		0.405 &
		0.212 &
		0.307 &
		0.309 &
		0.308 &
		0.241 &
		0.098 &
		0.143 &
		0.204 &
		0.172 &
		0.339 \\
		Sen &
		0.895 &
		0.898 &
		0.697 &
		0.513 &
		0.751 &
		0.685 &
		0.788 &
		0.927 &
		0.655 &
		0.764 &
		0.576 &
		0.919 &
		0.663 &
		0.434 &
		0.648 &
		0.808 &
		0.911 &
		0.732 &
		0.584 &
		0.759 &
		0.731 \\
		Spe &
		1.000 &
		0.997 &
		1.000 &
		1.000 &
		0.999 &
		0.999 &
		0.922 &
		1.000 &
		0.999 &
		0.980 &
		1.000 &
		0.995 &
		1.000 &
		1.000 &
		0.998 &
		0.999 &
		0.995 &
		1.000 &
		1.000 &
		0.998 &
		0.994
		 \\
			\hline 
				\rowcolor{white}\multicolumn{22}{l}{Organ indexes are numbers on the straight right side of the name of method.} \\
				\rowcolor{white}\multicolumn{22}{l}{Model performance is better when metrics are higher, except for HD and RVE.} \\
			\hline
	   \end{tabular}
	}
\end{table*}

\section{Generalization for different class divsion}
Experiments in the main paper are based on one kind of setting for the class division. To prove that UFPS is universal under different class divisions, we conduct more experiments on the non-IID benchmark. As in our main paper, intersection between annotated classes is still null so the demand for labels is reduced to the lowest. 

\noindent\textbf{Setting 2.}
In setting 2, client 1 is only annotated for spleen, client 2 is only annotated for liver, and client 3 is annotated for kidney and pancreas. From \hyperref[Table S10]{Table S10}, it can be seen that although the class division has changed, our proposed method is still solid for its universal design for the class division and strong generalization ability to different degrees of data heterogeneity.
\begin{table*}[]
	\label{Table S10}
	\rowcolors {3}{grey}{white}
	\renewcommand\arraystretch{1.2}
	\resizebox{\columnwidth}{!}{
	\centering
	\begin{tabular}{lllllllllllllllllllllll}
		
		  \arrayrulecolor{red}

		  \hline
		  \multicolumn{23}{l}{\textcolor{red}{\textbf{Table S10. Comparison with SOTAs under setting 2.}}}  \\
		  \hline
		  
		  & \multicolumn{21}{l}{\textcolor{red}{Client index (the next line) \& Organ index   (after name of method)}} \\ 
		  \cmidrule[0.5mm]{2 -21}
		  
		  \multirow{-2}{*}{\textcolor{red}{Method \& Metric}} &
		  \multicolumn{5}{l}{\textcolor{red}{1}} &
		  \multicolumn{5}{l}{\textcolor{red}{2}} &
		  \multicolumn{5}{l}{\textcolor{red}{3}} &
		  \multicolumn{5}{l}{\textcolor{red}{4}} &
		  \textcolor{red}{Mean} & \textcolor{red}{Post} \\ 
		  \hline
		  SOLO partial &
		  1 &
		  2 &
		  3 &
		  4 &
		  mean &
		  1 &
		  2 &
		  3 &
		  4 &
		  mean &
		  1 &
		  2 &
		  3 &
		  4 &
		  mean &
		  1 &
		  2 &
		  3 &
		  4 &
		  mean &
		   &
		   \\
		Dice &
		  86.597 &
		  80.353 &
		  78.451 &
		  63.418 &
		  77.205 &
		  80.222 &
		  96.593 &
		  64.549 &
		  41.904 &
		  70.817 &
		  83.817 &
		  86.602 &
		  47.438 &
		  57.571 &
		  68.857 &
		  86.267 &
		  87.972 &
		  49.282 &
		  43.355 &
		  66.719 &
		  70.899 &
		  70.944 \\
		HD &
		  0.869 &
		  1.916 &
		  0.705 &
		  0.704 &
		  1.049 &
		  2.536 &
		  2.312 &
		  2.006 &
		  2.215 &
		  2.267 &
		  1.111 &
		  2.640 &
		  1.463 &
		  0.941 &
		  1.539 &
		  1.414 &
		  2.383 &
		  1.599 &
		  1.246 &
		  1.661 &
		  1.629 &
		  1.449 \\
		JC &
		  0.155 &
		  0.190 &
		  0.112 &
		  0.071 &
		  0.132 &
		  0.333 &
		  0.570 &
		  0.182 &
		  0.099 &
		  0.296 &
		  0.183 &
		  0.233 &
		  0.070 &
		  0.075 &
		  0.140 &
		  0.224 &
		  0.336 &
		  0.093 &
		  0.064 &
		  0.179 &
		  0.187 &
		  0.187 \\
		RVE &
		  0.111 &
		  0.124 &
		  0.145 &
		  0.179 &
		  0.140 &
		  0.313 &
		  0.019 &
		  0.476 &
		  0.374 &
		  0.296 &
		  0.068 &
		  0.104 &
		  0.453 &
		  0.231 &
		  0.214 &
		  0.118 &
		  0.077 &
		  0.393 &
		  0.345 &
		  0.233 &
		  0.221 &
		  0.226 \\
		Sen &
		  0.813 &
		  0.768 &
		  0.779 &
		  0.612 &
		  0.743 &
		  0.756 &
		  0.969 &
		  0.626 &
		  0.385 &
		  0.684 &
		  0.816 &
		  0.844 &
		  0.442 &
		  0.543 &
		  0.661 &
		  0.850 &
		  0.870 &
		  0.453 &
		  0.392 &
		  0.641 &
		  0.682 &
		  0.680 \\
		Spe &
		  1.000 &
		  0.998 &
		  1.000 &
		  1.000 &
		  1.000 &
		  0.999 &
		  0.999 &
		  0.999 &
		  0.999 &
		  0.999 &
		  1.000 &
		  0.997 &
		  1.000 &
		  1.000 &
		  0.999 &
		  1.000 &
		  0.997 &
		  1.000 &
		  1.000 &
		  0.999 &
		  0.899 &
		  \textcolor{red}{0.999} \\
		FedAvg* &
		  1 &
		  2 &
		  3 &
		  4 &
		  mean &
		  1 &
		  2 &
		  3 &
		  4 &
		  mean &
		  1 &
		  2 &
		  3 &
		  4 &
		  mean &
		  1 &
		  2 &
		  3 &
		  4 &
		  mean &
		   &
		   \\
		Dice &
		  87.491 &
		  85.710 &
		  75.439 &
		  64.302 &
		  78.235 &
		  81.843 &
		  94.902 &
		  80.121 &
		  46.119 &
		  75.746 &
		  81.613 &
		  87.483 &
		  63.385 &
		  56.393 &
		  72.218 &
		  86.713 &
		  87.892 &
		  64.366 &
		  48.071 &
		  71.761 &
		  74.490 &
		  74.879 \\
		HD &
		  0.835 &
		  1.930 &
		  0.885 &
		  0.773 &
		  1.106 &
		  2.422 &
		  2.583 &
		  1.635 &
		  2.234 &
		  2.218 &
		  1.334 &
		  2.658 &
		  1.652 &
		  1.234 &
		  1.720 &
		  1.627 &
		  2.473 &
		  1.549 &
		  1.277 &
		  1.731 &
		  1.694 &
		  \textcolor{red}{1.426} \\
		JC &
		  0.157 &
		  0.203 &
		  0.105 &
		  0.071 &
		  0.134 &
		  0.341 &
		  0.554 &
		  0.233 &
		  0.110 &
		  0.309 &
		  0.177 &
		  0.237 &
		  0.095 &
		  0.072 &
		  0.145 &
		  0.224 &
		  0.333 &
		  0.127 &
		  0.071 &
		  0.189 &
		  0.194 &
		  0.196 \\
		RVE &
		  0.083 &
		  0.100 &
		  0.181 &
		  0.126 &
		  0.122 &
		  0.292 &
		  0.038 &
		  0.204 &
		  0.386 &
		  0.230 &
		  0.097 &
		  0.132 &
		  0.326 &
		  0.225 &
		  0.195 &
		  0.300 &
		  0.090 &
		  0.230 &
		  0.249 &
		  0.217 &
		  0.191 &
		  0.192 \\
		Sen &
		  0.832 &
		  0.842 &
		  0.720 &
		  0.650 &
		  0.761 &
		  0.776 &
		  0.942 &
		  0.794 &
		  0.434 &
		  0.737 &
		  0.801 &
		  0.867 &
		  0.591 &
		  0.572 &
		  0.708 &
		  0.877 &
		  0.876 &
		  0.611 &
		  0.467 &
		  0.708 &
		  0.729 &
		  0.727 \\
		Spe &
		  1.000 &
		  0.998 &
		  1.000 &
		  1.000 &
		  0.999 &
		  1.000 &
		  0.999 &
		  0.999 &
		  0.999 &
		  0.999 &
		  1.000 &
		  0.996 &
		  0.999 &
		  1.000 &
		  0.999 &
		  0.999 &
		  0.996 &
		  1.000 &
		  1.000 &
		  0.999 &
		  0.999 &
		  \textcolor{red}{0.999} \\
		UFPS(Ours) &
		  1 &
		  2 &
		  3 &
		  4 &
		  mean &
		  1 &
		  2 &
		  3 &
		  4 &
		  mean &
		  1 &
		  2 &
		  3 &
		  4 &
		  mean &
		  1 &
		  2 &
		  3 &
		  4 &
		  mean &
		   &
		   \\
		Dice &
		  86.947 &
		  84.396 &
		  80.748 &
		  61.723 &
		  78.454 &
		  86.442 &
		  94.902 &
		  89.666 &
		  45.739 &
		  79.187 &
		  84.747 &
		  87.565 &
		  72.425 &
		  57.972 &
		  75.677 &
		  87.138 &
		  87.976 &
		  69.928 &
		  51.247 &
		  74.072 &
		  76.848 &
		  \textcolor{red}{77.678} \\
		HD &
		  1.930 &
		  3.362 &
		  1.563 &
		  2.395 &
		  2.312 &
		  2.285 &
		  2.678 &
		  1.378 &
		  2.150 &
		  2.123 &
		  2.309 &
		  2.975 &
		  2.335 &
		  1.884 &
		  2.376 &
		  1.955 &
		  2.724 &
		  2.035 &
		  1.698 &
		  2.103 &
		  2.229 &
		  1.454 \\
		JC &
		  0.156 &
		  0.199 &
		  0.115 &
		  0.068 &
		  0.134 &
		  0.362 &
		  0.553 &
		  0.270 &
		  0.108 &
		  0.323 &
		  0.184 &
		  0.238 &
		  0.111 &
		  0.075 &
		  0.152 &
		  0.226 &
		  0.334 &
		  0.142 &
		  0.077 &
		  0.195 &
		  0.201 &
		  \textcolor{red}{0.204} \\
		RVE &
		  0.119 &
		  0.160 &
		  0.188 &
		  0.581 &
		  0.262 &
		  0.317 &
		  0.039 &
		  0.122 &
		  0.390 &
		  0.217 &
		  0.117 &
		  0.122 &
		  0.382 &
		  0.361 &
		  0.245 &
		  0.423 &
		  0.088 &
		  0.272 &
		  0.236 &
		  0.255 &
		  0.245 &
		  \textcolor{red}{0.186} \\
		Sen &
		  0.851 &
		  0.852 &
		  0.805 &
		  0.605 &
		  0.778 &
		  0.824 &
		  0.957 &
		  0.901 &
		  0.412 &
		  0.774 &
		  0.853 &
		  0.884 &
		  0.720 &
		  0.599 &
		  0.764 &
		  0.889 &
		  0.887 &
		  0.715 &
		  0.488 &
		  0.745 &
		  0.765 &
		  \textcolor{red}{0.763} \\
		Spe &
		  0.999 &
		  0.997 &
		  1.000 &
		  0.999 &
		  0.999 &
		  1.000 &
		  0.998 &
		  0.999 &
		  0.999 &
		  0.999 &
		  0.999 &
		  0.996 &
		  0.999 &
		  0.999 &
		  0.998 &
		  0.999 &
		  0.996 &
		  0.999 &
		  0.999 &
		  0.998 &
		  0.999 &
		  \textcolor{red}{0.999}
		 \\
		 \hline 
			 \rowcolor{white}\multicolumn{22}{l}{Organ indexes are numbers on the straight right side of the name of method.} \\
			 \rowcolor{white}\multicolumn{22}{l}{Model performance is better when metrics are higher, except for HD and RVE.} \\
		 \hline
	   \end{tabular}
	}
\end{table*}

\noindent\textbf{Setting 3.}
In setting 3, client 1 is annotated for liver and pancreas, client 2 is only annotated for kidney, and client 3 is only annotated for spleen. Noise in this setting is the largest among three settings. From \hyperref[Table S11]{Table S11}, it can be concluded that model performance is mainly affected by pancreas, which obviously is due to the domain gap. Even though the noise for pancreas is quite large, our method is capable of denoising this hard class for all clients by WS compared with other methods.
\begin{table*}[]
	\label{Table S11}
	\rowcolors {3}{grey}{white}
	\renewcommand\arraystretch{1.2}
	\resizebox{\columnwidth}{!}{
	\centering
	\begin{tabular}{lllllllllllllllllllllll}
		
		  \arrayrulecolor{red}

		  \hline
		  \multicolumn{23}{l}{\textcolor{red}{\textbf{Table S11. Comparison with SOTAs under setting 3.}}}  \\
		  \hline
		  
		  & \multicolumn{21}{l}{\textcolor{red}{Client index (the next line) \& Organ index   (after name of method)}} \\ 
		  \cmidrule[0.5mm]{2 -21}
		  
		  \multirow{-2}{*}{\textcolor{red}{Method \& Metric}} &
		  \multicolumn{5}{l}{\textcolor{red}{1}} &
		  \multicolumn{5}{l}{\textcolor{red}{2}} &
		  \multicolumn{5}{l}{\textcolor{red}{3}} &
		  \multicolumn{5}{l}{\textcolor{red}{4}} &
		  \textcolor{red}{Mean} & \textcolor{red}{Post} \\ 
		  \hline
	   SOLO partial &
		 1 &
		 2 &
		 3 &
		 4 &
		 mean &
		 1 &
		 2 &
		 3 &
		 4 &
		 mean &
		 1 &
		 2 &
		 3 &
		 4 &
		 mean &
		 1 &
		 2 &
		 3 &
		 4 &
		 mean &
		  &
		  \\
	   Dice &
		 83.130 &
		 91.808 &
		 66.233 &
		 71.113 &
		 78.071 &
		 94.319 &
		 77.109 &
		 77.739 &
		 24.397 &
		 68.391 &
		 67.871 &
		 85.653 &
		 81.382 &
		 22.628 &
		 64.384 &
		 86.879 &
		 83.395 &
		 69.573 &
		 26.200 &
		 66.512 &
		 69.339 &
		 69.693 \\
	   HD &
		 1.003 &
		 1.312 &
		 1.387 &
		 0.644 &
		 1.087 &
		 1.629 &
		 5.959 &
		 2.638 &
		 2.345 &
		 3.143 &
		 1.544 &
		 2.322 &
		 1.735 &
		 1.376 &
		 1.744 &
		 1.506 &
		 2.871 &
		 1.853 &
		 1.306 &
		 1.884 &
		 1.964 &
		 1.706 \\
	   JC &
		 0.150 &
		 0.225 &
		 0.086 &
		 0.083 &
		 0.136 &
		 0.413 &
		 0.424 &
		 0.221 &
		 0.056 &
		 0.279 &
		 0.147 &
		 0.230 &
		 0.126 &
		 0.027 &
		 0.133 &
		 0.225 &
		 0.308 &
		 0.141 &
		 0.037 &
		 0.178 &
		 0.181 &
		 0.183 \\
	   RVE &
		 0.112 &
		 0.059 &
		 0.366 &
		 0.135 &
		 0.168 &
		 0.056 &
		 0.842 &
		 0.844 &
		 0.547 &
		 0.572 &
		 0.318 &
		 0.141 &
		 0.318 &
		 0.551 &
		 0.332 &
		 0.480 &
		 0.098 &
		 0.267 &
		 0.574 &
		 0.355 &
		 0.357 &
		 0.341 \\
	   Sen &
		 0.816 &
		 0.930 &
		 0.633 &
		 0.693 &
		 0.768 &
		 0.946 &
		 0.762 &
		 0.908 &
		 0.223 &
		 0.710 &
		 0.679 &
		 0.873 &
		 0.806 &
		 0.204 &
		 0.641 &
		 0.901 &
		 0.833 &
		 0.709 &
		 0.230 &
		 0.668 &
		 0.697 &
		 0.694 \\
	   Spe &
		 1.000 &
		 0.999 &
		 0.999 &
		 1.000 &
		 1.000 &
		 1.000 &
		 0.968 &
		 0.996 &
		 0.999 &
		 0.991 &
		 1.000 &
		 0.997 &
		 0.999 &
		 1.000 &
		 0.999 &
		 0.999 &
		 0.996 &
		 0.999 &
		 1.000 &
		 0.999 &
		 0.997 &
		 \textcolor{red}{0.997} \\
	   FedAvg* &
		 1 &
		 2 &
		 3 &
		 4 &
		 mean &
		 1 &
		 2 &
		 3 &
		 4 &
		 mean &
		 1 &
		 2 &
		 3 &
		 4 &
		 mean &
		 1 &
		 2 &
		 3 &
		 4 &
		 mean &
		  &
		  \\
	   Dice &
		 86.362 &
		 90.381 &
		 77.539 &
		 65.820 &
		 80.026 &
		 92.694 &
		 79.490 &
		 83.124 &
		 25.580 &
		 70.222 &
		 73.052 &
		 86.870 &
		 78.673 &
		 30.003 &
		 67.150 &
		 86.030 &
		 85.495 &
		 69.783 &
		 28.162 &
		 67.368 &
		 71.192 &
		 71.523 \\
	   HD &
		 1.035 &
		 2.080 &
		 1.617 &
		 0.691 &
		 1.356 &
		 1.814 &
		 5.631 &
		 2.175 &
		 2.319 &
		 2.985 &
		 1.864 &
		 2.458 &
		 1.923 &
		 1.286 &
		 1.883 &
		 1.649 &
		 2.938 &
		 1.924 &
		 1.389 &
		 1.975 &
		 2.050 &
		 \textcolor{red}{1.685} \\
	   JC &
		 0.155 &
		 0.220 &
		 0.107 &
		 0.075 &
		 0.139 &
		 0.401 &
		 0.438 &
		 0.244 &
		 0.059 &
		 0.286 &
		 0.157 &
		 0.234 &
		 0.122 &
		 0.036 &
		 0.137 &
		 0.221 &
		 0.318 &
		 0.142 &
		 0.040 &
		 0.180 &
		 0.186 &
		 0.187 \\
	   RVE &
		 0.093 &
		 0.093 &
		 0.324 &
		 0.157 &
		 0.167 &
		 0.081 &
		 0.831 &
		 0.605 &
		 0.493 &
		 0.503 &
		 0.240 &
		 0.139 &
		 0.387 &
		 0.451 &
		 0.304 &
		 0.401 &
		 0.099 &
		 0.302 &
		 0.427 &
		 0.307 &
		 0.320 &
		 \textcolor{red}{0.294} \\
	   Sen &
		 0.853 &
		 0.919 &
		 0.779 &
		 0.650 &
		 0.800 &
		 0.929 &
		 0.784 &
		 0.917 &
		 0.237 &
		 0.717 &
		 0.734 &
		 0.891 &
		 0.780 &
		 0.279 &
		 0.671 &
		 0.907 &
		 0.864 &
		 0.711 &
		 0.262 &
		 0.686 &
		 0.719 &
		 0.716 \\
	   Spe &
		 1.000 &
		 0.998 &
		 0.999 &
		 1.000 &
		 0.999 &
		 0.999 &
		 0.969 &
		 0.997 &
		 0.999 &
		 0.991 &
		 0.999 &
		 0.996 &
		 0.999 &
		 1.000 &
		 0.999 &
		 0.999 &
		 0.995 &
		 0.999 &
		 1.000 &
		 0.998 &
		 0.997 &
		 \textcolor{red}{0.997} \\
	   UFPS(Ours) &
		 1 &
		 2 &
		 3 &
		 4 &
		 mean &
		 1 &
		 2 &
		 3 &
		 4 &
		 mean &
		 1 &
		 2 &
		 3 &
		 4 &
		 mean &
		 1 &
		 2 &
		 3 &
		 4 &
		 mean &
		  &
		  \\
	   Dice &
		 84.849 &
		 83.967 &
		 77.615 &
		 67.276 &
		 78.427 &
		 92.129 &
		 82.454 &
		 84.455 &
		 32.139 &
		 72.794 &
		 77.418 &
		 85.045 &
		 74.852 &
		 35.710 &
		 68.256 &
		 83.735 &
		 86.671 &
		 69.981 &
		 30.618 &
		 67.751 &
		 71.807 &
		 \textcolor{red}{72.601} \\
	   HD &
		 3.194 &
		 3.500 &
		 2.875 &
		 1.176 &
		 2.686 &
		 1.879 &
		 6.047 &
		 1.954 &
		 2.418 &
		 3.075 &
		 3.587 &
		 3.235 &
		 2.678 &
		 2.000 &
		 2.875 &
		 2.736 &
		 3.392 &
		 2.261 &
		 1.857 &
		 2.562 &
		 2.800 &
		 1.951 \\
	   JC &
		 0.150 &
		 0.197 &
		 0.107 &
		 0.076 &
		 0.132 &
		 0.397 &
		 0.461 &
		 0.251 &
		 0.073 &
		 0.296 &
		 0.163 &
		 0.227 &
		 0.114 &
		 0.043 &
		 0.137 &
		 0.212 &
		 0.324 &
		 0.142 &
		 0.044 &
		 0.180 &
		 0.186 &
		 \textcolor{red}{0.190} \\
	   RVE &
		 0.503 &
		 0.374 &
		 0.910 &
		 0.172 &
		 0.490 &
		 0.103 &
		 1.363 &
		 0.587 &
		 0.504 &
		 0.639 &
		 0.350 &
		 0.229 &
		 0.591 &
		 0.293 &
		 0.366 &
		 0.712 &
		 0.121 &
		 0.296 &
		 0.363 &
		 0.373 &
		 0.467 &
		 0.366 \\
	   Sen &
		 0.869 &
		 0.921 &
		 0.787 &
		 0.672 &
		 0.812 &
		 0.923 &
		 0.830 &
		 0.904 &
		 0.290 &
		 0.737 &
		 0.794 &
		 0.899 &
		 0.767 &
		 0.348 &
		 0.702 &
		 0.898 &
		 0.900 &
		 0.716 &
		 0.296 &
		 0.703 &
		 0.739 &
		 \textcolor{red}{0.735} \\
	   Spe &
		 0.998 &
		 0.995 &
		 0.998 &
		 1.000 &
		 0.998 &
		 0.999 &
		 0.952 &
		 0.997 &
		 0.999 &
		 0.987 &
		 0.998 &
		 0.994 &
		 0.998 &
		 0.999 &
		 0.997 &
		 0.998 &
		 0.994 &
		 0.999 &
		 0.999 &
		 0.997 &
		 0.995 &
		 0.996
		 \\
		 \hline 
			 \rowcolor{white}\multicolumn{22}{l}{Organ indexes are numbers on the straight right side of the name of method.} \\
			 \rowcolor{white}\multicolumn{22}{l}{Model performance is better when metrics are higher, except for HD and RVE.} \\
		 \hline
	   \end{tabular}
	}
\end{table*}

\section{Privacy Analysis}
Due to extra communication compared with FedAvg, i.e., the data package and teacher models, it is crucial to discuss the underlying privacy leakage. First of all, uncertainty values in the data package are merely two numbers, so the information revealed is negligible. Local masks in data packages leak no more privacy than the local model because of its generation process.

However, these pretained teacher models are sent to all clients, which means either the server or other clients may leverage them to perform an attack, like membership inference attack \upcite{shokri2017membership}. On account of the fact that local models are uploaded to the server for model aggregation, it makes little difference in the result of the server's attack whether teacher models are distributed or not. Besides, local models can reflect more information relevant to membership because its prediction is available considering all classes. Therefore, we only concentrate on the circumstance for MIA from client sides. 

\begin{figure*}
	\centering
	\subfloat[teacher 1, no aug]{\includegraphics[width = 0.5\textwidth]{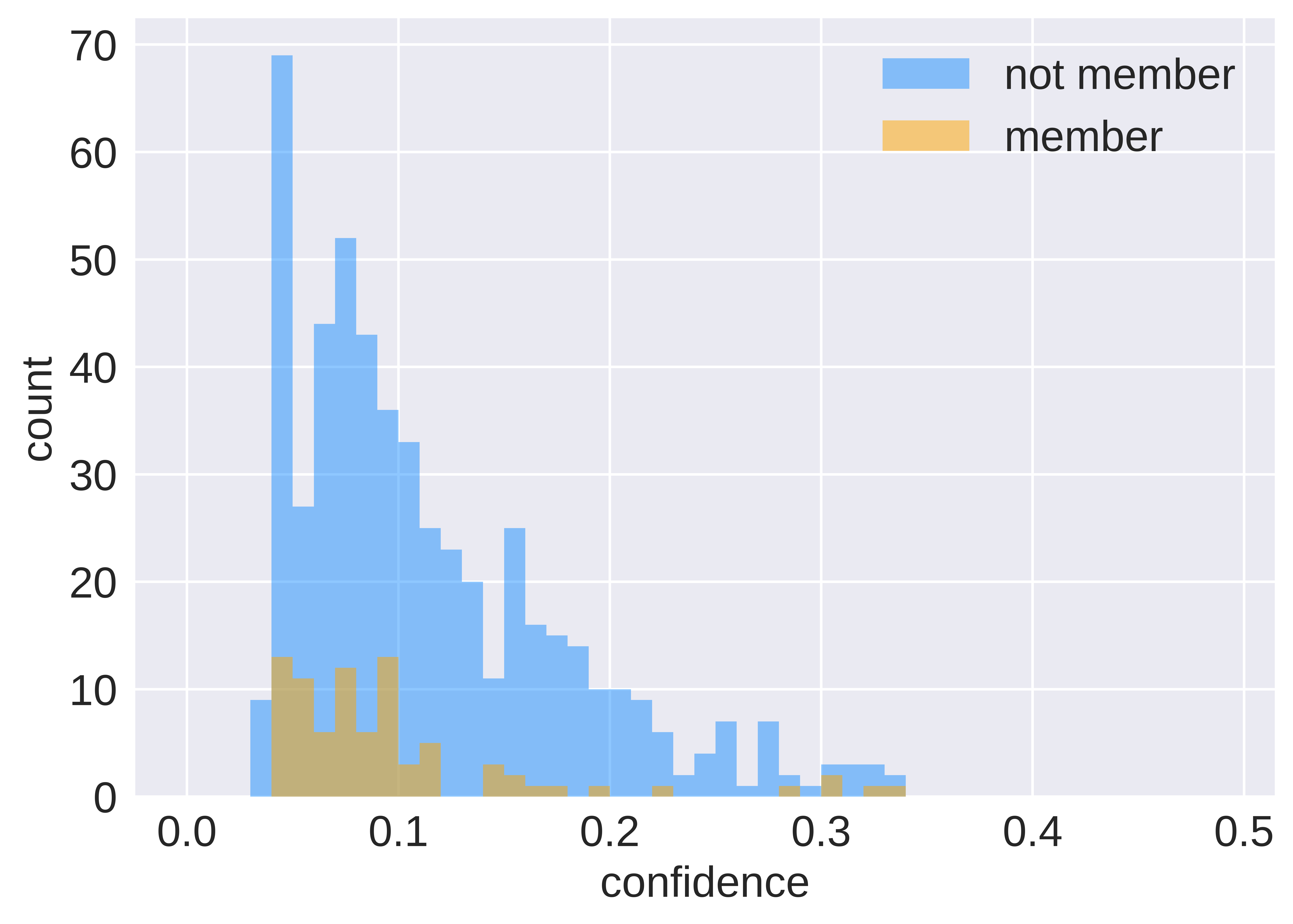}}
	\subfloat[teacher 1, aug]{\includegraphics[width = 0.5\textwidth]{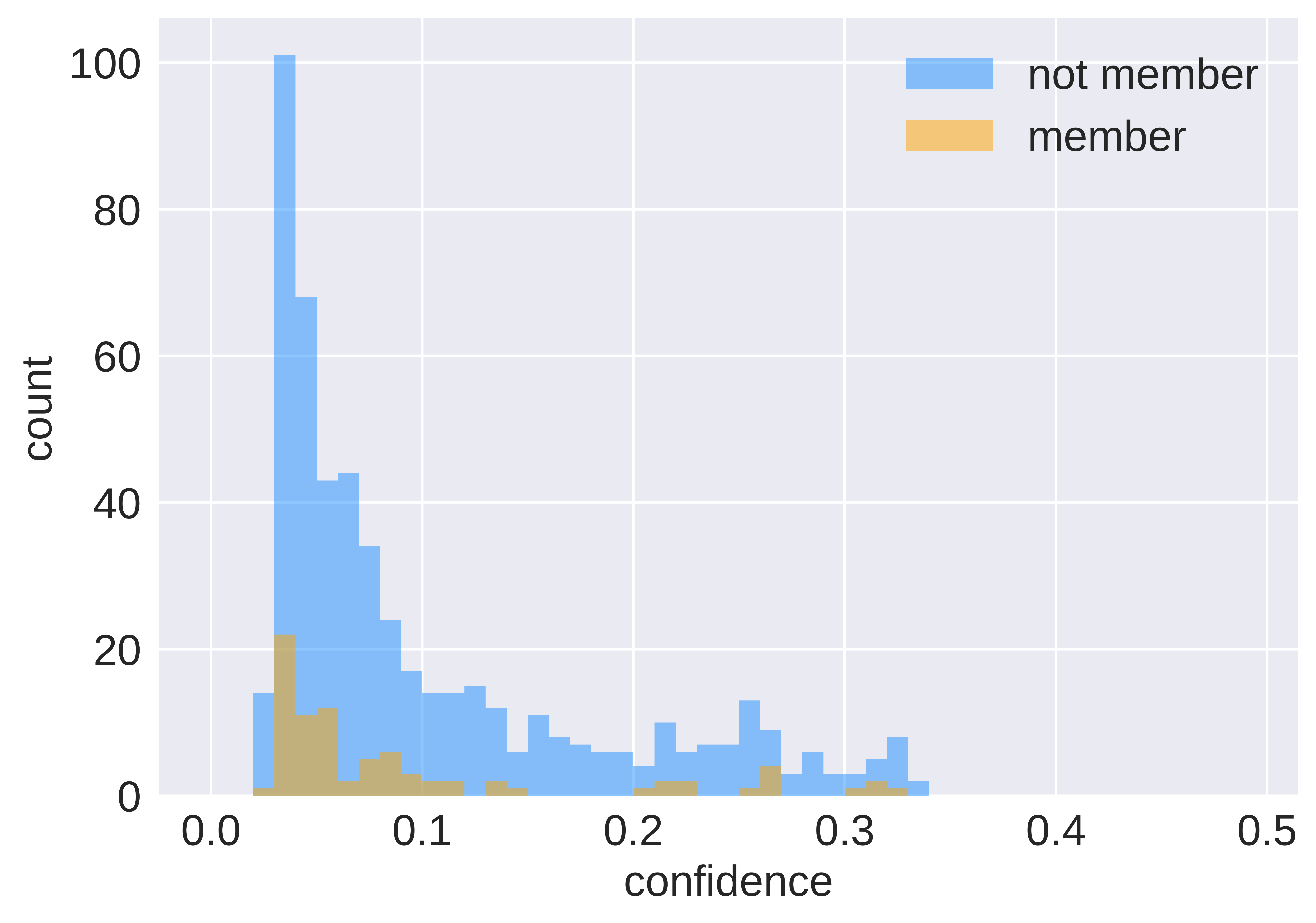}}

	\subfloat[teacher 2, no aug]{\includegraphics[width = 0.5\textwidth]{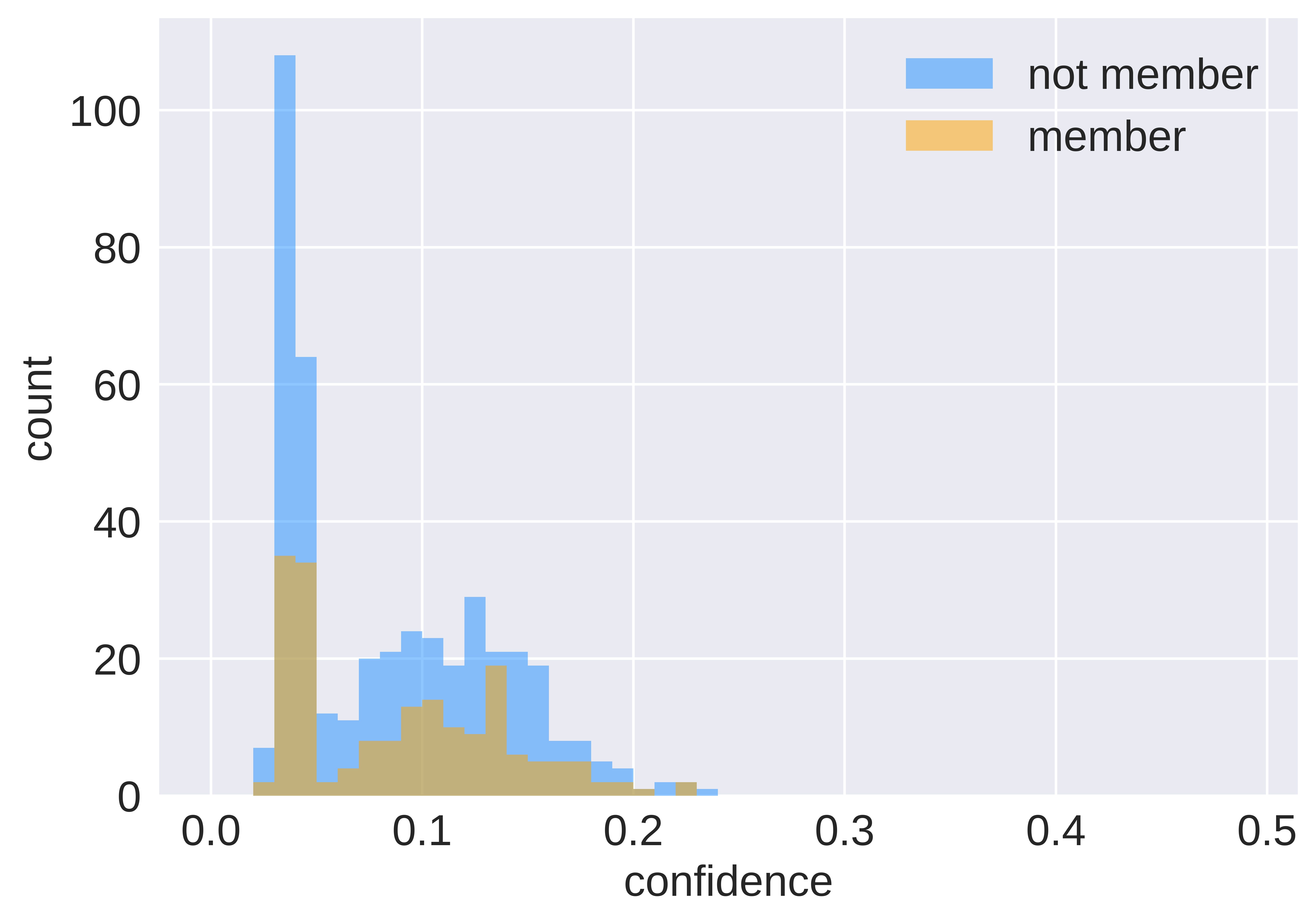}}
	\subfloat[teacher 2, aug]{\includegraphics[width = 0.5\textwidth]{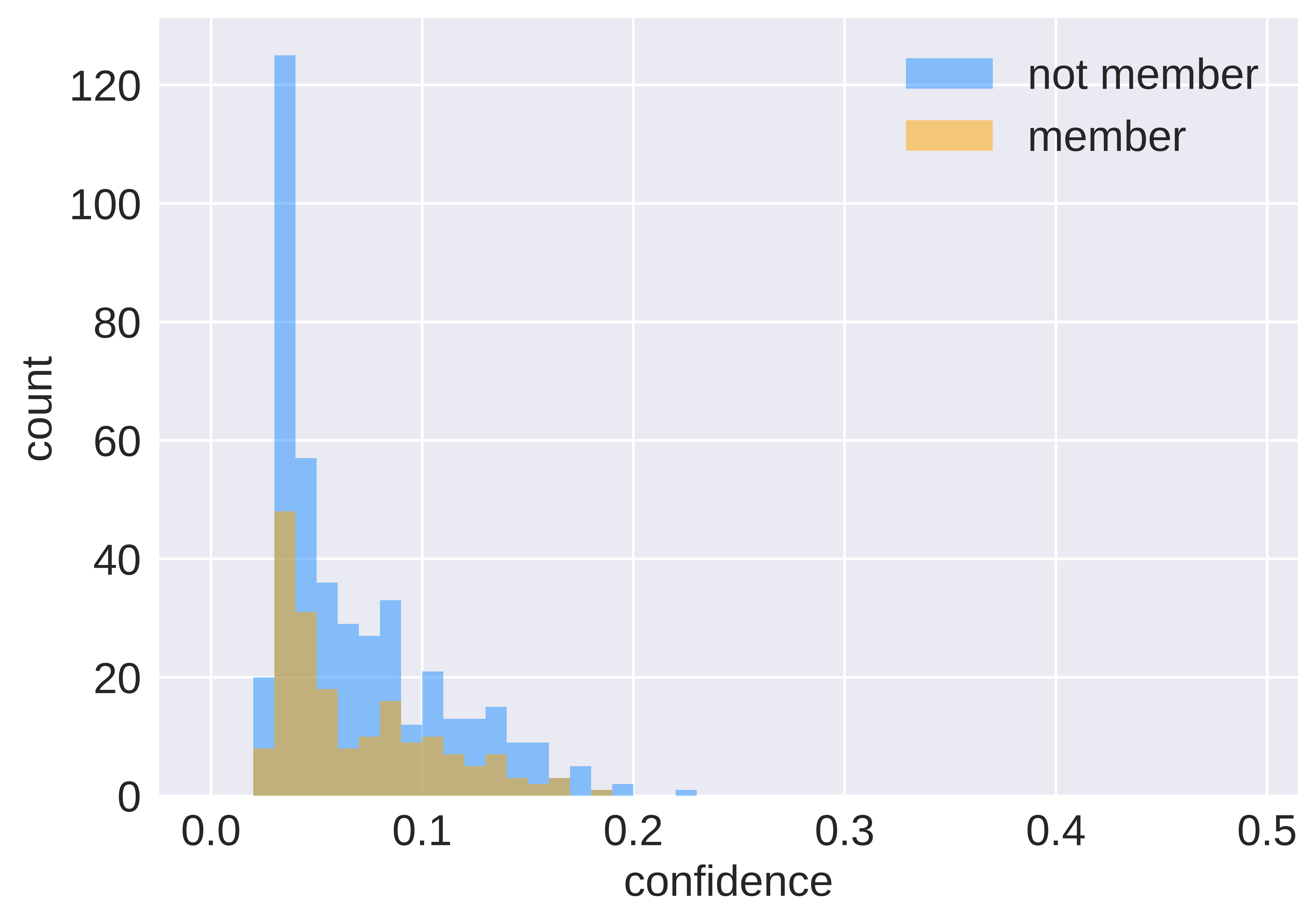}}

	\subfloat[teacher 3, no aug]{\includegraphics[width = 0.5\textwidth]{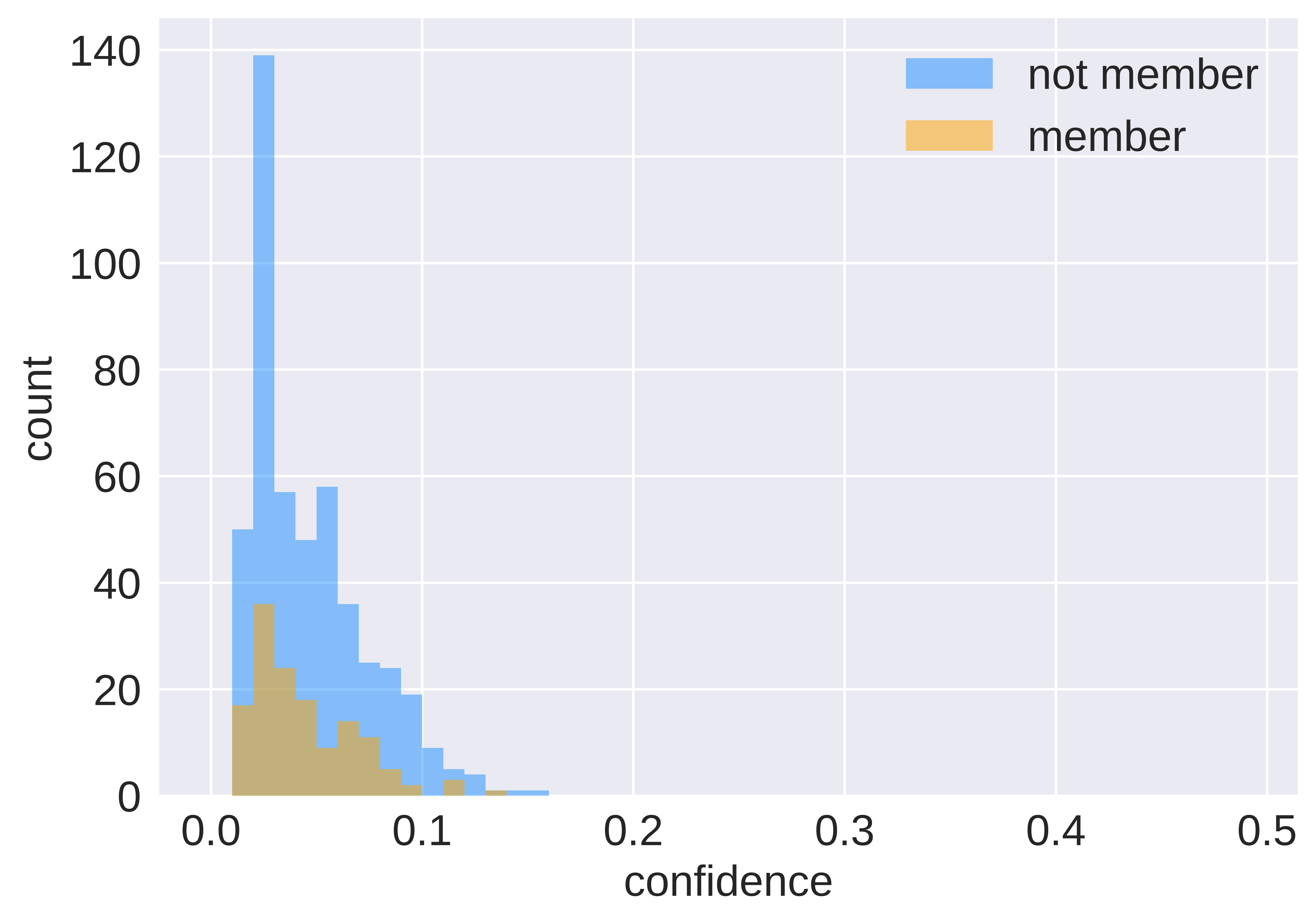}}
	\subfloat[teacher 3, aug]{\includegraphics[width = 0.5\textwidth]{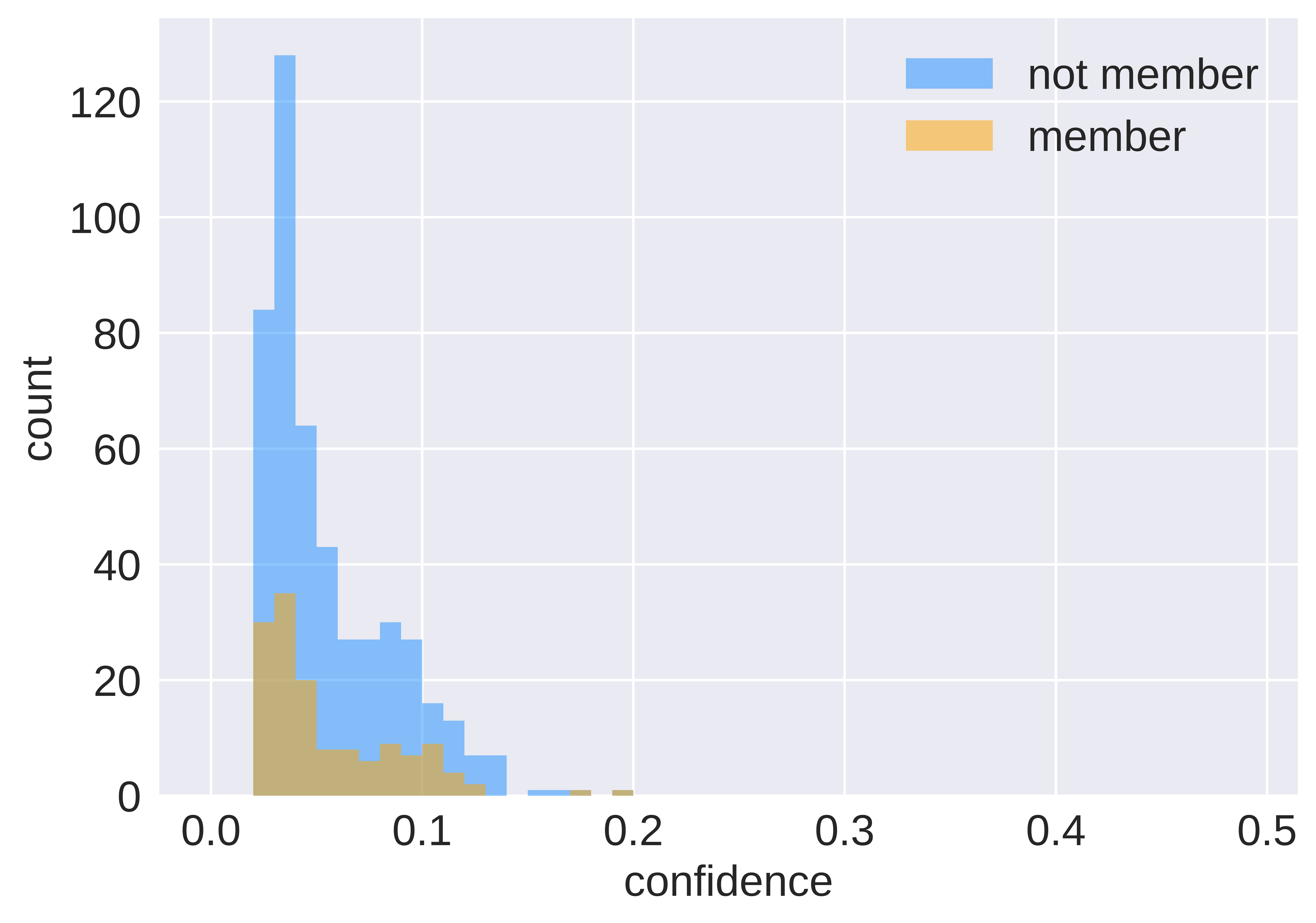}}
	\caption{\textbf{Comparison of uncertainty.}\\Comparison of uncertainty for all clients when teacher models are trained with or without augmentation (aug).}
	\label{Figure S5}
\end{figure*}

As we have mentioned in the experimental setting, we use a patch cropping strategy for training. It can drastically alleviate the risk of data leakage, as proven in \upcite{chobola2022membership}. Another factor influencing the vulnerability to MIA is data augmentation. To give an analysis, we start from the key assumption of MIA, that is, the target model suffers a large generalization gap between training and unseen datasets. Data in the training set of a teacher model are regarded as members while these in the validation set, testing set, and other clients' datasets are marked as 'not member'. \hyperref[Figure S5]{Figure S5} demonstrates uncertainty for three teacher models with or without data augmentation. As can be seen, when data augmentation is added to train a teacher model, the overall uncertainty is reduced. Besides, no matter whether data augmentation is added, in our experimental setting, the overall shape of uncertainty distribution between membership and non-membership is approximate, which can be adverse for MIA. We owe the main reason to the patch cropping training strategy.

Attributed to the above difficulty for MIA, we assume that there is an honest but curious client who is able to get half of the training dataset from the target model to train the shadow model, and that all ground-truth labels are accessible as extra input to the attack model. Note that it is a rather strong assumption, which is hard to satisfy in practice. Here we only use it as a rough upper bound of the privacy risk for our method.

\subsection{Experimental setup for MIA}
For the shadow model, the training set is composed of half of the training dataset from the target model, 60\% data with the lowest uncertainty in the 75\% rest data from the global dataset union, and pseudo labels from the target (teacher) model. The testing set is composed of 40\% data with the highest uncertainty in the 75\% rest data from the global dataset union. Whether data augmentation is used identifies with the setting of the target model. The training loss is Dice and BCE and the metric is Dice and HD between predictions from the shadow model and pseudo labels from the target model. 

For the attack model, the training set is composed of all data for the shadow model, predictions from the shadow model, ground-truth segmentation labels, and binary membership labels. Whether data augmentation is used identifies with the setting of the target model. The training loss is CE and metric is class-wise accuracy and F1 score. The testing set consists of another half of the training dataset from the target model, remaining 25\% data from the global dataset union, predictions from the shadow model, and ground-truth segmentation labels.

\subsection{Performance comparison between teacher models}
The uncertainty comparison only reflects prediction distribution, so we supplement experiments on the performance for in-distribution and out-distribution datasets in \hyperref[Table S12]{Table S12}, where DP is shorthand for differential privacy \upcite{dwork2006differential} discussed in later analysis. It can be concluded that the domain gap for the global dataset union is extremely large but is significantly alleviated through abundant data augmentation. Besides, DP's protection comes with performance degradation.
\begin{table*}[]
	\label{Table S12}
	\centering
	\rowcolors {3}{grey}{white}
	\renewcommand\arraystretch{1.2}
	\resizebox{\columnwidth}{!}{
		\begin{tabular}{llllll}
		\arrayrulecolor{red}
		\hline
		\multicolumn{6}{l}{\textcolor{red}{\textbf{Table S12. Performance comparison between teacher models. }}}  \\
		\hline
		\multicolumn{1}{c}{\textcolor{red}{No aug}} &
		\multicolumn{1}{c}{\textcolor{red}{Kidney}} &
		\multicolumn{1}{c}{\textcolor{red}{Liver}} &
		\multicolumn{1}{c}{\textcolor{red}{Spleen}} &
		\multicolumn{1}{c}{\textcolor{red}{Pancreas}} &
		\multicolumn{1}{c}{\textcolor{red}{Mean}} \\
		\hline
			Client 1 &
			91.43 / 0.77 &
			86.71 / 1.44 &
			0 / 1.61 &
			0.51 / 1.11 &
			47.01 / 2.23 \\
			Client 2 &
			51.41 / 3.66 &
			33.29 / 7.83 &
			93.64 / 0.93 &
			70.59 / 1.53 &
			62.23 / 3.49 \\
			Client 3 &
			43.56 / 2.81 &
			93.73 / 1.71 &
			0.21 / 2.19 &
			10.73 / 1.40 &
			37.06 / 2.03 \\
			Client 4 &
			55.40 / 2.27 &
			90.54 / 2.08 &
			0.99 / 2.84 &
			29.51 / 1.51 &
			44.11 / 2.18 \\
			Mean &
			60.45 / 2.37 &
			76.06 / 3.26 &
			23.41 / 1.89 &
			27.83 / 1.38 &
			47.01 / 2.23 \\
			\hline
			\multicolumn{1}{c}{\textcolor{red}{Aug}} &
			\multicolumn{1}{c}{\textcolor{red}{Kidney}} &
			\multicolumn{1}{c}{\textcolor{red}{Liver}} &
			\multicolumn{1}{c}{\textcolor{red}{Spleen}} &
			\multicolumn{1}{c}{\textcolor{red}{Pancreas}} &
			\multicolumn{1}{c}{\textcolor{red}{Mean}} \\
			\hline
			Client 1 &
			91.48 / 0.80 &
			85.92 / 1.21 &
			53.94 / 1.22 &
			45.35 / 0.85 &
			69.17 / 1.02 \\
			Client 2 &
			68.78 / 3.02 &
			72.38 / 6.34 &
			91.84 / 1.05 &
			70.01 / 1.53 &
			75.75 / 3.03 \\
			Client 3 &
			53.09 / 2.11 &
			90.20 / 1.74 &
			64.66 / 1.27 &
			34.43 / 1.35 &
			60.59 / 1.62 \\
			Client 4 &
			77.48 / 1.57 &
			86.16 / 2.28 &
			75.39 / 1.12 &
			59.76 / 1.11 &
			74.70 / 1.52 \\
			Mean &
			72.70 / 1.87 &
			83.66 / 2.89 &
			71.45 / 1.16 &
			52.38 / 1.21 &
			70.05 / 1.80 \\
			\hline
			\multicolumn{1}{c}{\textcolor{red}{Aug+DP}} &
			\multicolumn{1}{c}{\textcolor{red}{Kidney}} &
			\multicolumn{1}{c}{\textcolor{red}{Liver}} &
			\multicolumn{1}{c}{\textcolor{red}{Spleen}} &
			\multicolumn{1}{c}{\textcolor{red}{Pancreas}} &
			\multicolumn{1}{c}{\textcolor{red}{Mean}} \\
			\hline
			Client 1 &
			90.79 / 0.70 &
			83.76 / 1.20 &
			46.84 / 0.98 &
			38.20 / 0.76 &
			64.89 / 0.91 \\
			Client 2 &
			51.93 / 3.42 &
			68.73 / 5.96 &
			86.12 / 1.17 &
			68.12 / 1.52 &
			68.72 / 3.01 \\
			Client 3 &
			40.64 / 1.96 &
			88.58 / 1.65 &
			59.30 / 1.15 &
			27.10 / 1.22 &
			53.90 / 1.49 \\
			Client 4 &
			70.59 / 1.63 &
			84.50 / 2.26 &
			72.73 / 1.08 &
			55.26 / 1.07 &
			70.77 / 1.51 \\
			Mean &
			63.48 / 1.92 &
			81.39 / 2.76 &
			66.24 / 1.09 &
			47.17 / 1.14 &
			64.57 / 1.73\\
			\hline
			\rowcolor{white}\multicolumn{6}{l}{Here we show Dice / HD (higher / lower numbers are better) on test sets.} \\
			\hline
			\end{tabular}
		}
	\end{table*}

\subsection{MIA attack on teacher models}

The imitation performance and attack performance under the strong attacking assumption are shown in \hyperref[Table S13]{Table S13}. The stimulation performance of shadow models is in positive correlation with performance of teacher model except for client 2, which proves the reasonability of predictions from teacher models. Note that when data augmentation is not added to the teacher model from client 2, it completely fails to generalize on other clients. When segmentation results are meaningless, like fixed simple patterns, the mode is likely to be easier to imitate as a consequence.

As for the attack performance, when data augmentation is employed, the accuracy for class 1 is sharply reduced by 24.09, indicating the attacker cannot correctly distinguish these data in the training set from others and all clients get privacy protection considering this metric. However, the F1 score for class 1 increases by 5.93, meaning the attack model considers a bit more positive predictions into consideration. On the other hand, two metrics in the row 'Attack class mean' is elevated for 'Aug' and 'Aug + DP', suggesting the prediction for the background class is precise due to the better stimulation ability. When DP is used for privacy protection, metrics except for the accuracy of class 1 significantly decline. It demonstrates that with a slight increasement in the accuracy of class 1 (1.07), the attack model is restricted in distinguishing both classes due to ambiguity of prediction distribution.

In conclusion, even under a strong assumption for the MIA attack, data augmentation is helpful to safeguard the membership privacy for these training data. With DP, the overall protection effect can be promoted for all data. Other methods, like knowledge distillation mentioned in \upcite{chobola2022membership}, may also have a great impact and can be thoroughly analyzed in future studies.
\begin{table*}[]
	\label{Table S13}
	\centering
	\rowcolors {3}{grey}{white}
	\renewcommand\arraystretch{1.2}
	\resizebox{\columnwidth}{!}{
		\begin{tabular}{lllll}
		\arrayrulecolor{red}
		\hline
		\multicolumn{5}{l}{\textcolor{red}{\textbf{Table S13. MIA attack performance on teacher models. }}}  \\
		\hline
		\multicolumn{1}{c}{\textcolor{red}{No aug}} &
		\multicolumn{1}{c}{\textcolor{red}{Client 1}} &
		\multicolumn{1}{c}{\textcolor{red}{Client 2}} &
		\multicolumn{1}{c}{\textcolor{red}{Client 3}} &
		\multicolumn{1}{c}{\textcolor{red}{Mean}} \\
		\hline
		Shadow            & 50.39 / 2.588 & 60.36 / 1.115 & 71.43 / 5.189 & 60.72 / 2.964 \\
		Attack class mean & 68.06 / 57.75 & 69.61 / 66.11 & 53.05 / 36.40 & 63.57 / 53.42 \\
		Attack class 1    & 82.92 / 48.22 & 97.87 / 74.49 & 98.63 / 59.01 & 93.14 / 60.57 \\
		\hline
		\multicolumn{1}{c}{\textcolor{red}{Aug}} &
		\multicolumn{1}{c}{\textcolor{red}{Client 1}} &
		\multicolumn{1}{c}{\textcolor{red}{Client 2}} &
		\multicolumn{1}{c}{\textcolor{red}{Client 3}} &
		\multicolumn{1}{c}{\textcolor{red}{Mean}} \\
		\hline
		Shadow            & 64.45 / 2.614 & 46.29 / 1.304 & 82.21 / 5.760 & 64.31 / 3.226 \\
		Attack class mean & 78.32 / 78.56 & 68.12 / 67.18 & 76.23 / 72.74 & 74.22 / 72.82 \\
		Attack class 1    & 65.85 / 66.66 & 46.80 / 59.06 & 94.52 / 73.79 & 69.05 / 66.50 \\
		\hline
		\multicolumn{1}{c}{\textcolor{red}{Aug + DP}} &
		\multicolumn{1}{c}{\textcolor{red}{Client 1}} &
		\multicolumn{1}{c}{\textcolor{red}{Client 2}} &
		\multicolumn{1}{c}{\textcolor{red}{Client 3}} &
		\multicolumn{1}{c}{\textcolor{red}{Mean}} \\
		\hline
		Shadow            & 59.11 / 2.722 & 45.46 / 1.396 & 80.72 / 5.465 & 61.76 / 3.194 \\
		Attack class mean & 70.01 / 66.91 & 69.28 / 68.73 & 57.29 / 45.82 & 65.52 / 60.48 \\
		Attack class 1    & 63.41 / 51.99 & 51.06 / 61.93 & 95.89 / 60.86 & 70.12 / 58.26 \\
		\hline
		\rowcolor{white}\multicolumn{5}{l}{Here we show Dice / HD (higher / lower numbers indicate better stimulation) for} \\
		\rowcolor{white}\multicolumn{5}{l}{the shadow model and accuracy / F1 score (higher numbers indicate better attacks)} \\
		\rowcolor{white}\multicolumn{5}{l}{for the attack model.} \\
		\rowcolor{white}\multicolumn{5}{l}{Class 1 refers to the label 'member'.} \\
		\hline
		\end{tabular}
		}
	\end{table*}

\clearpage


\bibliographystyle{unsrt}
\bibliography{refs}

\end{sloppypar}